\documentclass[10pt,twocolumn,letterpaper]{article}

\usepackage{cvpr}              % To produce the CAMERA-READY version
\usepackage{graphicx}
\graphicspath{ {images/} }
\usepackage{amsmath}
\usepackage{amssymb}
\usepackage{mathtools}
\usepackage{bm}
\usepackage{bbm}
\usepackage{booktabs}
\usepackage{lipsum}
\usepackage{multirow}
\usepackage{tabularx}
\usepackage{makecell}
\usepackage{amssymb}% http://ctan.org/pkg/amssymb
\usepackage{pifont}% http://ctan.org/pkg/pifont
\newcommand{\cmark}{\ding{51}}%
\newcommand{\xmark}{\ding{55}}%
\usepackage{adjustbox}
\usepackage[percent]{overpic}
\usepackage[title]{appendix}
\usepackage[accsupp]{axessibility}

\usepackage{enumitem}
\setlist[itemize]{topsep={0pt},partopsep={0pt}}

\usepackage{caption}
\renewcommand{\arraystretch}{1.1}

\usepackage{xcolor}
\usepackage{colortbl}
\definecolor{gray}{gray}{0.95}

\DeclarePairedDelimiter\floor{\lfloor}{\rfloor}

\usepackage[pagebackref,breaklinks,colorlinks]{hyperref}

\usepackage[capitalize]{cleveref}
\crefname{section}{Sec.}{Secs.}
\Crefname{section}{Section}{Sections}
\Crefname{table}{Table}{Tables}
\crefname{table}{Tab.}{Tabs.}

\def\cvprPaperID{2552} % *** Enter the CVPR Paper ID here
\def\confName{CVPR}
\def\confYear{2022}

\begin{document}

\title{StyleSwin: Transformer-based GAN for High-resolution Image Generation }

\author{Bowen Zhang$^{1}$\thanks{Author did this work during his internship at Microsoft Research Asia.} \qquad Shuyang Gu$^{1}$ \qquad Bo Zhang$^{2}$\thanks{}  \qquad Jianmin Bao$^{2}$ \qquad Dong Chen$^{2}$ \vspace{1pt}\\
Fang Wen$^{2}$ \qquad Yong Wang$^{1}$ \qquad Baining Guo$^{2}$  \vspace{1pt}\\
$^{1}$University of Science and Technology of China  \qquad $^{2}$Microsoft Research Asia
% \qquad
% \\
% \hspace{0.1in}{\tt\small \{zhangbowen, gsy777\}@mail.ustc.edu.cn} \quad {\tt\small yongwang@ustc.edu.cn}\\ \quad {\tt\small \{zhanbo, jianbao, doch, fangwen,  bainguo\}@microsoft.com} \\
}

\twocolumn[{
\renewcommand\twocolumn[1][]{#1}
\maketitle
\centering
\vspace{-0.4cm}
\includegraphics[width=\textwidth]{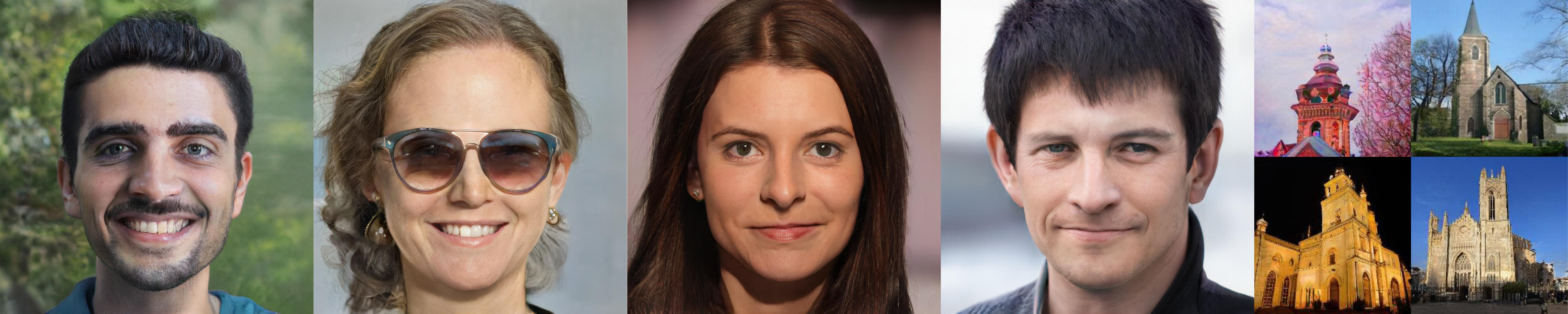}
\vspace{-0.5cm}
\captionsetup{type=figure}
\caption{Image samples generated by our StyleSwin on FFHQ $1024\times 1024$ and LSUN Church $256\times 256$ respectively.}
\label{fig:teaser}

\vspace{0.7cm}
}]

{
  \renewcommand{\thefootnote}%
    {\fnsymbol{footnote}}
  \footnotetext[1]{Author did this work during his internship at Microsoft Research Asia.}
  \footnotetext[2]{Corresponding author.}
}

%%%%%%%%% ABSTRACT
\begin{abstract}
Despite the tantalizing success in a broad of vision tasks, transformers have not yet demonstrated on-par ability as ConvNets in high-resolution image generative modeling. In this paper, we seek to explore using pure transformers to build a generative adversarial network for high-resolution image synthesis. To this end, we believe that local attention is crucial to strike the balance between computational efficiency and modeling capacity. Hence, the proposed generator adopts Swin transformer in a style-based architecture. To achieve a larger receptive field, we propose double attention which simultaneously leverages the context of the local and the shifted windows, leading to improved generation quality. Moreover, we show that offering the knowledge of the absolute position that has been lost in window-based transformers greatly benefits the generation quality. The proposed StyleSwin is scalable to high resolutions, with both the coarse geometry and fine structures benefit from the strong expressivity of transformers. However, blocking artifacts occur during high-resolution synthesis because performing the local attention in a block-wise manner may break the spatial coherency. To solve this, we empirically investigate various solutions, among which we find that employing a wavelet discriminator to examine the spectral discrepancy effectively suppresses the artifacts. Extensive experiments show the superiority over prior transformer-based GANs, especially on high resolutions, \eg, $1024\times 1024$. The StyleSwin, without complex training strategies, excels over StyleGAN on CelebA-HQ $1024$, and achieves on-par performance on FFHQ-$1024$, proving the promise of using transformers for high-resolution image generation. The code and pretrained models are available at \url{https://github.com/microsoft/StyleSwin}.
\end{abstract}

%%%%%%%%% BODY TEXT
\vspace{-0.5em}
\section{Introduction}
\label{sec:intro}
The state of image generative modeling has seen dramatic advancement in recent years, among which generative adversarial networks~\cite{Goodfellow2014GenerativeAN,liu2020generative} (GANs) offer arguably the most compelling quality on synthesizing high-resolution images. While early attempts focus on stabilizing the training dynamics via proper regularization~\cite{Miyato2018SpectralNF,Gulrajani2017ImprovedTO,Kurach2019ALS,Mescheder2018WhichTM,gu2020giqa} or adversarial loss designs~\cite{JolicoeurMartineau2019TheRD,Mao2017LeastSG,Lim2017GeometricG,Arjovsky2017WassersteinG}, remarkable performance leaps in recent prominent works mainly attribute to the architectural modifications that aim for stronger modeling capacity, such as adopting self-attention~\cite{Zhang2019SelfAttentionGA}, aggressive model scaling~\cite{Brock2019LargeSG}, or style-based generators~\cite{karras2019stylebased,Karras2020AnalyzingAI}. Recently, drawn by the broad success of transformers in discriminative models~\cite{Dosovitskiy2021AnII,liu2021swin,Khan2021TransformersIV}, a few works~\cite{jiang2021transgan,Lee2021ViTGANTG,Zhao2021ImprovedTF,Xu2021STransGANAE} attempt to use pure transformers to build generative networks in the hope that the increased expressivity and the ability to model long-range dependencies can benefit the generation of complex images, yet high-quality image generation, especially on high resolutions, remains challenging. 

This paper aims to explore key ingredients when using transformers to constitute a competitive GAN for high-resolution image generation. The first obstacle is to tame the quadratic computational cost so that the network is scalable to high resolutions, \eg, $1024\times 1024$. We propose to leverage Swin transformers~\cite{liu2021swin} as the basic building block since the window-based local attention strikes a balance between computational efficiency and modeling capacity. As such, we could take advantage of the increased expressivity to characterize all the image scales, as opposed to reducing to point-wise multi-layer perceptrons (MLP) for higher scales~\cite{Zhao2021ImprovedTF}, and the synthesis is scalable to high resolution,
\eg, $1024\times 1024$, with delicate details. Besides, the local attention introduces locality inductive bias so there is no need for the generator to relearn the regularity of images from scratch. These merits make a simple transformer network substantially outperform the convolutional baseline. 

In order to compete with the state of the arts, we further propose three instrumental architectural adaptations. First, we strengthen the generative model capacity by employing the local attention in a style-based architecture~\cite{karras2019stylebased}, during which we empirically compare various style injection approaches for our transformer GAN. 
Second, we propose double attention in order to enlarge the limited receptive field brought by the local attention, where each layer attends to both the local and the shifted windows,
effectively improving the generator capacity without much computational overhead. 
Moreover, we notice that Conv-based GANs implicitly utilize zero padding to infer the absolute positions, a crucial clue for generation, yet such feature is missing in the window-based transformers. We propose to fix this by introducing sinusoidal positional encoding~\cite{tancik2020fourier} to each layer such that absolute positions can be leveraged for image synthesis. Equipped with the above techniques, the proposed network, dubbed as \emph{StyleSwin}, starts to show advantageous generation quality on $256\times 256$ resolution.

Nonetheless, we observe blocking artifacts when synthesizing high-resolution images. We conjecture that these disturbing artifacts arise because computing the attention independently in a block-wise manner breaks the spatial coherency. That is, while proven successful in discriminative tasks~\cite{vaswani2021scaling,liu2021swin}, the block-wise attention requires special treatment when applied in synthesis networks. To tackle these blocking artifacts, we empirically investigate various solutions, among which we find that a wavelet discriminator~\cite{gal2021swagan} examining the artifacts in spectral domain could effectively suppress the artifacts, making our transformer-based GAN yield visually pleasing outputs.

The proposed \emph{StyleSwin}, achieves state-of-the-art quality on multiple established benchmarks, \eg, FFHQ, CelebA-HQ, and LSUN Church. In particular, our approach shows compelling quality on high-resolution image synthesis (Figure~\ref{fig:teaser}), achieving competitive quantitative performance relative to the leading ConvNet-based methods without complex training strategies. On CelebA-HQ $1024$, our approach achieves an FID of $4.43$, outperforming all the prior works including StyleGAN~\cite{karras2019stylebased}; whereas on FFHQ-$1024$, we obtain an FID of $5.07$, approaching the performance of  StyleGAN2~\cite{Karras2020AnalyzingAI}. 

\begin{figure*}[t]
    \centering
    \begin{overpic}
        [scale=0.38]{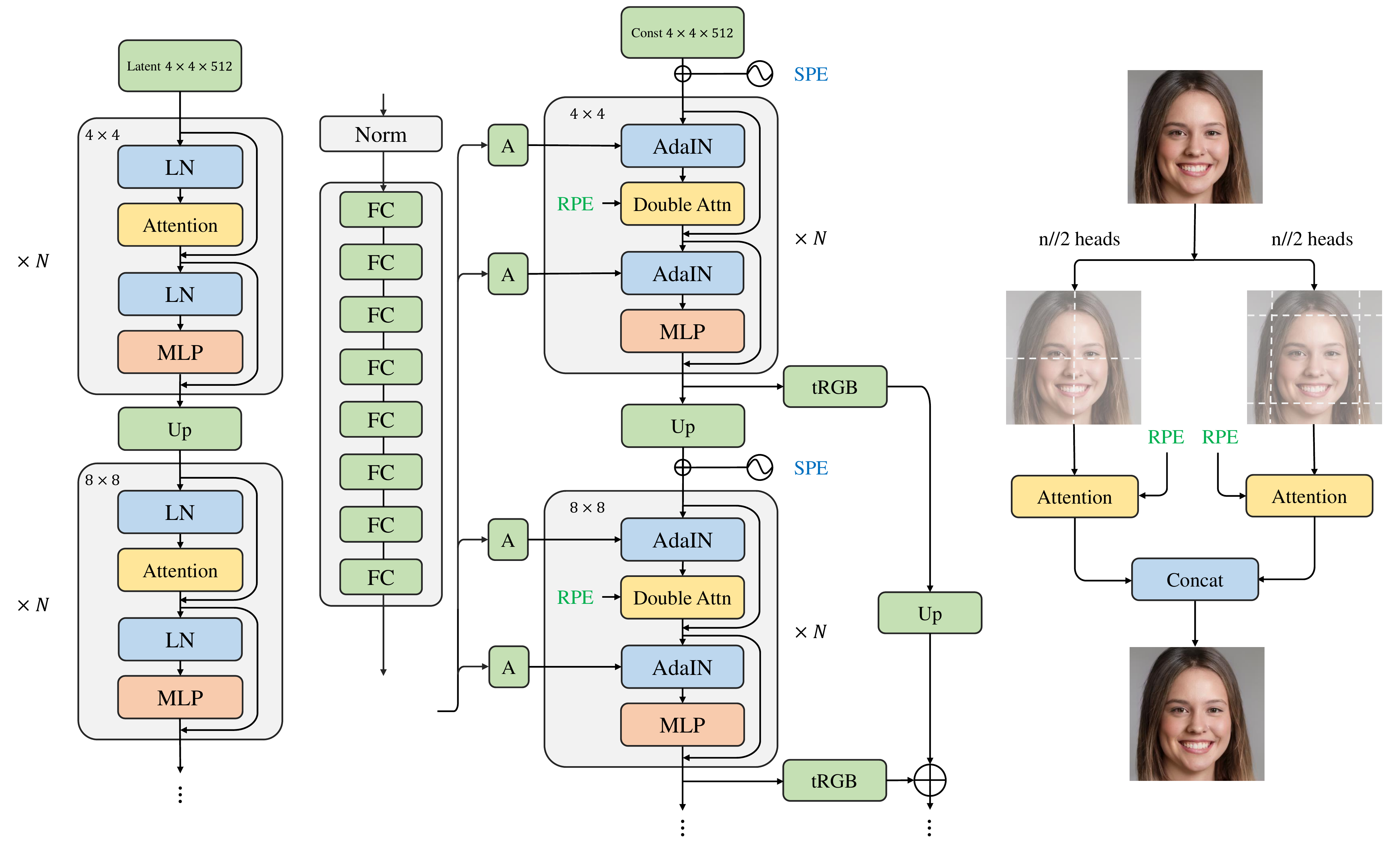} 
        \put(25, 55.3){\small $\bm{z} \in \mathcal{Z}$}
        \put(24, 9.5){\small $\bm{w} \in \mathcal{W}$}
        \put(84.5, 56.8){$\bm{x}$}
        \put(82, 34.8){\small$\bm{x}_w$}
        \put(85.6, 34.8){\small$\bm{x}_{sw}$}
        \put(12,-1.3){\small(a)} 
        \put(47.7,-1.3){\small(b)}
        \put(84.5,-1.3){\small(c)}
    \end{overpic}
    \vspace{1.0em}
    \caption{{The architectures we investigate.} (a) {The baseline architecture is comprised of a series of transformer blocks hierarchically.}  (b) The proposed \emph{StyleSwin} adopts style-based architecture, where the style codes derived from the latent code $\bm z$ modulate the feature maps of transformer blocks through style injection.  (c) The proposed double attention enlarges the receptive field of transformer blocks by using split heads attending to the local and the shifted windows respectively. }
    \label{fig:overall_framework}
\end{figure*}

\vspace{-0.5em}
\section{Related Work}
\label{sec:related_work}
% \vspace{0.5em}
\noindent\textbf{High-resolution image generation.} 
Image generative modeling has improved rapidly in the past decade~\cite{kingma2013auto,Goodfellow2014GenerativeAN,van2016pixel,ho2020denoisingDP,kingma2018glow,liu2020generative}. Among various solutions, generative adversarial networks (GANs) offer competitive generation quality. While early methods~\cite{Arjovsky2017WassersteinG,radford2015unsupervised,Miyato2018SpectralNF} focus on stabilizing the adversarial training, recent prominent works~\cite{Brock2019LargeSG,karras2019stylebased, Karras2020AnalyzingAI, karras2021aliasFG} rely on designing architectures with enhanced capacity, which considerably improves generation quality. However, contemporary GAN-based methods adopt convolutional backbones which are now deemed to be inferior to transformers in terms of modeling capacity. In this paper, we are interested in applying the emerging vision transformers to GANs for high-resolution image generation.

% \vspace{0.5em}
\noindent\textbf{Vision transformers.} Recent success of transformers~\cite{Vaswani2017AttentionIA,brown2020language} in NLP tasks inspires the research of vision transforms. The seminal work ViT~\cite{Dosovitskiy2021AnII} proposes a pure transformer-based architecture for image classification and demonstrates the great potential of transformers for vision tasks. Later, transformers dominate the benchmarks in a broad of discriminative tasks~\cite{liu2021swin, vaswani2021scaling, dong2021cswin,wu2021cvt, wang2021pyramid,touvron2021training, yuan2021tokens, han2021transformer}. However, the self-attention in transformer blocks brings quadratic computational complexity, which limits its application for high-resolution inputs. A few recent works~\cite{liu2021swin, vaswani2021scaling, dong2021cswin} tackle this problem by proposing to compute self-attention in local windows, so that linear computational complexity can be achieved. Moreover, the hierarchical architecture makes them suitable to serve as general purpose backbones. 

% \vspace{0.5em}
\noindent\textbf{Transformer-based GANs.} 
Recently, the research community begins to explore using transformers for generative tasks in the hope that the increased expressivity can benefit the generation of complex images. One natural way is to use transformers to synthesize pixels in an auto-regressive manner~\cite{chen2020generative,esser2020taming}, but the slow inference speed limits their practical usage. Recently a few works~\cite{jiang2021transgan,Lee2021ViTGANTG,Zhao2021ImprovedTF,Xu2021STransGANAE} attempt to propose transformer-based GANs, yet most of these methods only support the synthesis up to $256\times 256$ resolution. Notably, the HiT~\cite{Zhao2021ImprovedTF} successfully generates $1024\times 1024$ images at the cost of reducing to MLPs in its high-resolution stages, hence unable to synthesize high-fidelity details as the Conv-based counterpart~\cite{karras2019stylebased}. In comparison, our \emph{StyleSwin} can synthesize fine structures using transformers, leading to comparable quality as the leading ConvNets on high-resolution synthesis.

\section{Method}
\subsection{Transformer-based GAN architecture}

We start from a simple generator architecture, as shown in Figure~\ref{fig:overall_framework}(a), which receives a latent variable $\bm{z}\sim \mathcal{N}(0,\bm{I})$ as input and gradually upsamples the feature maps through a cascade of transformer blocks. 

Due to the quadratic computational complexity, it is unaffordable to compute full-attention on high-resolution feature maps. We believe that local attention is a good way to achieve trade-off between computational efficiency and modeling capacity. We adopt Swin transformer~\cite{liu2021swin} as the basic building block which computes multi-head self-attention (MSA)~\cite{Vaswani2017AttentionIA} locally in non-overlapping windows. To advocate the information interaction across adjacent windows, Swin transformer uses shifted window partition in alternative blocks. Specifically, given the input feature map $\bm{x}^l\in\mathbb{R}^{H \times W \times C}$ of layer $l$, the consecutive Swin blocks operate as follows:
\begin{equation}
    \begin{split}
        &\begin{rcases*}
        \hat{\bm{x}}^l = \text{W-MSA}(\text{LN}(\bm{x}^l)) + \bm{x}^l\\
        \bm{x}^{l+1} = \text{MLP}(\text{LN}(\hat{\bm{x}}^l)) + \hat{\bm{x}}^l
        \end{rcases*}\ \text{regular window},\\
        &\begin{rcases*}
        \hat{\bm{x}}^{l+1} = \text{SW-MSA}(\text{LN}(\bm{x}^{l+1})) + \bm{x}^{l+1}\\
        \bm{x}^{l+2} = \text{MLP}(\text{LN}(\hat{\bm{x}}^{l+1})) + \hat{\bm{x}}^{l+1}
        \end{rcases*}\ \text{shifted window},
    \end{split}
\end{equation}
where {W-MSA} and {SW-MSA} denote the window-based multi-head self-attention under the regular and shifted window partitioning respectively, and LN stands for layer normalization. Since such block-wise attention induces linear computational complexity relative to the image size, the network is scalable to the high-resolution generation where the fine structures can be modeled by these capable transformers as well. 

Since the discriminator severely affects the stability of adversarial training, we opt to use a Conv-based discriminator directly from~\cite{karras2019stylebased}. In our experiment, we find that simply replacing the convolution with transformer blocks under this baseline architecture yields more stabilized training due to the improved model capacity. However, such naive architecture cannot make our transformer-based GAN compete with the state of the arts, so we make further studies which we introduce as follows.

\vspace{0.5em}
\noindent\textbf{Style injection.} We first strengthen the model capability by adapting the generator to a style-based architecture~\cite{karras2019stylebased, Karras2020AnalyzingAI} as shown in Figure~\ref{fig:overall_framework}(b). We learn a non-linear mapping $f:\mathcal{Z}\rightarrow\mathcal{W}$ to map the latent code $\bm{z}$ from $\mathcal{Z}$ space to $\mathcal{W}$ space, which specifies the styles that are injected into the main synthesis network. We investigate the following style injection approaches:
\begin{itemize}[leftmargin=*]
    \itemsep=-0.9mm
    \item \emph{AdaNorm} modulates the statistics (\ie, mean and variance) of feature maps after normalization. We study multiple normalization variants, including instance normalization (IN)~\cite{ulyanov2016instance}, batch normalization (BN)~\cite{ioffe2015batch}, layer normalization (LN)~\cite{ba2016layer} and the recently proposed RMSnorm~\cite{Zhang2019RootMS}. Since the RMSNorm removes the mean-centering of LN,  we only predict the variance from the $\mathcal{W}$ code.  
    \item \emph{Modulated MLP}: Instead of modulating feature maps, one can also modulate the weights of linear layers. Specifically, we rescale the channel-wise weight magnitude of the feed-forward network (FFN) within transformer blocks. According to~\cite{Karras2020AnalyzingAI}, such style injection admits faster speed than AdaNorm.
    \item \emph{Cross-attention}: Motivated by the decoder transformer~\cite{Vaswani2017AttentionIA}, we explore a transformer-specific style injection in which the transformers additionally attend to the style tokens derived from the $\mathcal{W}$ space. The effectiveness of this cross-attention is also validated in~\cite{Zhao2021ImprovedTF}. 
\end{itemize}

\begin{table}[tb]
    \small
    \centering
    \begin{tabular}{c|c}
        \toprule
         Style injection methods & FID $\downarrow$ \\
         \midrule
         Baseline & 15.03 \\
         \midrule
         AdaIN & \textbf{6.34}\\
         AdaLN & 6.95\\
         AdaBN & $>$ 100 \\
         AdaRMSNorm & 7.43 \\
         \midrule
         Modulated MLP & 7.09\\
         Cross attention & 6.59 \\
         \bottomrule
    \end{tabular}
    \caption{{Comparison of different style injection methods on FFHQ-$256$.} The style injection methods considerably improve the FID, among which the AdaIN leads to the best generation quality.}
    \label{tab:style_injection}
\end{table}

Table~\ref{tab:style_injection} shows that all the above style injection methods significantly boost the generative modeling capacity except that the training with AdaBN does not converge because the batch size is compromised for high-resolution synthesis. In comparison, AdaNorm brings more sufficient style injection possibly because the network could take advantage of the style information twice --- in either the attention block and the FFN, whereas the modulated MLP and cross-attention make use of the style information once. We did not further study the hybrid of modulated MLP and cross-attention due to efficiency considerations. Furthermore, compared to AdaBN and AdaLN, AdaIN offers finer and more sufficient feature modulation as feature maps are normalized and modulated independently, so we choose AdaIN by default for our following experiments. 

\vspace{0.5em}
\noindent\textbf{Double attention.} Using local attention, nonetheless, sacrifices the ability to model long-range dependencies, which is pivotal to capture geometry~\cite{Zhang2019SelfAttentionGA,Brock2019LargeSG}. Let the window size used by the Swin block be $\kappa\times \kappa$, then due to the shifted window strategy, the receptive field increases by~$\kappa$ in each dimension using one more Swin block. Suppose we use Swin blocks to process a $64\times 64$ feature map and we by default choose $\kappa=8$, then it takes $64/\kappa=8$ transformer blocks to span over the entire feature map. 

In order to achieve an enlarged receptive field, we propose \emph{double attention} which allows a single transformer block to simultaneously attend to the context of the local and shifted windows. As illustrated in Figure~\ref{fig:overall_framework}(c), we split $h$ attention heads into two groups: the first half of heads perform the regular window attention whereas the second half compute the shifted window attention, both of whose results are further concatenated to form the output. Specifically, we denote with $\bm{x}_{{w}}$ and $\bm{x}_{{sw}} $ the non-overlapping patches under the regular and shifted window partitioning respectively, \ie $\bm{x}_{{w}},\bm{x}_{{sw}} \in \mathcal{R}^{\frac{HW}{\kappa^2}\times \kappa \times \kappa \times C}$, then the double attention is formulated as,
\begin{equation}
    \text{Double-Attention} = \text{Concat}(\text{head}_1,..., \text{head}_h)\bm{W}^O
    \label{eq:double_attention}
\end{equation}
where $\bm{W}^O\in \mathcal{R}^{C\times C}$ is the projection matrix used to mix the heads to output. The attention heads in Equation~\ref{eq:double_attention} can be computed as:
\begin{equation}
    \thinmuskip=1mu
    \thickmuskip=1mu
    \text{head}_i = \begin{cases}
        \text{Attn}(\bm{x}_{w}\bm{W}_i^Q,\bm{x}_{w}\bm{W}_i^K,\bm{x}_{w}\bm{W}_i^V) & i\leq\floor*{\frac{h}{2}}\\
        \text{Attn}(\bm{x}_{sw}\bm{W}_i^Q,\bm{x}_{sw}\bm{W}_i^K,\bm{x}_{sw}\bm{W}_i^V) & i>\floor*{\frac{h}{2}}
    \end{cases}
\end{equation}
where $\bm{W}_i^Q, \bm{W}_i^K, \bm{W}_i^V \in \bm{R}^{C\times (C/h)}$ are query, key and value projection matrix for $i$-th head respectively. One can derive that the receptive field of each dimension increases by $2.5\kappa$ with one additional double attention block, which allows capturing larger context more efficiently. Still, for a $64\times 64$ input, it now takes 4 transformer blocks to cover the entire feature map.

\vspace{0.5em}
\noindent\textbf{Local-global positional encoding.} Relative positional encoding (RPE) adopted by the default Swin blocks encodes the relative position of pixels and has proven crucial for discriminative tasks~\cite{liu2021swin,dai2021coatnet}. Theoretically, a multi-head local attention layer with RPE can express any convolutional layer of window-sized kernels~\cite{Cordonnier2020OnTR,Li2021CanVT}. However, when substituting the convolutional layers with transformers that use RPE, one thing is rarely noticed: ConvNets could infer the absolute positions by leveraging the clues from the zero paddings~\cite{islam2020much,kayhan2020translation} yet such feature is missing in Swin blocks using RPE. On the other hand, it is essential to let the generator be aware of the absolute positions because the synthesis of a specific component, \eg, mouth, highly depends on its spatial coordinate~\cite{lin2019coco,anokhin2020image}.   

In view of this, we propose to introduce sinusoidal position encoding~\cite{Vaswani2017AttentionIA,xu2021positional,choi2021toward} (SPE) on each  scale, as shown in Figure~\ref{fig:overall_framework}(b). Specifically, after the scale upsampling, the feature maps are added with the following encoding:
\begin{equation}
\thinmuskip=1.2mu
\thickmuskip=1.2mu
[\underbrace{\sin(\omega_0 {i}),\, \cos(\omega_0 {i}),\, \cdots}_{\rm horizontal\ dimension},\: \underbrace{\sin(\omega_0 {j}),\, \cos(\omega_0 {j}),\, \cdots}_{\rm vertical\ dimension}]\in \mathbb{R}^{C},
\end{equation}
where  and $\omega_k = 1/10000^{2k}$ and $(i,j)$ denotes the 2D location. We use SPE rather than learnable absolute positional encoding~\cite{Dosovitskiy2021AnII} because SPE admits translation invariance~\cite{wang2020position}. In practice, we make the best of RPE and SPE by employing them altogether: the RPE applied within each transformer block offers the relative positions within the local context, whereas the SPE introduced on each scale informs the global position. 

\begin{figure}[t]
    \center
    \small
    \setlength\tabcolsep{1pt}
    \renewcommand{\arraystretch}{0.6}
    {
    \begin{tabular}{@{}cc@{}}
         \includegraphics[width=0.45\columnwidth]{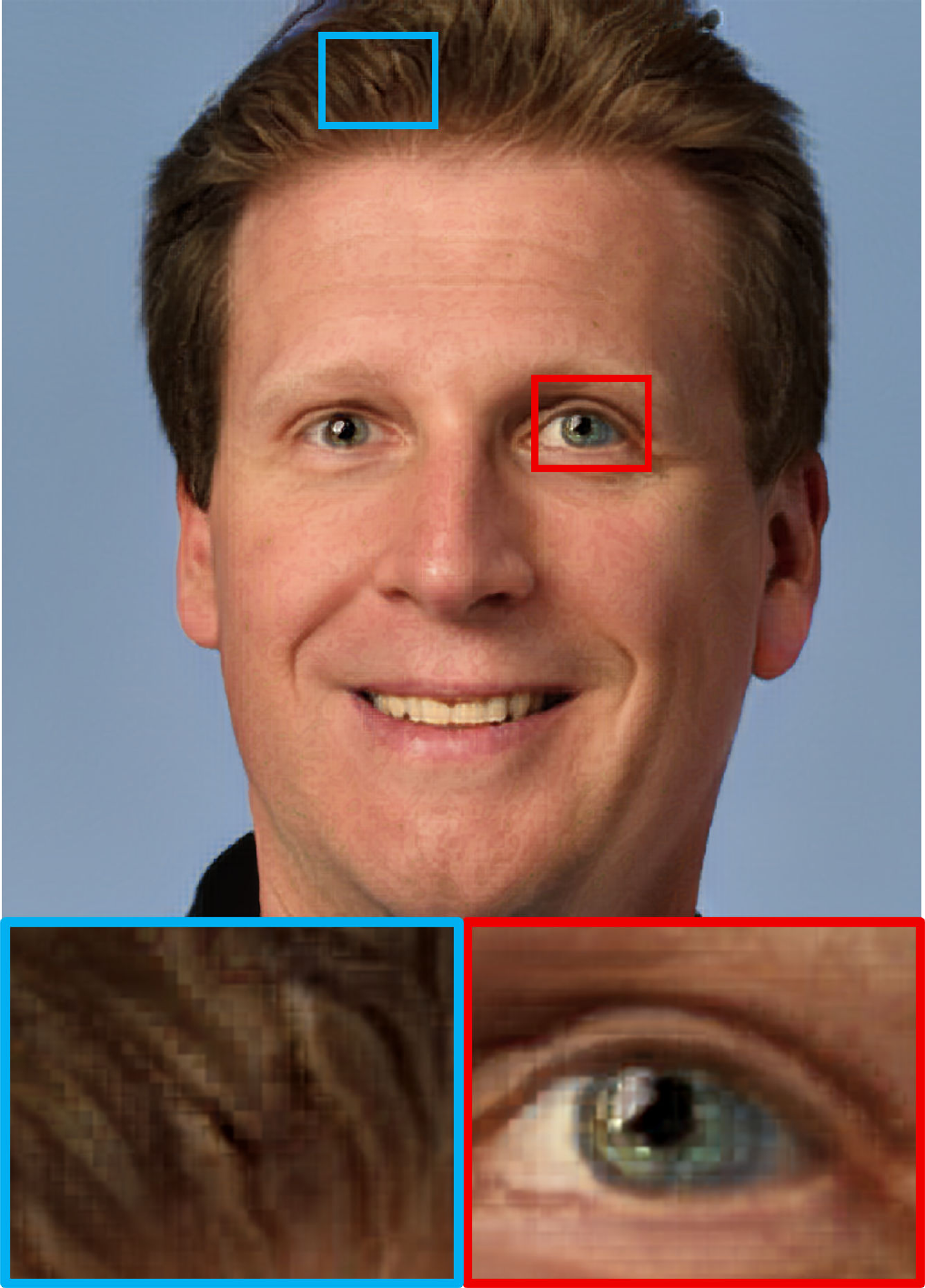} & \includegraphics[width=0.45\columnwidth]{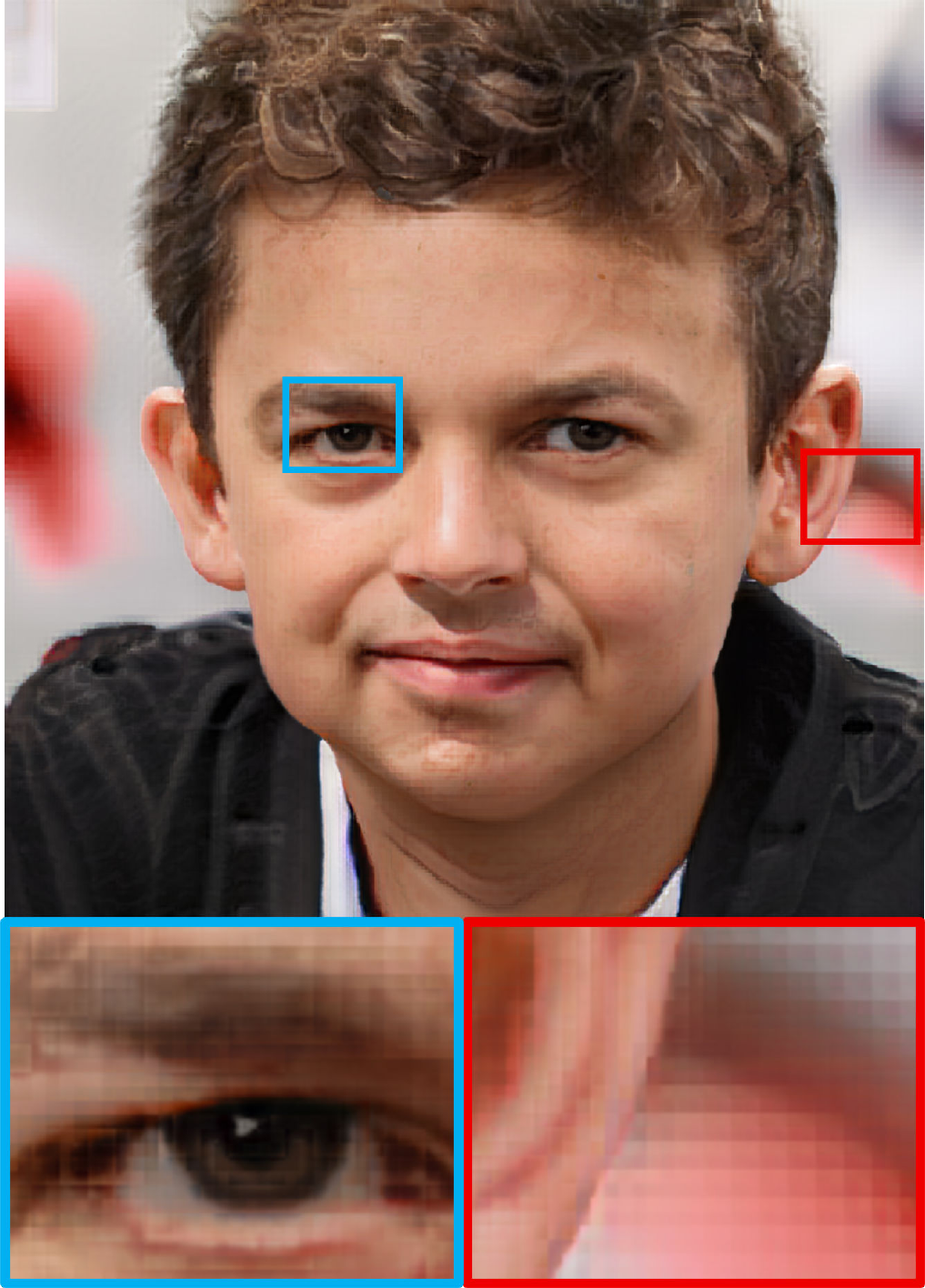} 
    \end{tabular}
    }
    \caption{{Blocking artifacts become obvious on $1024\times 1024$ resolution.} These artifacts correlate with the window size of local attentions. }
    \label{fig:artifacts}
\end{figure}

\subsection{Blocking artifact in high-resolution synthesis}
While achieving state-of-the-art quality on synthesizing $256\times 256$ images with the above architecture, directly applying it for higher resolution synthesis, \eg, $1024\times 1024$, brings blocking artifacts as shown in Figure~\ref{fig:artifacts}, which severely affects the perceptual quality. Note that these are by no means the checkboard artifacts caused by the transposed convolution~\cite{odena2016deconvolution} as we use bilinear upsampling followed by anti-aliasing filters as~\cite{karras2019stylebased}. 

\begin{figure}[tb]
    \centering
    \small
    \begin{overpic}
        [scale=0.31]{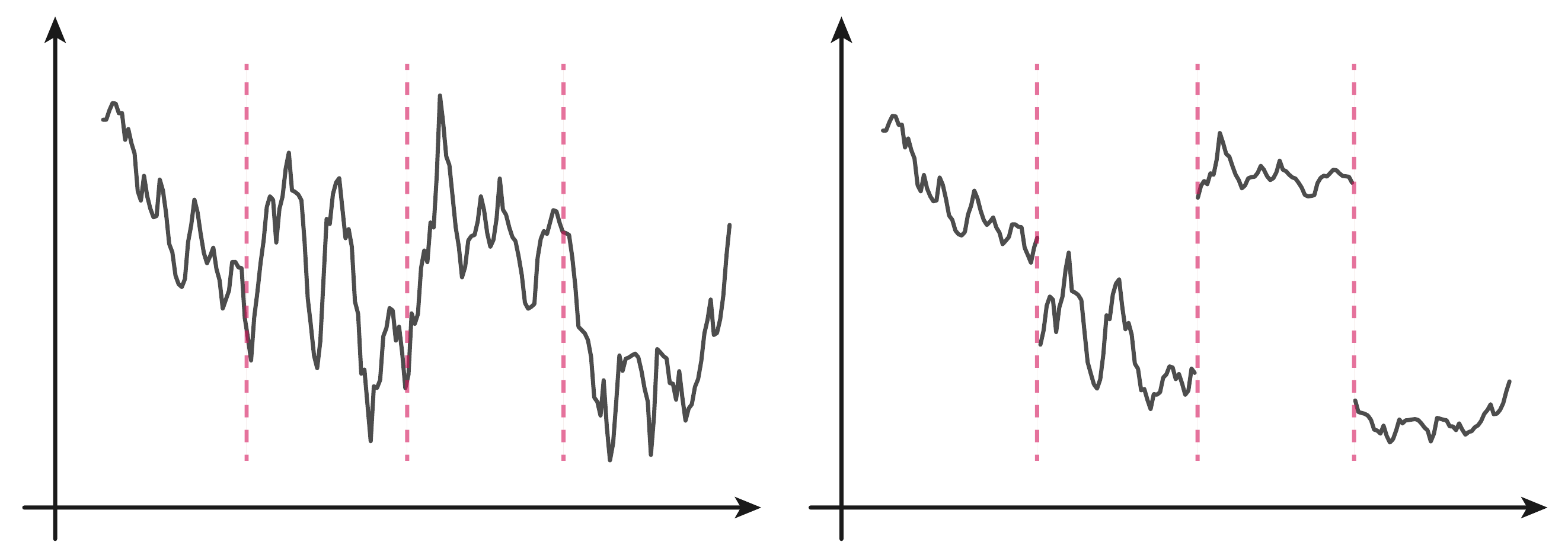} 
        \put(23,-2){\small(a)} 
        \put(73.5,-2){\small(b)}
    \end{overpic}
    \vspace{0em}
    \caption{{A 1D example illustrates that the window-wise local attention causes blocking artifacts.}  (a) Input continuous signal along with partitioning windows. (b) Output discontinuous signal after window-wise attention. For simplicity, we adopt one attention head with random projection matrices.}
    \label{fig:attention}
\end{figure}

We conjecture that the blocking artifacts are caused by the transformers. To verify this, we remove the attention operators starting from $64\times 64$ and employ only MLPs to characterize the high-frequency details. This time we obtain artifact-free results. To be further, we find that these artifacts exhibit periodic patterns with a strong correlation with the window size of local attention. Hence, we are certain it is the window-wise processing that breaks the spatial coherency and causes the blocking artifacts. To simplify, one can consider the 1D case in Figure~\ref{fig:attention}, where attention is computed locally in strided windows. For a continuous signal, the window-wise local attention is likely to produce a discontiguous output because the values within the same window tend to become uniform after the softmax operation, so the outputs of neighboring windows appear rather distinct. The 2D case is analogous to the JPEG compression artifacts 
caused by the block-wise encoding~\cite{liu2002efficient}.

\subsection{Artifact suppression}
In the next, we discuss a few solutions to suppress the blocking artifacts. 

\vspace{0.5em}
\noindent\textbf{Artifact-free generator.}
We first attempt to reduce artifacts by improving the generator.
\begin{itemize}[leftmargin=*]
    \itemsep=-0.9mm
    \item \emph{Token sharing}. Blocking artifacts arise because there is an abrupt change of keys and values used by the attention computing across distinct windows, so we propose to make windows have shared tokens in a way like HaloNet~\cite{vaswani2021scaling}. However, artifacts are still noticeable since there always exist tokens exclusive to specific windows. 
    % \item \emph{Convolutions} can be introduced to remedy the window-like artifacts, but this requires to stack multiple layers or adopt a large kernel so as to filter the large window artifacts. Computational cost discourages this option.
    \item Theoretically, \emph{sliding window attention}~\cite{hu2019local} should lead to artifact-free results. Note that training the generator with sliding attention is too costly so we only adopt the sliding window for inference. 
    \item \emph{Reduce to MLPs on fine scales}. Just as~\cite{Zhao2021ImprovedTF}, one can remove self-attention  and purely rely on point-wise MLPs for fine structure synthesis at the cost of sacrificing the ability to model high-frequency details.
\end{itemize}

\vspace{0.5em}
\noindent\textbf{Artifact-suppression discriminator.} Indeed, we observe blocking artifacts in the early training phase on $256\times 256$ resolution, but they gradually fade out as training precedes. In other words, although the window-based attention is prone to produce artifacts, the generator does have the capability to offer an artifact-free solution. The artifacts plague the high-resolution synthesis because the discriminator fails to examine the high-frequency details. This enlightens us to resort to stronger discriminators for artifact suppression. 
\begin{figure}[t]
    \center
    \small
    \setlength\tabcolsep{1pt}
    {
    \renewcommand{\arraystretch}{0.6}
    \begin{tabular}{@{}ccc@{}}
         \includegraphics[width=0.28\columnwidth]{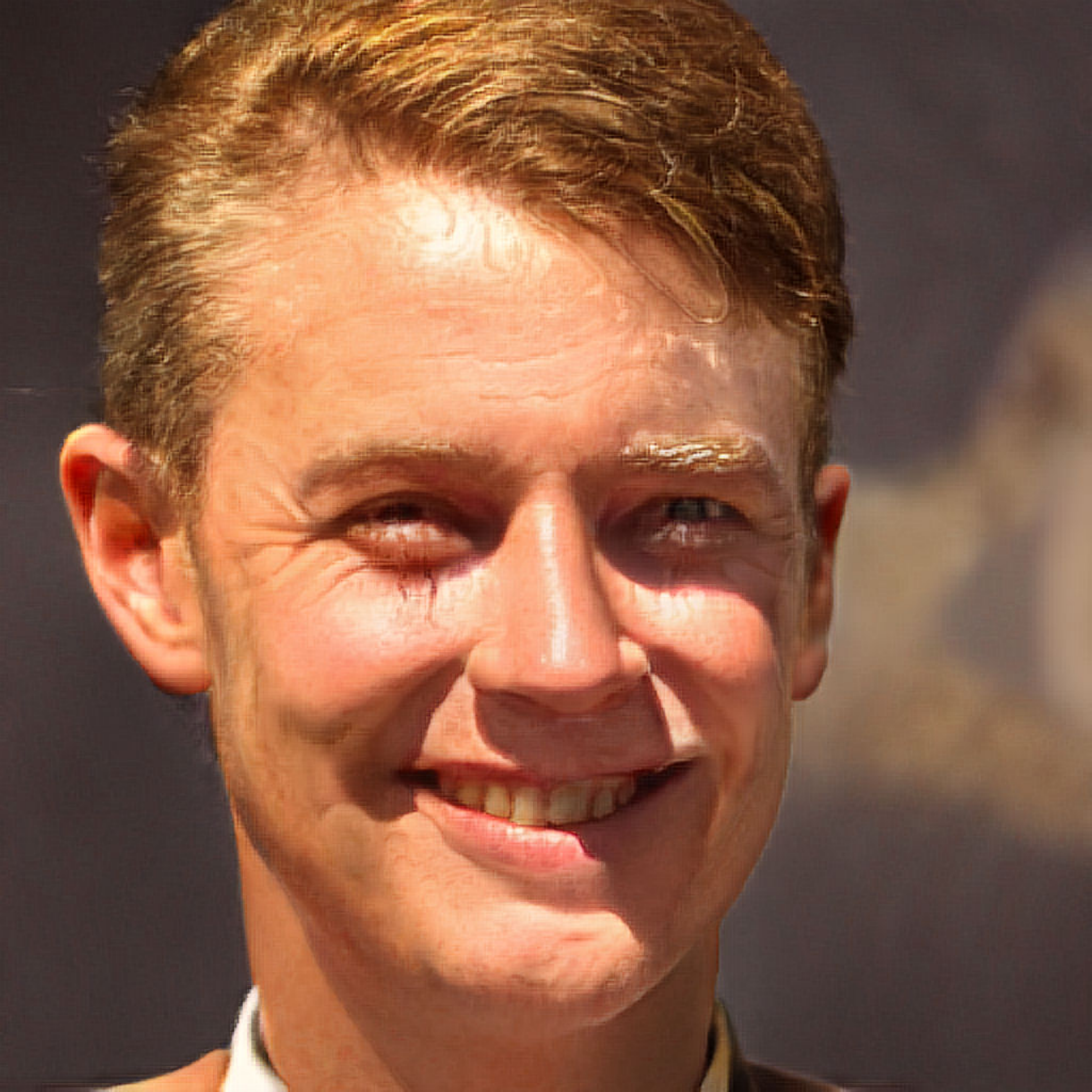} & \includegraphics[width=0.28\columnwidth]{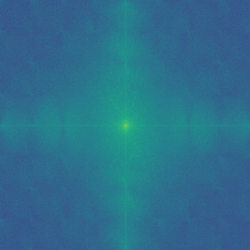} & \includegraphics[width=0.28\columnwidth]{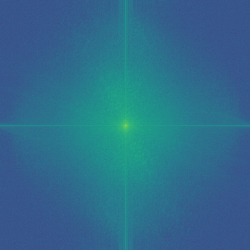} \\
         \includegraphics[width=0.28\columnwidth]{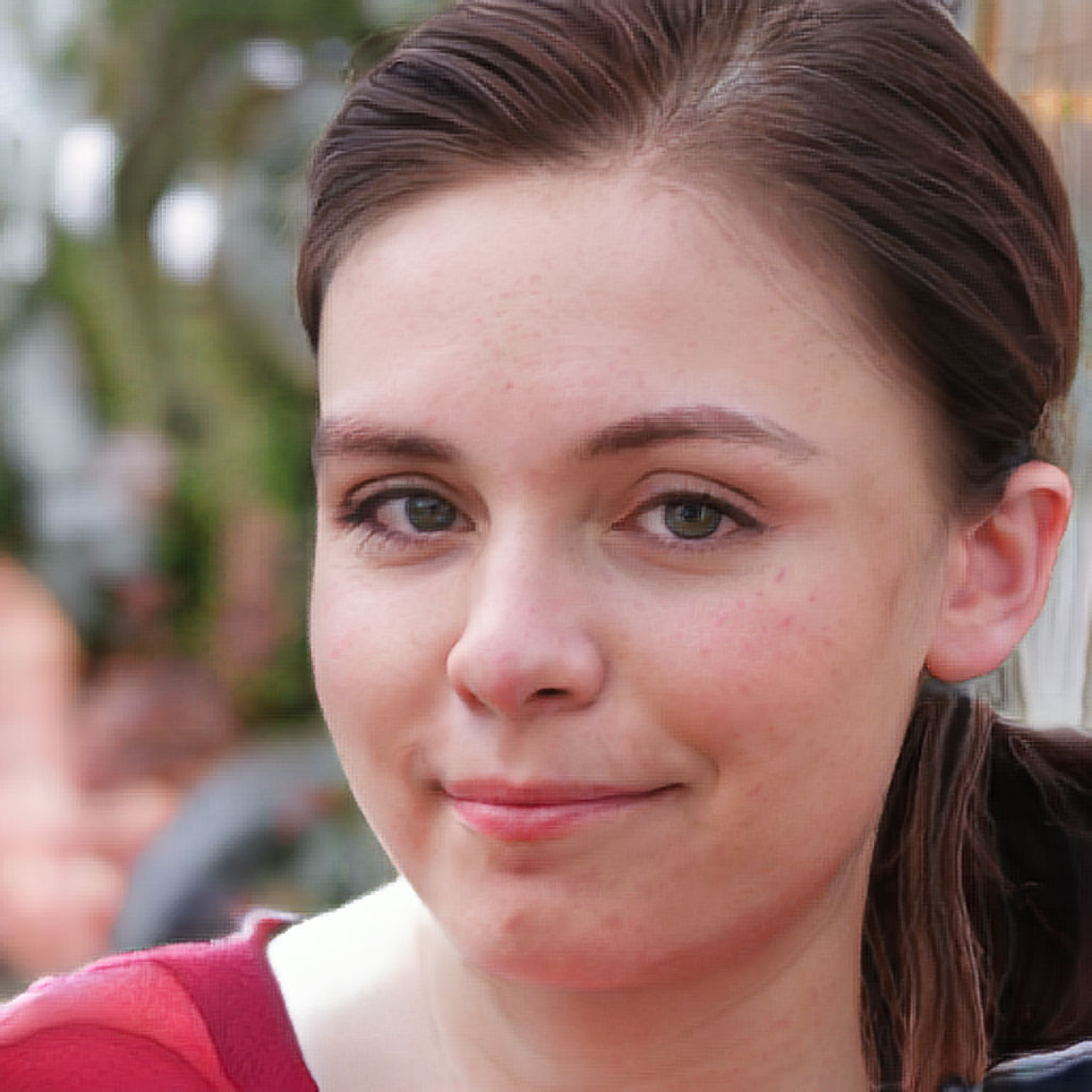} & \includegraphics[width=0.28\columnwidth]{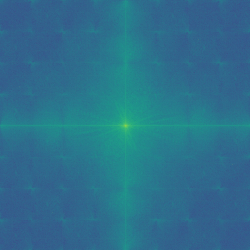} & \includegraphics[width=0.28\columnwidth]{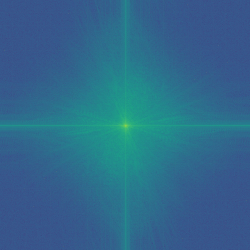} \\
         (a) & (b) & (c)
    \end{tabular}
    }
    \caption{{The Fourier spectrum of blocking artifacts.} (a) Images with blocking artifacts. (b) The artifacts with periodic patterns can be clearly distinguished in the spectrum. (c) The spectrum of artifact-free images derived from the sliding window inference.}
    \label{fig:spectrum}
\end{figure}

\begin{figure}[tb]
    \centering
    \vspace{-1.2em}
    \includegraphics[width=\columnwidth]{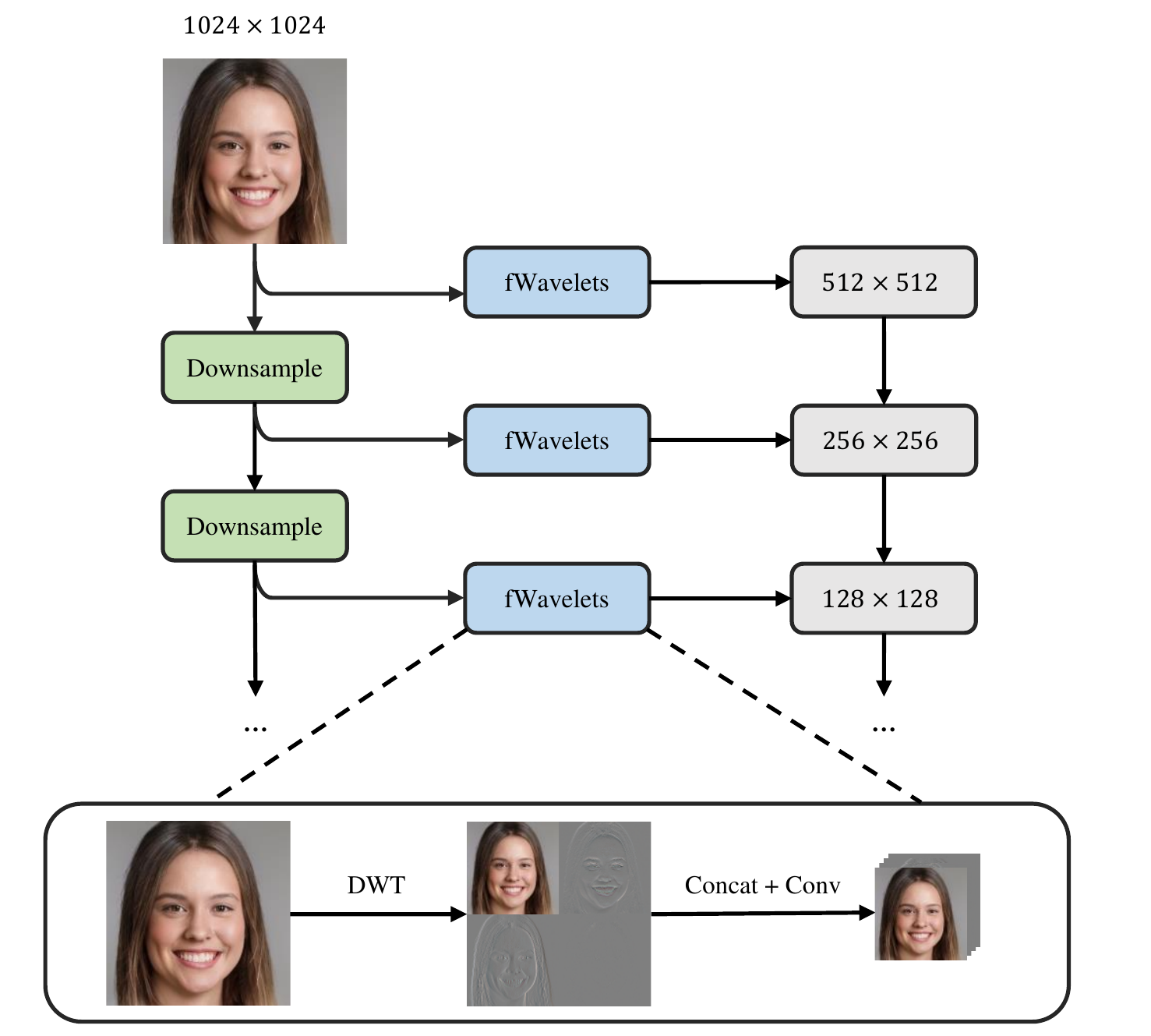}
    \vspace{-1.2em}
    \caption{{The wavelet discriminator suppresses the artifacts by examining the wavelet spectrum of the multi-scaled input.} }
    \label{fig:wavalet_d}
\end{figure}
\vspace{0.5em}
\begin{itemize}[leftmargin=*]
    \itemsep=-0.9mm
    \item \emph{Patch discriminator}~\cite{pix2pix2017} possesses limited receptive field and can be employed to specifically penalize the local structures. Experiments show partial suppression of the blocking artifacts using a patch discriminator.
    \item \emph{Total variation annealing}. To advocate smooth outputs, we apply a large total variation loss at the beginning of training, aiming to suppress the network's tendency to artifacts. The loss weight is then linearly decayed to zero towards the end of training. Though artifacts can be completely removed, such handcrafted constraint favors over-smoothed results and inevitably affects the distribution matching for high-frequency details. 
    \item \emph{Wavelet discriminator}. As shown in Figure~\ref{fig:spectrum}, the periodic artifact pattern can be easily distinguished in the spectral domain. Inspired by this, we resort to a wavelet discriminator~\cite{gal2021swagan} to complement our spatial discriminator and we illustrate its architecture in Figure~\ref{fig:wavalet_d}. The discriminator hierarchically downsamples the input image and on each scale examines the frequency discrepancy relative to real images after discrete wavelet decomposition. Such a wavelet discriminator works remarkably well in combating the blocking artifacts. Meanwhile, it does not bring any side-effects on distribution matching, effectively guiding the generator to produce rich details.   
\end{itemize}

\begin{table}[tb]
    \small
    \centering
    \begin{tabular}{l|c|c}
        \toprule
         Solutions & FID $\downarrow$ & Remove artifacts?\\
         \midrule
         % one of
         Window-based attention &
          8.39 & \xmark\\
          \midrule
         Sliding window inference & 12.08 & \cmark\\
          Token sharing & 8.95  & \xmark\\   
          MLPs after $64\times 64$ & 12.69 & \cmark\\
          Patch discriminator & 7.73 & \xmark \\
          Total variation annealing & 12.79 & \cmark \\
          Wavelet discriminator & \textbf{5.07} & \cmark\\
         \bottomrule
    \end{tabular}
    \caption{Comparison of the artifact suppression solutions on FFHQ-$1024$.}
    \label{tab:ablation_artifact}
\vspace{-1em}
\end{table}

Table~\ref{tab:ablation_artifact} compares the above artifact suppression methods, showing that there are four approaches that could totally remove the visual artifacts. However, sliding window inference suffers from the train-test gap, whereas MLPs fail to synthesize fine details on high-resolution stages, both of them leading to a higher FID score. On the other hand, the total variation with annealing still deteriorates the FID. In comparison, the wavelet-discriminator achieves the lowest FID score and yields the most visually pleasing results.

\section{Experiments}
\label{sec:exp}

\subsection{Experiment setup}
\noindent\textbf{Datasets.} We validate our StyleSwin on the following datasets: CelebA-HQ\cite{karras2017progressive}, LSUN Church\cite{yu15lsun}, and FFHQ\cite{karras2019stylebased}. CelebA-HQ is a high-quality version of CelebA dataset \cite{liu2015faceattributes} which contains 30,000 human face images of $1024\times 1024$ resolution. FFHQ \cite{karras2019stylebased} is a commonly used dataset for high-resolution image generation. It contains 70,000 high-quality human face images with more variation of  age, ethnicity and background, and has better coverage of accessories such as eyeglasses, sunglasses, hats, etc. We synthesize images on FFHQ and CelebA-HQ on either $256\times 256$ and $1024\times 1024$ resolutions. LSUN Church\cite{yu15lsun} contains around 126,000 church images in diverse architecture styles, on which we conduct experiments with $256\times 256$ resolution. 

\vspace{0.5em}
\noindent\textbf{Evaluation protocol.} We adopt Fréchet Inception Distance (FID) \cite{heusel2018gans} as the quantitative metric, which measures the distribution discrepancy between generated images and real ones. Lower FID scores indicate better generation quality. For FFHQ\cite{karras2019stylebased} and LSUN Church\cite{yu15lsun} datasets, we randomly sample 50,000 images from the original datasets as validation sets and calculate FID between the validation sets and 50,000 generated images. For CelebA-HQ\cite{karras2017progressive}, we calculated the FID between 30,000 generated images and all the training samples.

\subsection{Implementation details}

During training we use Adam solver~\cite{kingma2017adam} with $\beta_1=0.0$, $\beta_2=0.99$. Following TTUR~\cite{heusel2018gans}, we set imbalanced learning rates, $5e{-5}$ and $2e{-4}$, for the generator and discriminator respectively. We train our model using the standard non-saturating GAN loss with $R_1$ gradient penalty\cite{Karras2020AnalyzingAI} and stabilize the adversarial training by applying spectral normalization\cite{Miyato2018SpectralNF} on the discriminator. By default, we report all the results with the wavelet discriminator as shown in Figure~
\ref{fig:wavalet_d}. Using 8 32GB V100 GPUs, we are able to fit 32 images as one training batch for the training on $256\times 256$ resolution and the batch size reduces to $16$ on $1024\times 1024$ resolution. For fair comparison with prior works, we report the FID with balanced consistency regularization (bCR) \cite{zhao2020improved} on the FFHQ-$256$ and CelebA-HQ $256$ datasets with the loss weight $\lambda_{\text{real}} = \lambda_{\text{fake}} = 10$. Similar to~\cite{Zhao2021ImprovedTF}, we do not observe performance gain using bCR on higher resolutions.  
Note that we do not adopt complex training strategies, such as path length and mixing regularizations~\cite{karras2019stylebased}, as we wish to conduct studies on neat network architectures.

\subsection{Main results}

\begin{figure*}[t]
    \center
    \small
    \setlength\tabcolsep{1pt}
    \renewcommand{\arraystretch}{0.6}
    {
    \begin{tabular}{@{}cc@{}}
         \includegraphics[width=0.496\textwidth]{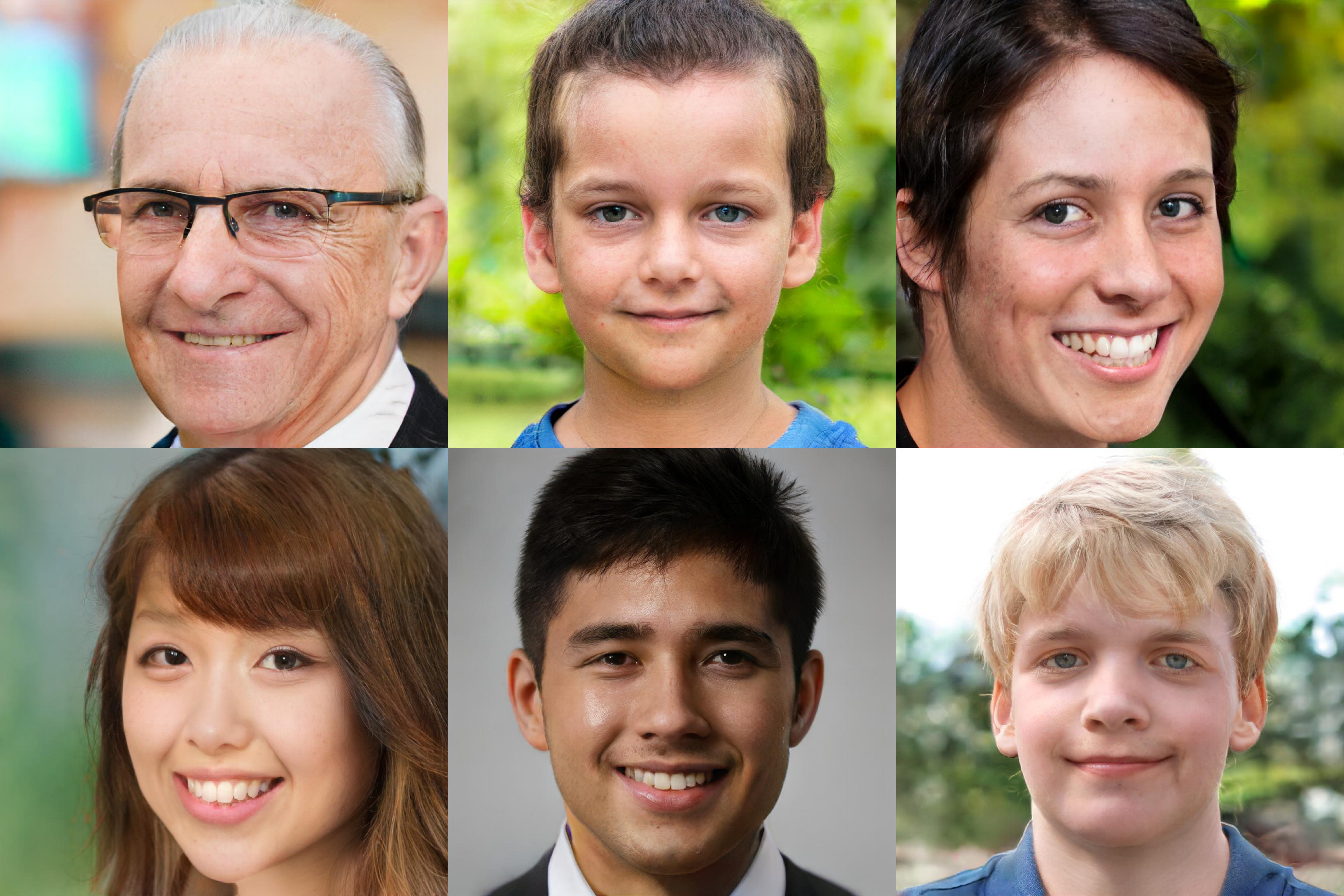} & \includegraphics[width=0.496\textwidth]{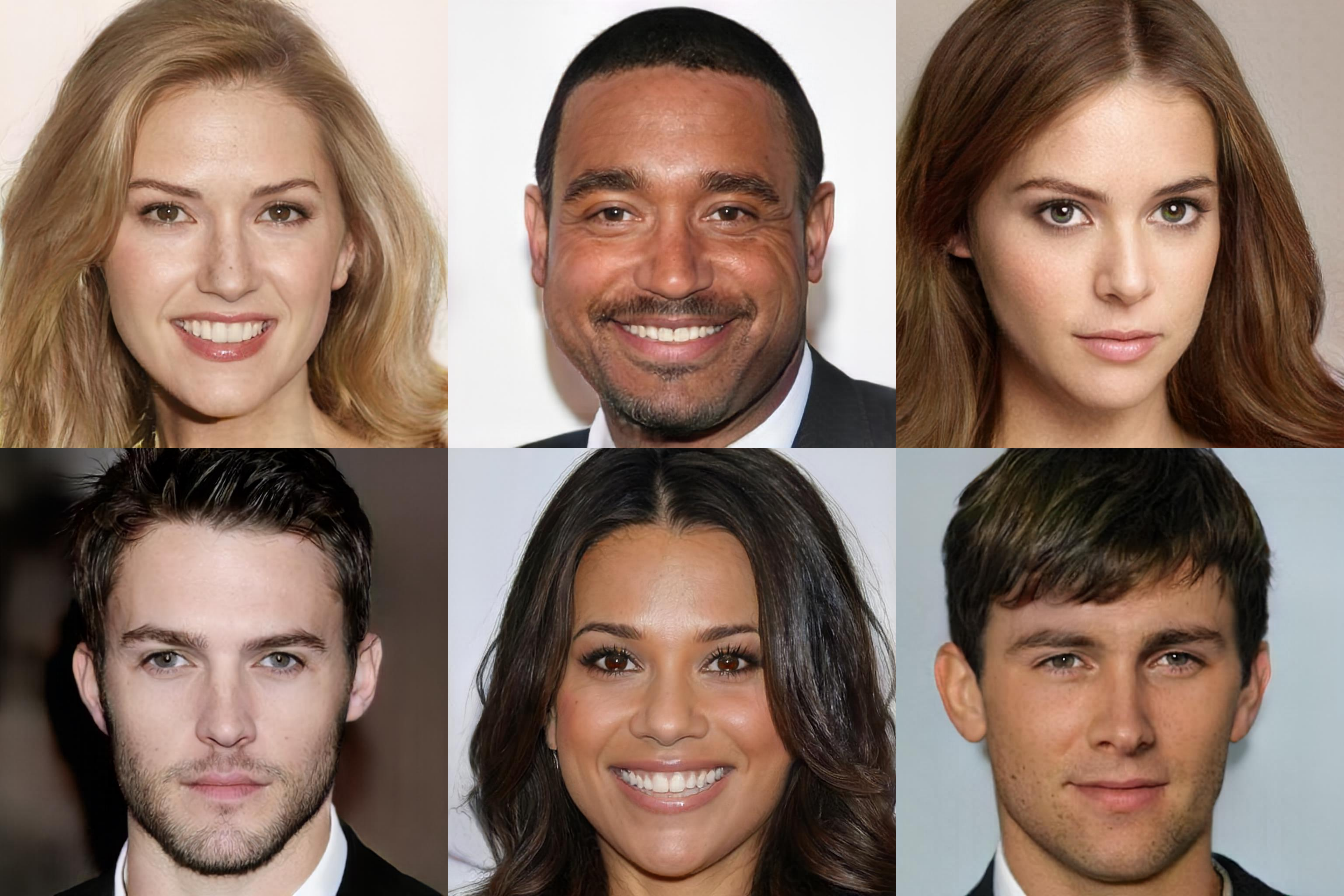} \\
         (a) & (b)
    \end{tabular}
    }
    \caption{Image samples generated by our StyleSwin on (a) FFHQ $1024\times 1024$ and (b) CelebA-HQ $1024\times 1024$.}
    \label{fig:res1}
\end{figure*}

\begin{table}[t]
    \small
    \centering
    \begin{tabular}{c|c|c|c}
        \toprule
        Methods & FFHQ & CelebA-HQ & LSUN Church \\
        \midrule
        StyleGAN2\cite{Karras2020AnalyzingAI} & 3.62$^\ast$ & - & 3.86\\
        PG-GAN\cite{karras2017progressive} & - & 8.03 & 6.42 \\
        U-Net GAN~\cite{schonfeld2020u} & 7.63 & - & -\\
        INR-GAN\cite{skorokhodov2021adversarial} & 9.57 & - & 5.09 \\
        MSG-GAN\cite{karnewar2019msg} & - & - & 5.20 \\
        CIPS\cite{anokhin2020image} & 4.38 & - & \textbf{2.92}\\
        \hline        
        TransGAN\cite{jiang2021transgan} & - & 9.60$^\ast$ & 8.94 \\
        VQGAN\cite{esser2020taming} & 11.40 & 10.70 & - \\
        HiT-B\cite{Zhao2021ImprovedTF} & 2.95$^\ast$ & 3.39$^\ast$ & -\\
        \cellcolor{gray}\emph{StyleSwin} &\cellcolor{gray}\textbf{2.81}$^\ast$  & \cellcolor{gray}\textbf{3.25}$^\ast$ & \cellcolor{gray} 2.95 \\
        \bottomrule
    \end{tabular}
    \caption{Comparison of state-of-the-art unconditional image generation methods on FFHQ, CelebA-HQ and LSUN Church of $256\times 256$ resolution in terms of FID score (lower is better). The subscript ($\ast$) indicates that bCR is applied during training.}
    \label{tab:256}
\end{table}

\begin{table}[tb]
    \small
    \centering
    \begin{tabular}{c|c|c}
        \toprule
         Methods & FFHQ-$1024$ & CelebA-HQ $1024$\\
         \midrule
         StyleGAN$^1$\cite{Karras2020AnalyzingAI}\cite{karras2019stylebased} & \textbf{4.41} & 5.06 \\
         COCO-GAN & - & 9.49 \\
         PG-GAN\cite{karras2017progressive} & - & 7.30 \\
         MSG-GAN\cite{karnewar2019msg} & 5.80 & 6.37\\
         INR-GAN\cite{skorokhodov2021adversarial} & 16.32 & - \\
         CIPS\cite{anokhin2020image} & 10.07 & -\\
         HiT-B\cite{Zhao2021ImprovedTF} & 6.37 & 8.83\\
         \cellcolor{gray}\emph{StyleSwin} & \cellcolor{gray}5.07 & \cellcolor{gray}\textbf{4.43}\\
         \bottomrule
    \end{tabular}
    \caption{Comparison of state-of-the-art unconditional image generation methods on FFHQ and CelebA-HQ of resolution $1024\times 1024$ in terms of FID score (lower is better). $^1$We report the FID score of StyleGAN2 on FFHQ-$1024$ and that of StyleGAN on CelebA-HQ $1024$. For fair comparison, we report results of StyleGAN2 without style-mixing and path regularization.}
    \label{tab:1024}
\vspace{-1em}
\end{table}

\begin{figure}[tb]
    \centering
    \includegraphics[width=\columnwidth]{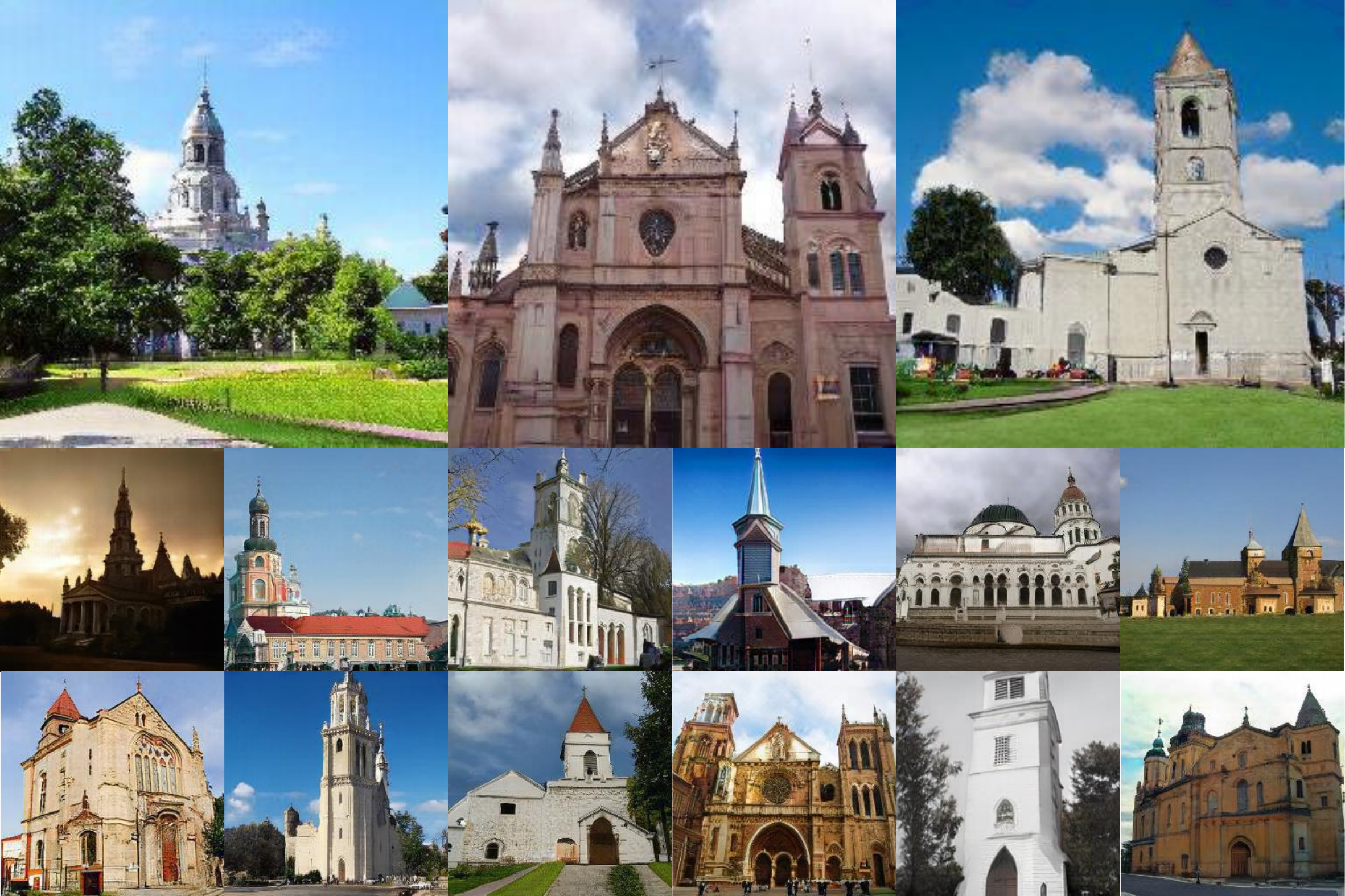}
    \caption{Image samples generated by our StyleSwin on LSUN Church $256\times 256$.}
    \label{fig:res2}
\vspace{-1em}
\end{figure}

\noindent\textbf{Quantitative results.} We compare with state-of-the-art Conv-based GANs as well as the recent transformer-based methods. As shown in Table~\ref{tab:256}, our StyleSwin achieves state-of-the-art FID scores on all the $256\times 256$ synthesis. In particular, on both FFHQ and LSUN Church datasets, StyleSwin outperforms StyleGAN2\cite{Karras2020AnalyzingAI}. Besides the impressive results on resolution $256\times 256$, the proposed StyleSwin shows a strong capability on high-resolution image generation. As shown in Table \ref{tab:1024}, we evaluate models on FFHQ and CelebA-HQ on the resolution of $1024\times 1024$, where the proposed StyleSwin also demonstrates state-of-the-art performance. Notably, we obtain the record FID score of 4.43 on CelebA-HQ $1024$ dataset while considerably closing the gap with the leading approach StyleGAN2 without involving complex training strategies or additional regularization. Also, StyleSwin outperforms the transformer-based approach HiT by a large margin on $1024\times 1024$ resolution, proving that the  self-attention on high-resolution stages is beneficial to high-fidelity detail synthesis.

\vspace{0.5em}
\noindent\textbf{Qualitative results.} Figure~\ref{fig:res1} shows the image samples generated by StyleSwin on FFHQ and CelebA-HQ of $1024\times 1024$ resolution. Our StyleSwin shows compelling quality on synthesizing diverse images of different ages, backgrounds and viewpoints on the resolution of $1024\times 1024$. On top of face modeling, we show generation results of LSUN Church in Figure~\ref{fig:res2}, showing StyleSwin is capable to model complex scene structures. Both the coherency of global geometry and the high-fidelity details all prove the advantages of using transformers among all the resolutions.

\vspace{0.5em}
\noindent\textbf{Ablation study.}
\begin{table}[tb]
    \small
    \centering
    \vspace{0.2em}
    \begin{tabular}{l|c}
        \toprule
         Model Configuration & FID $\downarrow$\\
         \midrule
         A. Swin baseline & 15.03 \\
         B. + Style injection & 8.40 \\
         C. + Double attention & 7.86 \\
         D. + Wavelet discriminator & 6.34 \\
         E. + SPE & 5.76 \\
         F. + Larger model & 5.50 \\
         G. + bCR &\textbf{2.81} \\
         \bottomrule
    \end{tabular}
    \caption{{Ablation study conducted on FFHQ-$256$.} Starting from the baseline architecture, we prove the effectiveness of each proposed component. }
    \label{tab:ablation_256}
    \vspace{-1em}
\end{table}
To validate the effectiveness of the proposed components, we conduct ablation studies in Table \ref{tab:ablation_256}. Compared with the baseline architecture, we observe significant FID improvement thanks to the enhanced model capacity brought by the style injection. The double attention makes each layer leverage larger context at one time and further reduces the FID score. Wavelet discriminator brings a large FID improvement because it effectively suppresses the blocking artifacts and meanwhile brings stronger supervision for high-frequencies. In our experiment, we observe faster adversarial training when adopting the wavelet discriminator. Further, introducing sinusoidal positional encoding (SPE) on each generation scale effectively reduces the FID. Employing a larger model brings slight improvement and it seems that the model capacity of the current transformer structure is not the bottleneck. From Table~\ref{tab:ablation_256} we see that bCR considerably improves the FID by $2.69$, which coincides with the recent findings~\cite{Zhao2021ImprovedTF,Lee2021ViTGANTG,jiang2021transgan} that data augmentation is still vital in transformer-based GAN since transformers are data-hungry and prone to overfitting. However, we do not observe its effectiveness on higher resolutions, \eg, $1024\times 1024$, and we leave  regularization schemes for high-resolution synthesis to future work.

\vspace{0.5em}
\noindent\textbf{Parameters and Throughput.} In Table~\ref{tab:params}, We compare the number of model parameters and FLOPs with StyleGAN2 for $1024\times 1024$ synthesis. Although our approach has a larger model size, it achieves lower FLOPs than StyleGAN2, which means the method achieves competitive generation quality with less theoretical computational cost.

\begin{table}[tb]
    \small
    \centering
    \begin{tabular}{c|c c}
        \toprule
         Methods & \makecell{\#params} & \makecell[c]{FLOPs} \\
         \midrule
         StyleGAN2~\cite{Karras2020AnalyzingAI} & 30.37M & 74.27B \\
         \cellcolor{gray}\emph{StyleSwin} & \cellcolor{gray}40.86M & \cellcolor{gray} 50.90B\\
         \bottomrule
    \end{tabular}
    \caption{Comparison of the network parameters and FLOPs with StyleGAN2.}
    \label{tab:params}
\vspace{-1em}
\end{table}

\section{Conclusion}
We propose StyleSwin, a transformer-based GAN for high-resolution image generation. The use of local attention is efficient to compute while attaining most modeling capability since the receptive field is largely compensated by double attention. Besides, we find one key feature is missing in transformer-based GANs --- the generator is not aware of the position for patches under synthesis, so we introduce SPE for global positioning. Thanks to the increased expressivity, the proposed StyleSwin consistently outperforms the leading Conv-based approaches on $256\times 256$ datasets. To solve the blocking artifacts on high-resolution synthesis, we propose to penalize the spectral discrepancy with a wavelet discriminator~\cite{gal2021swagan}. Ultimately, the proposed StyleSwin offers compelling quality on the resolution of $1024\times 1024$, which for the first time, approaches the best performed ConvNets. Our work hopefully incentives more studies on utilizing the capable transformers in generative modeling.

%%%%%%%%% REFERENCES
% \clearpage
{\small
\bibliographystyle{ieee_fullname}
\bibliography{egbib}
}

\appendix
\onecolumn{
\begin{appendices}
\section{Implementation Details}

We train the StyleSwin using the standard non-saturating logistic GAN loss~\cite{Goodfellow2014GenerativeAN} with $R_1$ gradient penalty~\cite{Mescheder2018WhichTM}. Specifically, the discriminator is trained to measure the realism of image samples whereas the generator is trained to generate samples that the discriminator mistakenly recognizes as real ones. In addition, the $R_1$ regularization term penalizes the gradient on real data to advocate the local stability. The training loss can be formulated as:
$$\begin{aligned}
\mathcal{L}_{D} &=-\mathbb{E}_{x \sim P_{x}}[\log (D(x))]-\mathbb{E}_{z \sim P_{z}}[\log (1-D(G(z)))]+\gamma \cdot \mathbb{E}_{x \sim P_{x}}[\left\|\nabla_{x} D(x)\right\|_{2}^{2}], \\
\mathcal{L}_{G} &=-\mathbb{E}_{z \sim P_{z}}[\log (D(G(z)))].
\end{aligned}$$
In practice, we perform $R_1$ gradient penalty every $16$ iterations and the corresponding weight $\gamma$ varies for different datasets.

The training follows the TTUR strategy~\cite{heusel2018gans} in which the discriminator adopts a $4\times$ larger learning rate than the generator. We linearly decay the learning rate to $0$ from the LR decay start iteration for training all datasets except CelebA-HQ 1024. We apply spectral normalization~\cite{Miyato2018SpectralNF} upon discriminator to ensure its Lipschitz continuity. The transformers are initialized with a truncated normal distribution~\cite{hanin2018start} with zero mean and standard deviation of $0.02$. For the convolution $1\times 1$ used in tRGB layers, we use Glorot initialization\cite{glorot2010understanding} with a gain of $0.02$. We use an exponential moving average of weights of generator~~\cite{karras2017progressive} when sampling image, with a decay rate of $0.9978$ following~\cite{karras2019stylebased}.
% \noindent\textbf{Network capacity.} We set the channels of feature maps to $\min (2^{15} / \sqrt{H\times W}, 512)$ when operating on resolution of $H\times W$ on datasets of resolution $256\times 256$. Considering the computational cost when applying StyleSwin on high-resolution image generation, we set the channels of feature maps to $\min (2^{14} / \sqrt{H\times W}, 512)$ when training datasets of resolution $1024\times 1024$.
% \noindent\textbf{Initialization.} We use a truncated normal distribution with zero mean and standard deviation of $0.02$ to initialize all of the weights in the generator backbone. For the convolution $1\times 1$ used in tRGB layers, we use Glorot initialization\cite{glorot2010understanding} with gain of $0.02$. 

When synthesizing $256\times 256$ resolution images of FFHQ and CelebA-HQ, the training benefits from balanced consistency regularization (bCR)~\cite{zhao2020improved}. Specifically, images are augmented by \{\textit{Flipping}, \textit{Color}, \textit{Translation}, \textit{Cutout}\} of probability \{$0.5$, 
$1.0$, $1.0$, $1.0$\} as in DiffAug~\cite{zhao2020differentiable}. \textit{Translation} is performed within [$-1/8$, $1/8$] of the image size, and random squares of half image size are masked when applying \textit{Cutout}.

We implement the StyleSwin using Pytorch and conduct experiments with Tesla V100 GPUs. Training on $1024\times 1024$ resolution takes about $14$ days using 8 32GB GPUs. The hyper-parameters in the experiments are summarized in Table~\ref{tab:hyperparam}. 

\begin{table*}[h]
    \small
    \centering
    \begin{tabular}{l|c c c c c}
        \toprule
          & FFHQ-256 & CelebA-HQ 256 & LSUN Church 256 & FFHQ-1024 & CelebA-HQ 1024\\
         \midrule
         Training iteration & 32.0M & 25.6M & 48M & 25.6M & 25.6M\\
         Number of GPUs & 8 & 8 & 8 & 16 & 16 \\
         Batch size & 32 & 32 & 32 & 32 & 32 \\
         Learning rate of D & $2{e-4}$ & $2{e-4}$ & $2{e-4}$ & $2{e-4}$ & $2{e-4}$\\
         Learning rate of G & $5{e-5}$ & $5{e-5}$ & $5{e-5}$ & $5{e-5}$ & $5{e-5}$ \\
         LR decay start iteration & 24.8M & 16M & 41.6M & 19.2M & - \\
         $R_1$ regularization $\gamma$ & 10 & 5 & 5 & 10 & 10 \\
         bCR & \cmark & \cmark & \xmark & \xmark & \xmark \\
         \bottomrule
    \end{tabular}
    \caption{Experiment settings for different datasets.}
    \label{tab:hyperparam}
\end{table*}

\section{Detailed Architecture}
StyleSwin starts from a constant input of size $4\times 4 \times 512$ and hierarchically upsamples the feature map with transformer blocks. We use two transformer blocks to model each resolution scale. The detailed model architecture is shown in Table~\ref{tab:arch_detail}. ``Double attn, 512-d, 4-w, 16-h'' indicates a double attention block with a channel dimension of $512$, window size of $4$, and $16$ attention heads. ``Bilinear upsampling, 512-d'' indicates a bilinear upsampling layer followed by feedforward MLPs with an output dimension of $512$. 

% We set the channels of feature maps to $\min (2^{15} / \sqrt{H\times W}, 512)$ when operating on resolution of $H\times W$ on datasets of resolution $256\times 256$. Considering the computational cost when applying StyleSwin on high-resolution image generation, we set the channels of feature maps to $\min (2^{14} / \sqrt{H\times W}, 512)$ when training datasets of resolution $1024\times 1024$.

\begin{table}[h]
    \small
    \centering
    \begin{tabular}{c | c | c}
    \toprule
        Input size & StyleSwin-256 & StyleSwin-1024 \\
    \midrule
        \multirow{2}*{4$\times$4} & \makecell[c]{$\left\{
                        \begin{array}{c}
                        % \text{AdaIN, 512-d} \\
                        \text{Double attn, 512-d, 4-w, 16-h} \\
                        % \text{AdaIN, 512-d} \\
                        \text{MLP, 512-d}
                        \end{array}\right\} \times 2$} & \makecell[c]{$\left\{
                        \begin{array}{c}
                        % \text{AdaIN, 512-d} \\
                        \text{Double attn, 512-d, 4-w, 16-h} \\
                        % \text{AdaIN, 512-d} \\
                        \text{MLP, 512-d}
                        \end{array}\right\} \times 2$}\\
        \cmidrule{2-3}
        ~ & Bilinear upsampling, 512-d & Bilinear upsampling, 512-d\\
    \midrule
        \multirow{2}*{8$\times$8} & \makecell[c]{$\left\{
                        \begin{array}{c}
                        % \text{AdaIN, 512-d} \\
                        \text{Double attn, 512-d, 8-w, 16-h} \\
                        % \text{AdaIN, 512-d} \\
                        \text{MLP, 512-d}
                        \end{array}\right\} \times 2$} & \makecell[c]{$\left\{
                        \begin{array}{c}
                        % \text{AdaIN, 512-d} \\
                        \text{Double attn, 512-d, 8-w, 16-h} \\
                        % \text{AdaIN, 512-d} \\
                        \text{MLP, 512-d}
                        \end{array}\right\} \times 2$}\\
        \cmidrule{2-3}
        ~ & Bilinear upsampling, 512-d & Bilinear upsampling, 512-d\\
    \midrule
        \multirow{2}*{16$\times$16} & \makecell[c]{$\left\{
                        \begin{array}{c}
                        % \text{AdaIN, 512-d} \\
                        \text{Double attn, 512-d, 8-w, 16-h} \\
                        % \text{AdaIN, 512-d} \\
                        \text{MLP, 512-d}
                        \end{array}\right\} \times 2$} & \makecell[c]{$\left\{
                        \begin{array}{c}
                        % \text{AdaIN, 512-d} \\
                        \text{Double attn, 512-d, 8-w, 16-h} \\
                        % \text{AdaIN, 512-d} \\
                        \text{MLP, 512-d}
                        \end{array}\right\} \times 2$}\\
        \cmidrule{2-3}
        ~ & Bilinear upsampling, 512-d & Bilinear upsampling, 512-d\\
    \midrule
        \multirow{2}*{32$\times$32} & \makecell[c]{$\left\{
                        \begin{array}{c}
                        % \text{AdaIN, 512-d} \\
                        \text{Double attn, 512-d, 8-w, 16-h} \\
                        % \text{AdaIN, 512-d} \\
                        \text{MLP, 512-d}
                        \end{array}\right\} \times 2$} & \makecell[c]{$\left\{
                        \begin{array}{c}
                        % \text{AdaIN, 512-d} \\
                        \text{Double attn, 512-d, 8-w, 16-h} \\
                        % \text{AdaIN, 512-d} \\
                        \text{MLP, 512-d}
                        \end{array}\right\} \times 2$}\\
        \cmidrule{2-3}
        ~ & Bilinear upsampling, 512-d & Bilinear upsampling, 256-d\\
    \midrule
        \multirow{2}*{64$\times$64} & \makecell[c]{$\left\{
                        \begin{array}{c}
                        % \text{AdaIN, 512-d} \\
                        \text{Double attn, 512-d, 8-w, 16-h} \\
                        % \text{AdaIN, 512-d} \\
                        \text{MLP, 512-d}
                        \end{array}\right\} \times 2$} & \makecell[c]{$\left\{
                        \begin{array}{c}
                        % \text{AdaIN, 256-d} \\
                        \text{Double attn, 256-d, 8-w, 8-h} \\
                        % \text{AdaIN, 256-d} \\
                        \text{MLP, 256-d}
                        \end{array}\right\} \times 2$}\\
        \cmidrule{2-3}
        ~ & Bilinear upsampling, 256-d & Bilinear upsampling, 128-d\\
    \midrule
        \multirow{2}*{128$\times$128} & \makecell[c]{$\left\{
                        \begin{array}{c}
                        % \text{AdaIN, 256-d} \\
                        \text{Double attn, 256-d, 8-w, 8-h} \\
                        % \text{AdaIN, 256-d} \\
                        \text{MLP, 256-d}
                        \end{array}\right\} \times 2$} & \makecell[c]{$\left\{
                        \begin{array}{c}
                        % \text{AdaIN, 128-d} \\
                        \text{Double attn, 128-d, 8-w, 4-h} \\
                        % \text{AdaIN, 128-d} \\
                        \text{MLP, 128-d}
                        \end{array}\right\} \times 2$}\\
        \cmidrule{2-3}
        ~ & Bilinear upsampling, 128-d & Bilinear upsampling, 64-d\\
    \midrule
        \multirow{2}*{256$\times$256} & \multirow{2}*{\makecell[c]{$\left\{
                        \begin{array}{c}
                        % \text{AdaIN, 128-d} \\
                        \text{Double attn, 128-d, 8-w, 4-h} \\
                        % \text{AdaIN, 128-d} \\
                        \text{MLP, 128-d}
                        \end{array}\right\} \times 2$}} & \makecell[c]{$\left\{
                        \begin{array}{c}
                        % \text{AdaIN, 64-d} \\
                        \text{Double attn, 64-d, 8-w, 4-h} \\
                        % \text{AdaIN, 64-d} \\
                        \text{MLP, 64-d}
                        \end{array}\right\} \times 2$}\\
        \cmidrule{3-3}
        ~ & ~ & Bilinear upsampling, 32-d\\
    \midrule
        \multirow{2}*{512$\times$512} & \multirow{2}*{} & \makecell[c]{$\left\{
                        \begin{array}{c}
                        % \text{AdaIN, 32-d} \\
                        \text{Double attn, 32-d, 8-w, 4-h} \\
                        % \text{AdaIN, 32-d} \\
                        \text{MLP, 32-d}
                        \end{array}\right\} \times 2$}\\
        \cmidrule{3-3}
        ~ & ~ & Bilinear upsampling, 16-d\\
    \midrule
        1024$\times$1024 &  & \makecell[c]{$\left\{
                        \begin{array}{c}
                        % \text{AdaIN, 16-d} \\
                        \text{Double attn, 16-d, 8-w, 4-h} \\
                        % \text{AdaIN, 16-d} \\
                        \text{MLP, 16-d}
                        \end{array}\right\} \times 2$}\\
    \bottomrule
    \end{tabular}
    \caption{The detailed generator architecture of StyleSwin-256 and StyleSwin-1024.}
    \label{tab:arch_detail}
\end{table}

\section{The Modeling Capacity of Double Attention}
In order to prove the improved expressivity of the proposed double attention, we train an autoencoder for image reconstruction. Specifically, we adopt a conv-based encoder --- a ResNet-50 pretrained from MoCo~\cite{he2020momentum} such that both the low-level and high-level information are well preserved in the $16\times 16$ feature map~\cite{zhao2020makes}. The latent feature map is further fed into the decoder for image reconstruction. The decoder adopts transformer blocks, which hierarchically upsamples the latent feature map and reconstructs the input. No style injection module is needed and we replace AdaIN  with layer normalization. The decoder adopts either the vanilla Swin attention block or the proposed double attention. The autoencoders are trained with $\mathcal{L}_1$ loss. Figure~\ref{fig:l1_loss} shows the training loss curve of the two autoencoders. One can see that the decoder with double attention shows faster convergence and yields lower reconstruction loss, indicating that the decoder that leverages enhanced receptive field shows stronger generative capacity.

\begin{figure}[h]
    \centering
    \includegraphics[width=0.5\columnwidth]{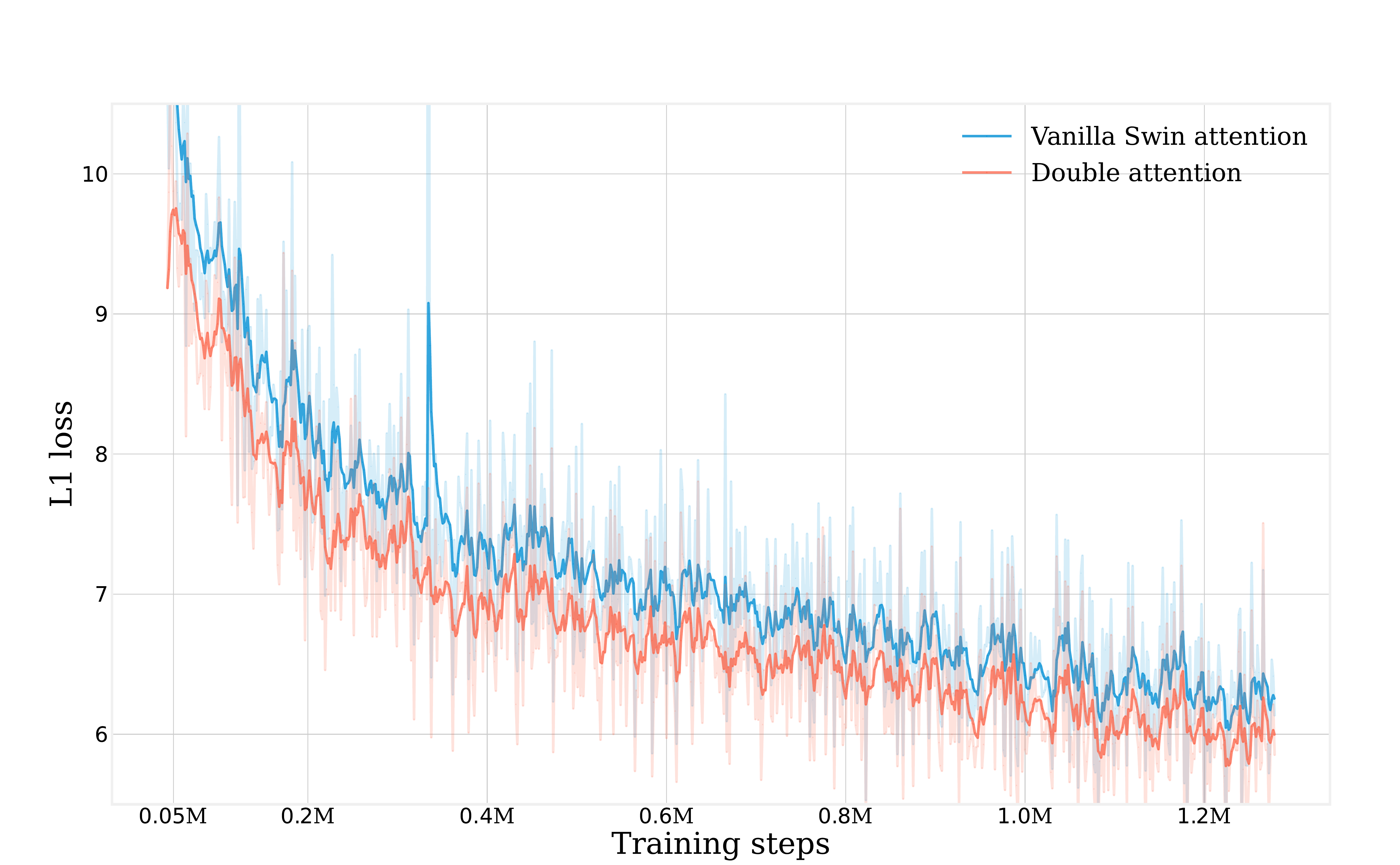}
    \caption{Image reconstruction training loss of autoencoders. The autoencoder adopts a fixed conv-based encoder and transformer-based decoder and is trained with $\mathcal{L}_1$ loss. The decoder with double attention shows improved modeling capacity over the vanilla Swin attention.}
    \label{fig:l1_loss}
\end{figure}

\section{Additional Quantitative Evaluation}

To further demonstrate StyleSwin's strong ability to model complex scenes and materials, we train our model on a subset of LSUN Car, which achieves comparable performance to  state-of-the-art StyleGAN2. We also present additional quantitative evaluation results in terms of KID~\cite{binkowski2018demystifying} and FID-Inf~\cite{chong2020effectively} on all evaluation datasets, comparing to StyleGAN2. The detailed measures are presented in Table~\ref{tab:metrics_256} and Table~\ref{tab:metrics_1024}.

\begin{table}[h]
    \scriptsize
    \centering
    \begin{tabular}{@{}c|c|c|c|c|c|c|c|c|c|c|c|c@{}}
        \toprule
         \multirow{2}*{Methods} & \multicolumn{3}{c|}{FFHQ-256} & \multicolumn{3}{c|}{Church-256} & \multicolumn{3}{c|}{CelebAHQ-256} & \multicolumn{3}{c}{Car-256} \\
         ~ & FID & KID$\times 10^{-3}$ & FID-Inf & FID & KID$\times 10^{-3}$ & FID-Inf & FID & KID$\times 10^{-3}$ & FID-Inf & FID & KID$\times 10^{-3}$ & FID-Inf  \\
          \midrule
          StyleGAN2 & 3.62 & 1.45 & 1.37 & 3.86 & 1.71 & 1.53 & - & - & - & \textbf{4.32} & 1.63 & \textbf{1.60}  \\
          \cellcolor{gray}\emph{StyleSwin}& \cellcolor{gray}\textbf{2.81} & \cellcolor{gray}\textbf{0.54} & \cellcolor{gray}\textbf{0.83} & \cellcolor{gray}\textbf{2.95} & \cellcolor{gray}\textbf{1.02} & \cellcolor{gray}\textbf{1.44} & \cellcolor{gray}\textbf{3.25} & \cellcolor{gray}\textbf{0.61} & \cellcolor{gray}\textbf{1.36} & \cellcolor{gray}4.35 & \cellcolor{gray}\textbf{1.53} & \cellcolor{gray}1.80  \\
         \bottomrule
    \end{tabular}
    \caption{Evaluation results comparing to StyleGAN2 on resolution 256 in terms of FID, KID and FID-Inf.}
    \label{tab:metrics_256}
\end{table}

\begin{table}[h]
    \small
    \centering
    \begin{tabular}{@{}c|c|c|c|c|c|c@{}}
        \toprule
         \multirow{2}*{Methods} & \multicolumn{3}{c|}{FFHQ-1024} &  \multicolumn{3}{c}{CelebAHQ-1024} \\
         ~ & FID & KID$\times 10^{-3}$ & FID-Inf & FID & KID$\times 10^{-3}$ & FID-Inf \\
          \midrule
          StyleGAN2$^1$ & \textbf{4.41} & \textbf{1.22} & \textbf{1.57} & 5.17 & 1.71 & 1.53 \\
          \cellcolor{gray}\emph{StyleSwin} & \cellcolor{gray}5.07 & \cellcolor{gray}2.07 & \cellcolor{gray}2.13 & \cellcolor{gray}\textbf{4.43} & \cellcolor{gray}\textbf{1.42} & \cellcolor{gray}\textbf{2.08} \\
         \bottomrule
    \end{tabular}
    \caption{Evaluation results comparing to StyleGAN2 on resolution 1024 in terms of FID, KID and FID-Inf. $^1$We report the metrics of StyleGAN2 on FFHQ-1024 and that of StyleGAN on CelebA-HQ 1024.}
    \label{tab:metrics_1024}
\end{table}

\section{More Qualitative Results}

\noindent\textbf{Latent code interpolation.} To explore the property of the learned latent space of StyleSwin, we randomly sample two latent codes in the latent space and perform linear interpolation between them. As shown in Figure~\ref{fig:latent_code_interpolation}, our StyleSwin could produce smooth, meaningful image morphing with respect to different styles like gender, poses, and eyeglasses.

\begin{figure*}[h]
    \center
    \small
    \setlength\tabcolsep{0pt}
    \renewcommand{\arraystretch}{0}
    {
    \begin{tabular}{@{}cccccccc@{}}
         \includegraphics[width=0.125\columnwidth]{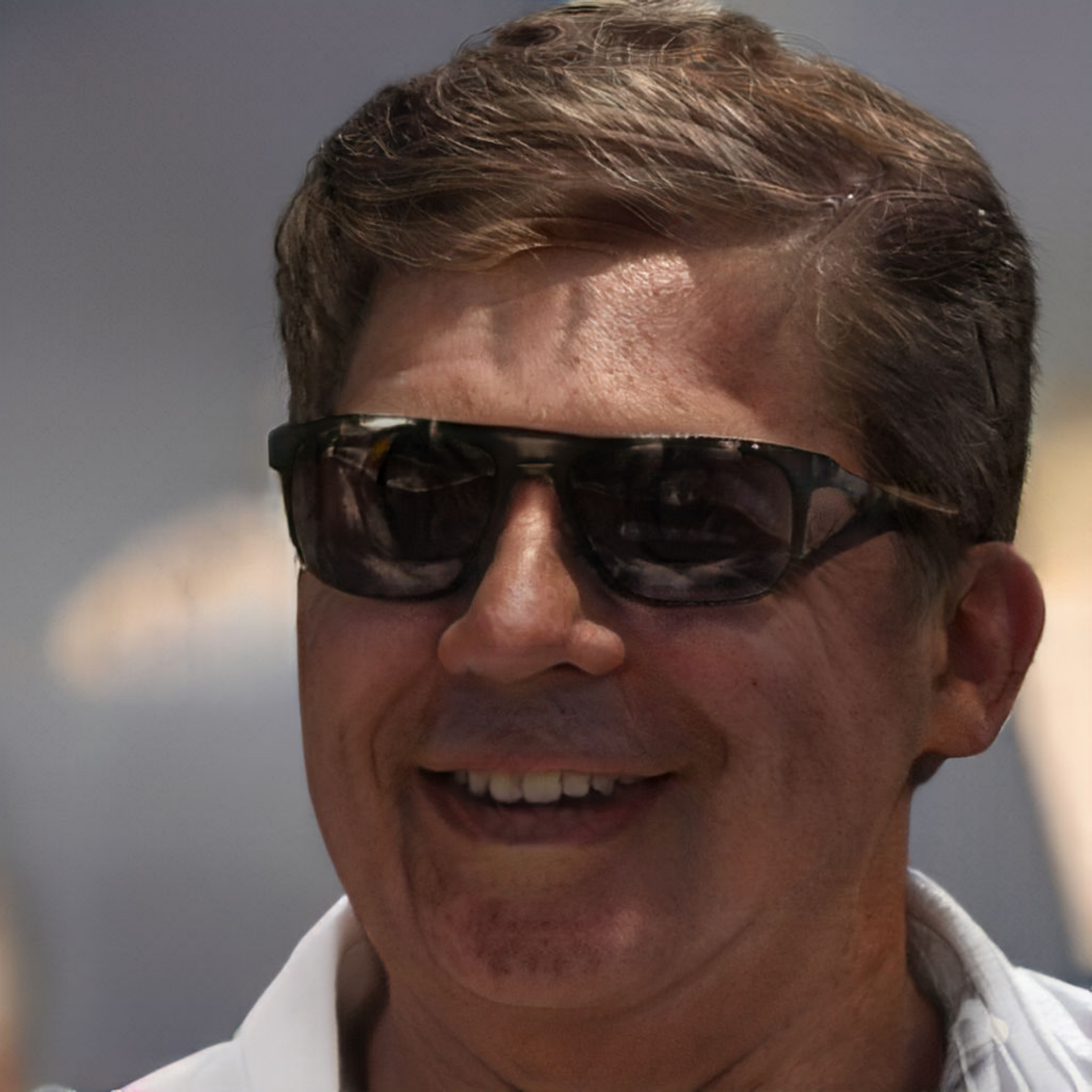} & \includegraphics[width=0.125\columnwidth]{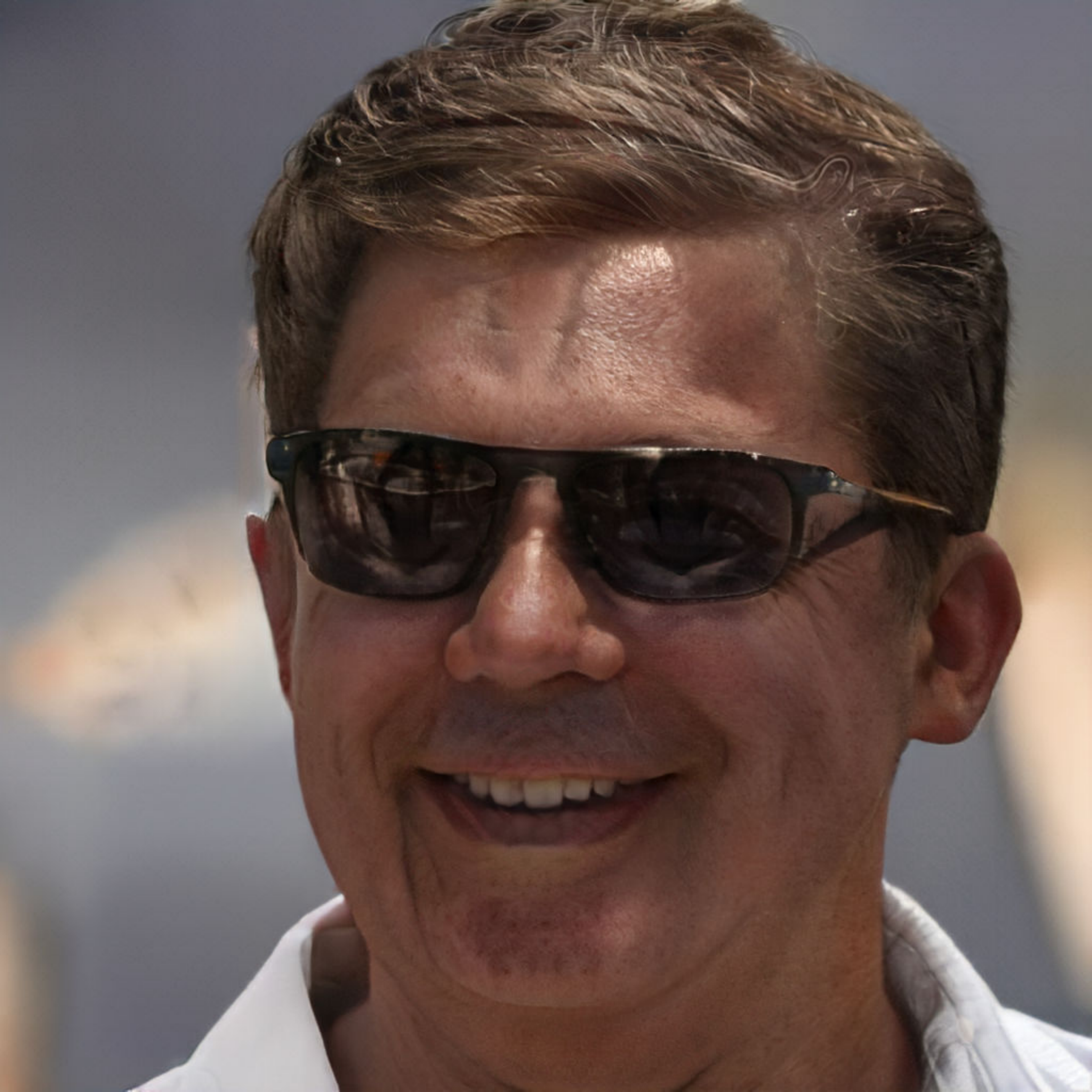} &
         \includegraphics[width=0.125\columnwidth]{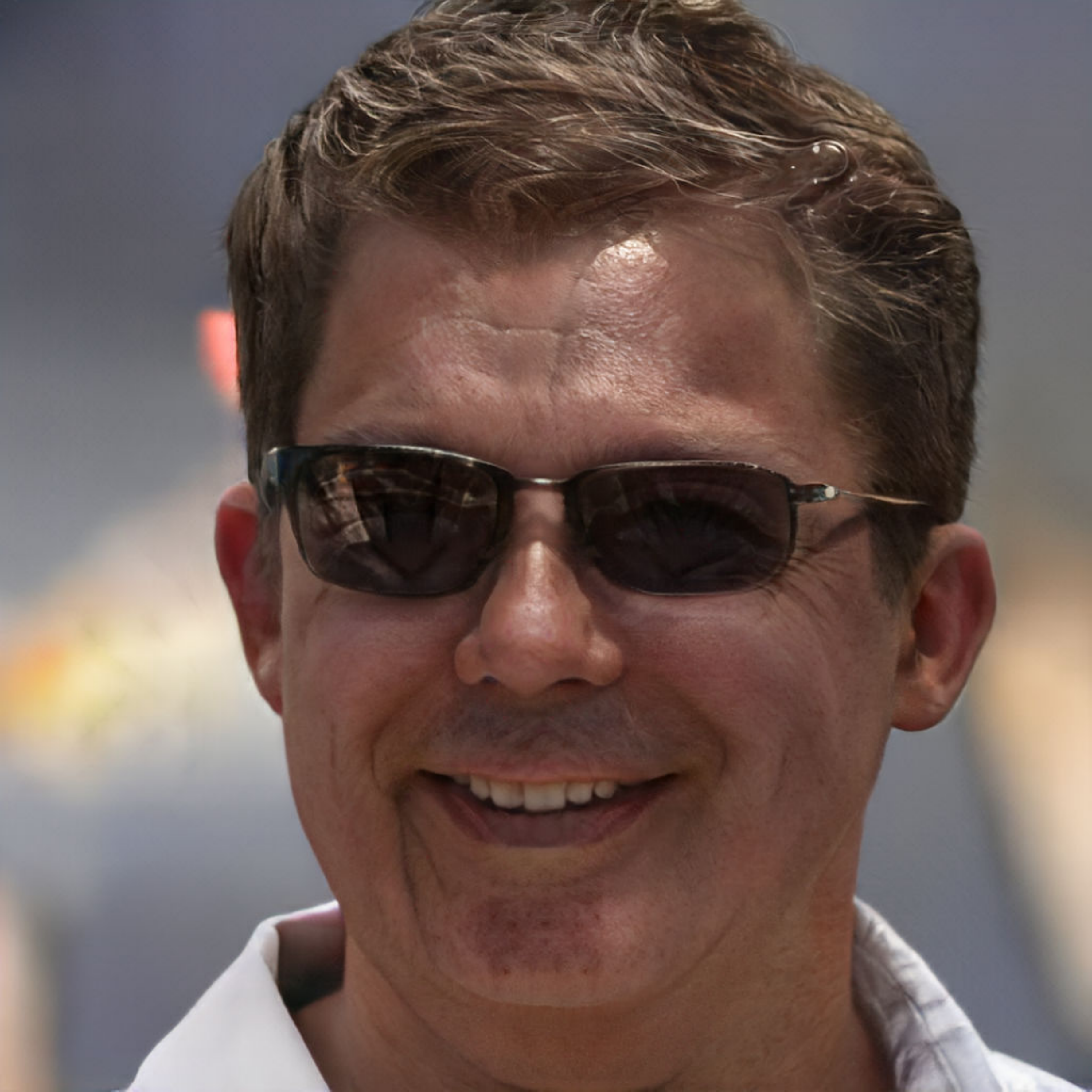} & \includegraphics[width=0.125\columnwidth]{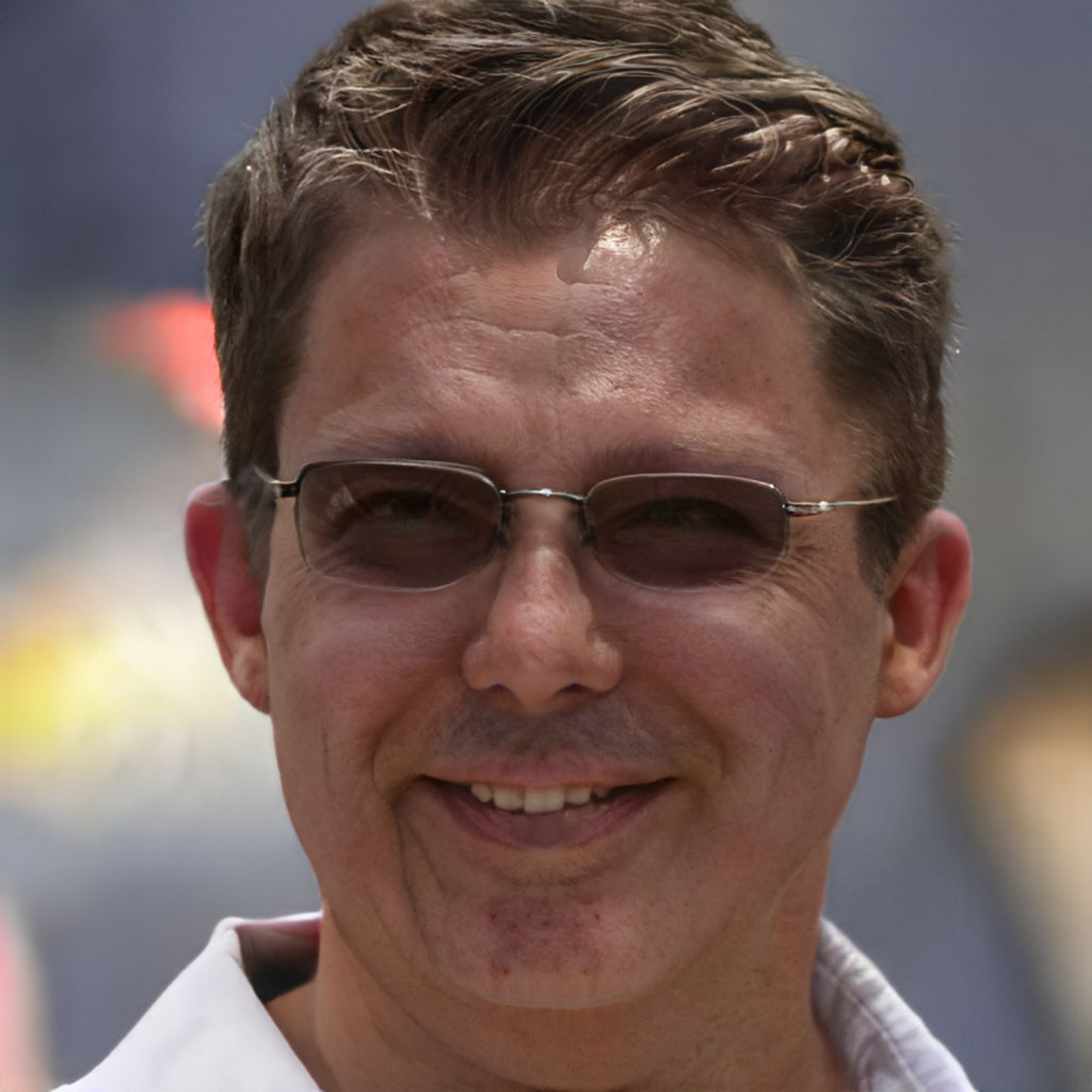} & \includegraphics[width=0.125\columnwidth]{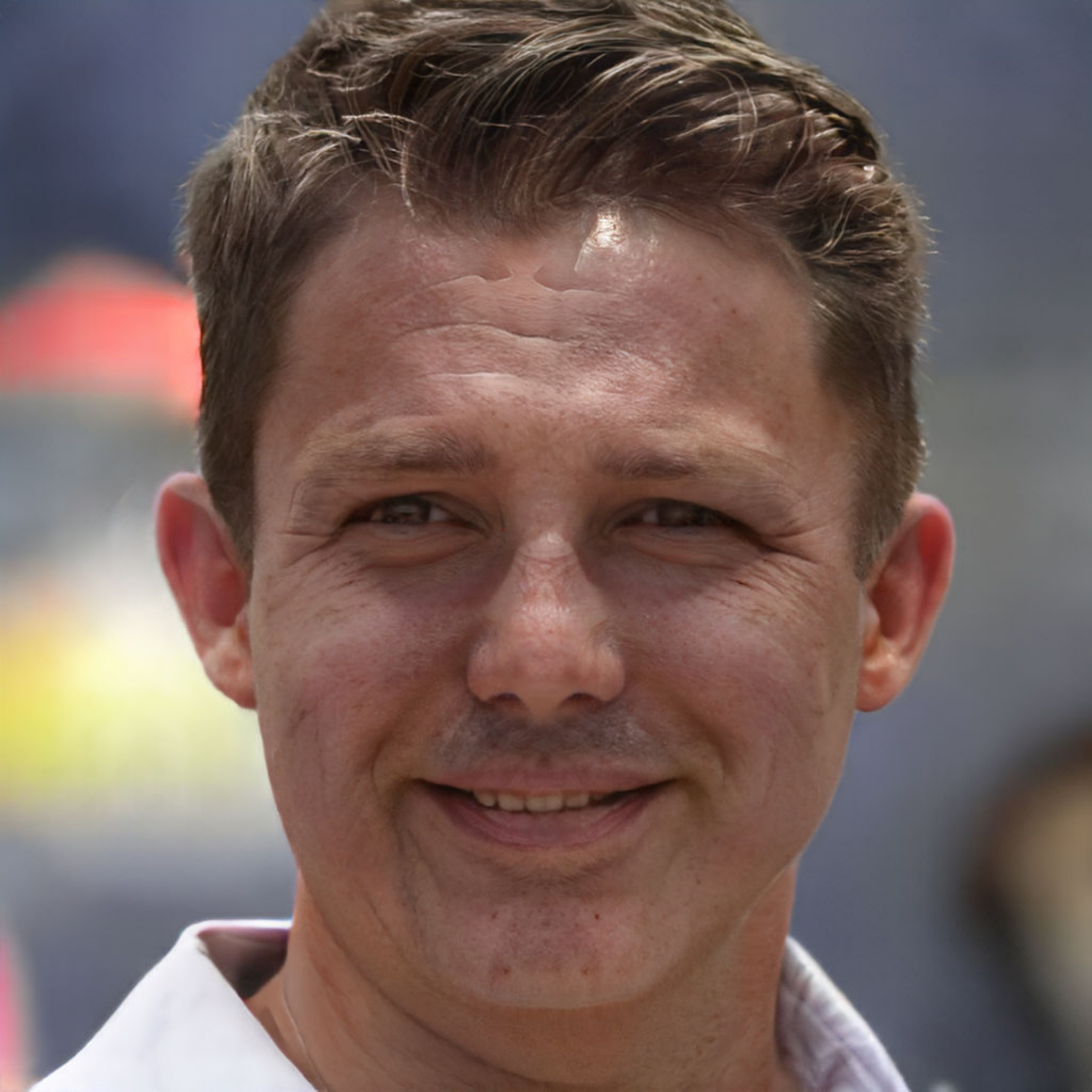} & \includegraphics[width=0.125\columnwidth]{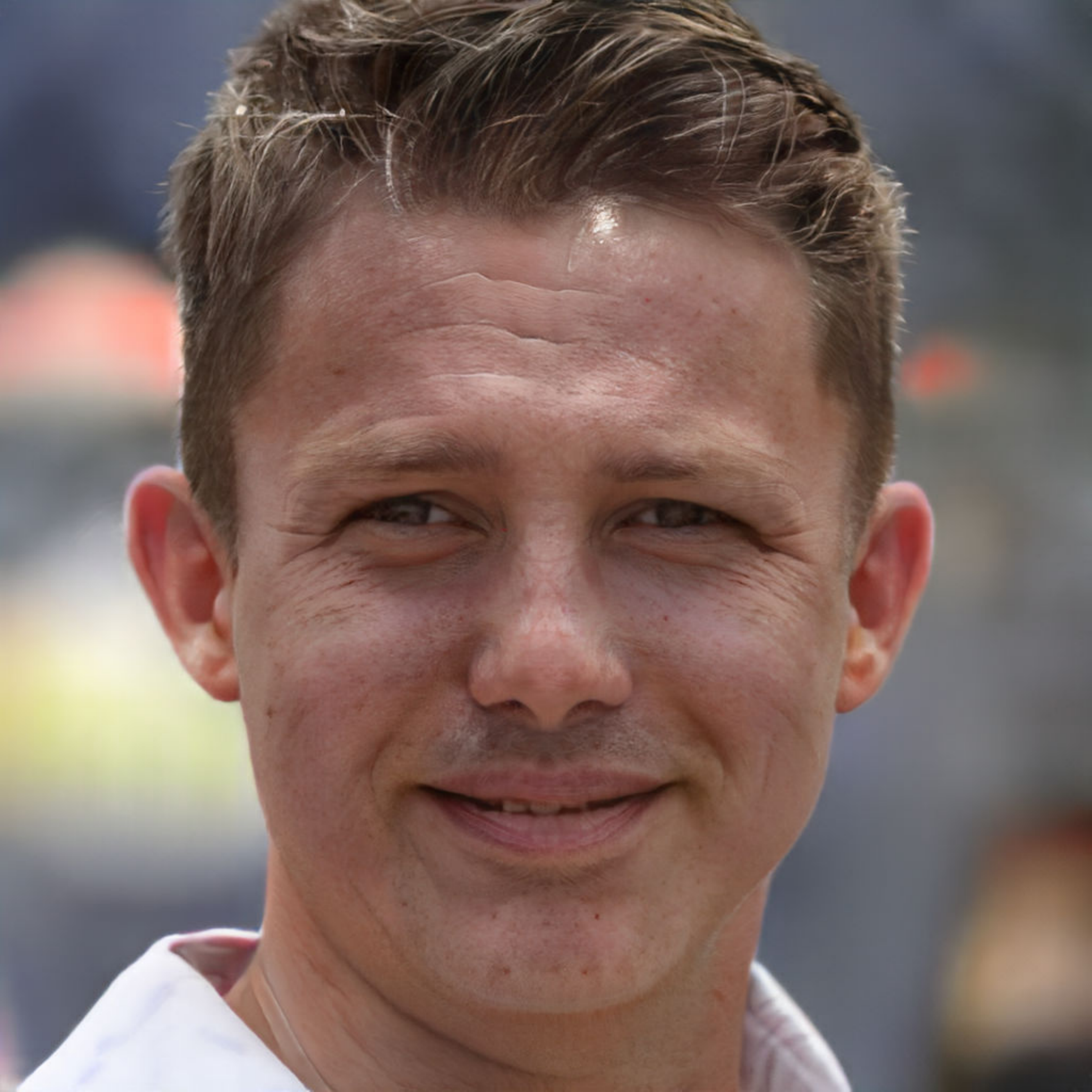} &
         \includegraphics[width=0.125\columnwidth]{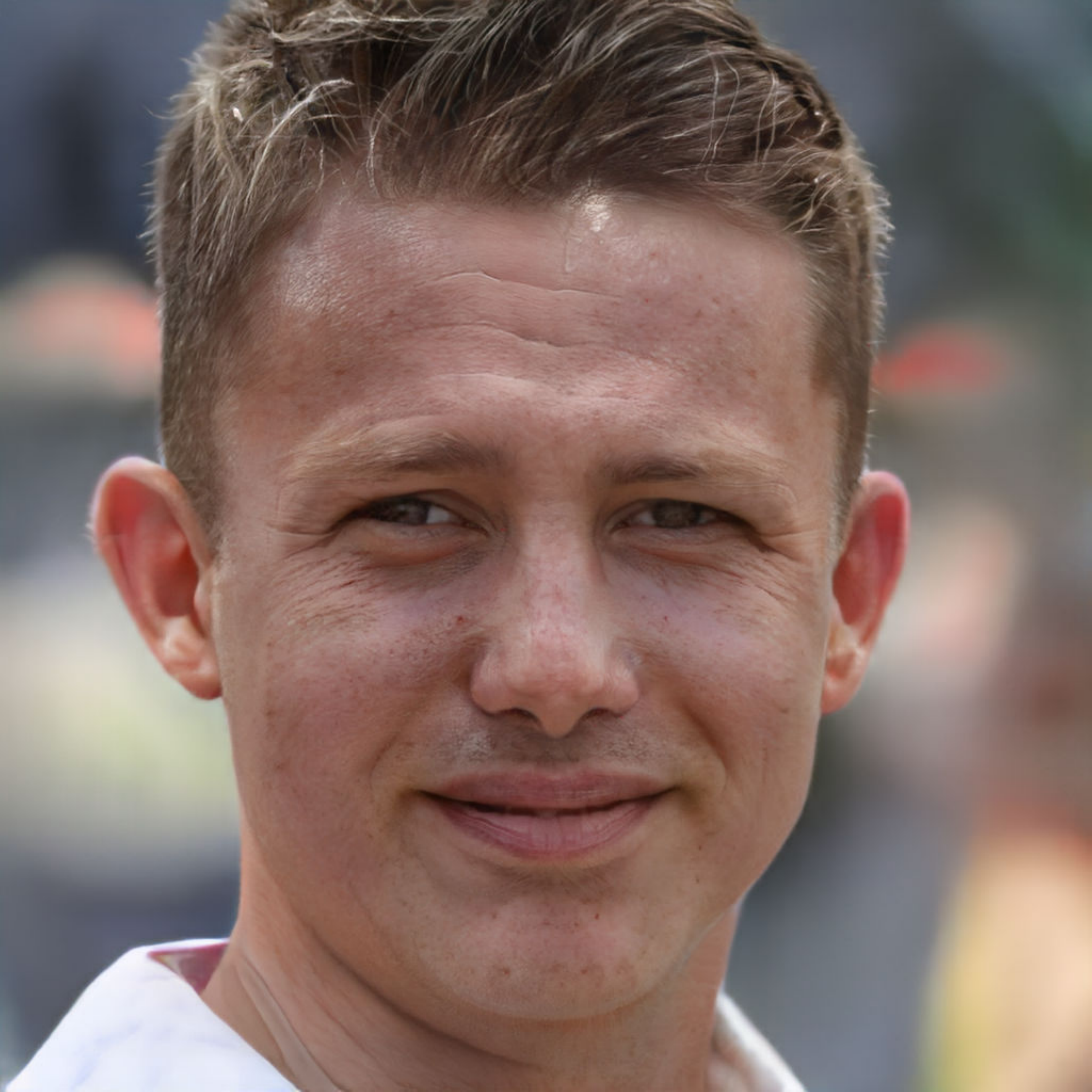} &
         \includegraphics[width=0.125\columnwidth]{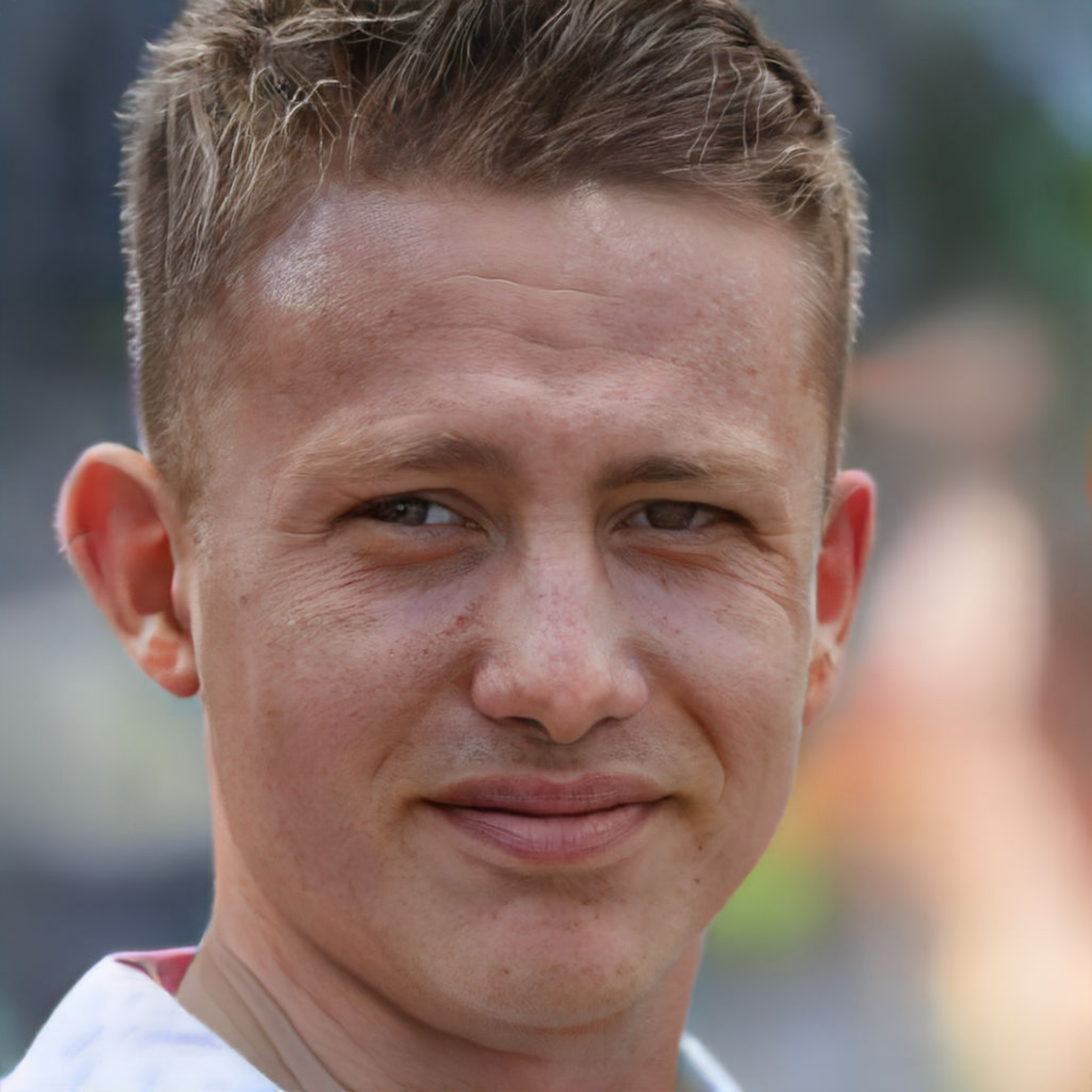}\\
         \includegraphics[width=0.125\columnwidth]{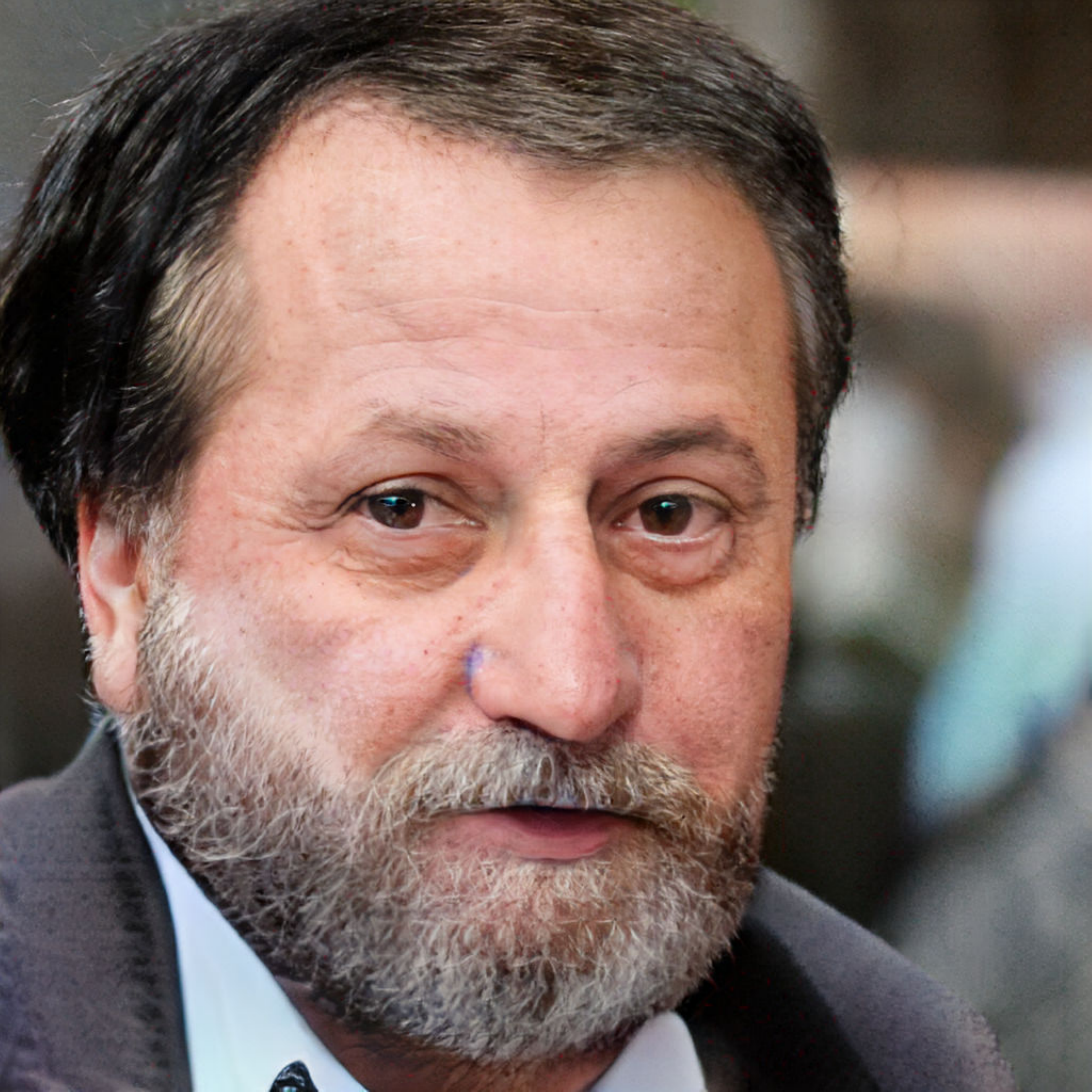} & \includegraphics[width=0.125\columnwidth]{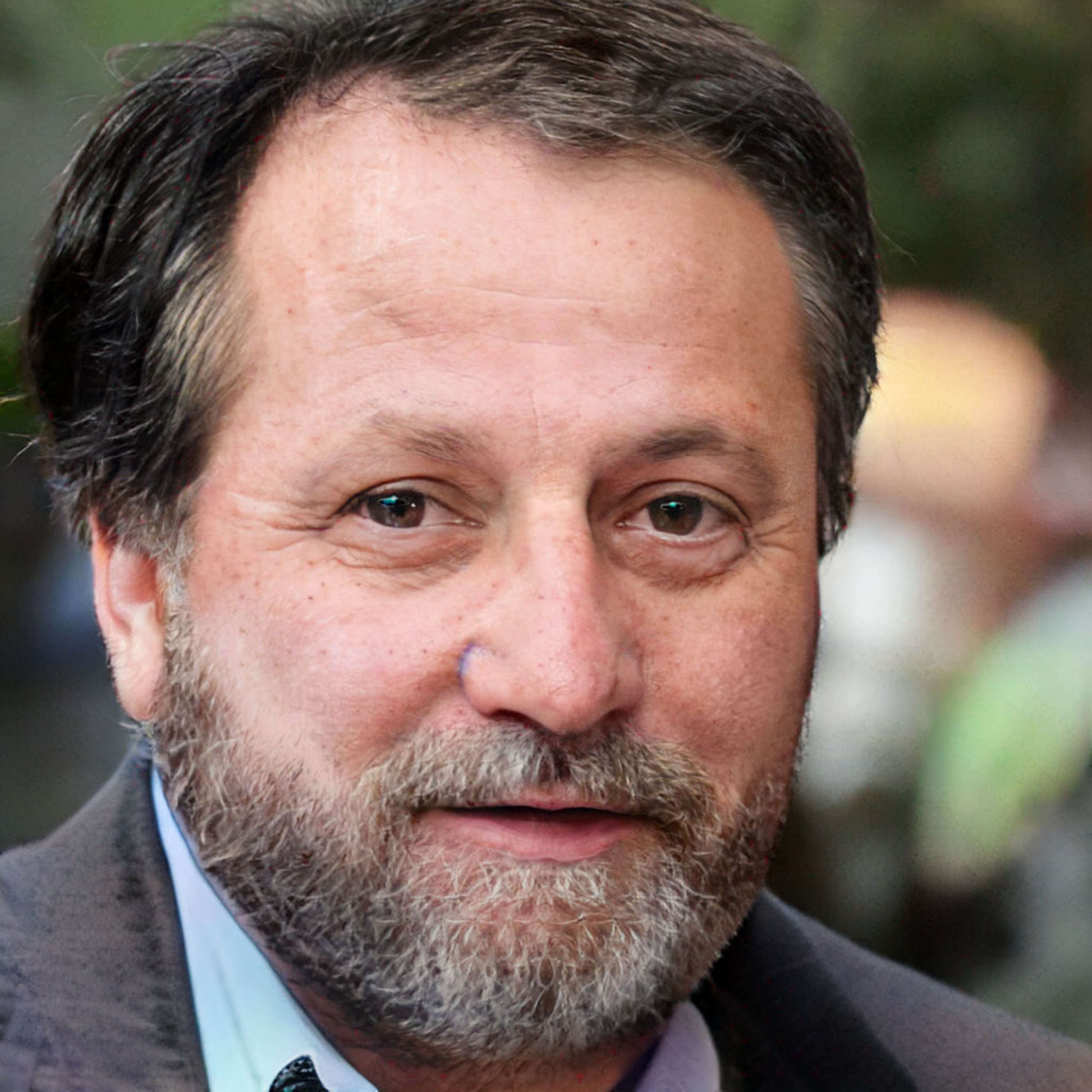} &
         \includegraphics[width=0.125\columnwidth]{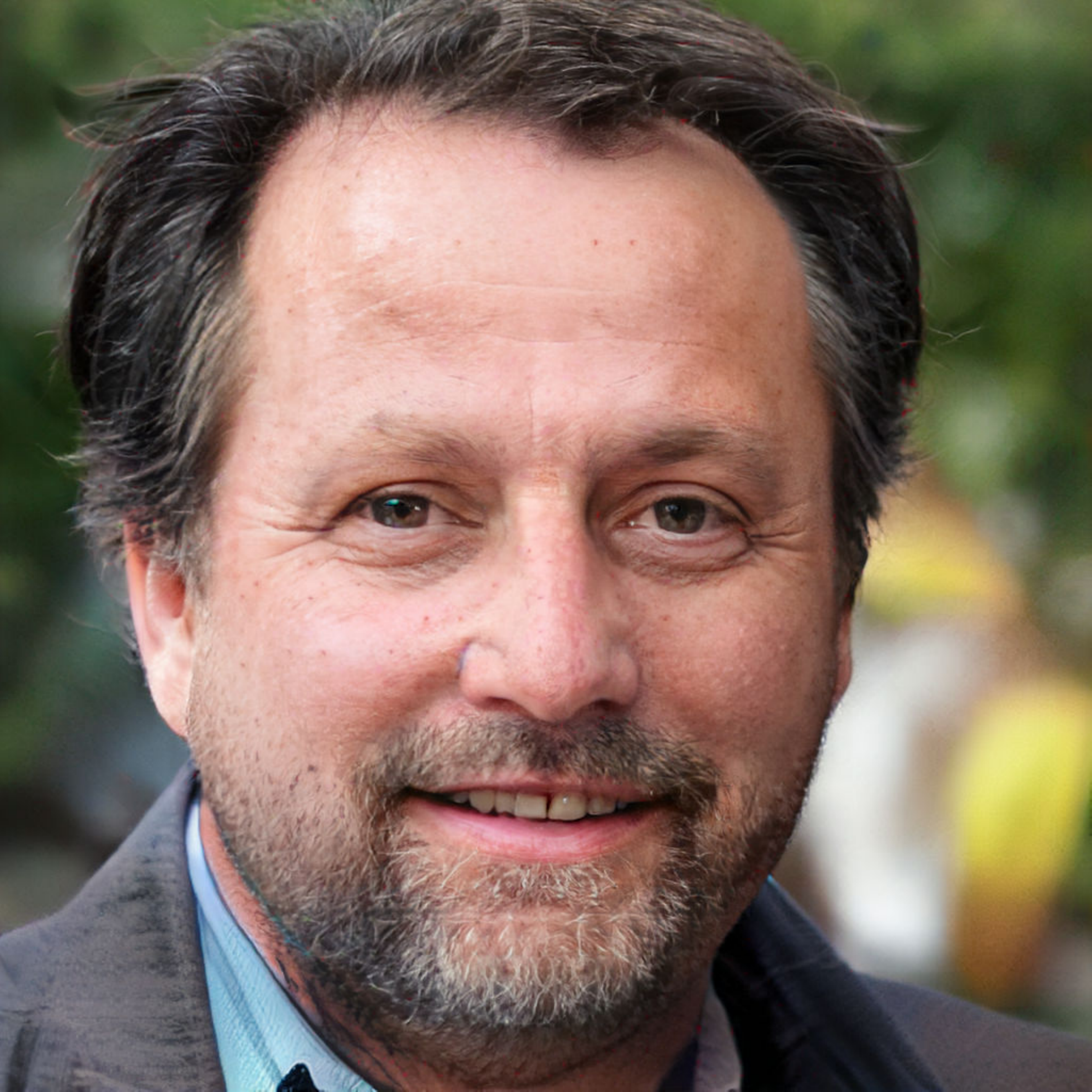} & \includegraphics[width=0.125\columnwidth]{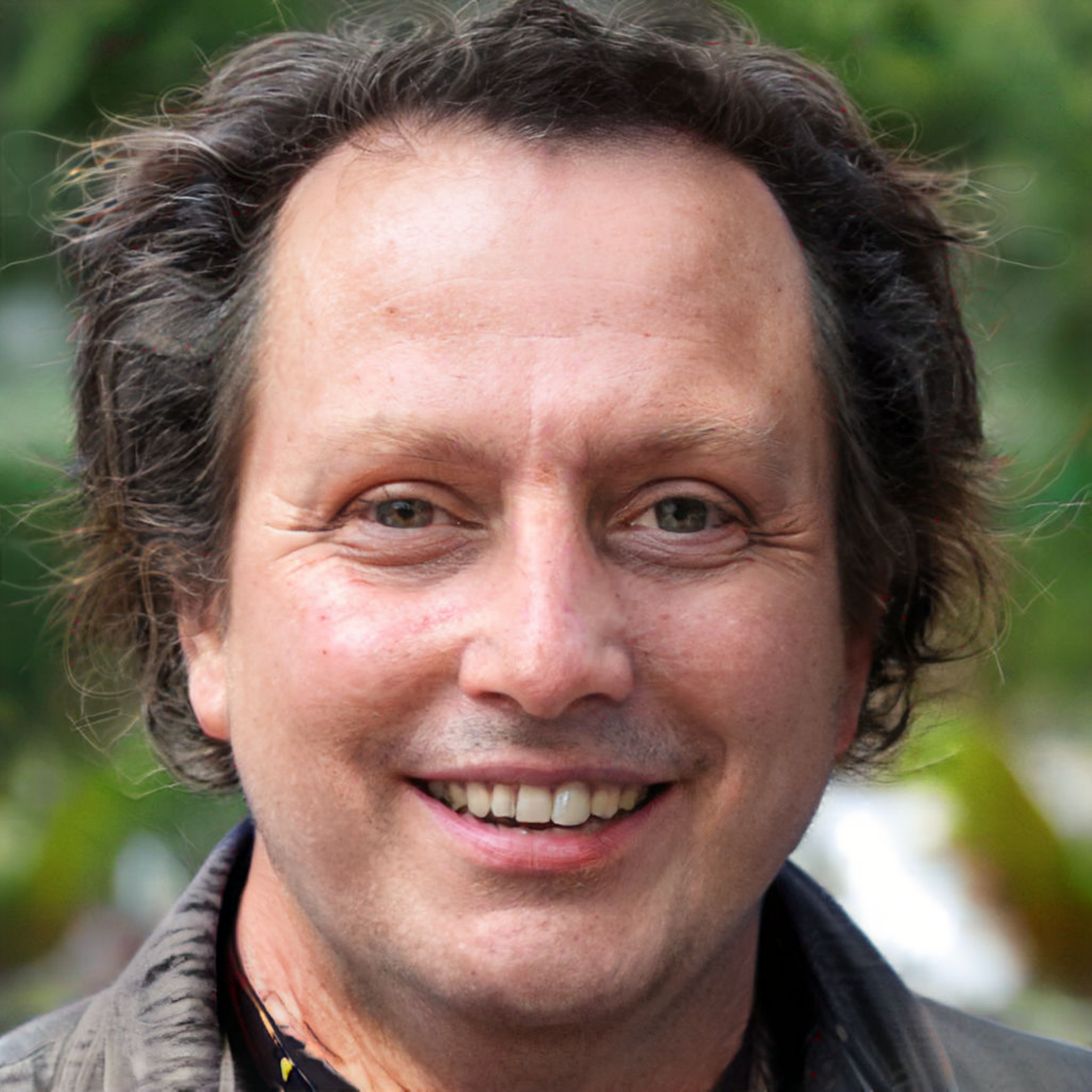} & \includegraphics[width=0.125\columnwidth]{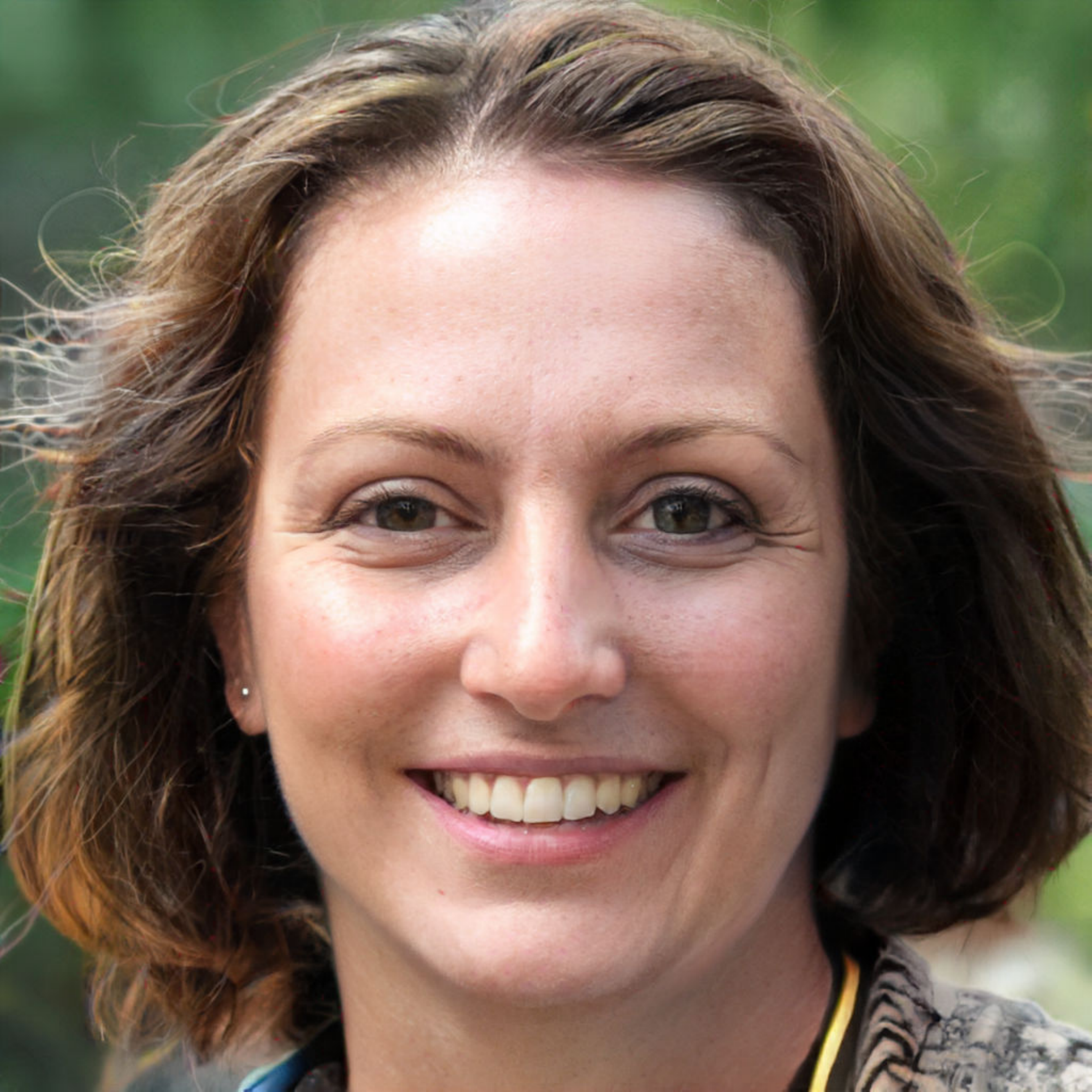} & \includegraphics[width=0.125\columnwidth]{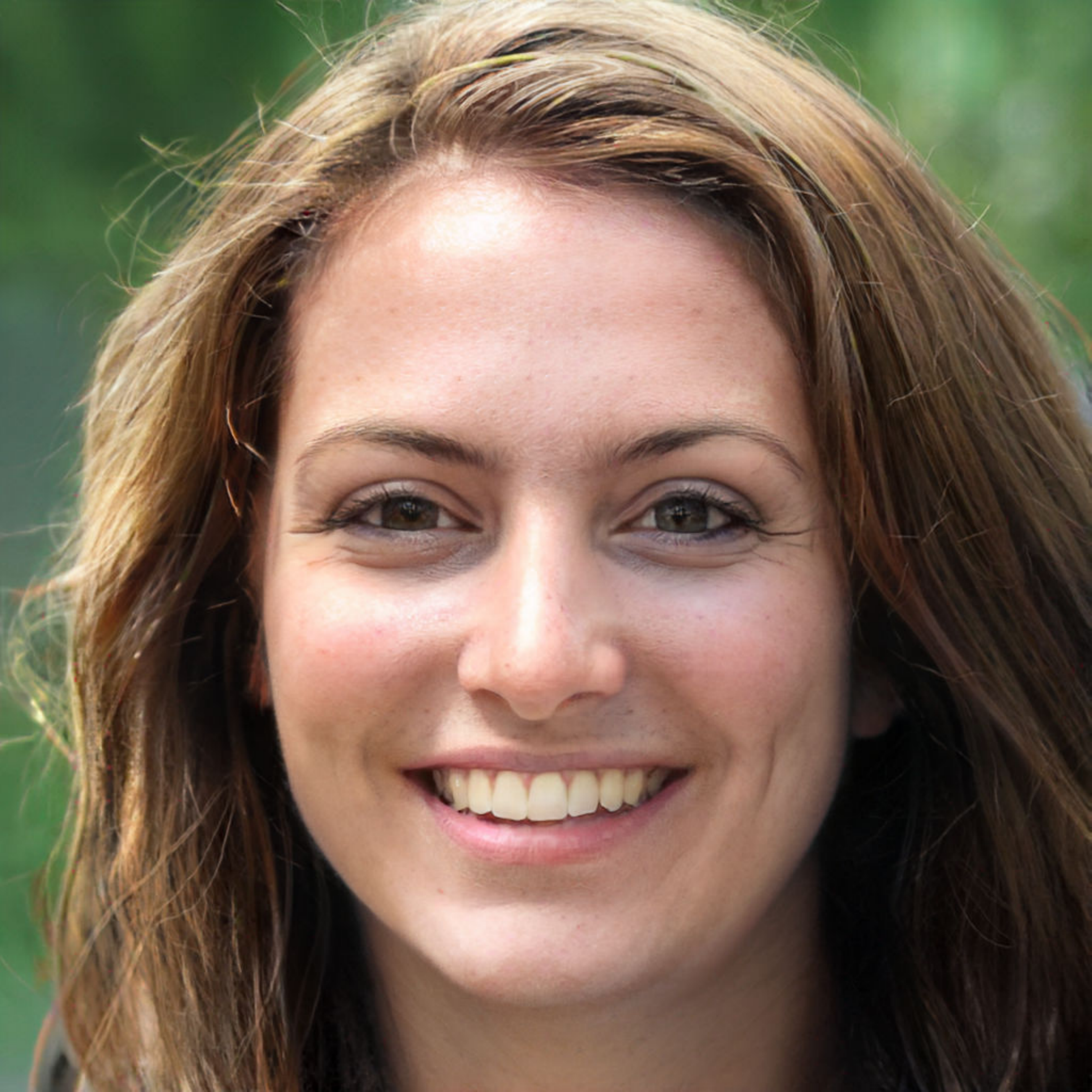} &
         \includegraphics[width=0.125\columnwidth]{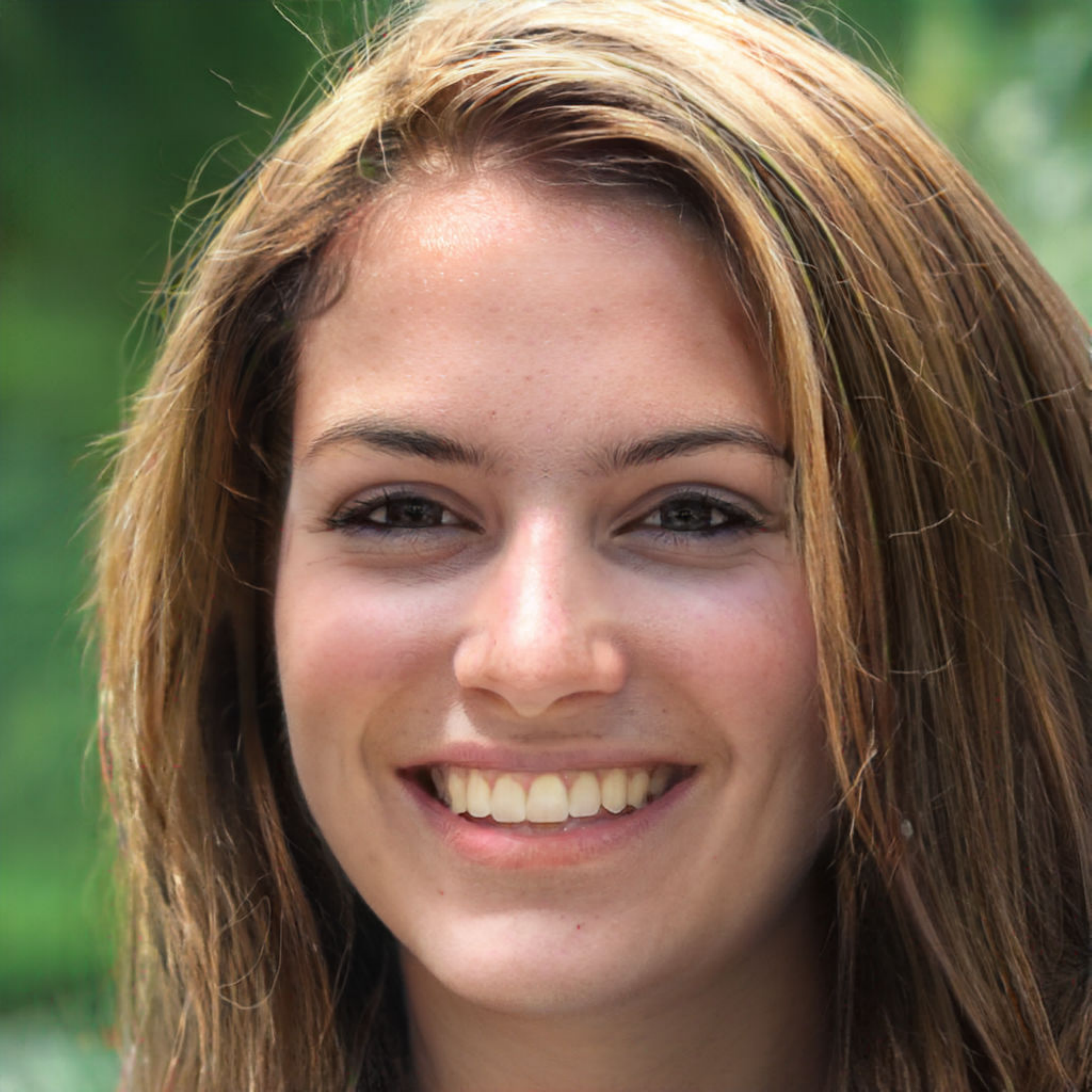} &
         \includegraphics[width=0.125\columnwidth]{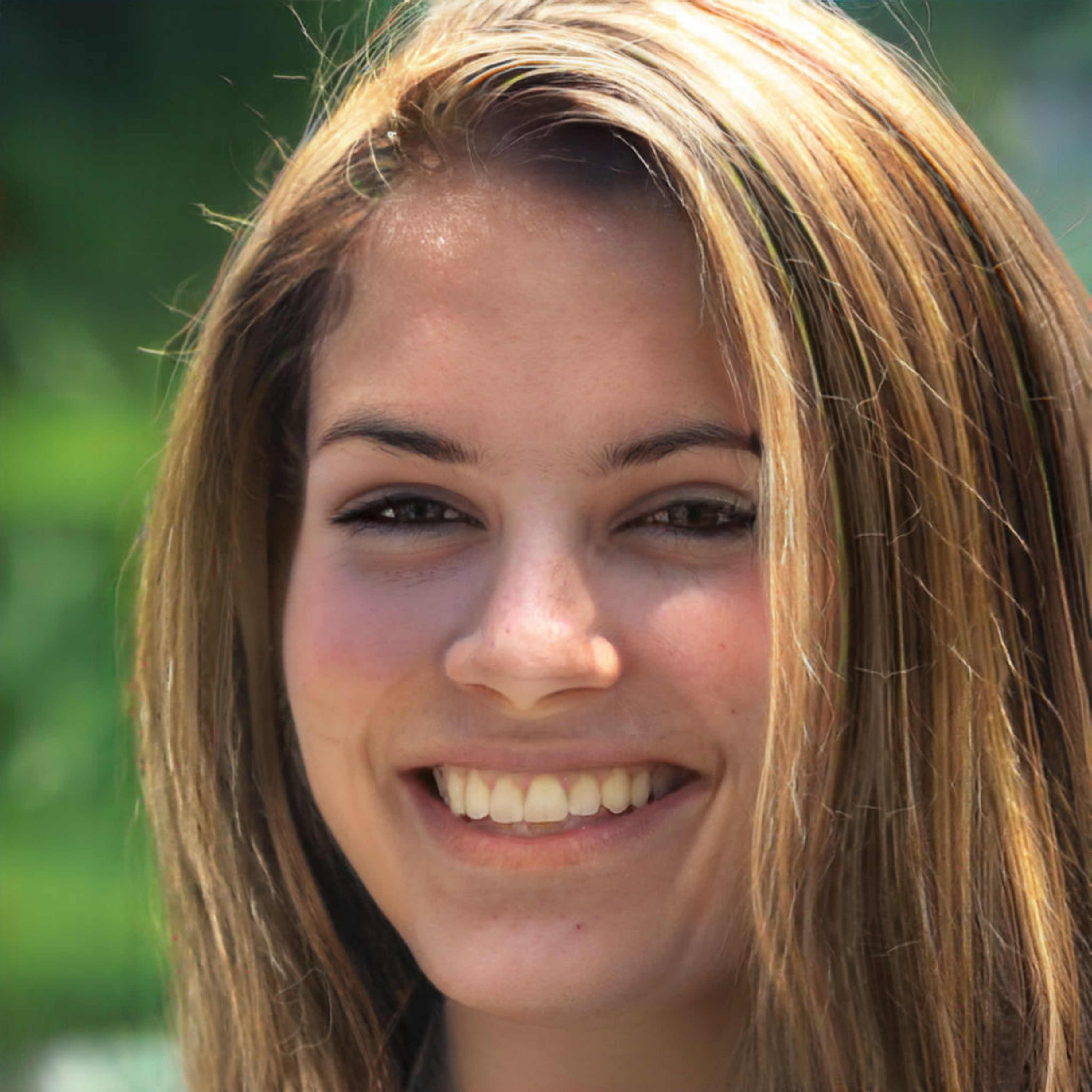}\\
         \includegraphics[width=0.125\columnwidth]{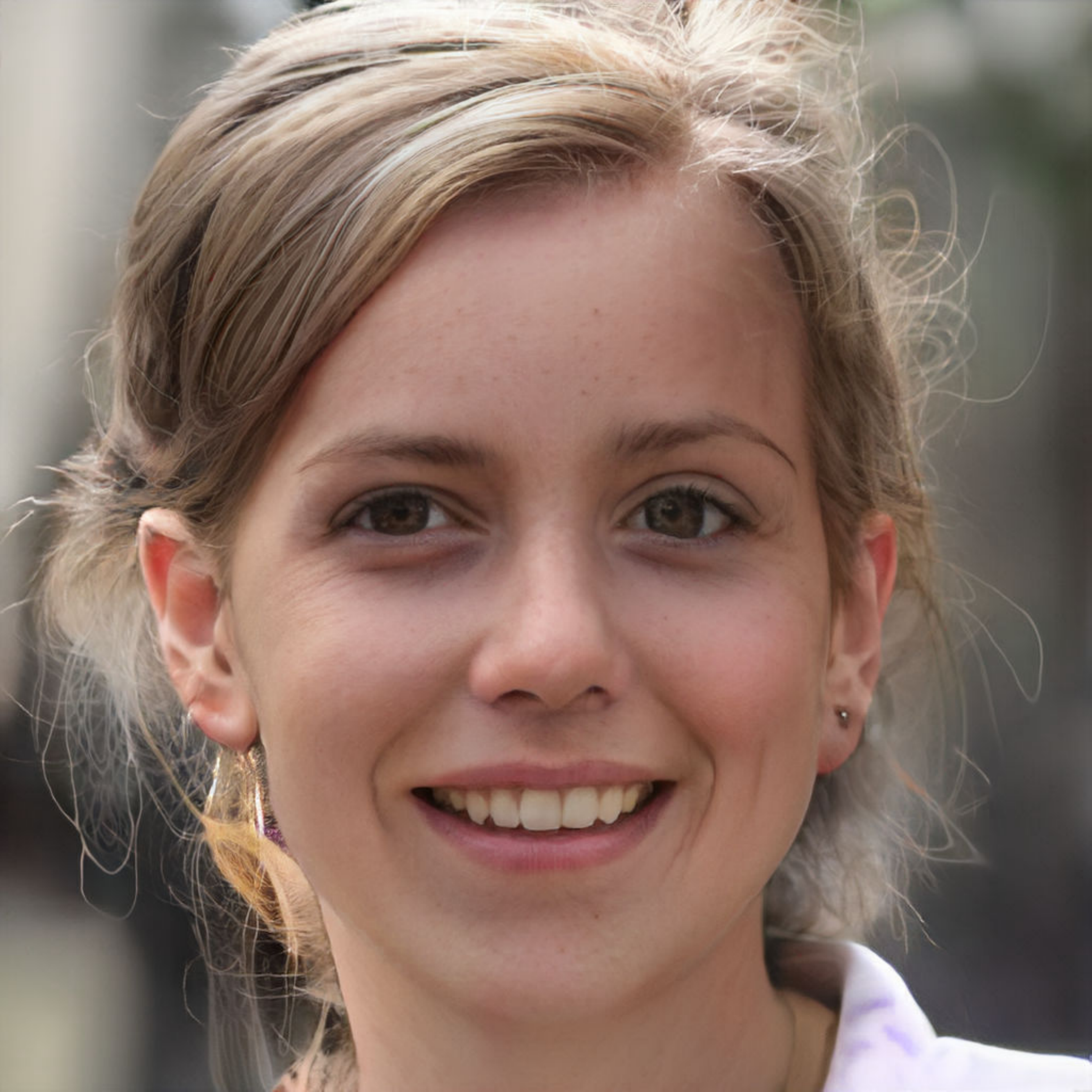} & \includegraphics[width=0.125\columnwidth]{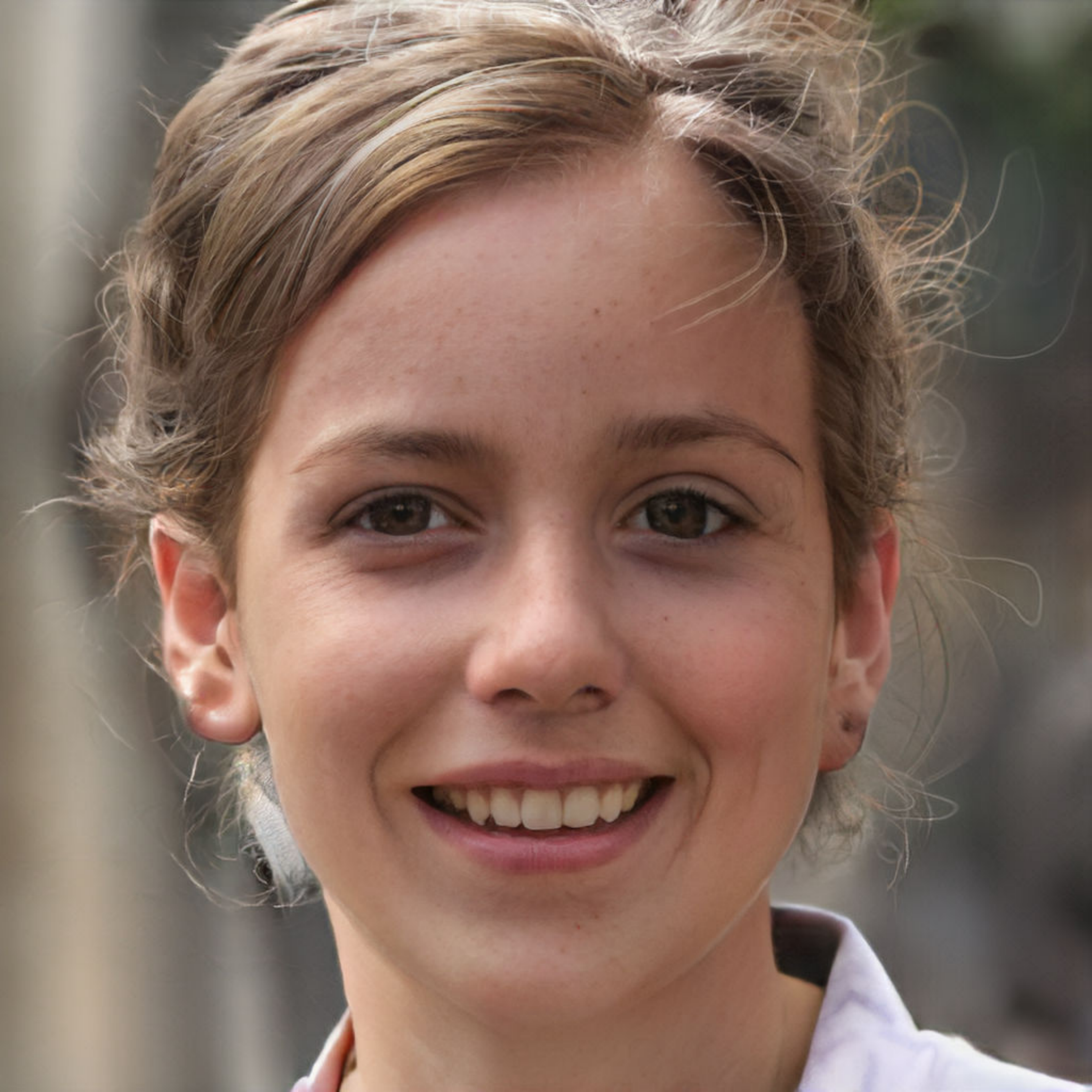} &
         \includegraphics[width=0.125\columnwidth]{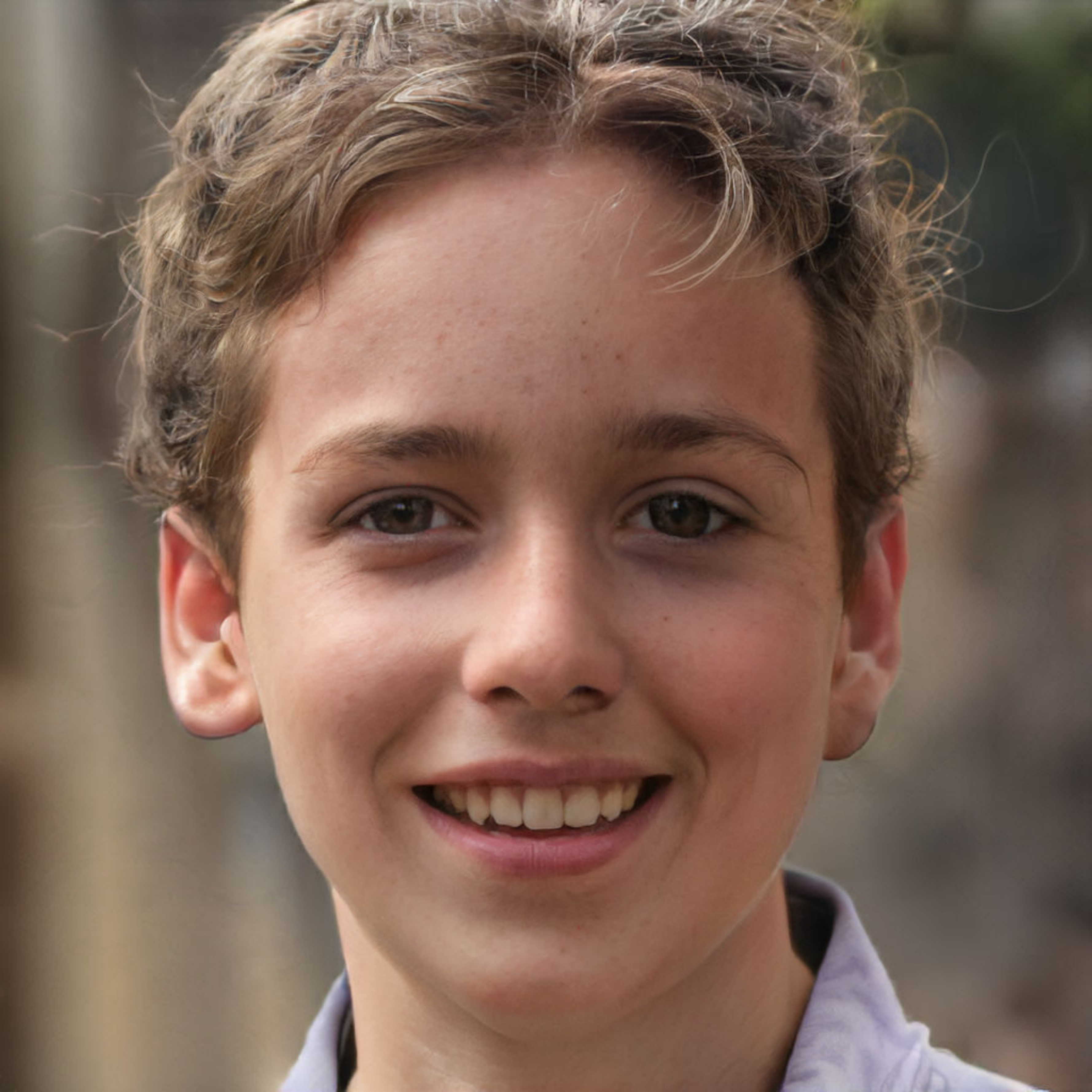} & \includegraphics[width=0.125\columnwidth]{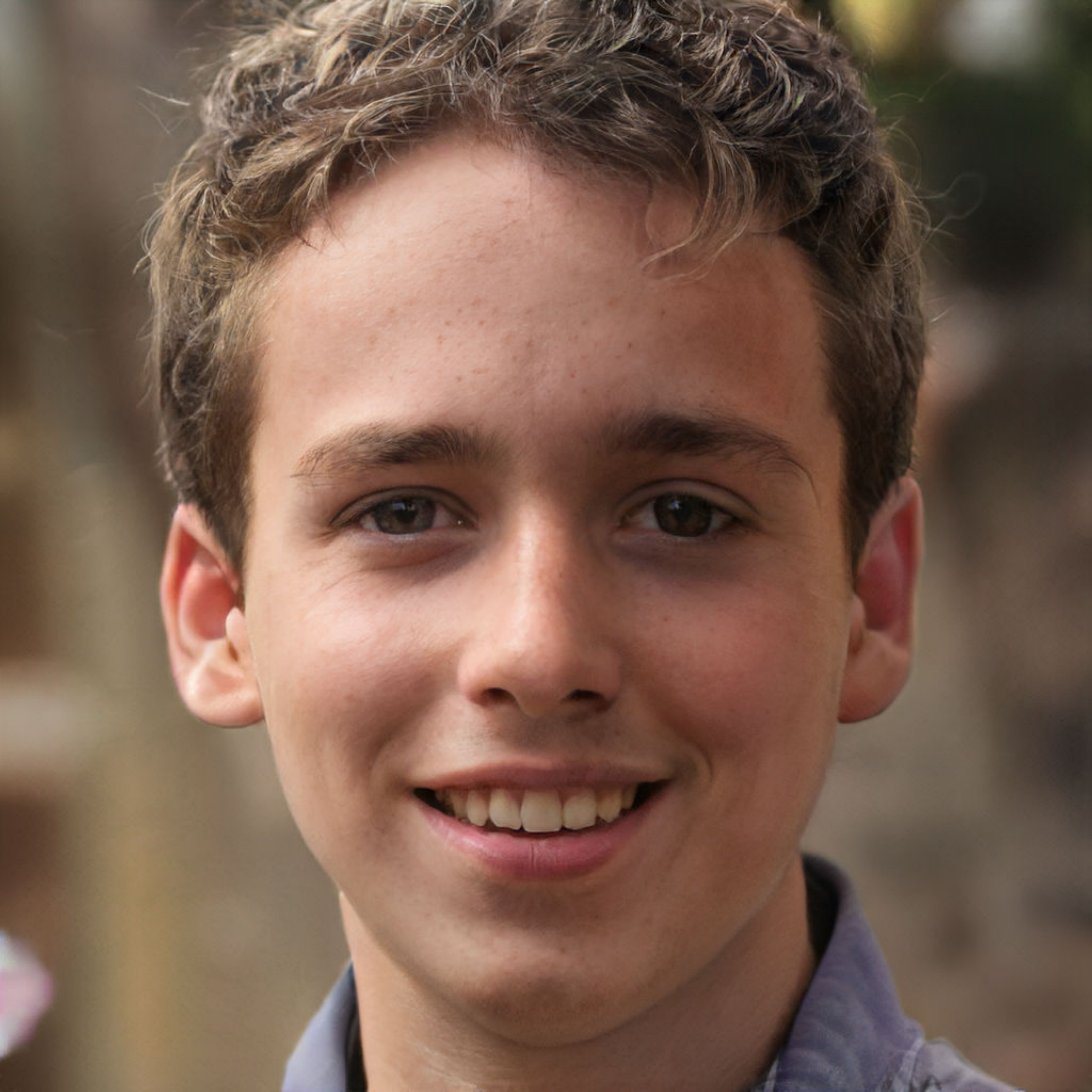} & \includegraphics[width=0.125\columnwidth]{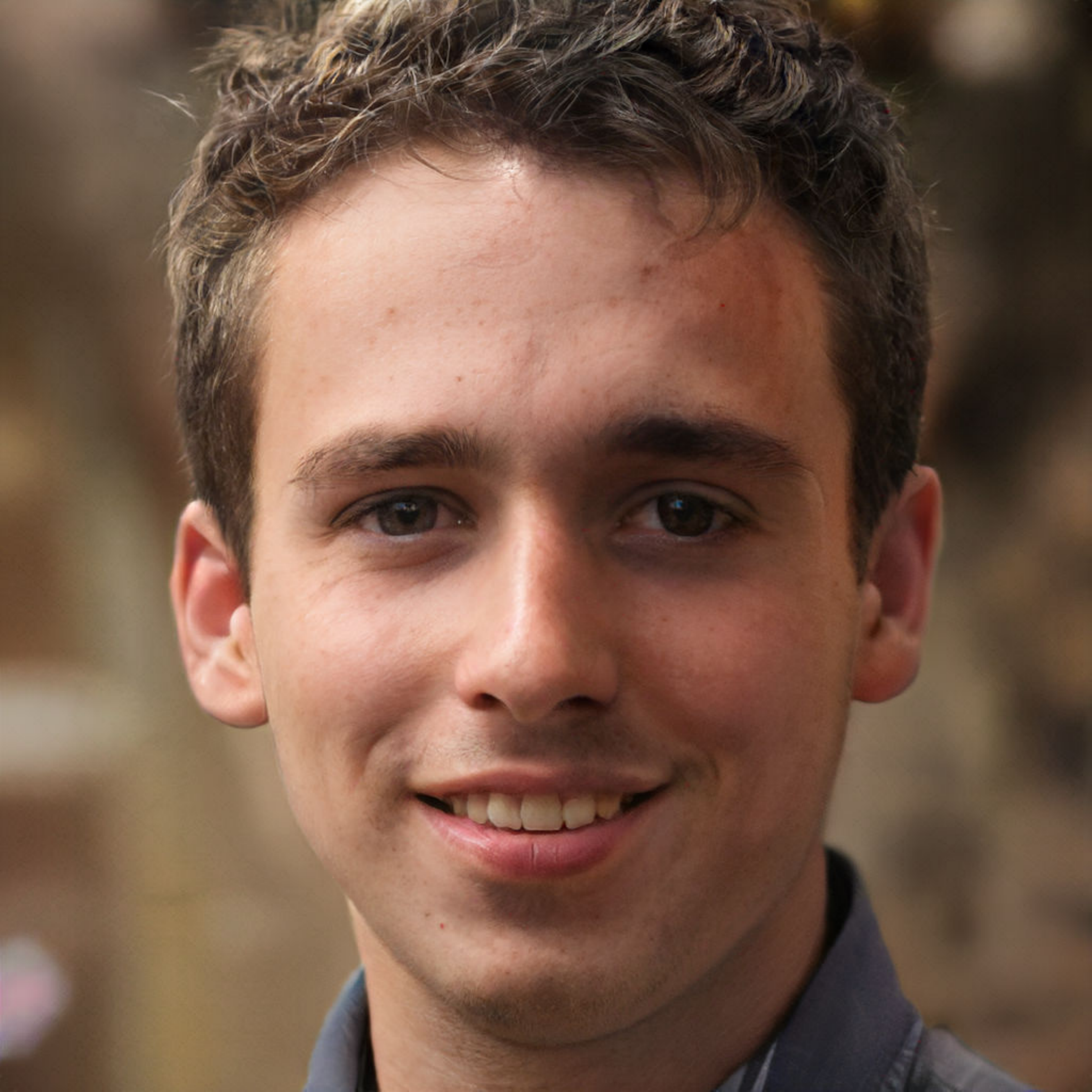} & \includegraphics[width=0.125\columnwidth]{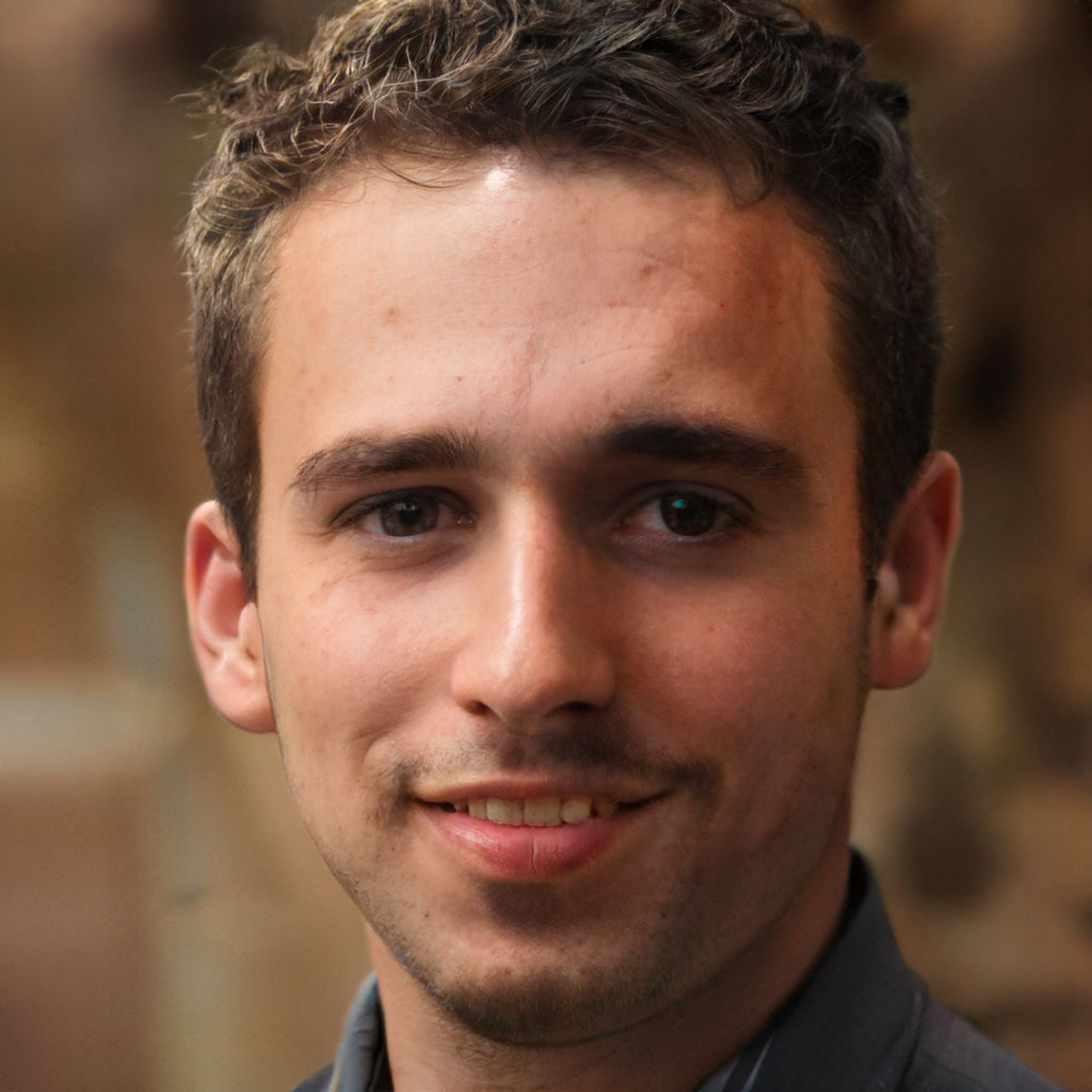} &
         \includegraphics[width=0.125\columnwidth]{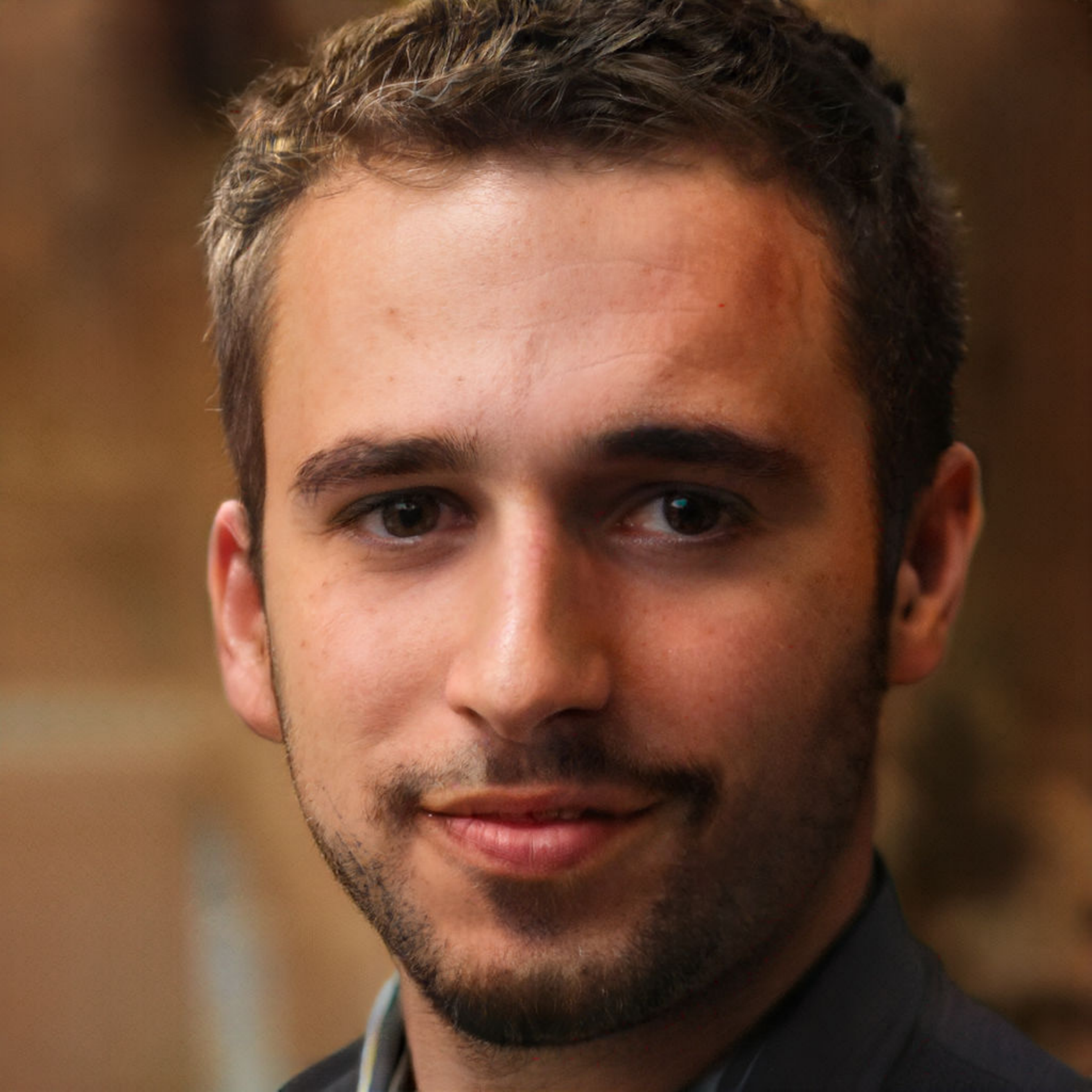} &
         \includegraphics[width=0.125\columnwidth]{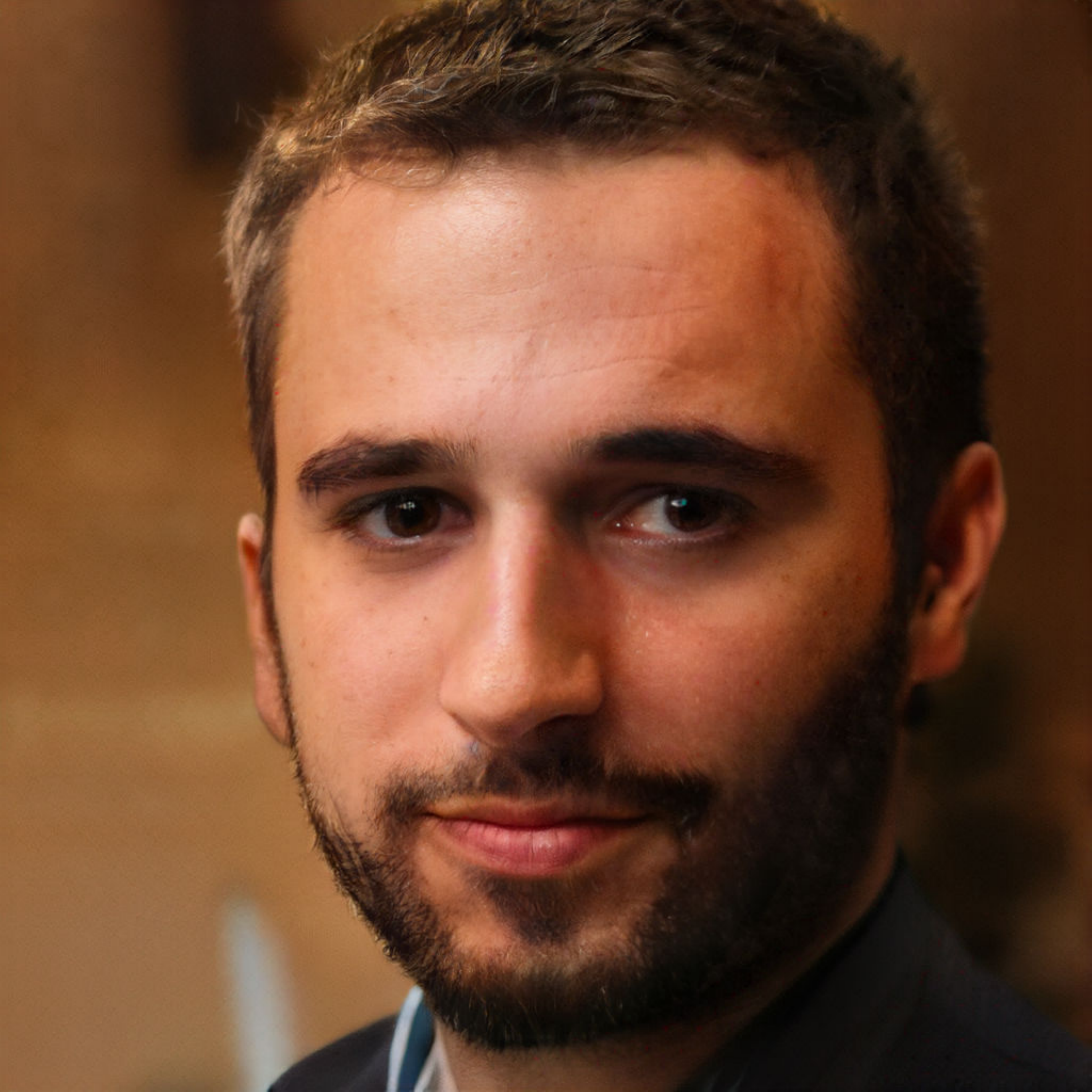}\\
         \includegraphics[width=0.125\columnwidth]{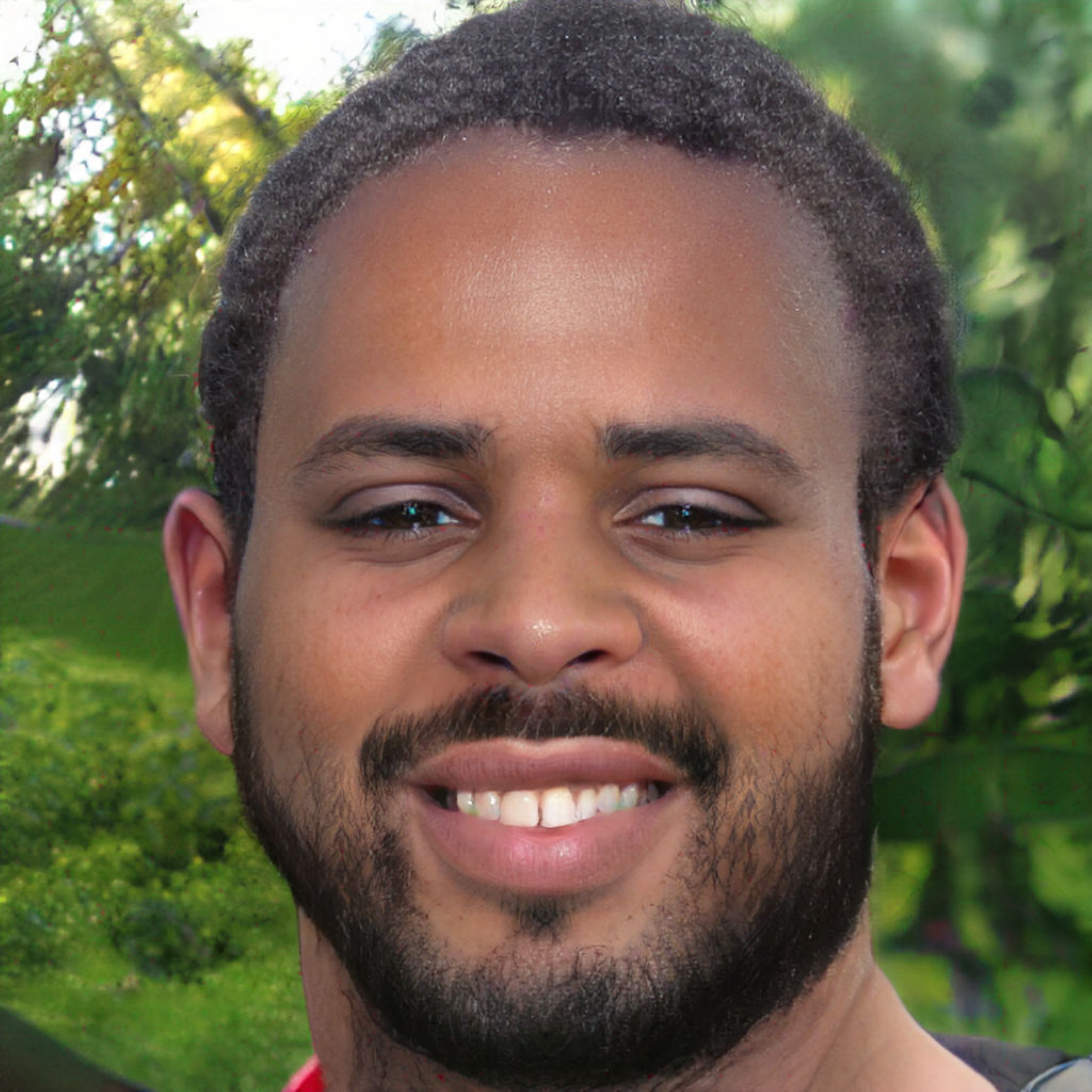} & \includegraphics[width=0.125\columnwidth]{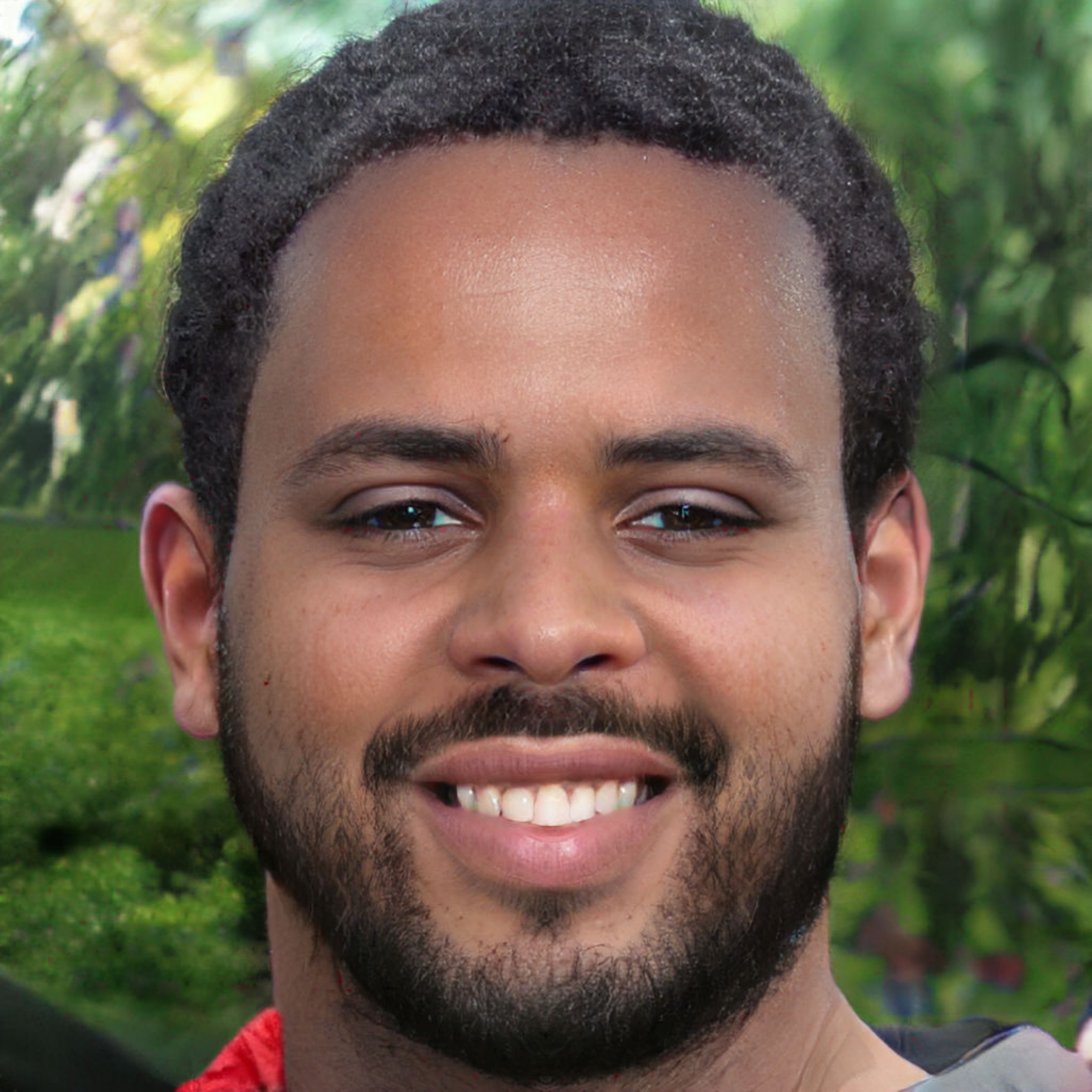} &
         \includegraphics[width=0.125\columnwidth]{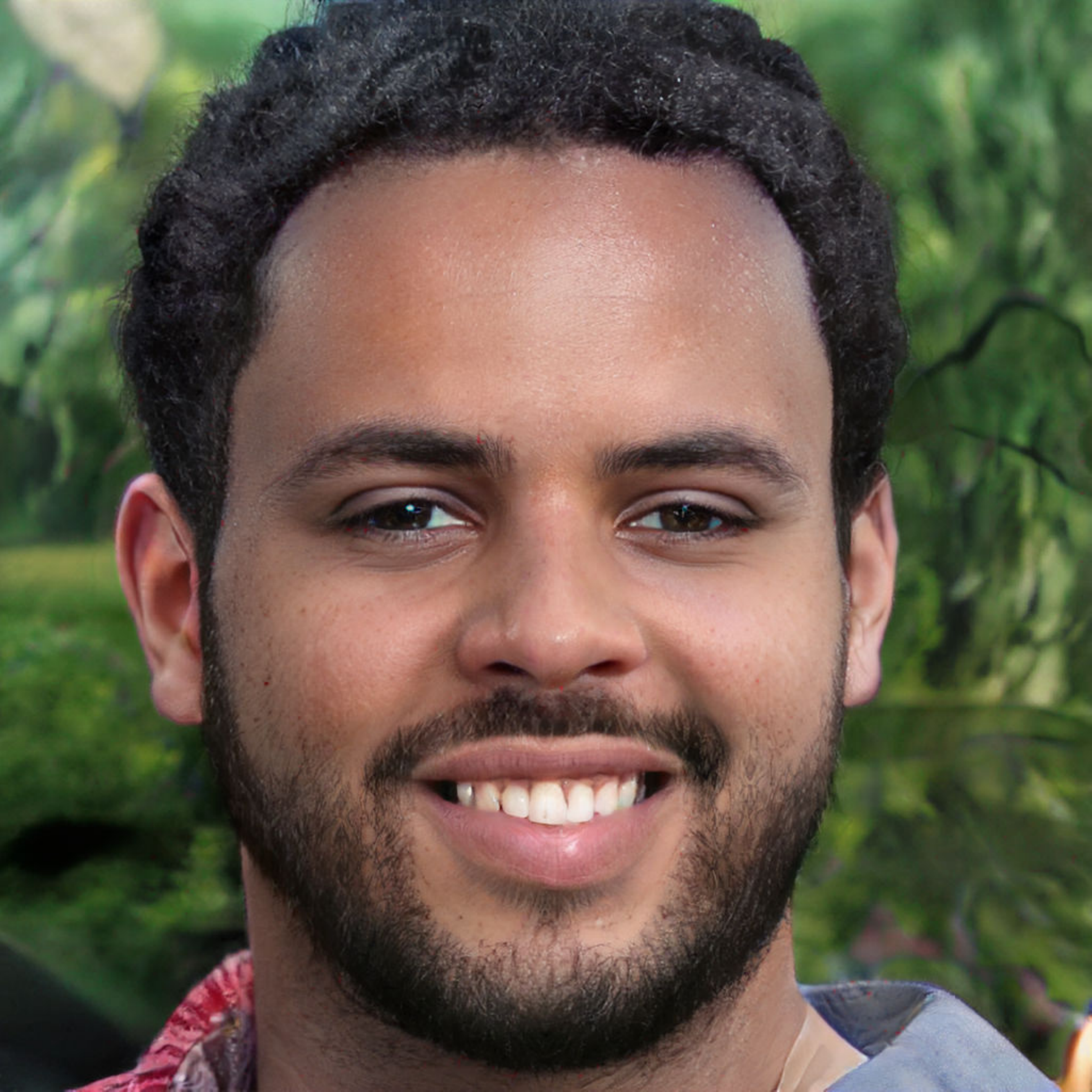} & \includegraphics[width=0.125\columnwidth]{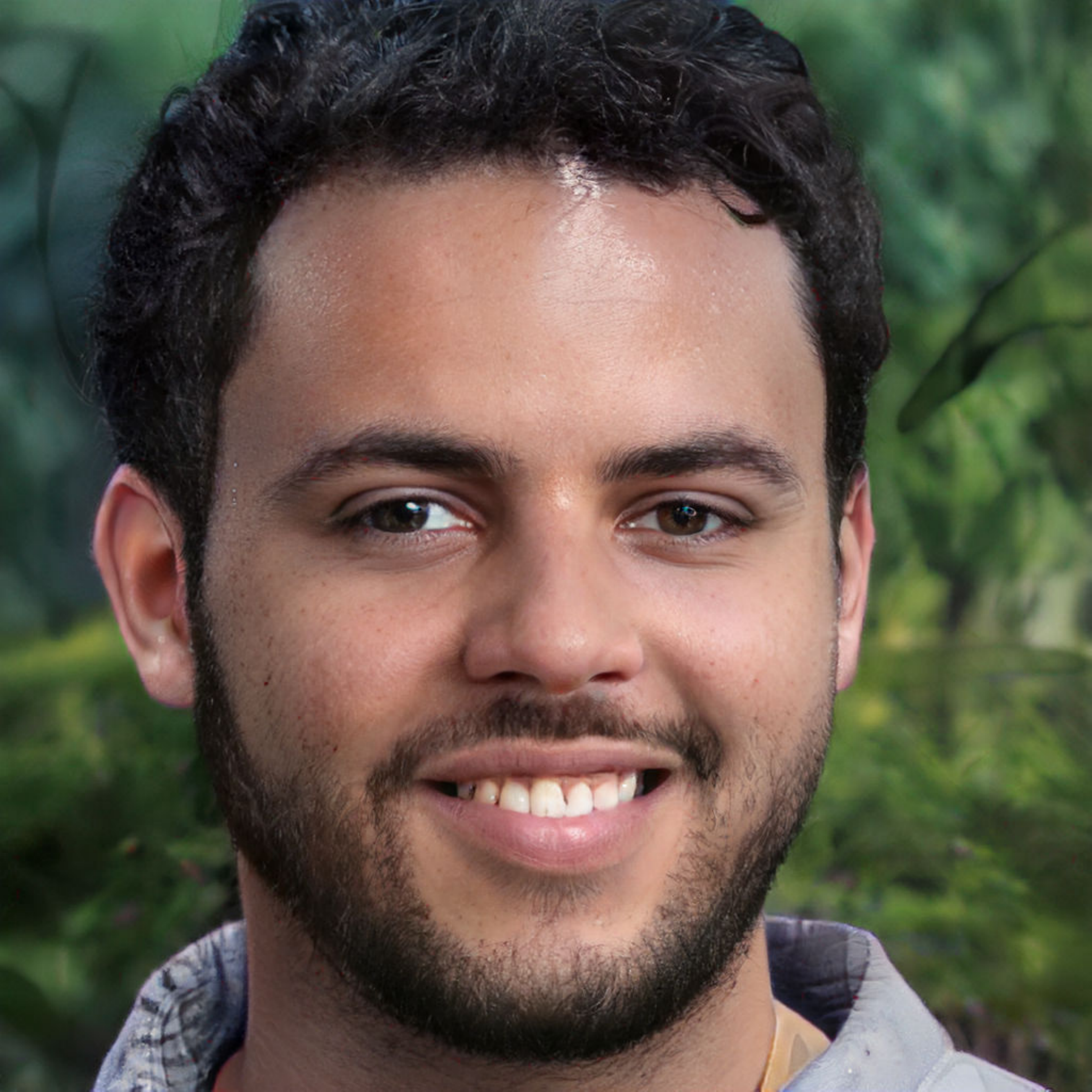} & \includegraphics[width=0.125\columnwidth]{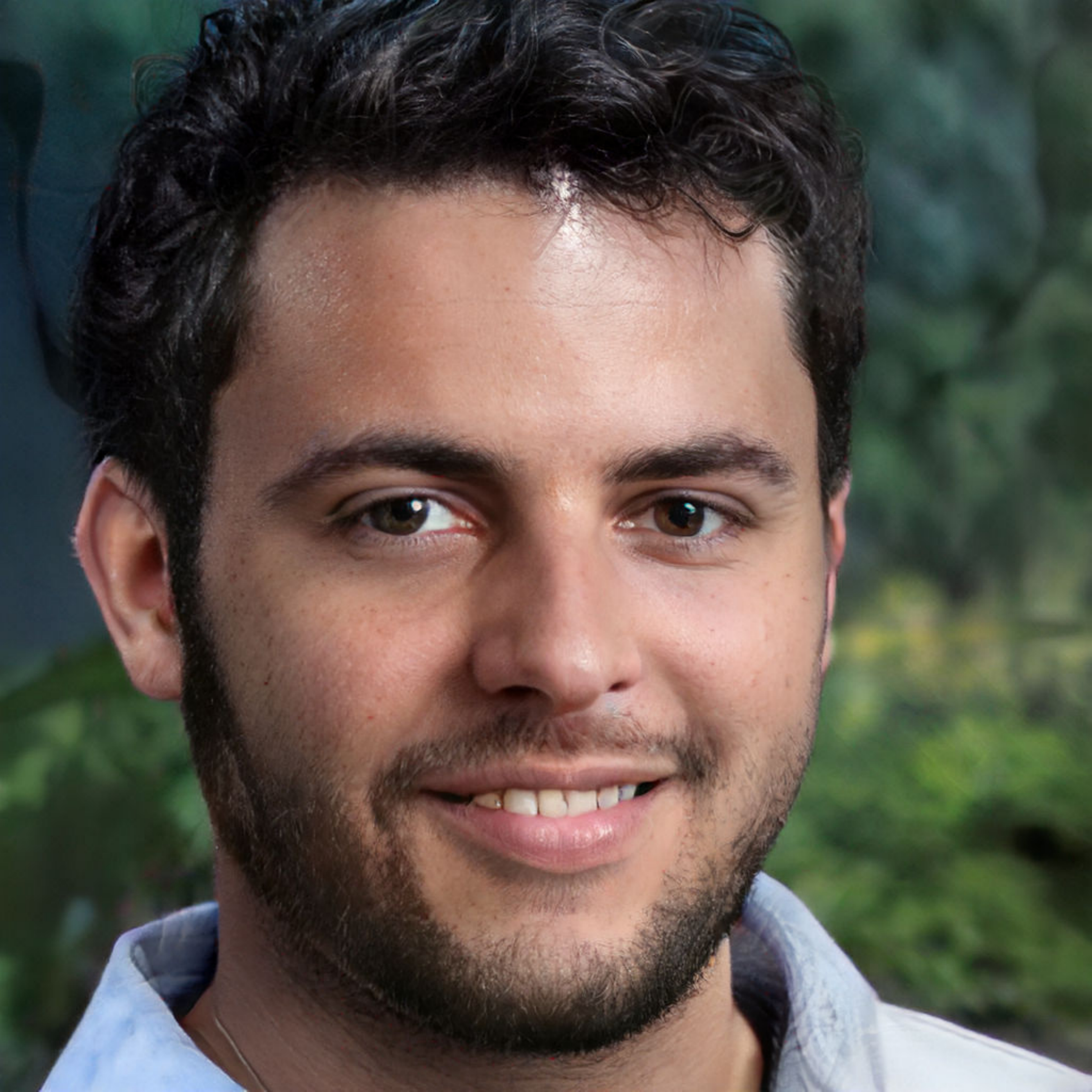} & \includegraphics[width=0.125\columnwidth]{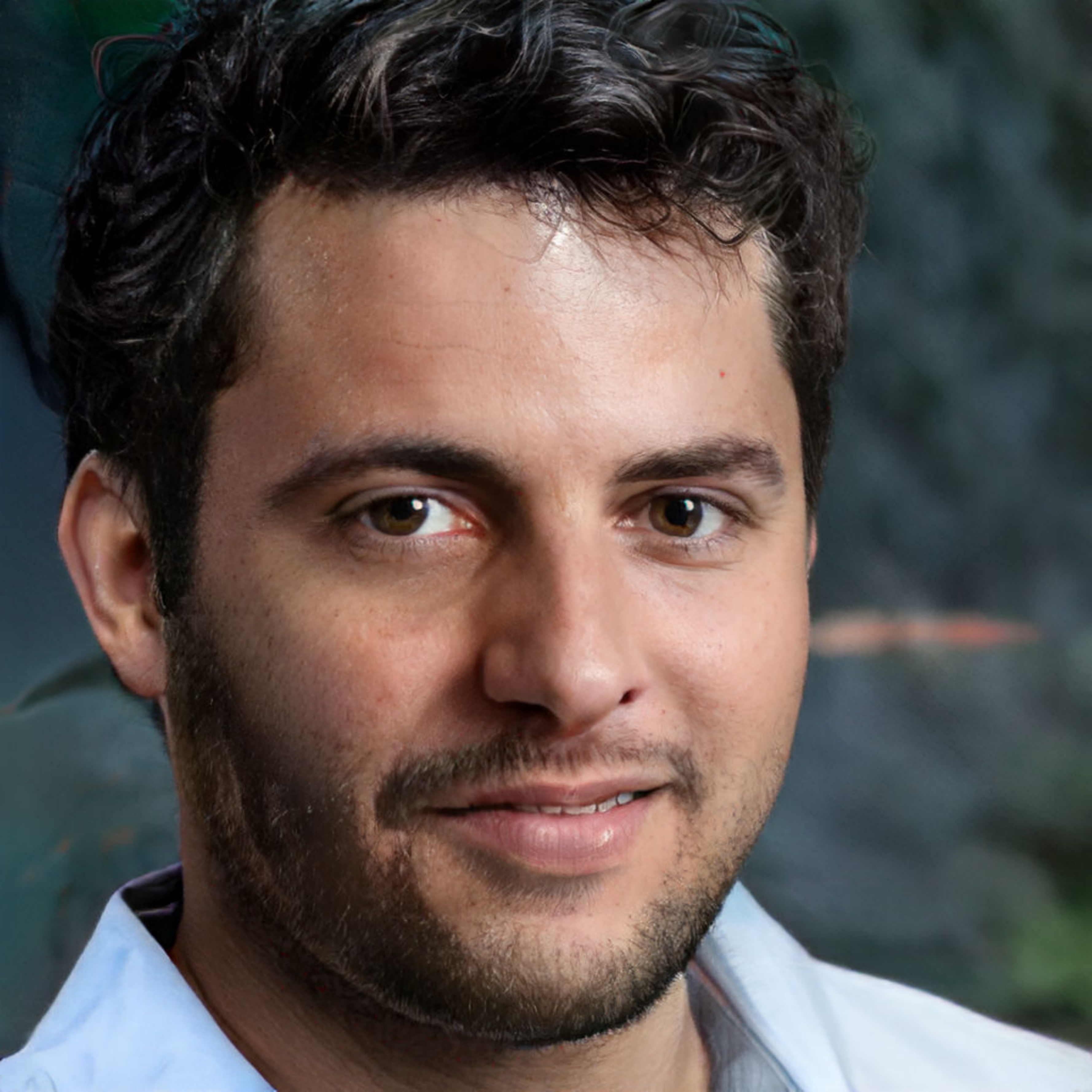} &
         \includegraphics[width=0.125\columnwidth]{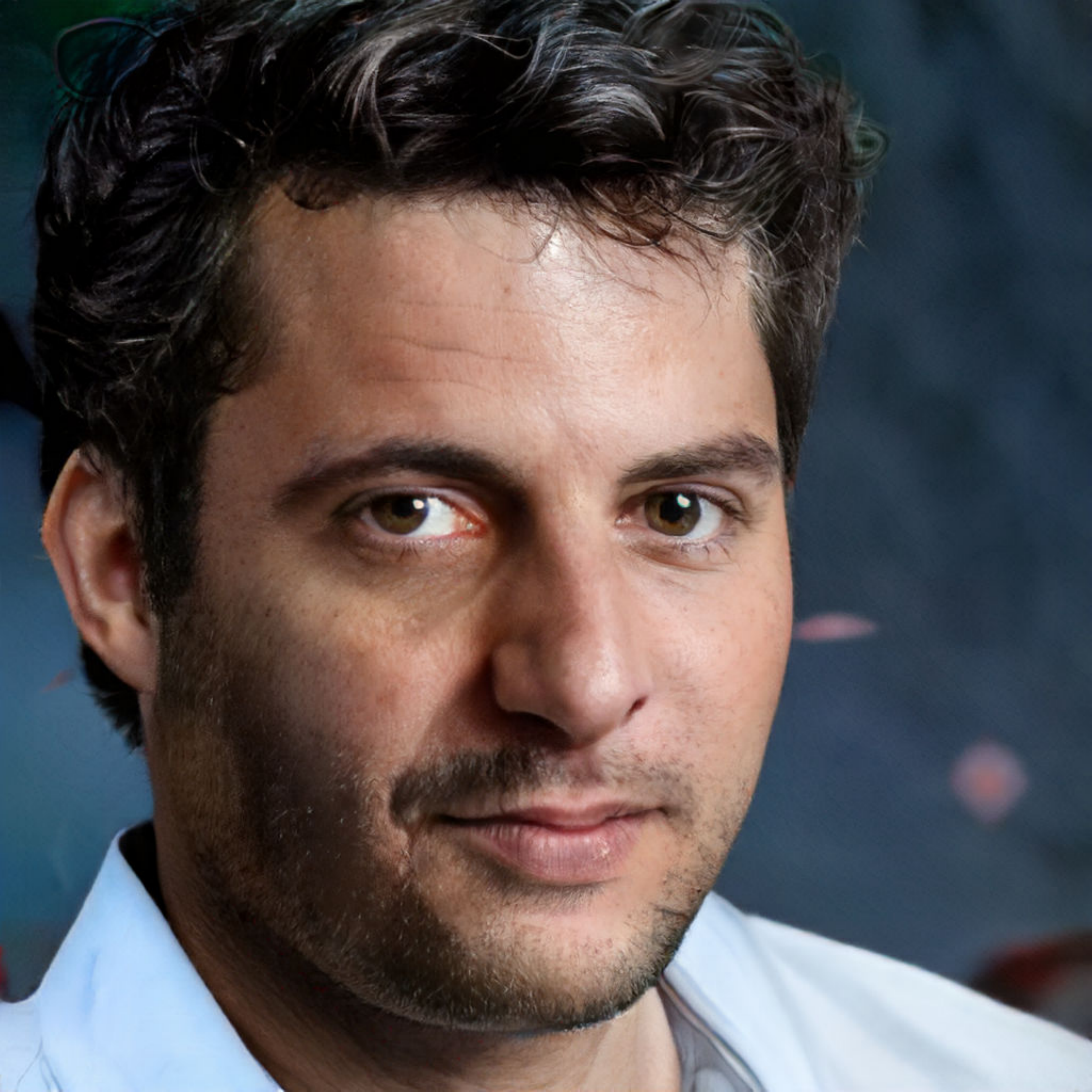} &
         \includegraphics[width=0.125\columnwidth]{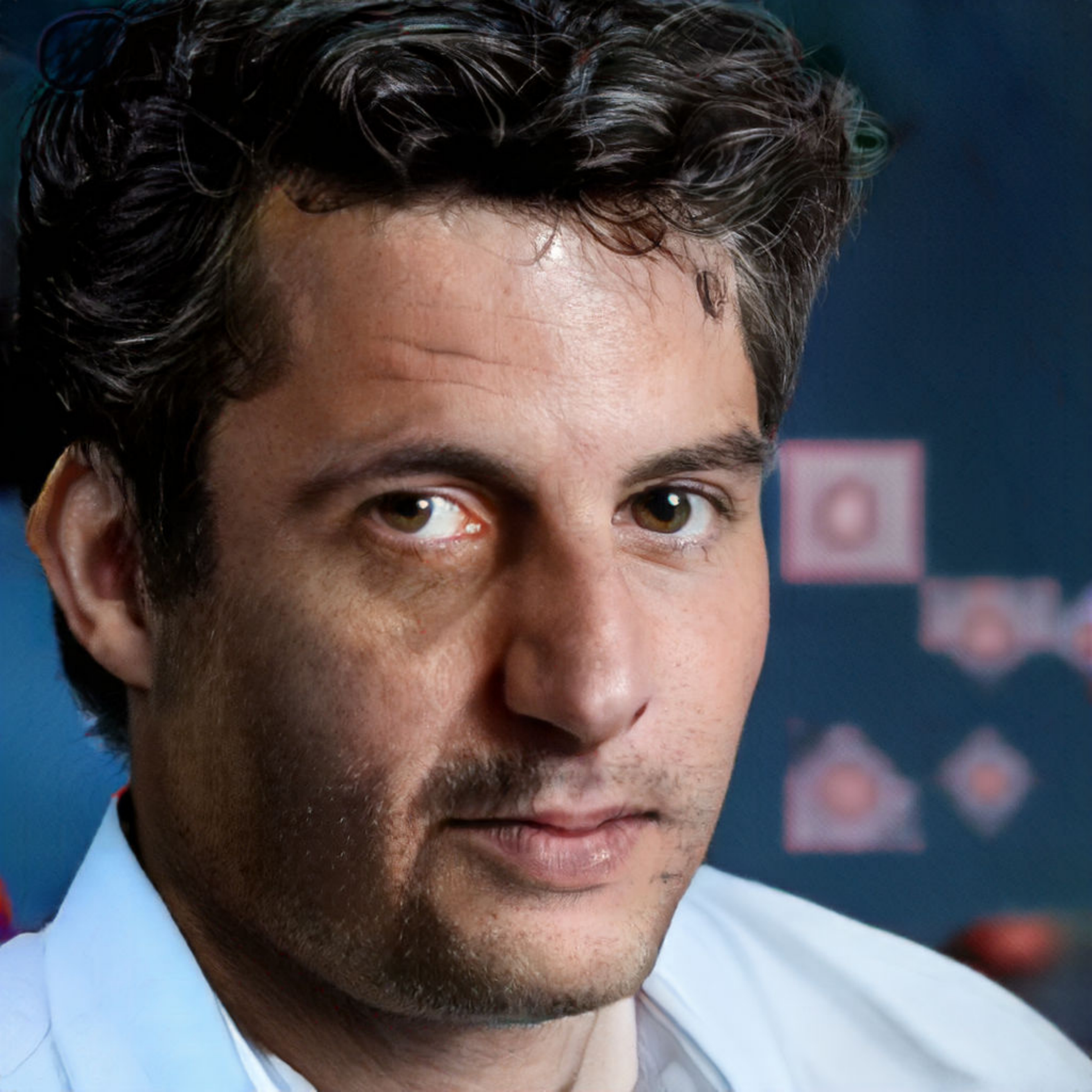}\\
    \end{tabular}
    }
    \caption{Latent interpolation results of the left-most and the right-most images on FFHQ $1024\times 1024$.}
    \label{fig:latent_code_interpolation}
\end{figure*}

\noindent\textbf{Additional image samples.} We provide additional image samples generated by our StyleSwin. Figure~\ref{fig:FFHQ1024_supp} and Figure \ref{fig:CelebAHQ1024_supp} show the impressive synthetic face images of FFHQ-1024 and CelebA-HQ 1024 with diverse viewpoints, backgrounds, and accessories, which illustrate the strong capacity of the proposed StyleSwin. Image samples of LSUN Church 256 and LSUN Car 256 are shown in Figure~\ref{fig:LSUNChurch256_supp} and Figure~\ref{fig:LSUNCar256_supp}, showing that our StyleSwin is capable to synthesize complex scenes with coherent structures and complicated materials with high-quality light effects.

\section{Responsible AI Considerations}

Our work does not directly modify the exiting images which may alter the identity or expression of the people. We discourage the use of our work in such applications as it is not designed to do so. We have quantitatively verified that the proposed method does not show evident disparity, on gender and ages as the model mostly follows the dataset distribution, however, we encourage additional care if you intend to use the system on certain demographic groups. We also encourage use of fair and representative data when training on customized data. We caution that the high-resolution images produced by our model may potentially be misused for impersonating humans and viable solutions so avoid this include adding tags or watermarks when distributing the generated photos.

\section{Discussion of Limitation}

Although, as stated in the main article, StyleSwin's theoretical FLOPs are smaller than StyleGAN2, there is a gap between the theoretical FLOPs and the throughput in
practice. The throughput of StyleGAN2 and StyleSwin are 40.05 imgs/sec and 11.05 imgs/sec respectively on a single V100 GPU. This is primarily due to the fact that vision transformers have not been sufficiently optimized as ConvNets (e.g. using CuDNN), and we believe future optimization will democratize the usage of transformers as they exhibit lower theoretical FLOPs. Besides, bCR is not effective on 1024 × 1024, which we leave for further study.

\newpage
\begin{figure*}[h]
    \center
    \small
    \setlength\tabcolsep{0pt}
    \renewcommand{\arraystretch}{0}
    {
    \begin{tabular}{@{}cccc@{}}
         \includegraphics[width=0.24\columnwidth]{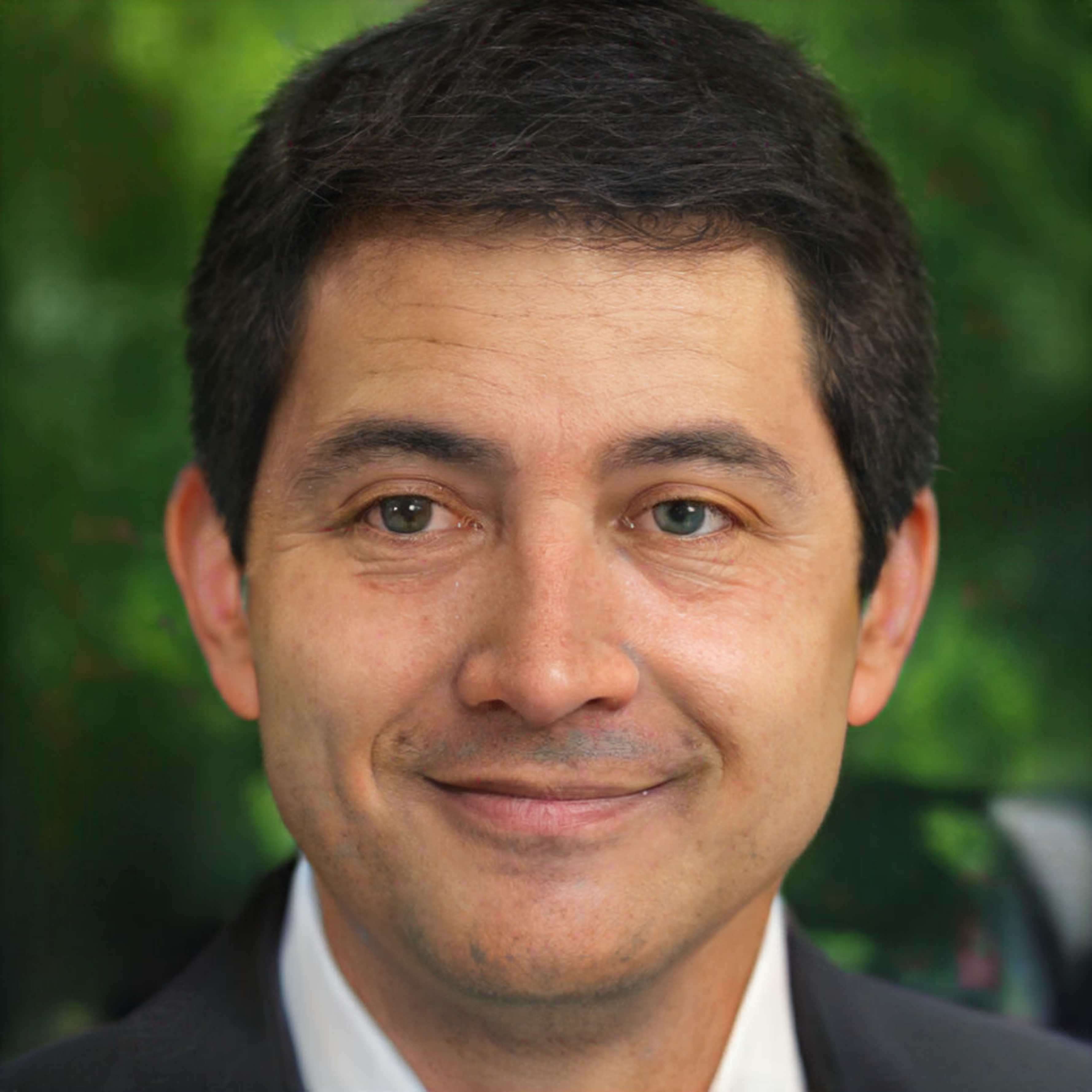} & \includegraphics[width=0.24\columnwidth]{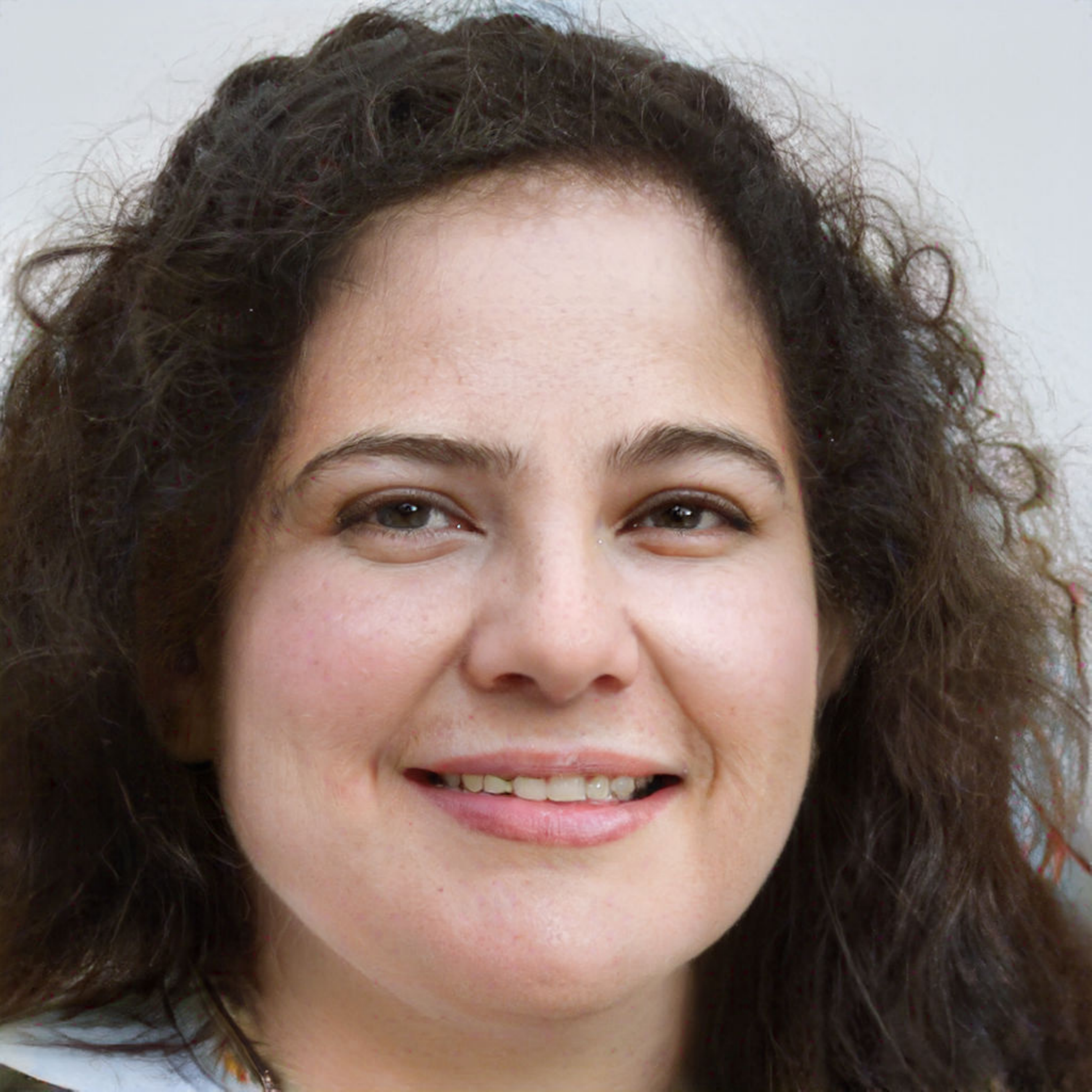} &
         \includegraphics[width=0.24\columnwidth]{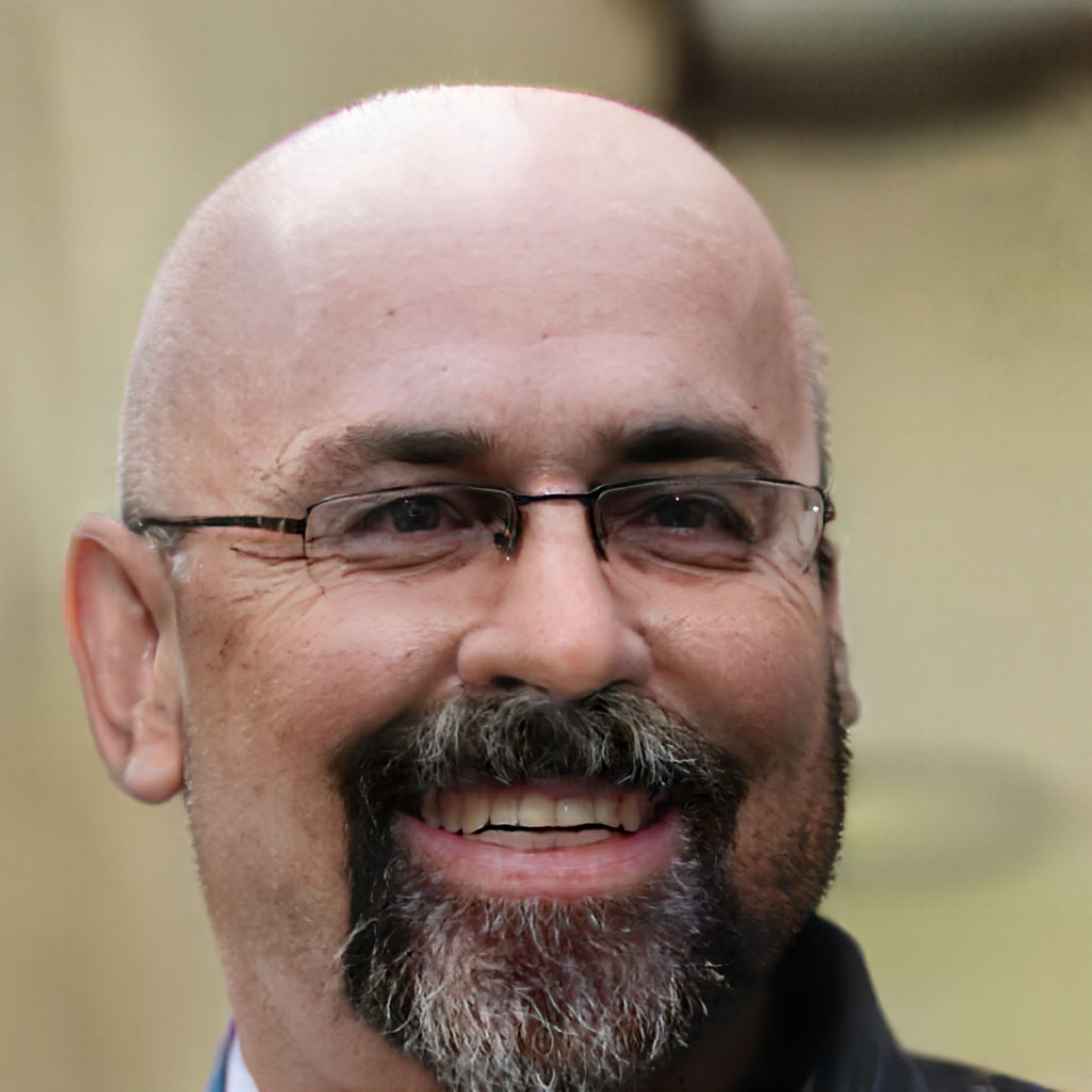} & \includegraphics[width=0.24\columnwidth]{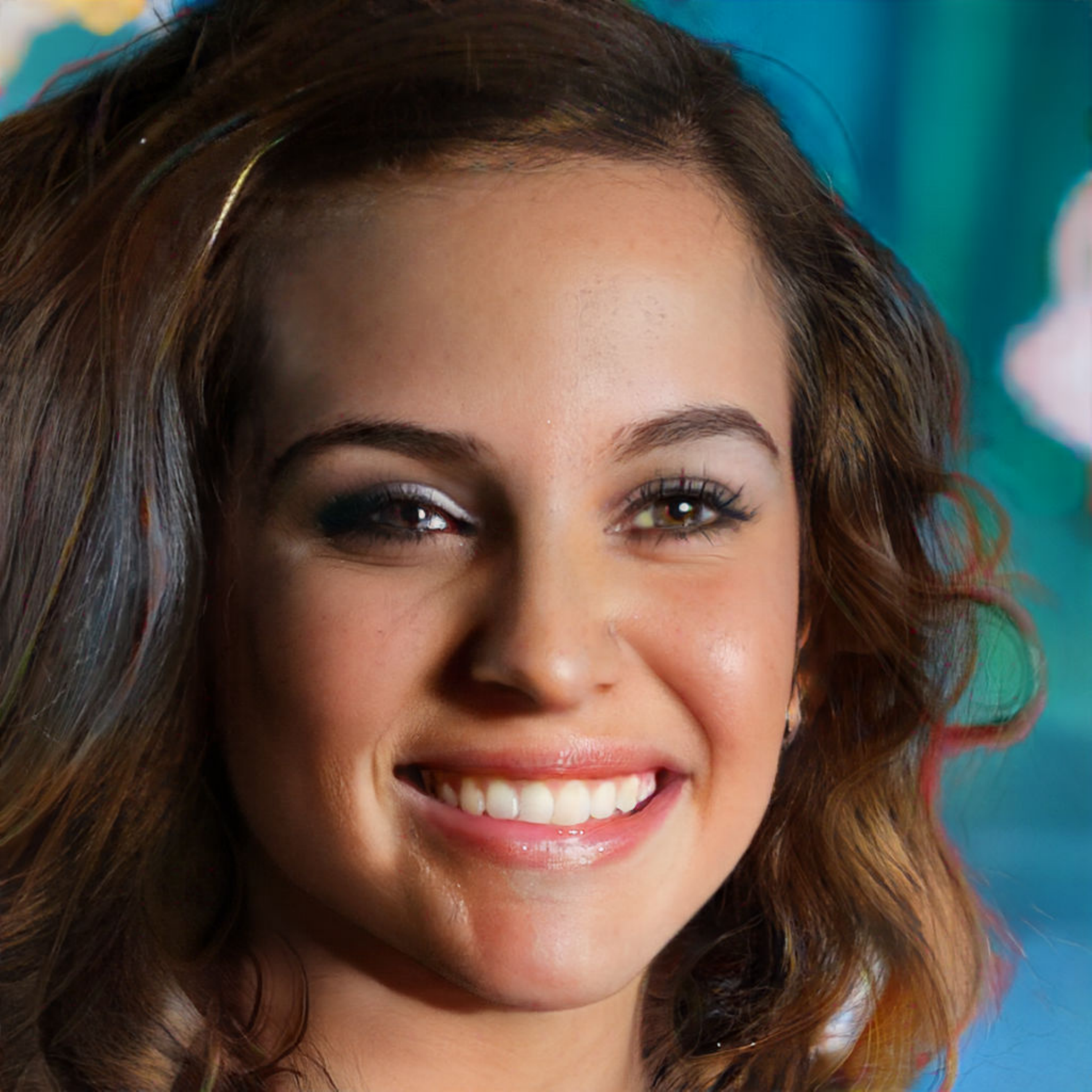}\\
         
         \includegraphics[width=0.24\columnwidth]{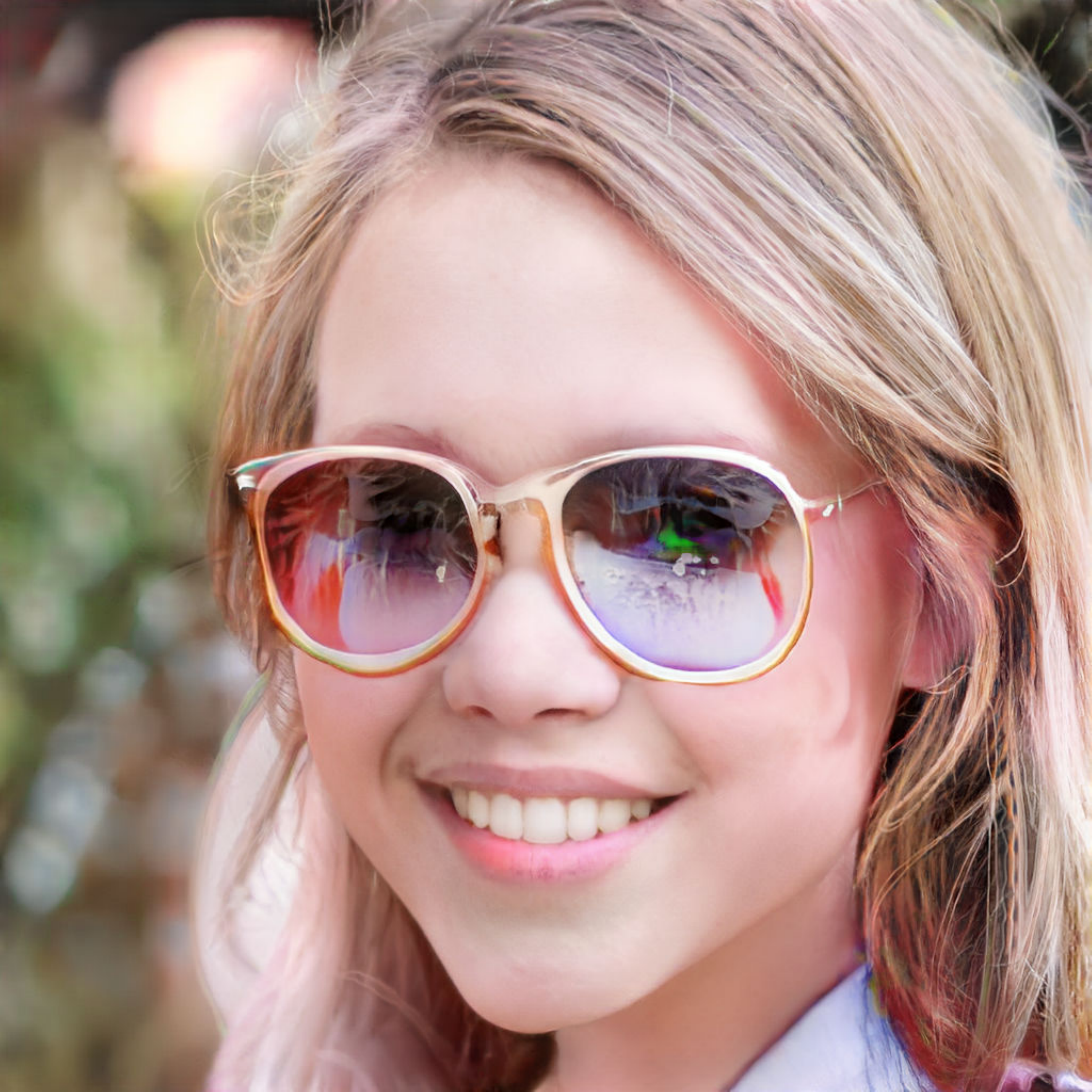} & \includegraphics[width=0.24\columnwidth]{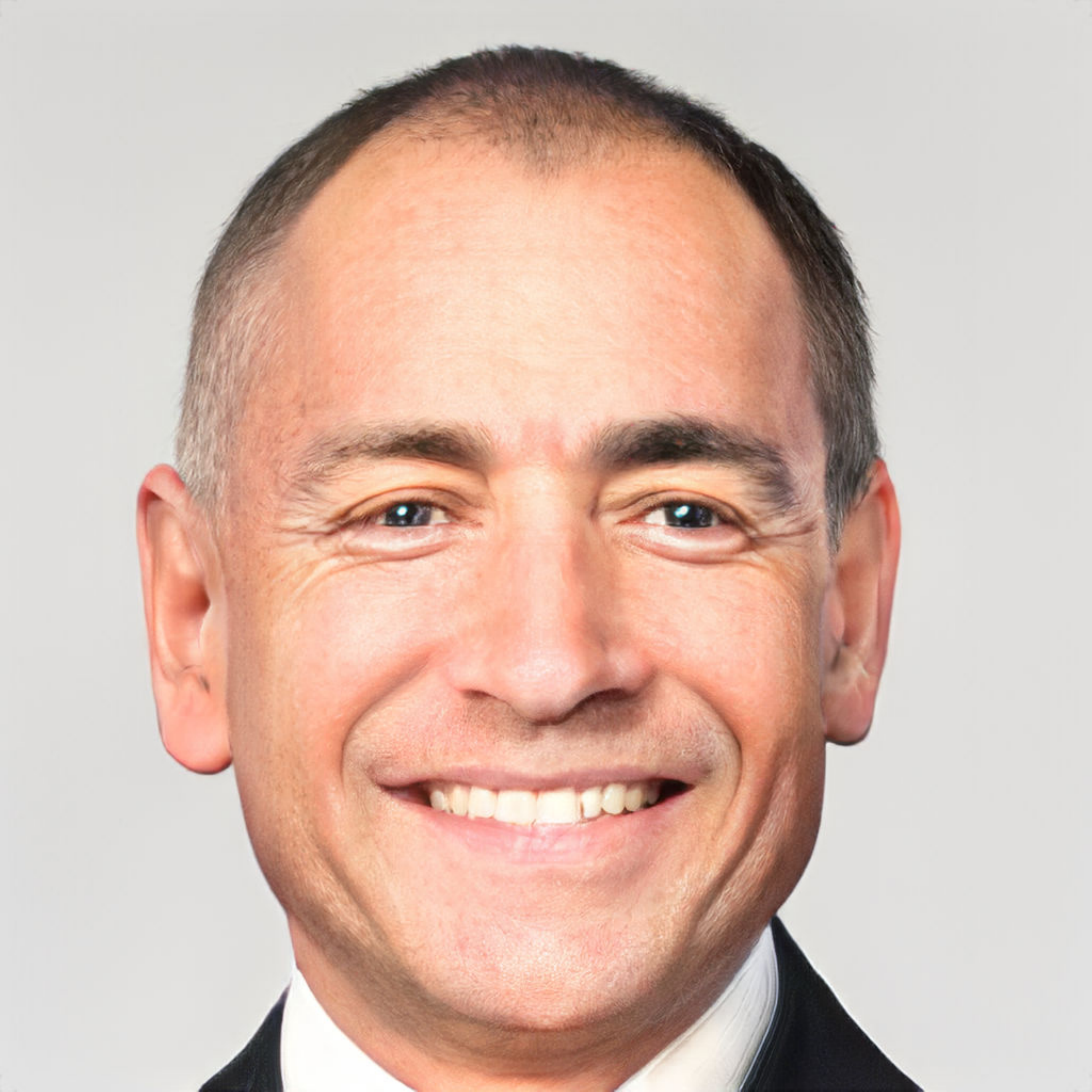} &
         \includegraphics[width=0.24\columnwidth]{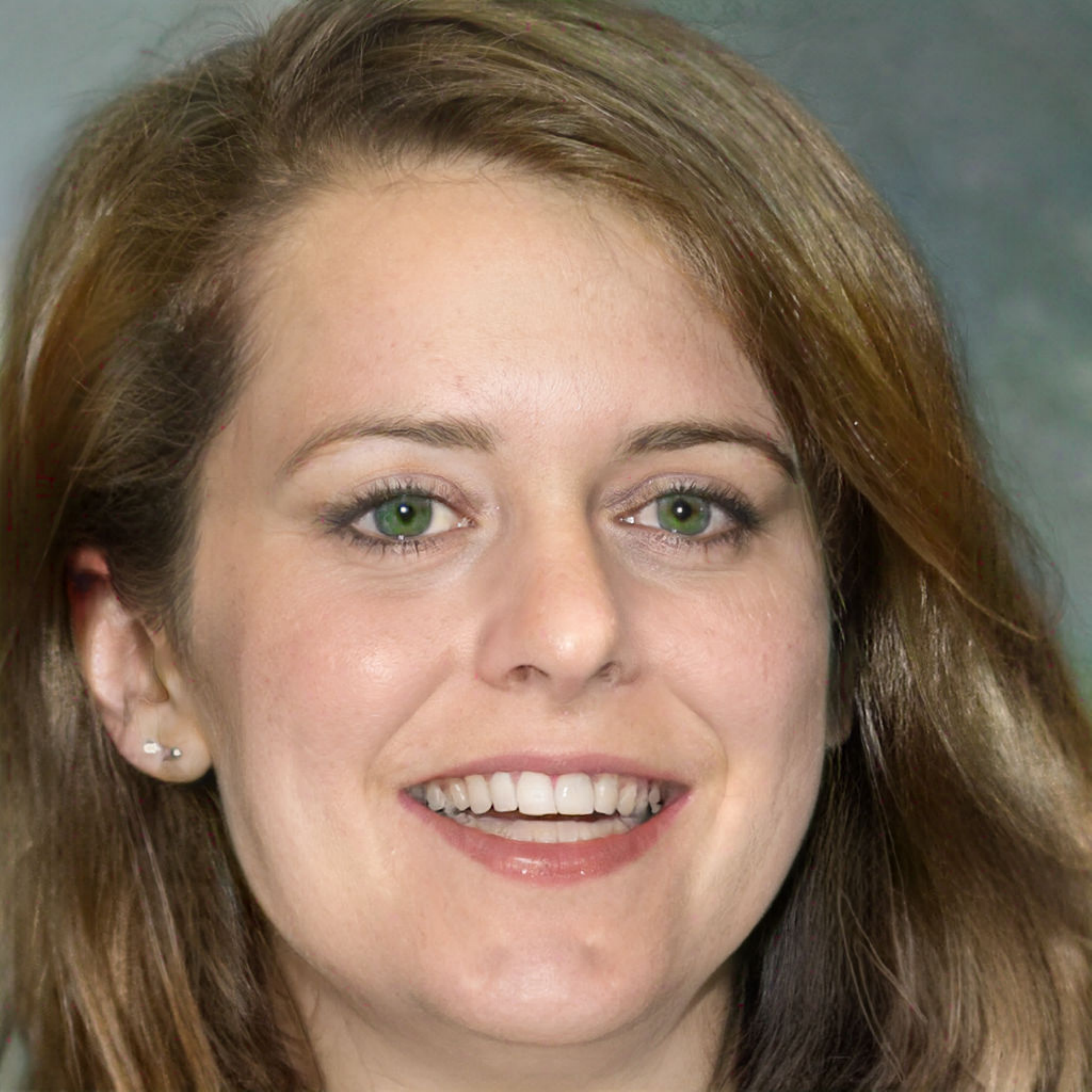} & \includegraphics[width=0.24\columnwidth]{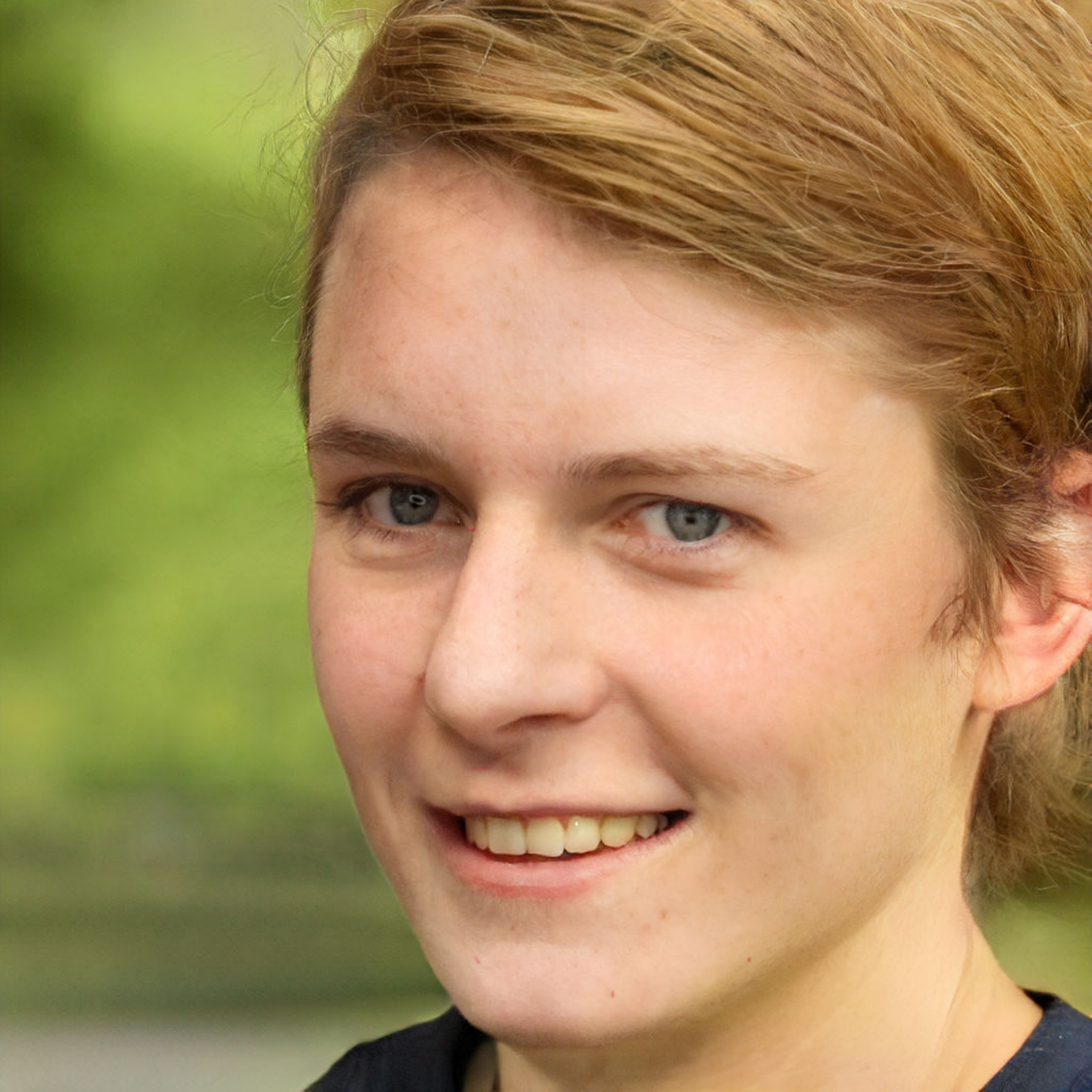}\\
         
         \includegraphics[width=0.24\columnwidth]{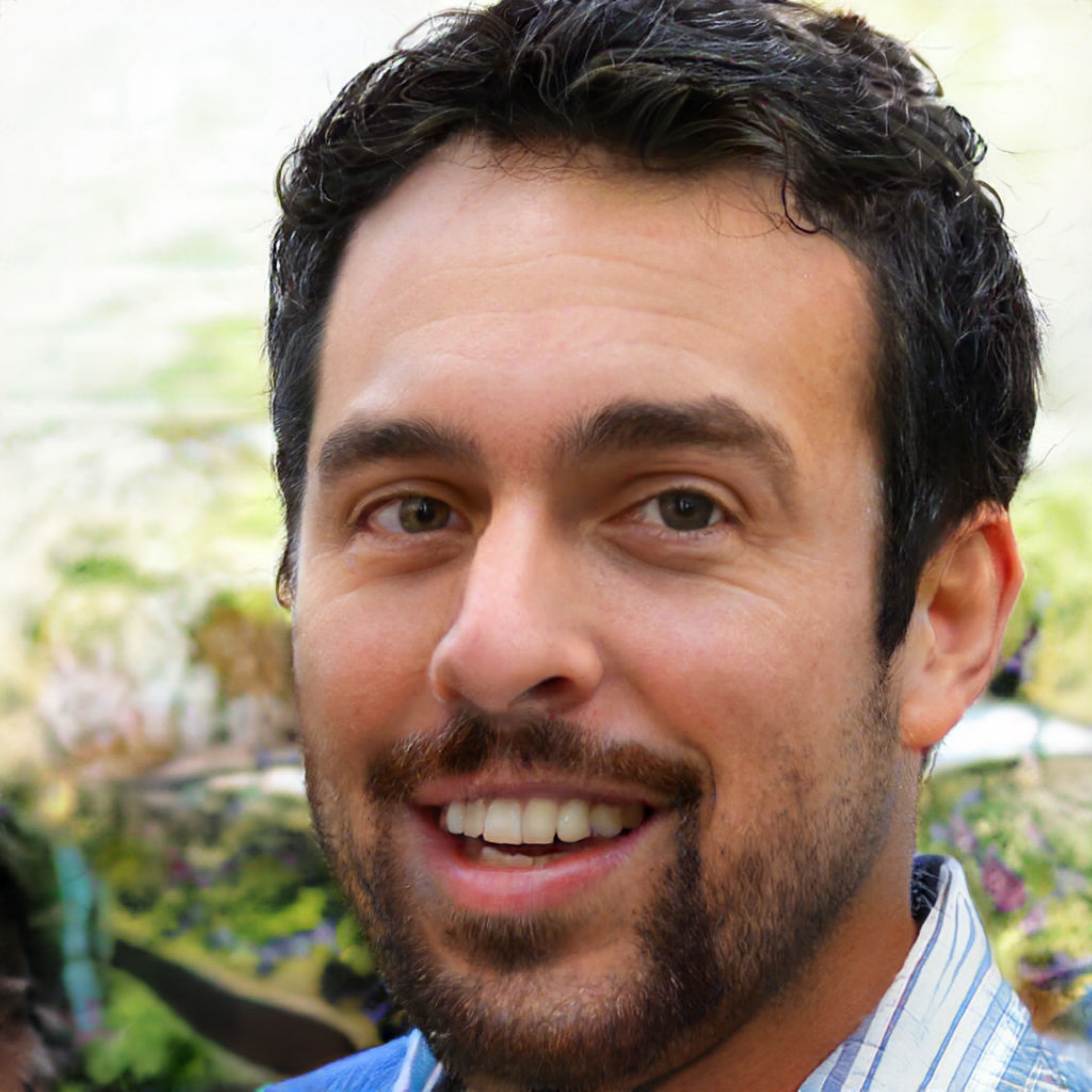} & \includegraphics[width=0.24\columnwidth]{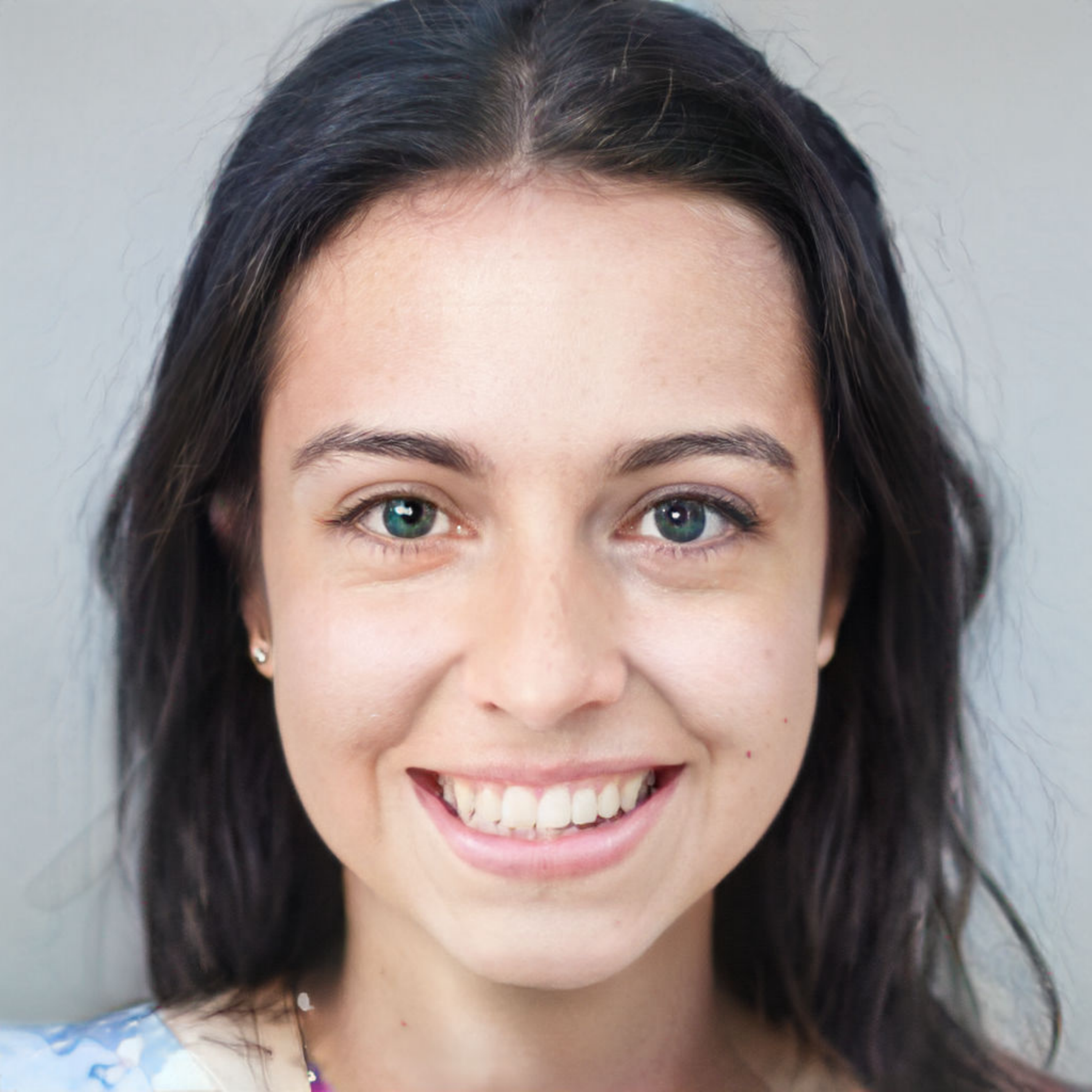} &
         \includegraphics[width=0.24\columnwidth]{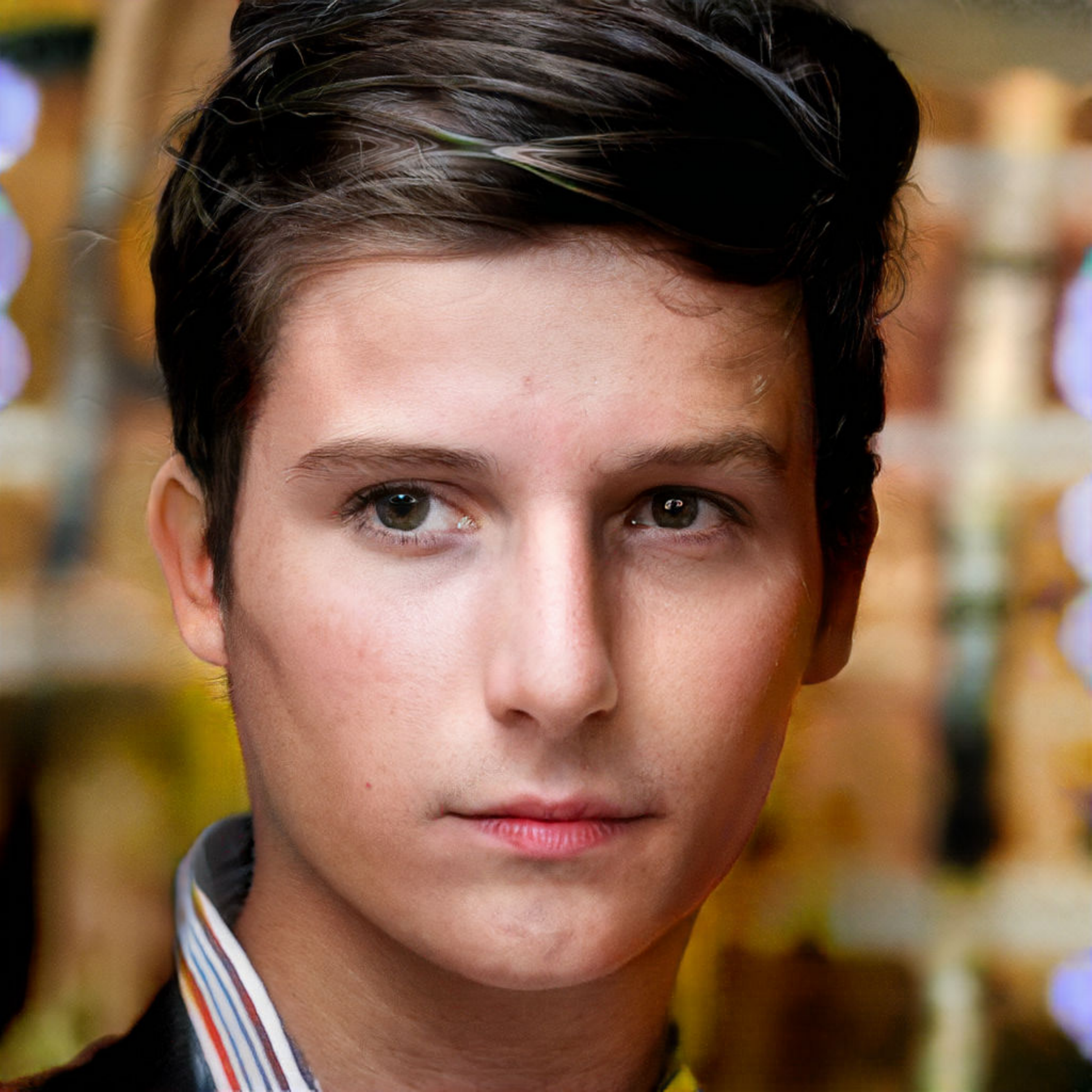} & \includegraphics[width=0.24\columnwidth]{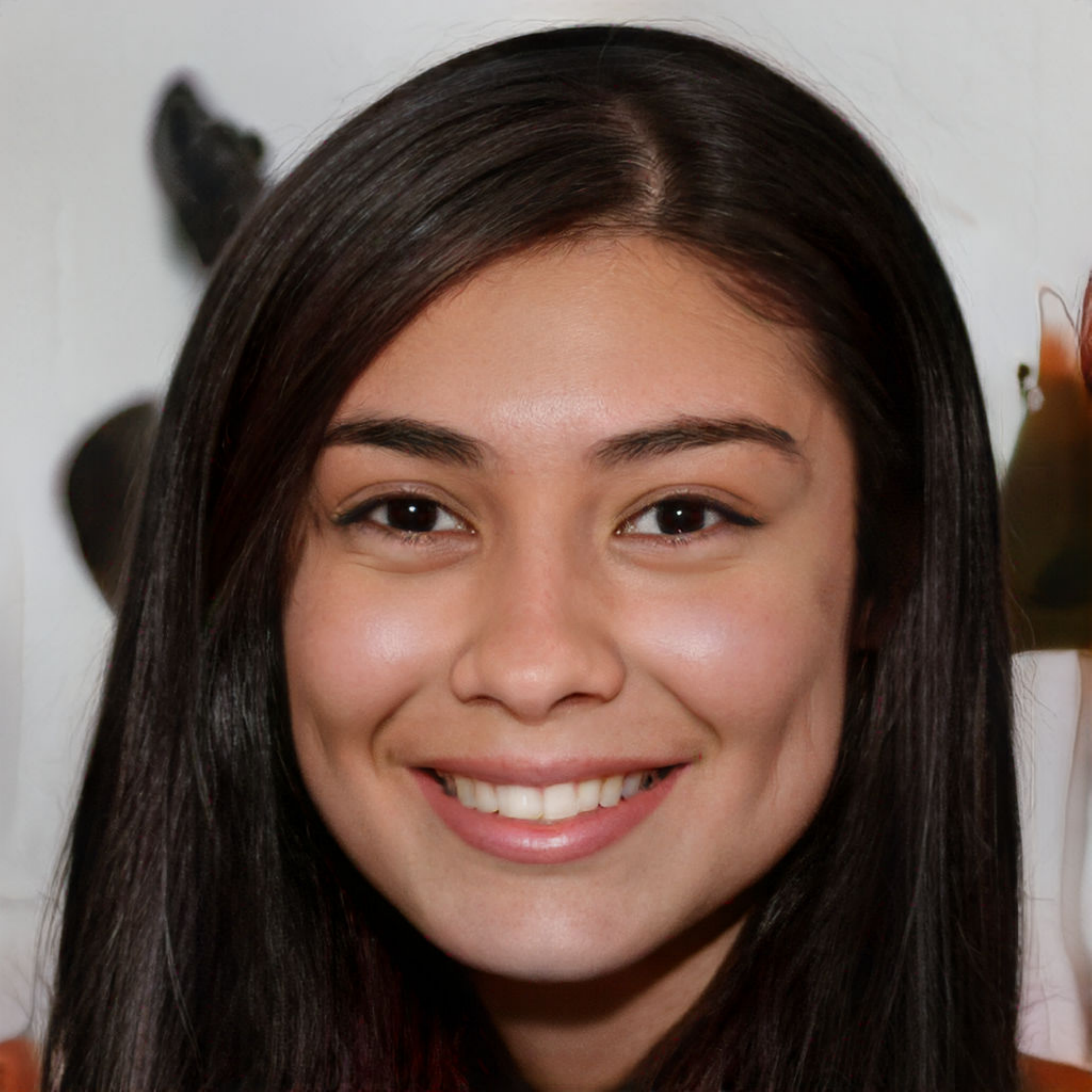}\\
         
         \includegraphics[width=0.24\columnwidth]{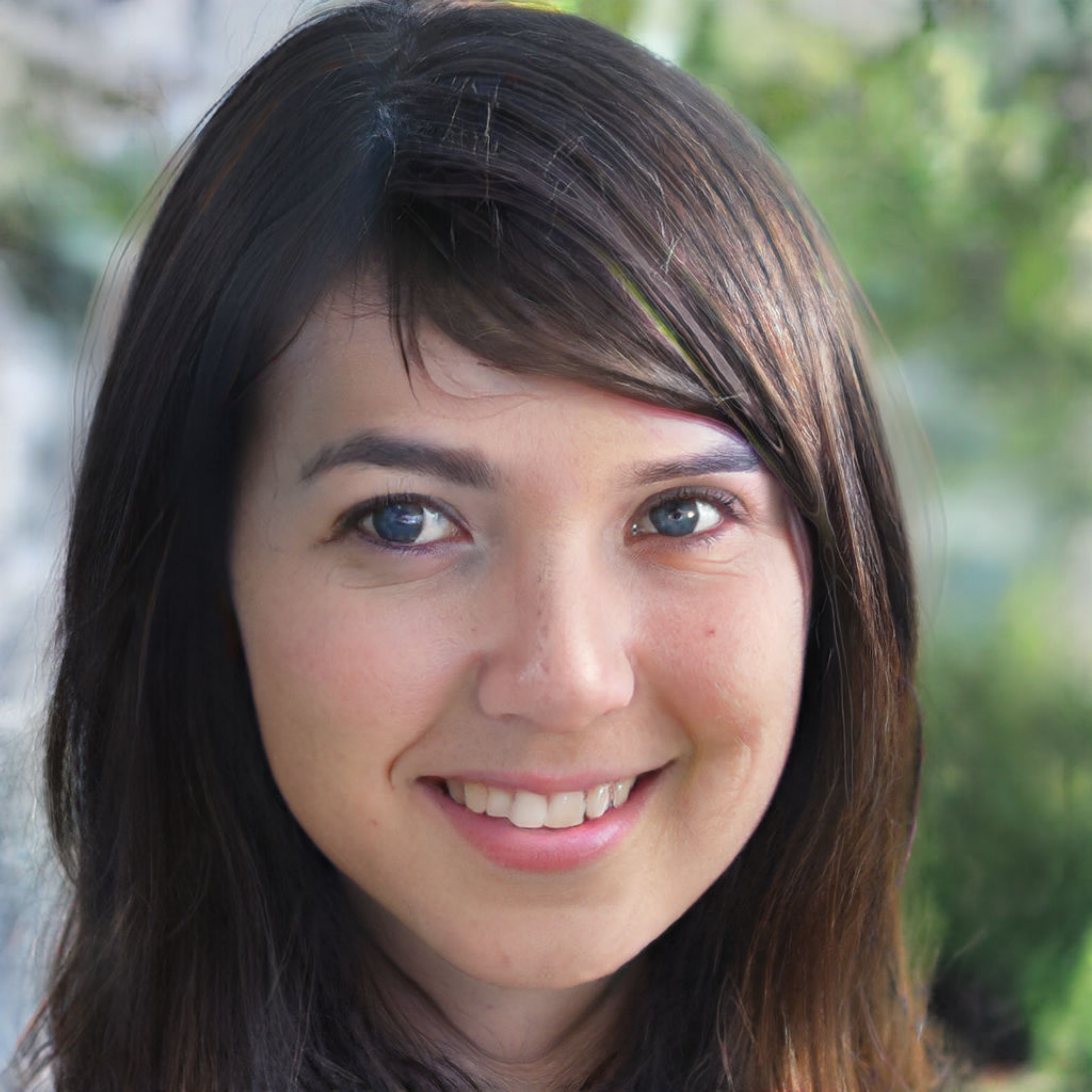} & \includegraphics[width=0.24\columnwidth]{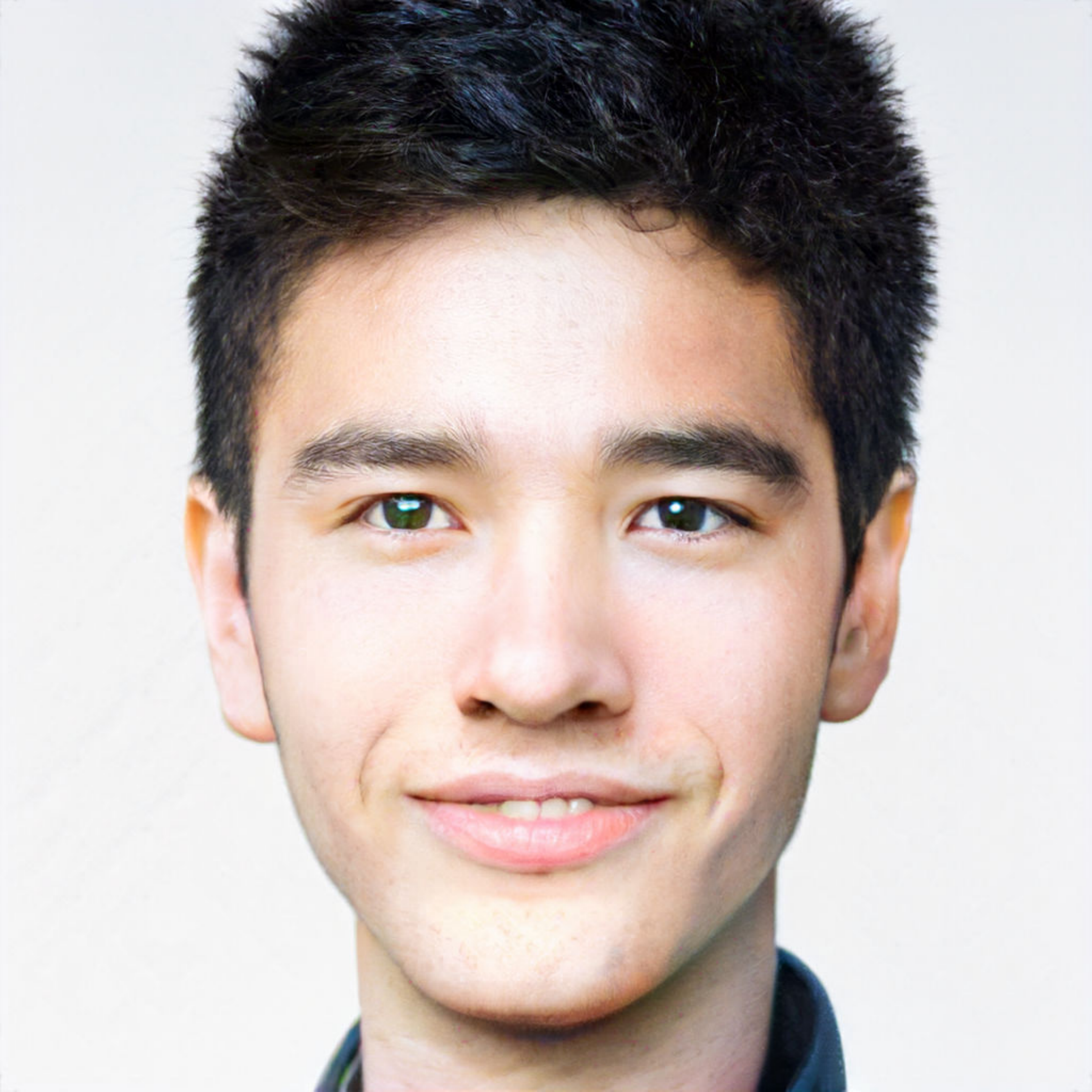} &
         \includegraphics[width=0.24\columnwidth]{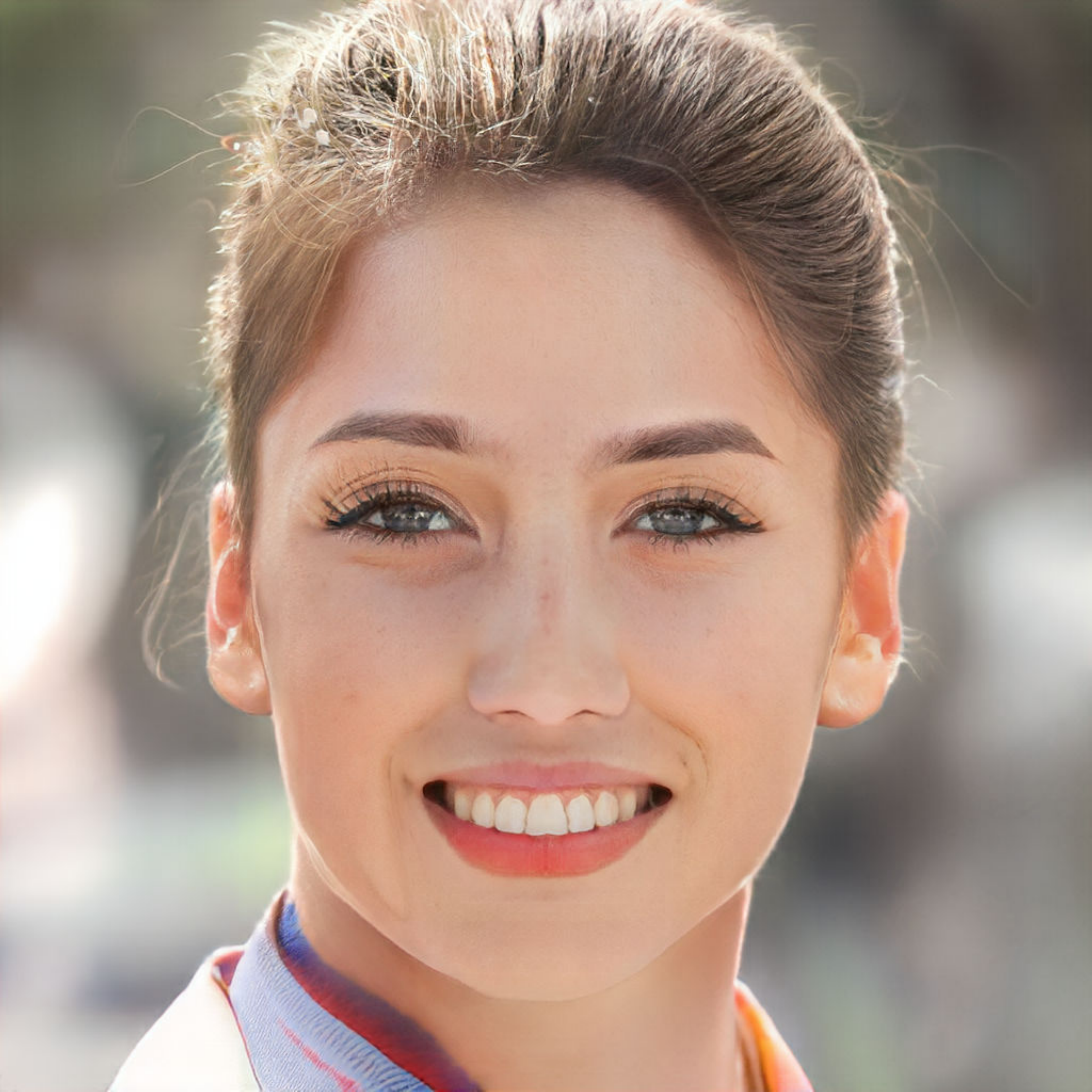} & \includegraphics[width=0.24\columnwidth]{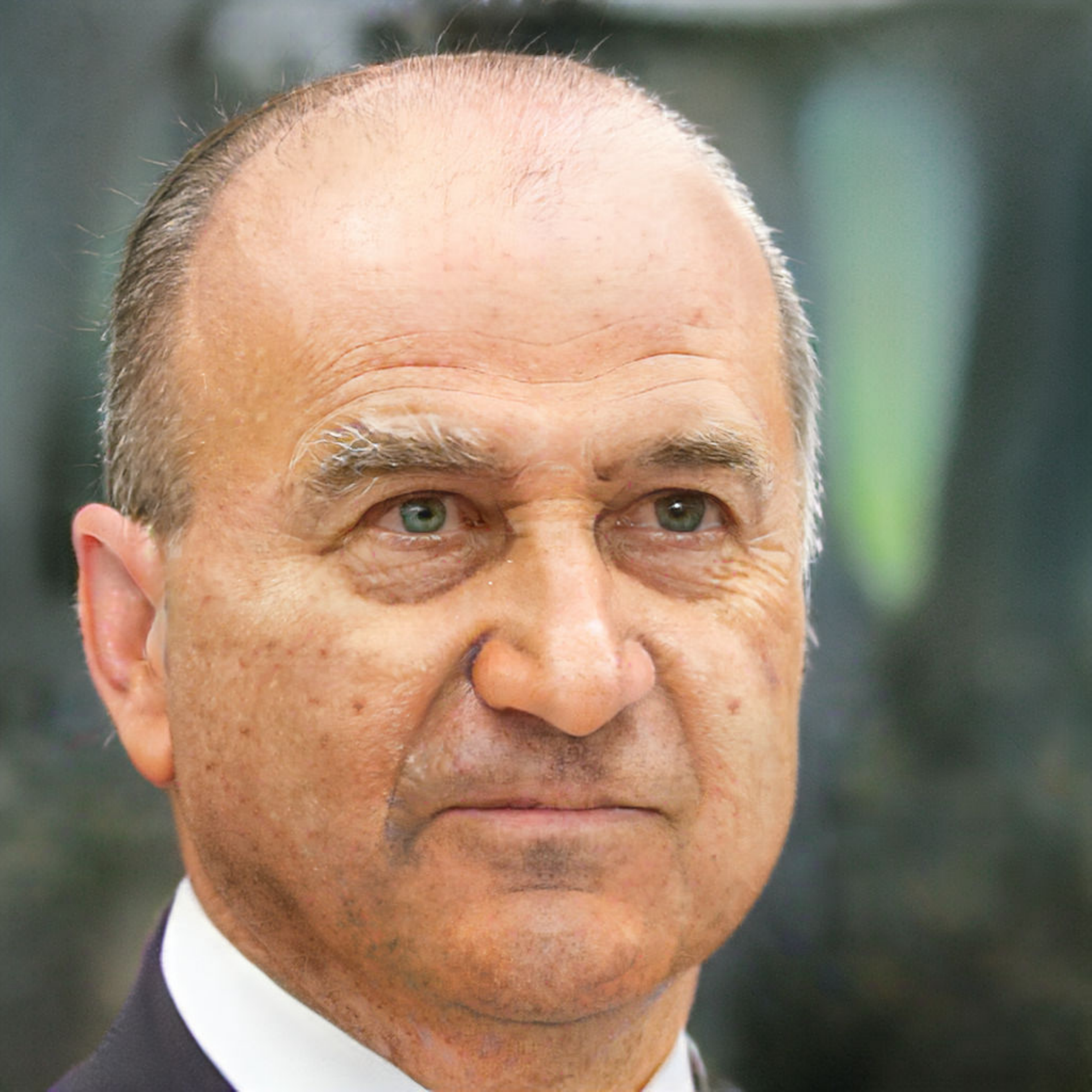}\\
         
         \includegraphics[width=0.24\columnwidth]{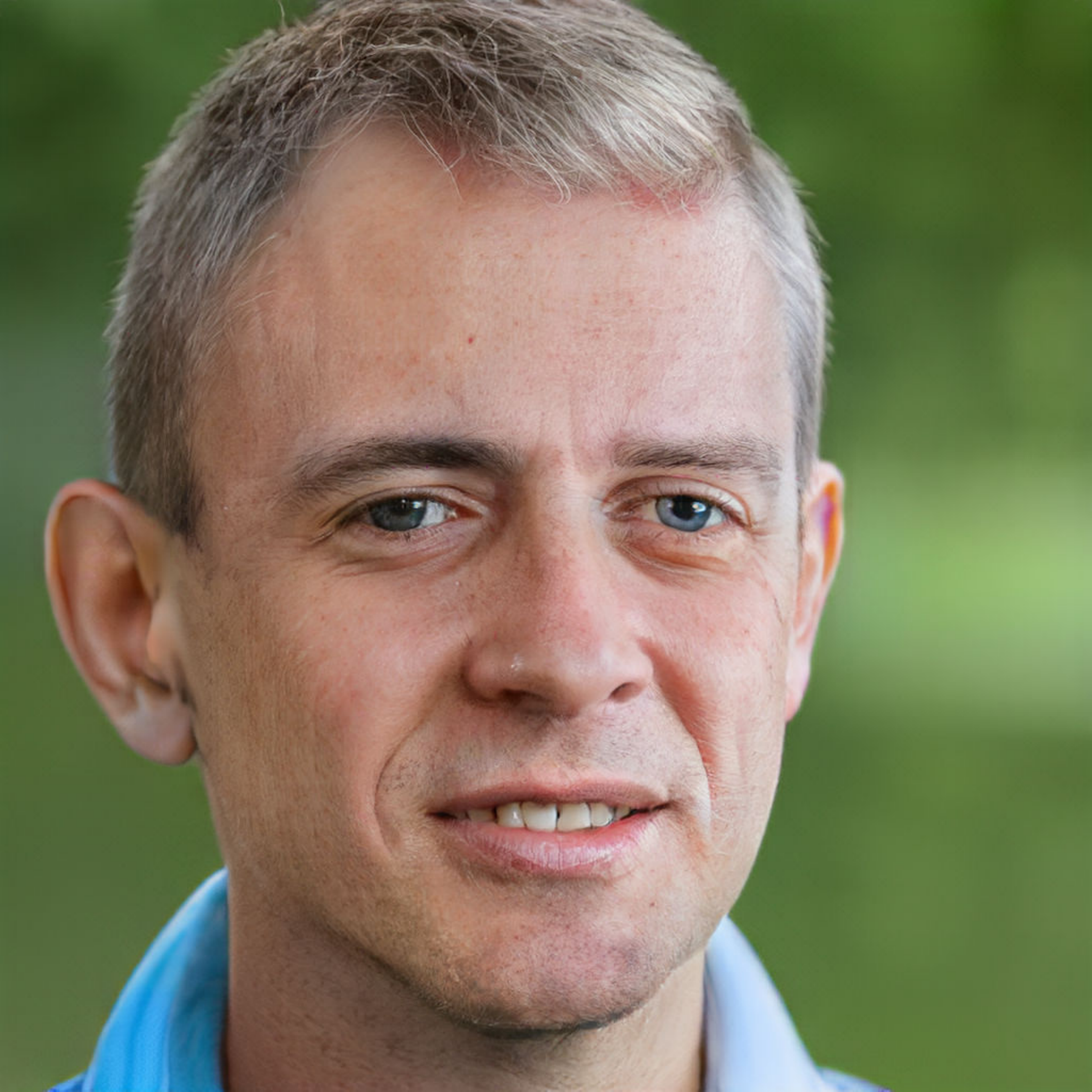} & \includegraphics[width=0.24\columnwidth]{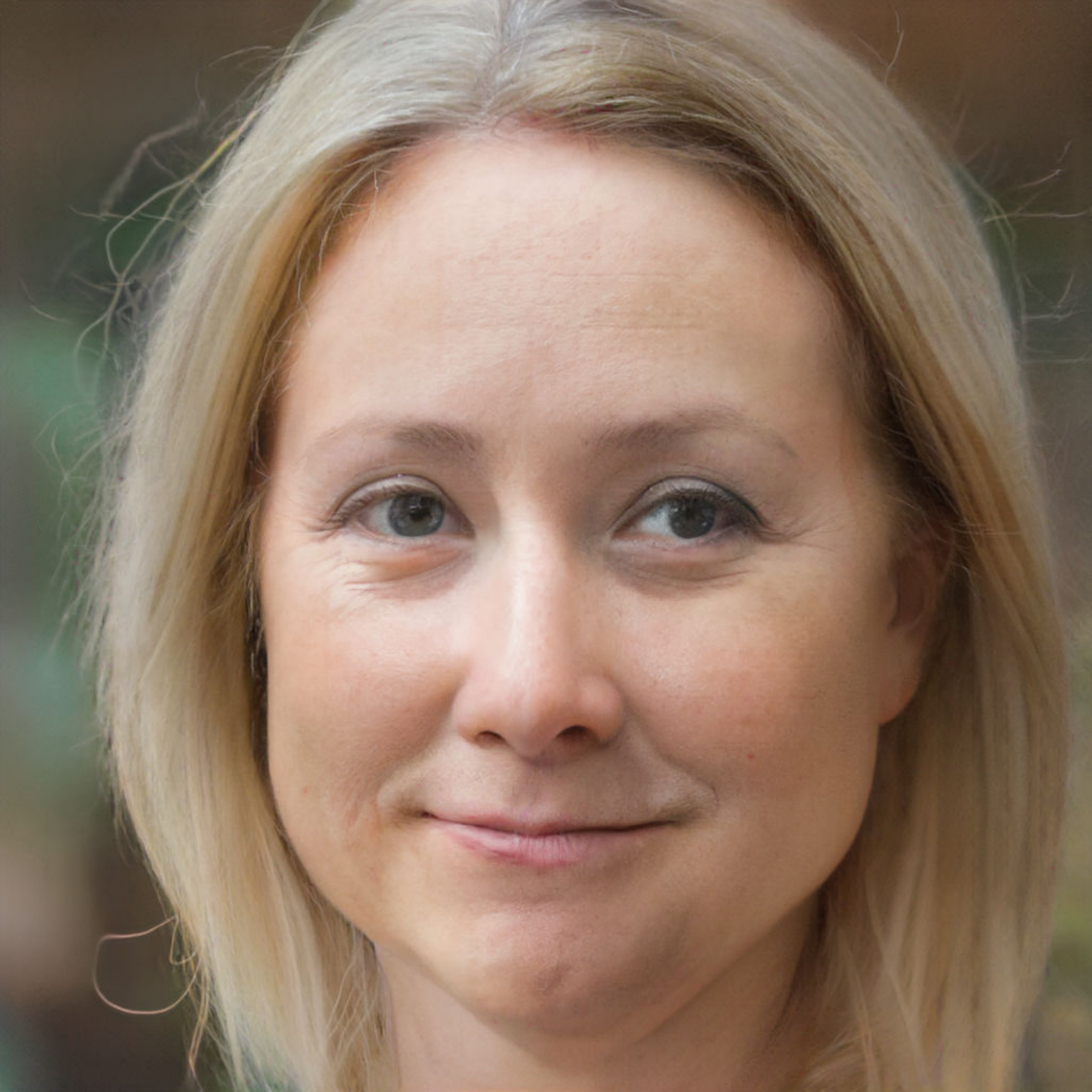} &
         \includegraphics[width=0.24\columnwidth]{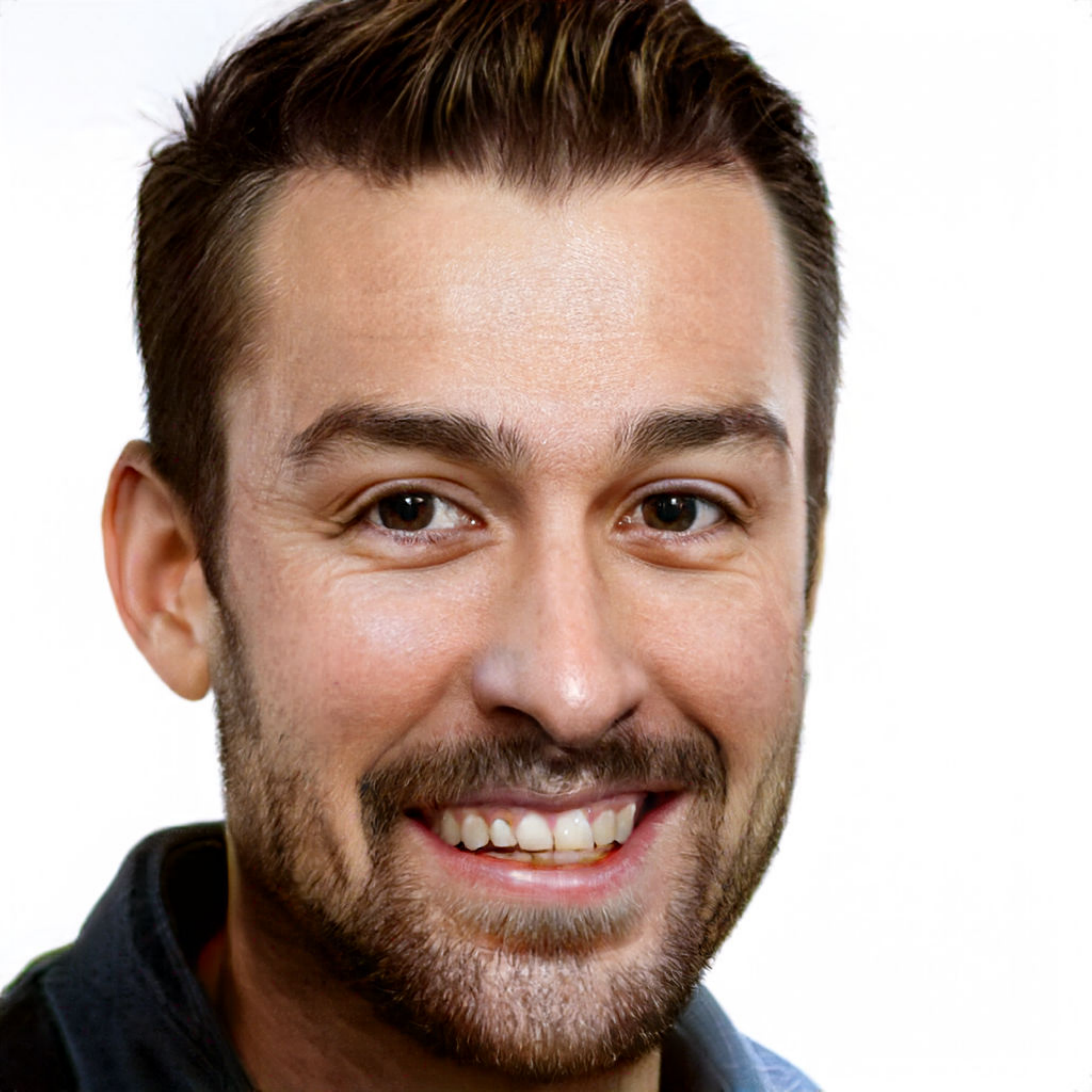} & \includegraphics[width=0.24\columnwidth]{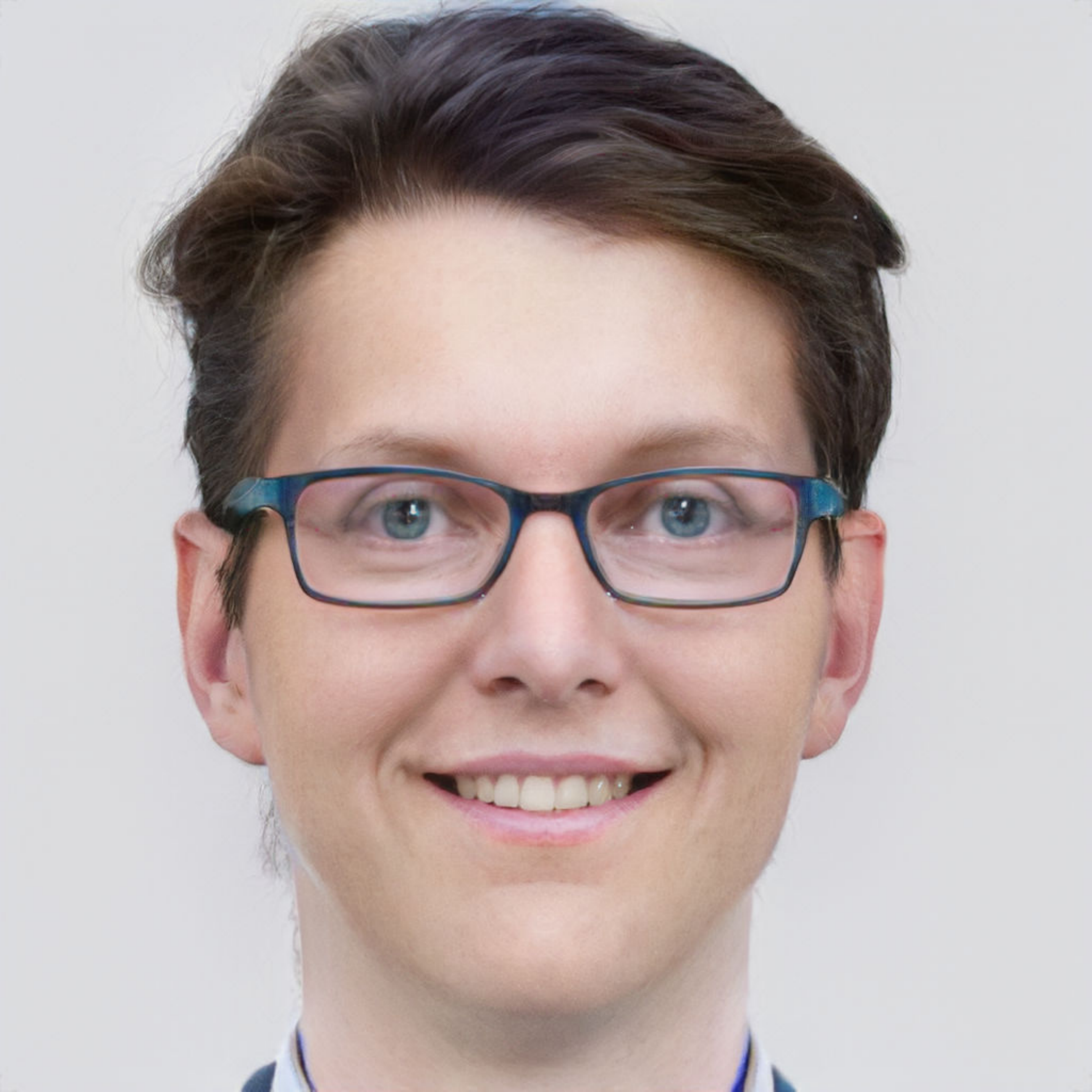}\\
    \end{tabular}
    }
    \caption{Image samples of FFHQ $1024\times 1024$.}
    \label{fig:FFHQ1024_supp}
\end{figure*}

\newpage
\begin{figure*}[h]
    \center
    \small
    \setlength\tabcolsep{0pt}
    \renewcommand{\arraystretch}{0}
    {
    \begin{tabular}{@{}cccc@{}}
         \includegraphics[width=0.24\columnwidth]{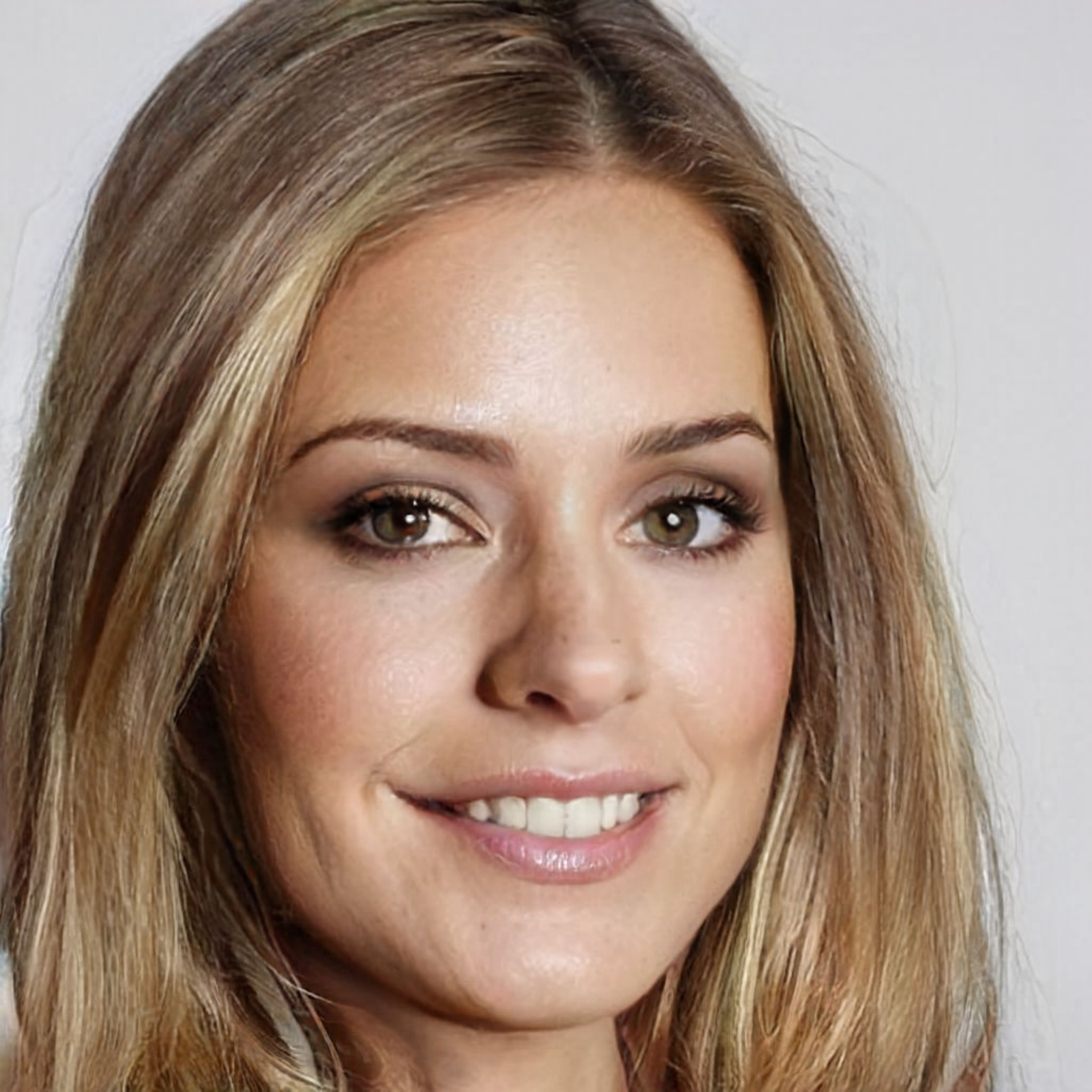} & \includegraphics[width=0.24\columnwidth]{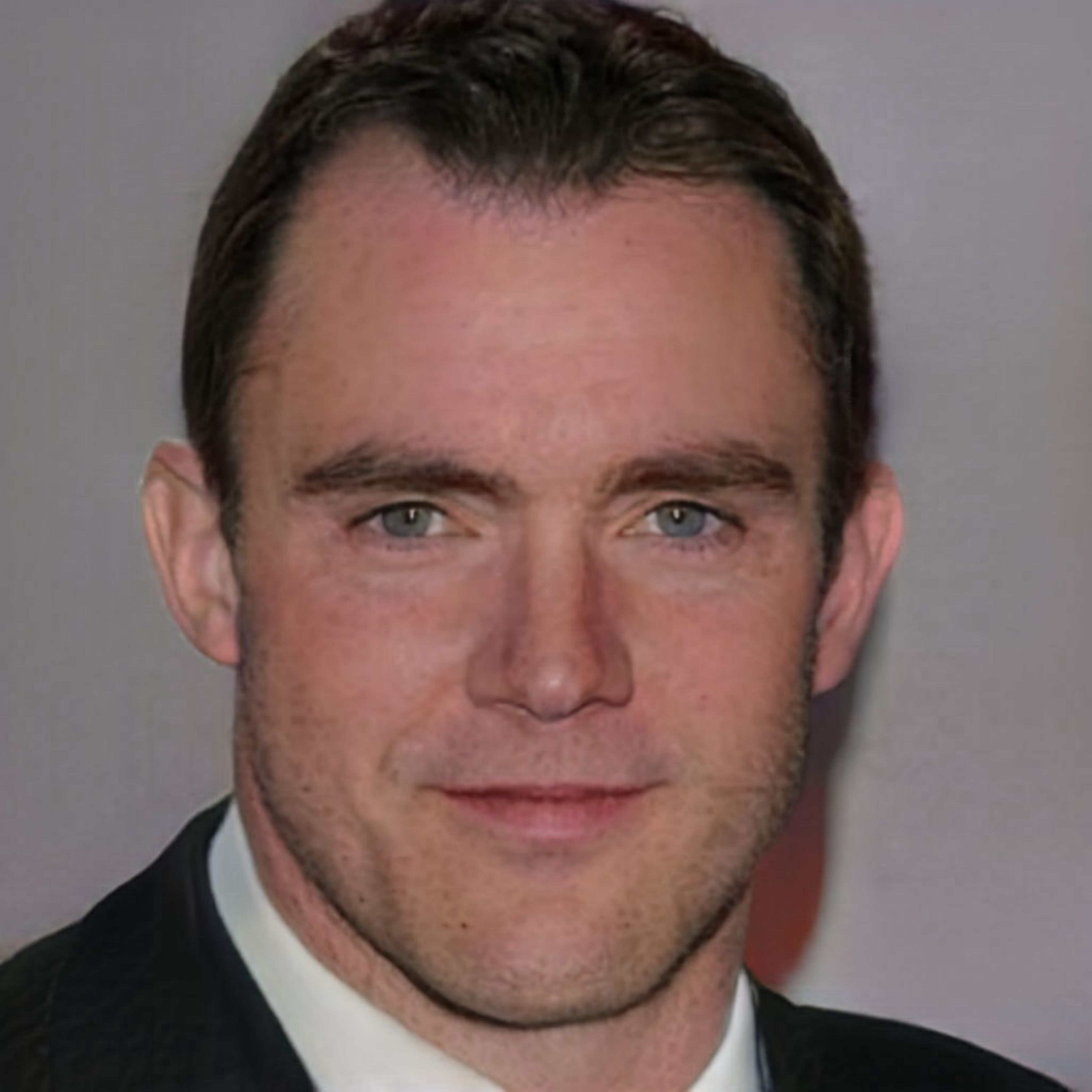} &
         \includegraphics[width=0.24\columnwidth]{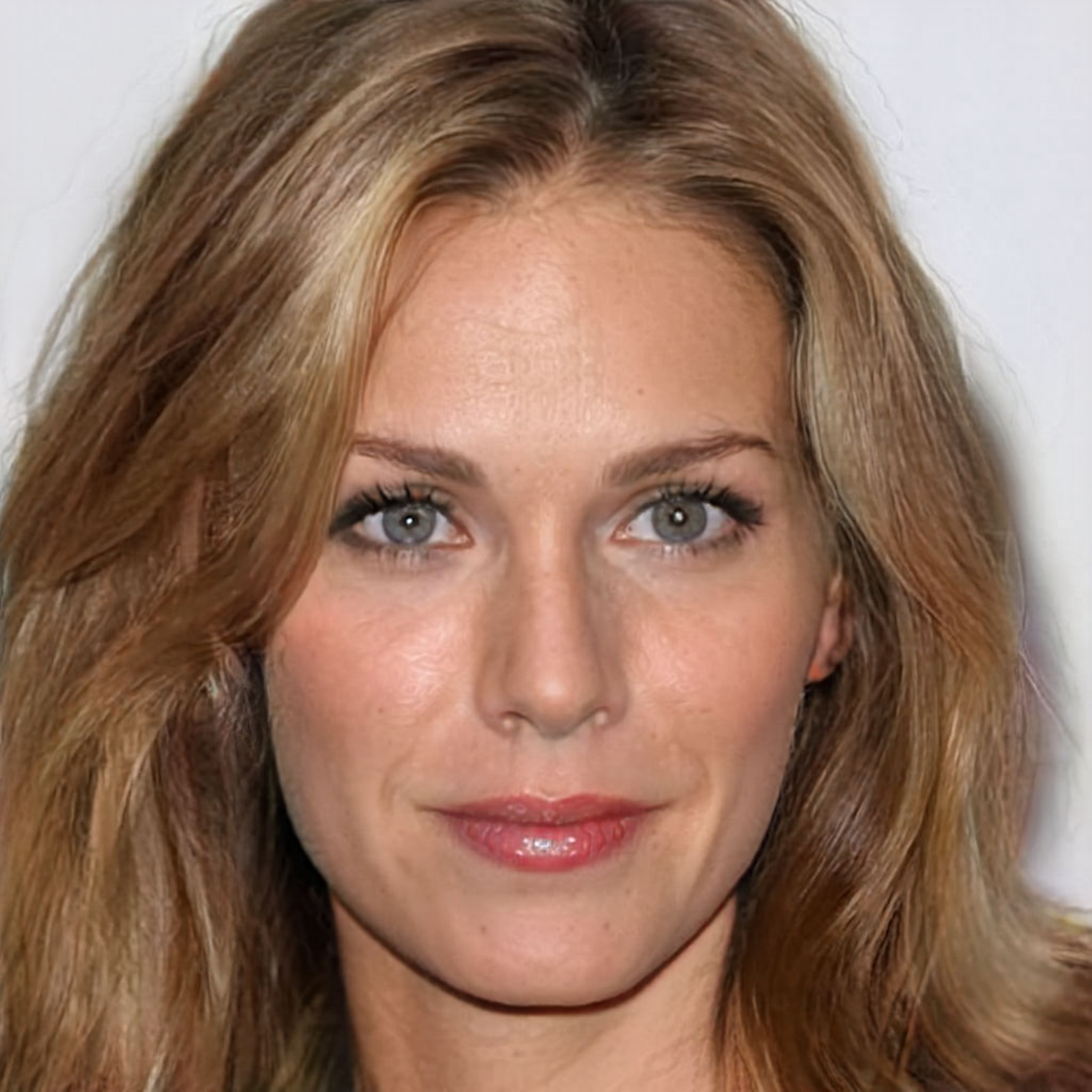} & \includegraphics[width=0.24\columnwidth]{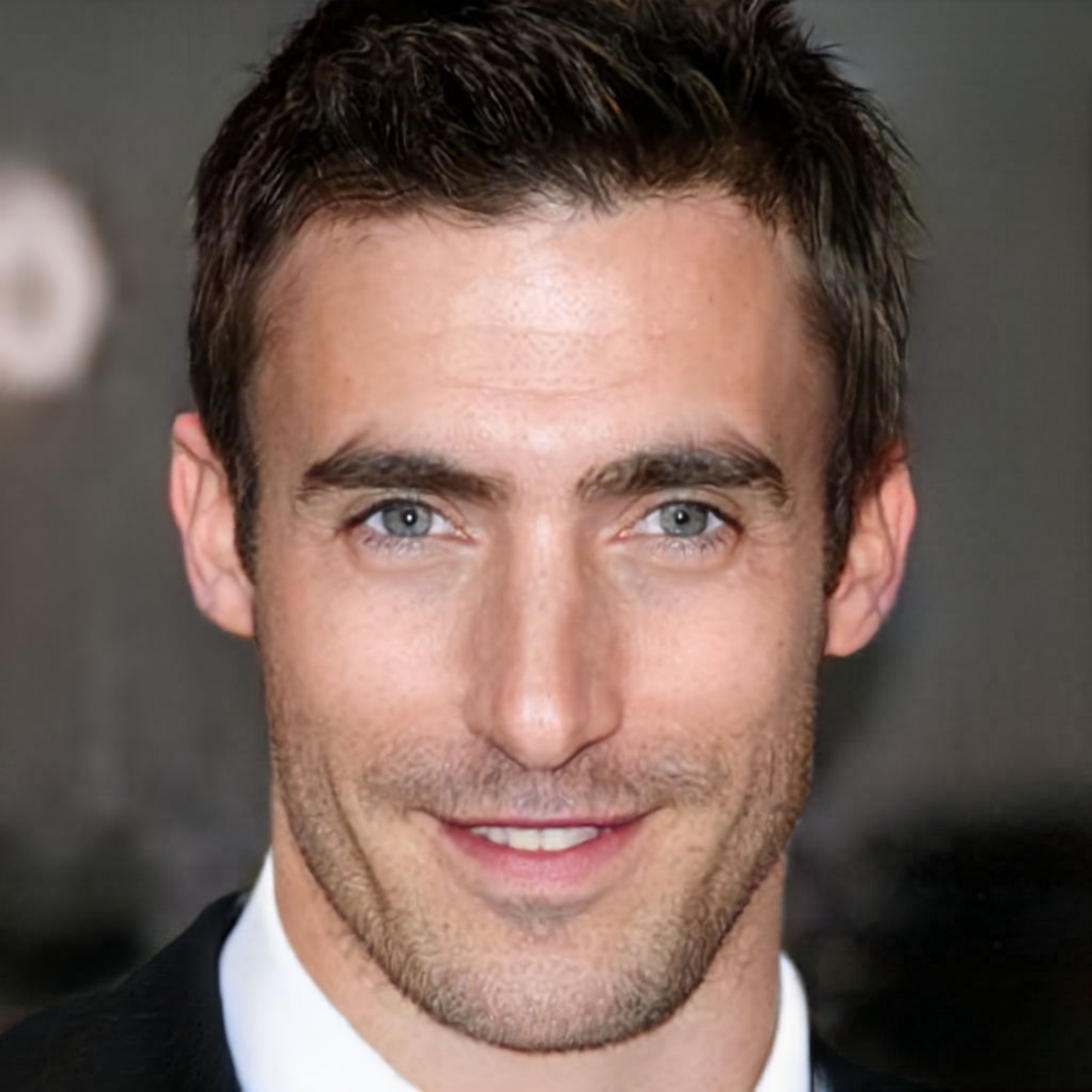}\\
         \includegraphics[width=0.24\columnwidth]{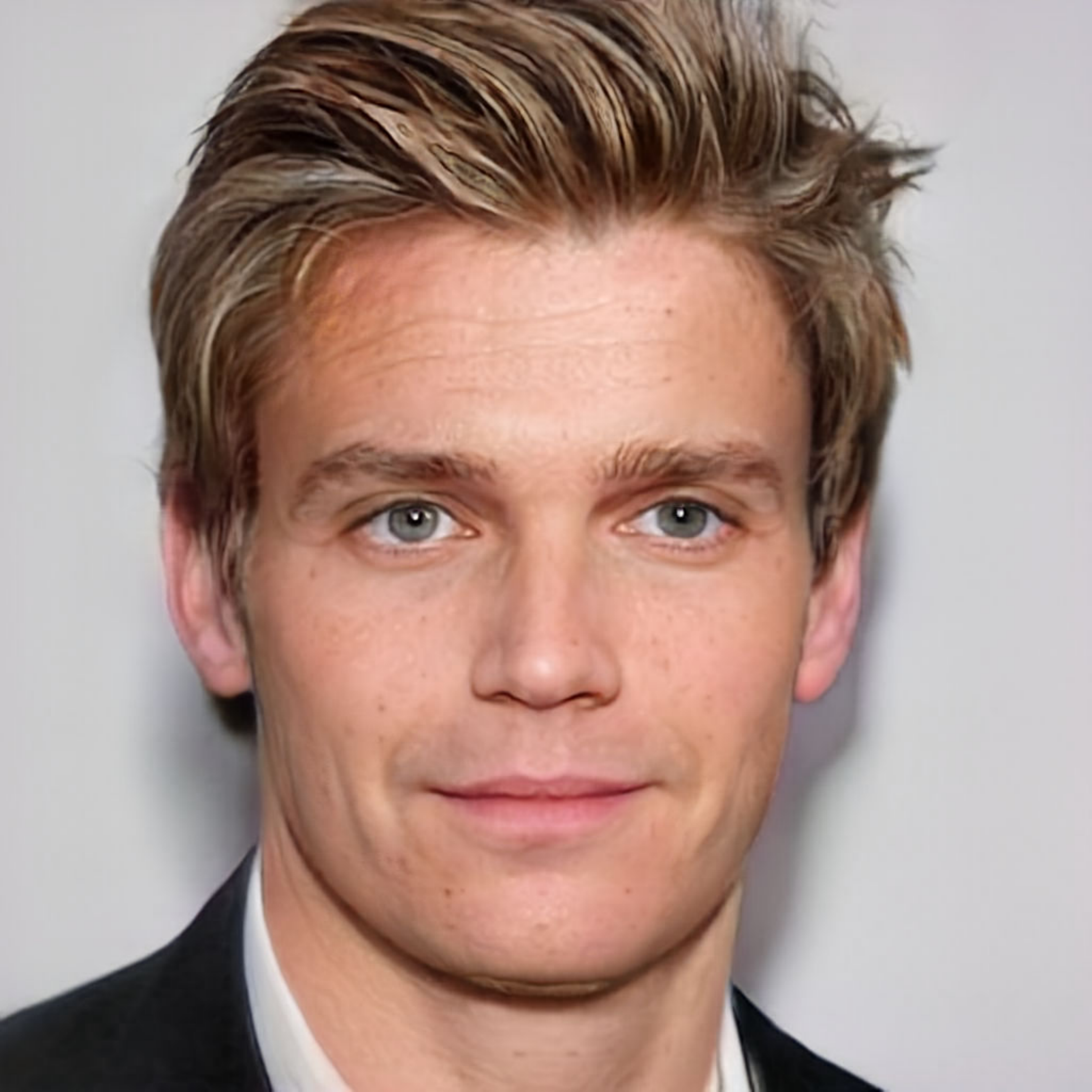} & \includegraphics[width=0.24\columnwidth]{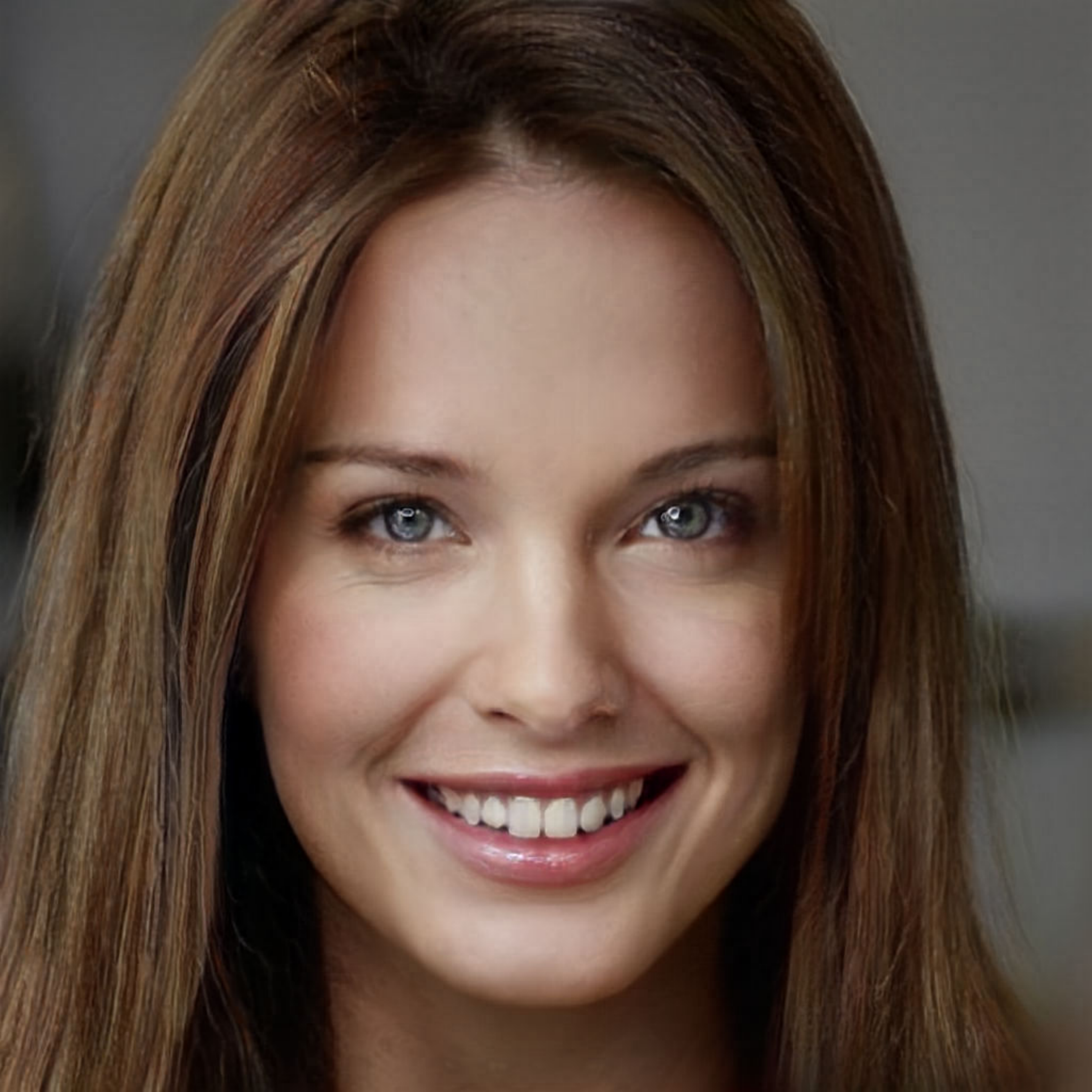} &
         \includegraphics[width=0.24\columnwidth]{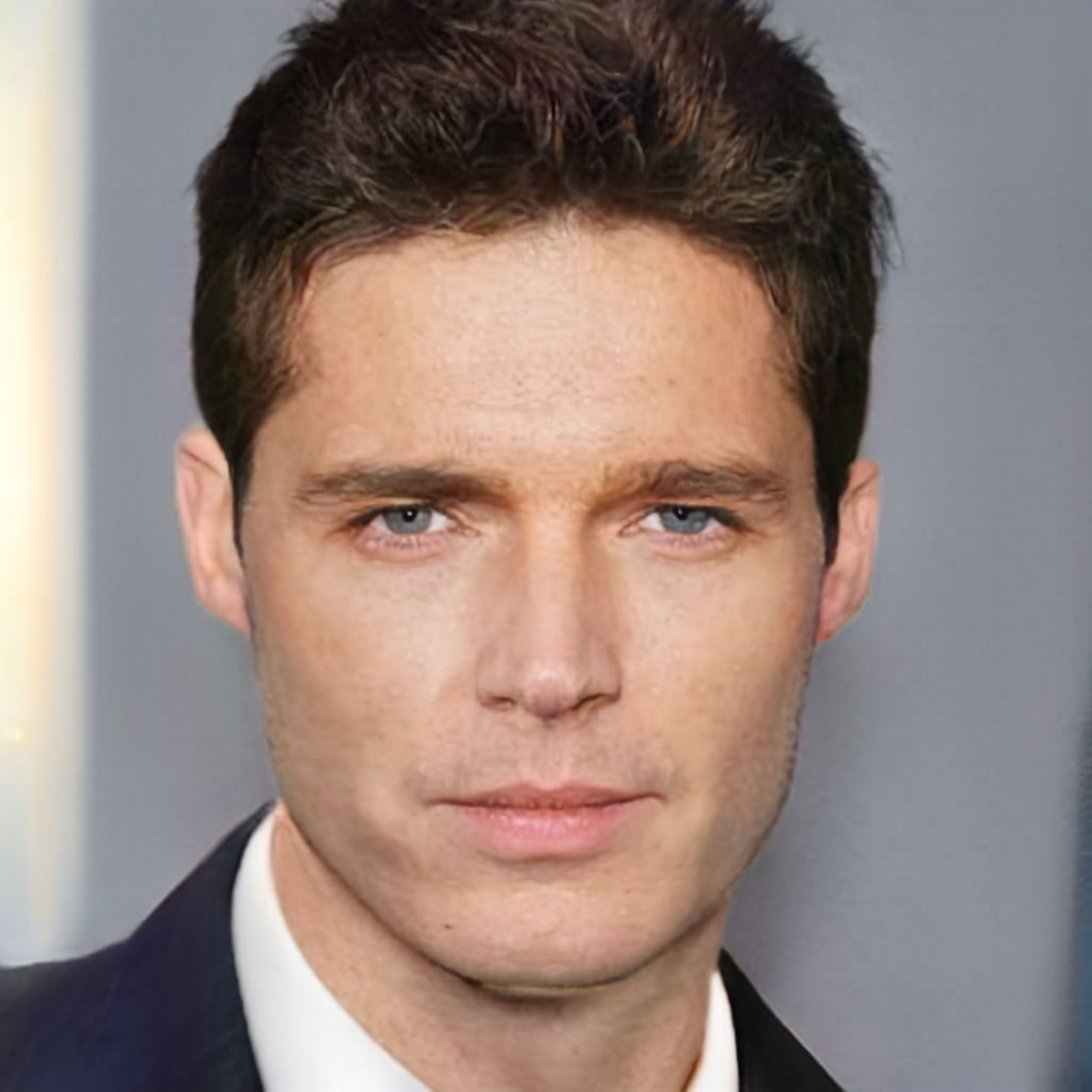} & \includegraphics[width=0.24\columnwidth]{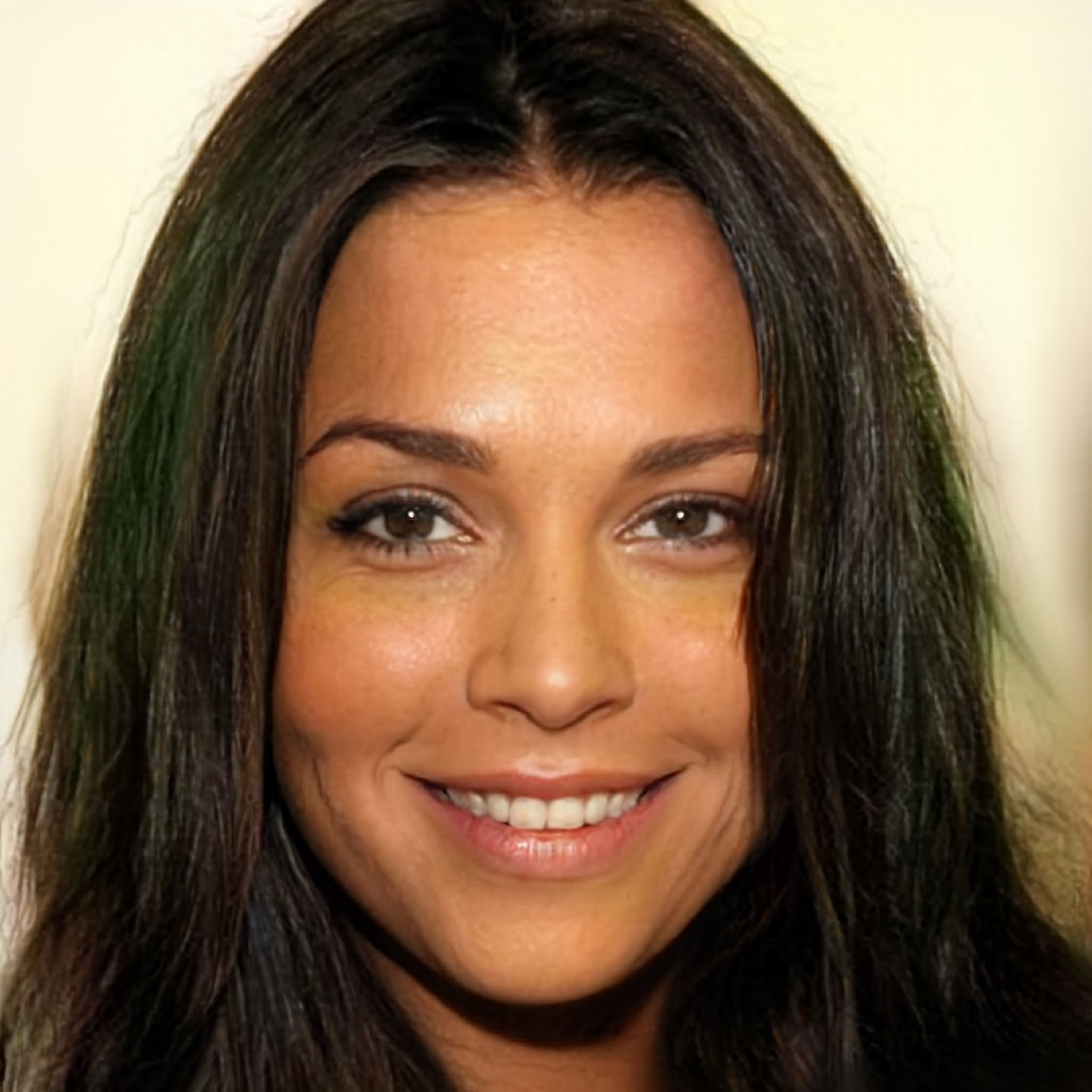}\\
         \includegraphics[width=0.24\columnwidth]{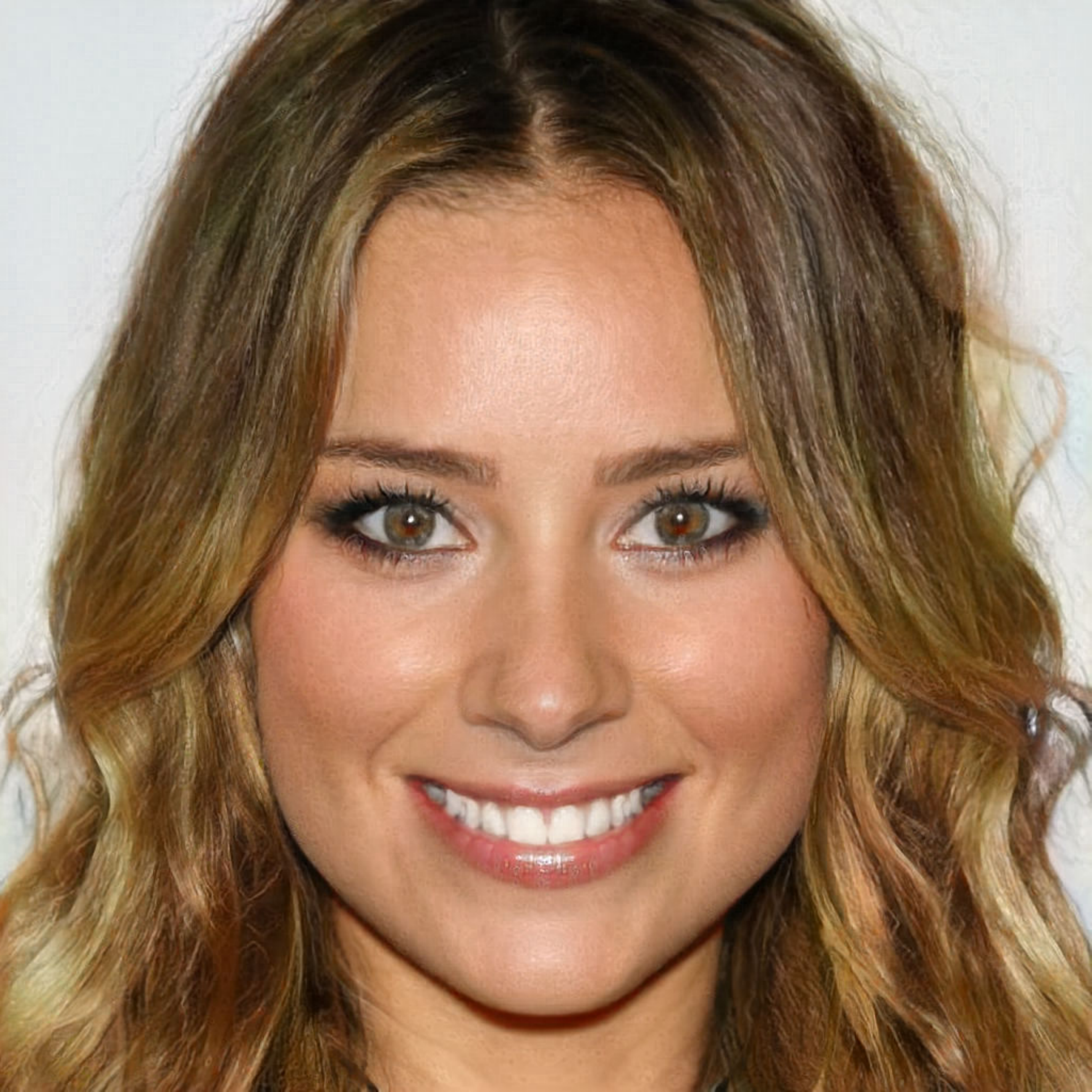} & \includegraphics[width=0.24\columnwidth]{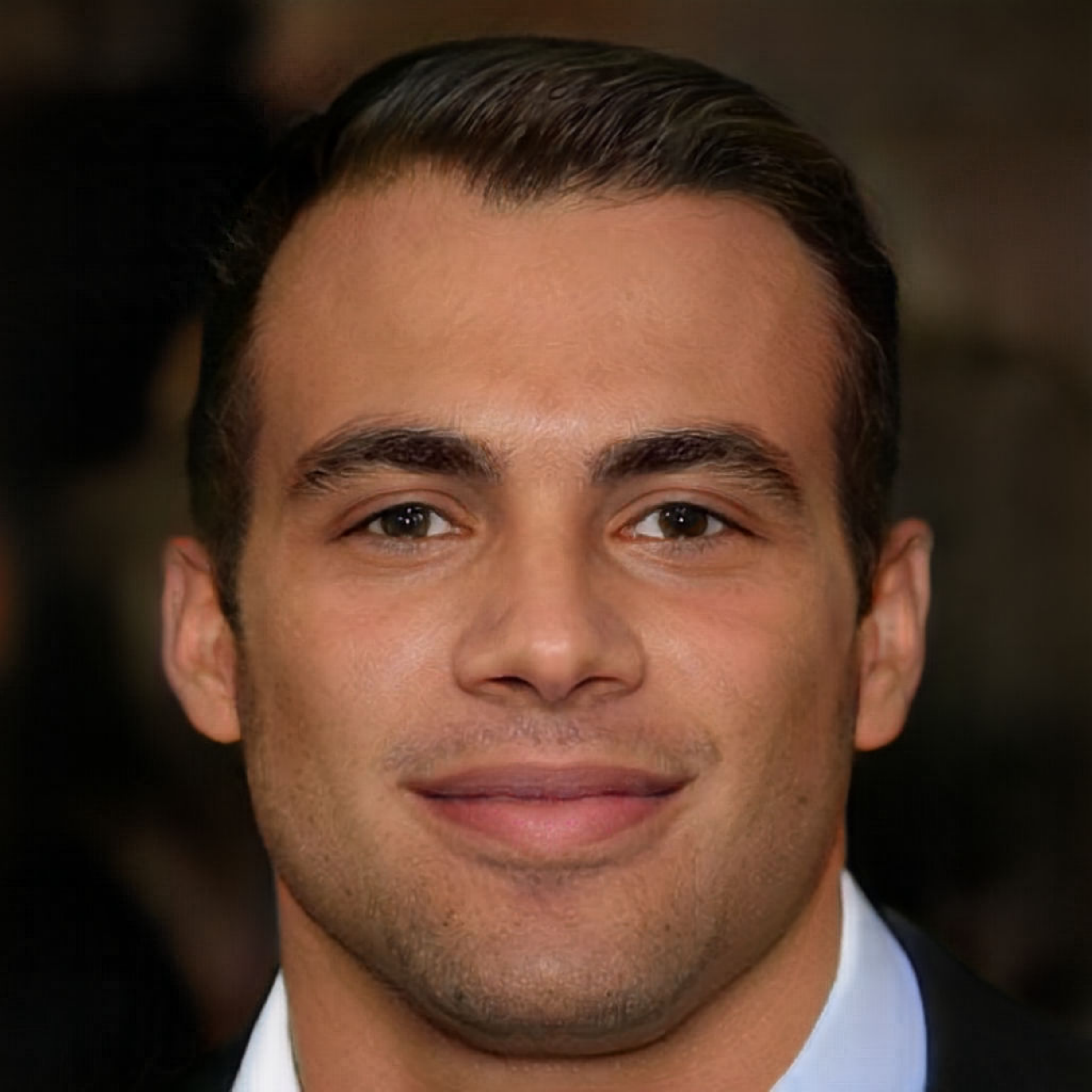} &
         \includegraphics[width=0.24\columnwidth]{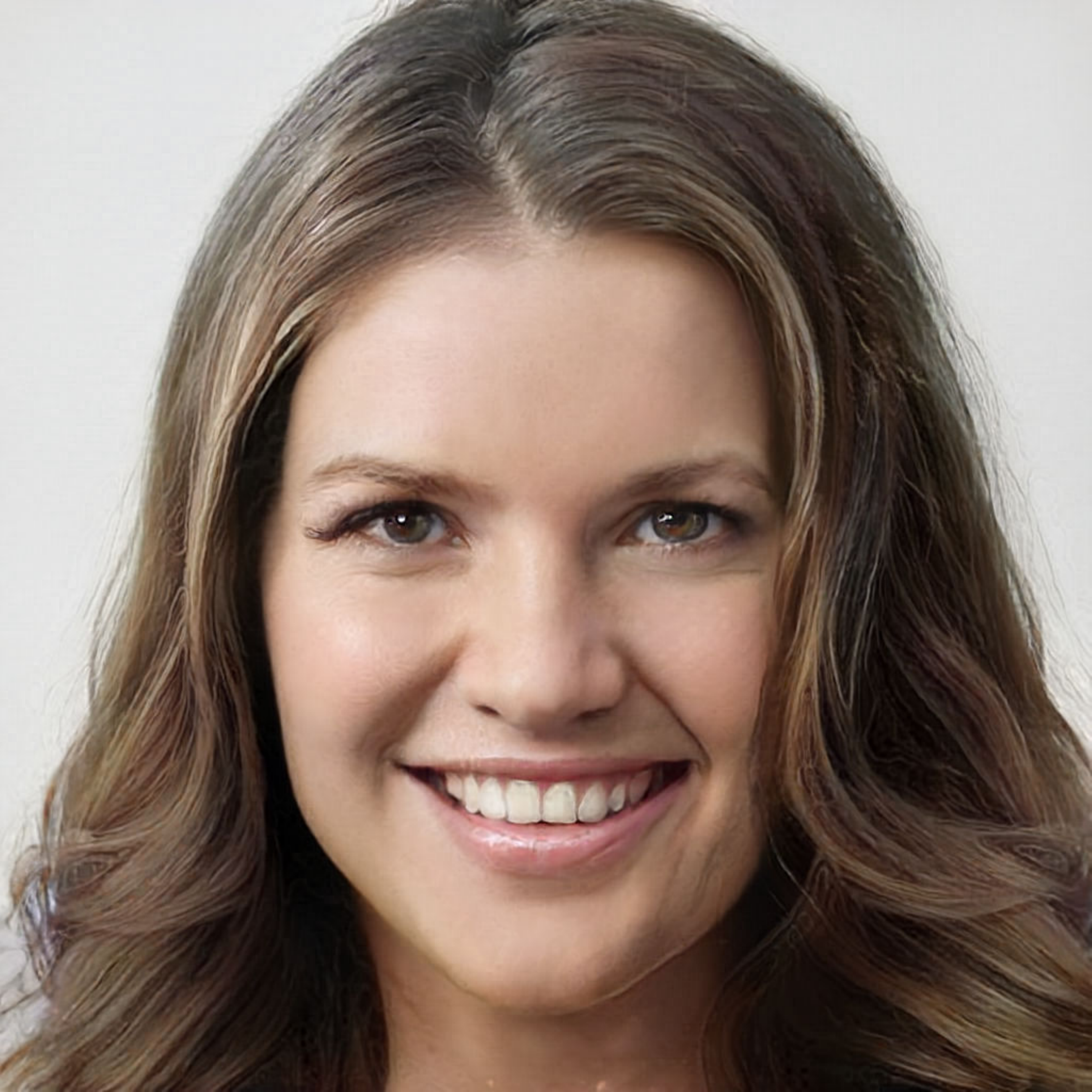} & \includegraphics[width=0.24\columnwidth]{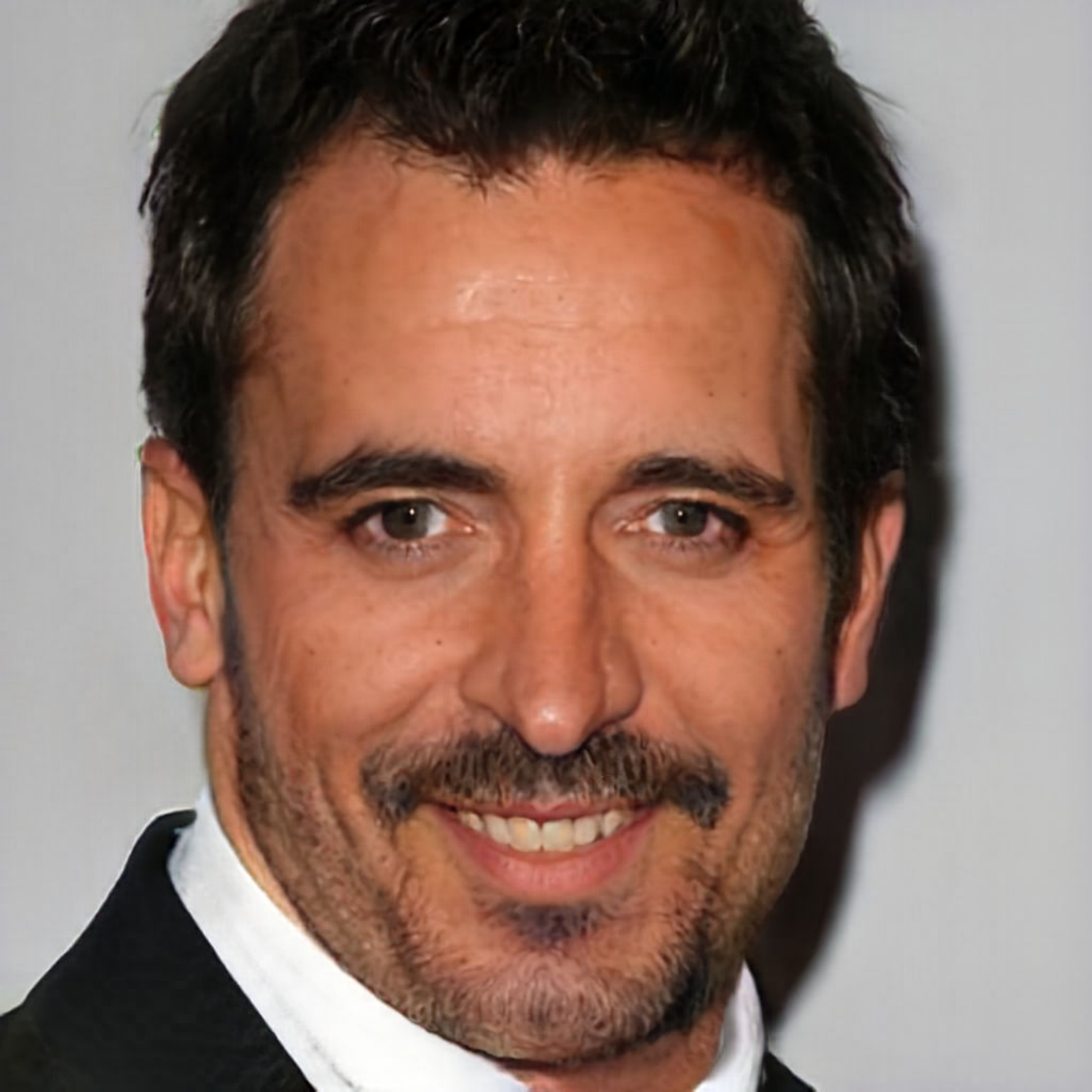}\\
         \includegraphics[width=0.24\columnwidth]{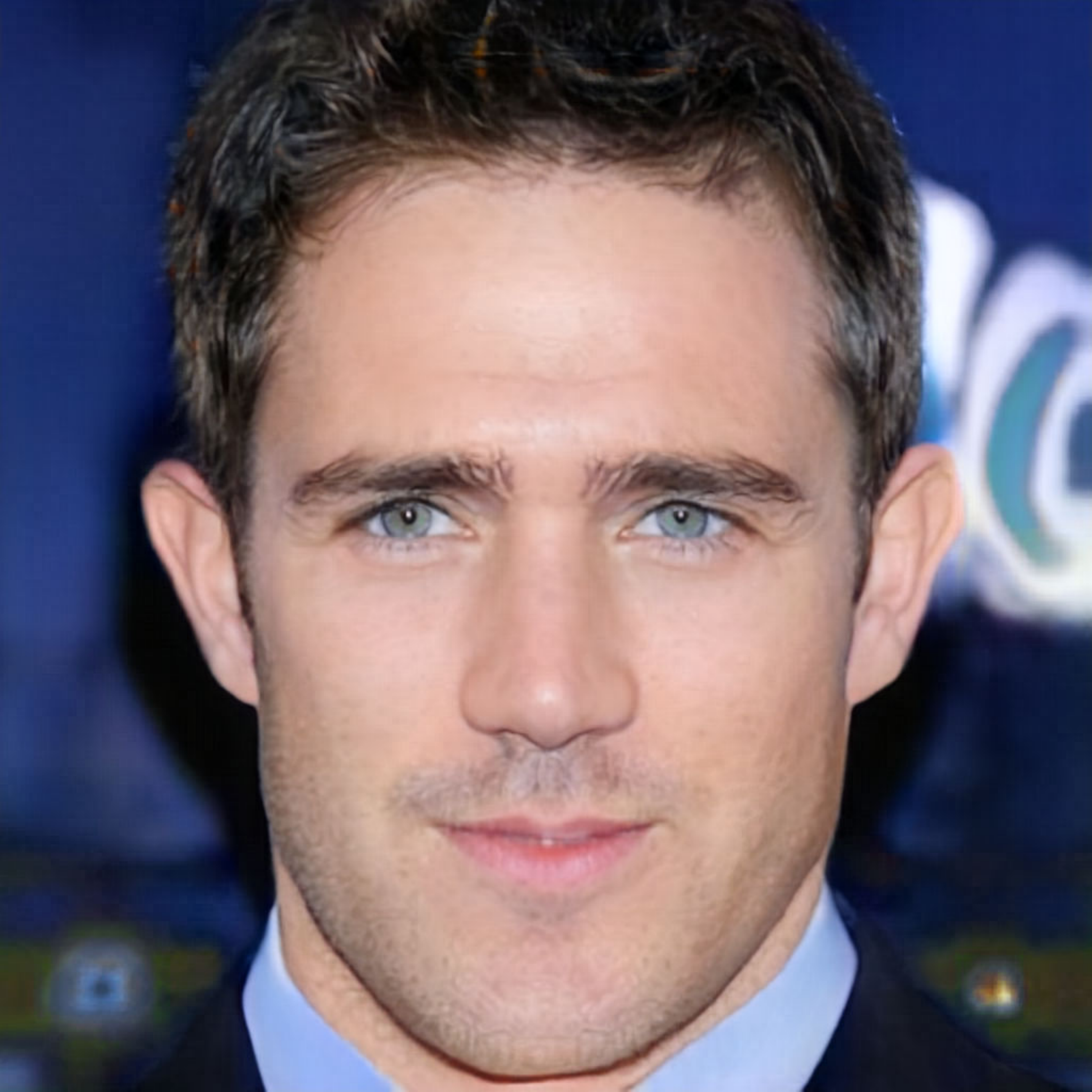} & \includegraphics[width=0.24\columnwidth]{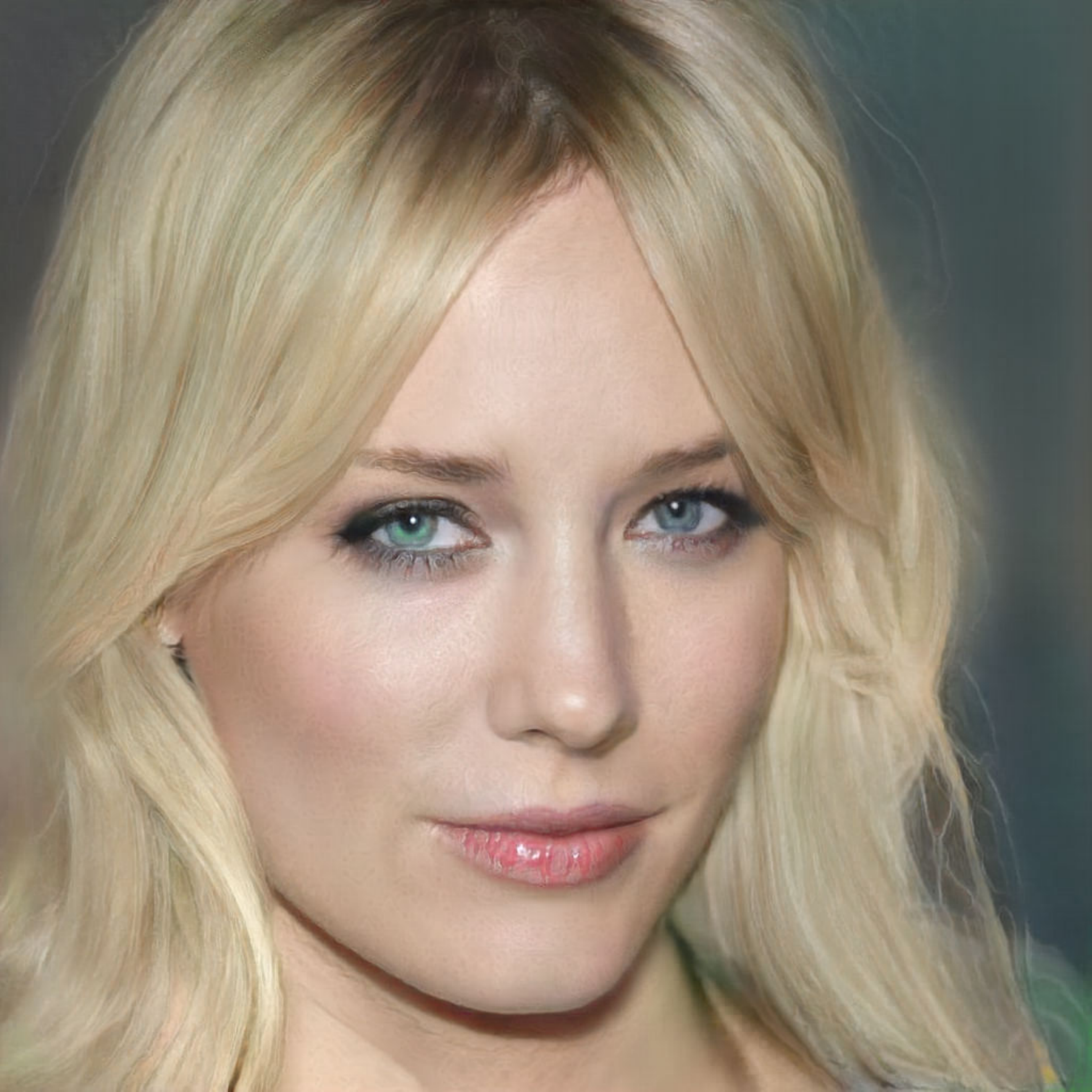} &
         \includegraphics[width=0.24\columnwidth]{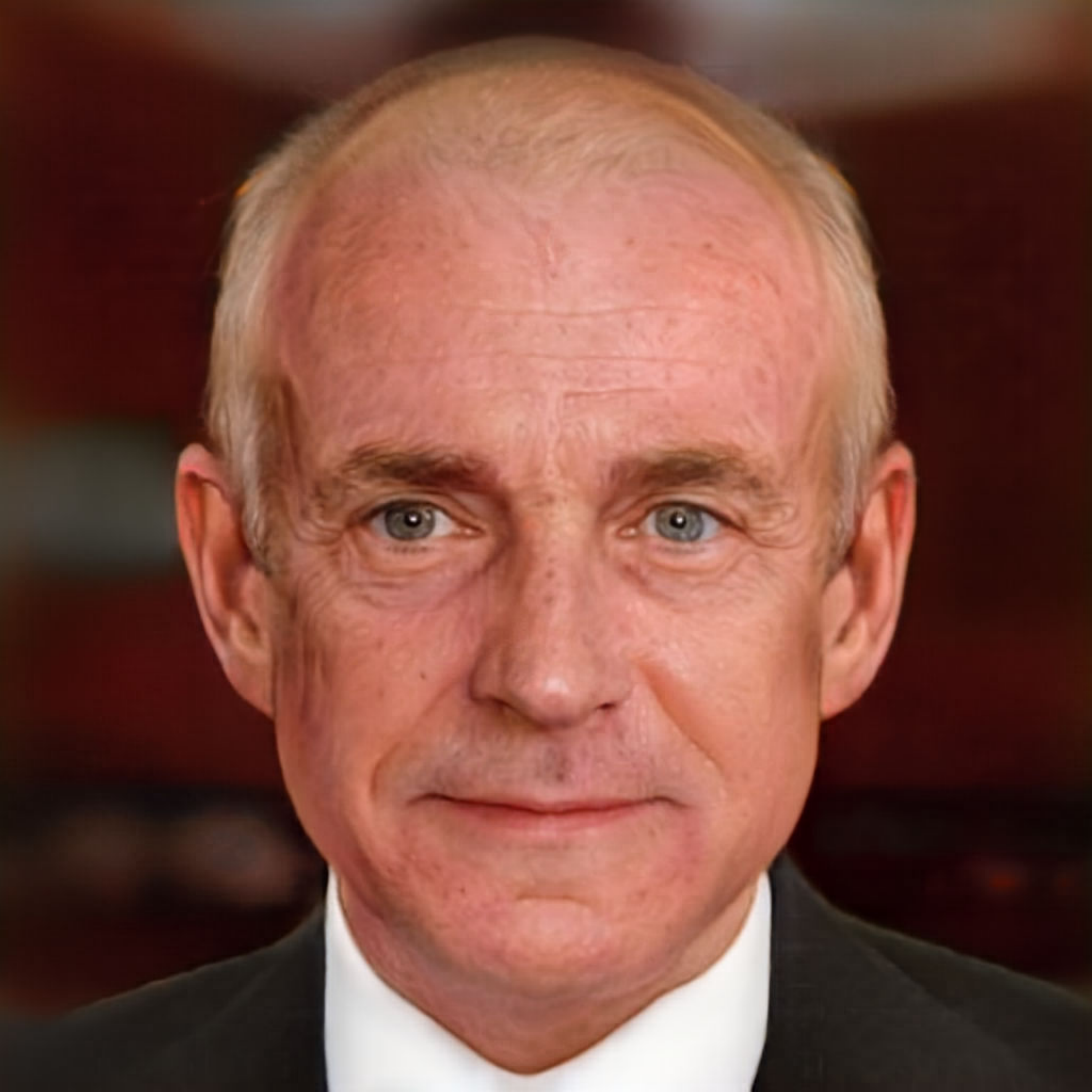} & \includegraphics[width=0.24\columnwidth]{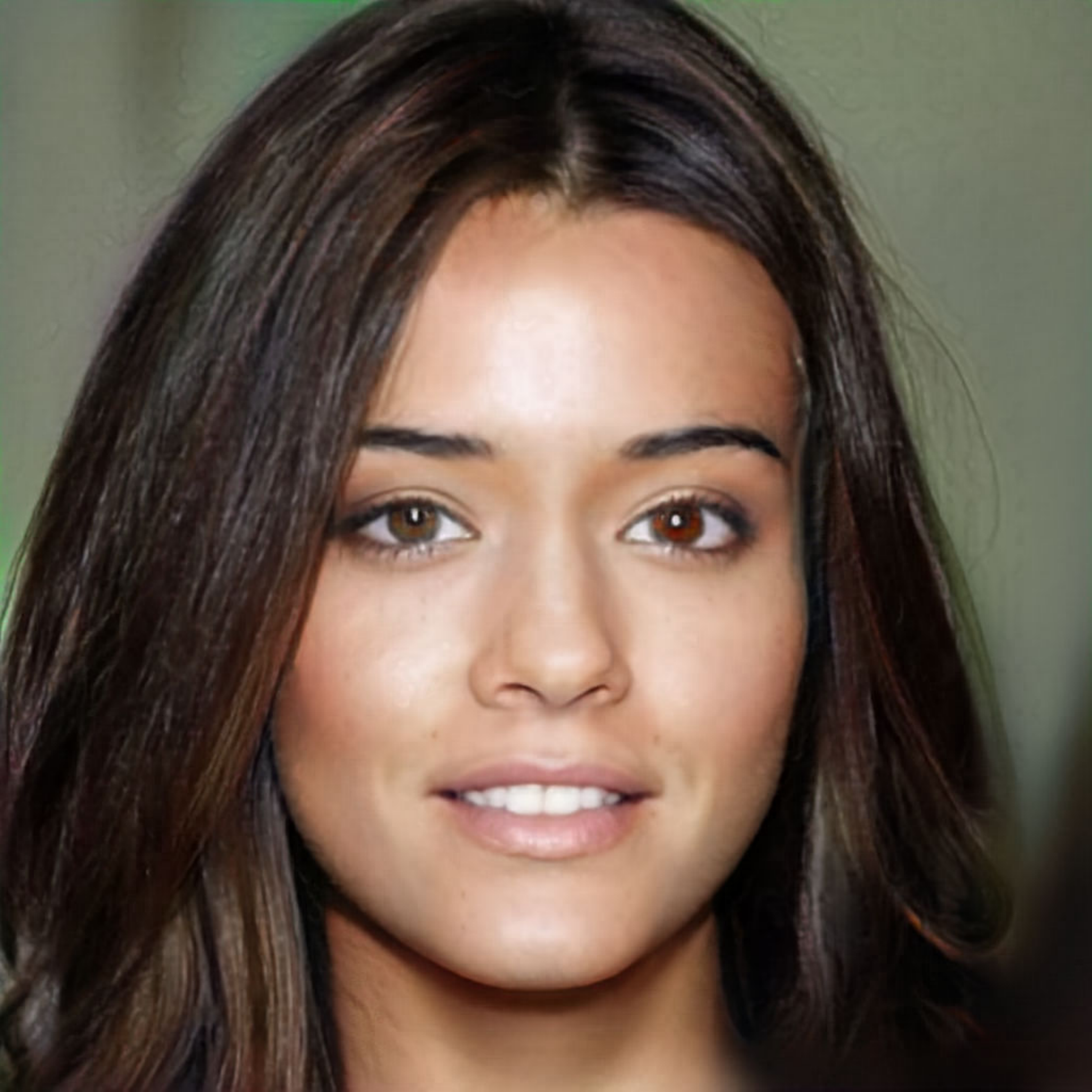}\\
         \includegraphics[width=0.24\columnwidth]{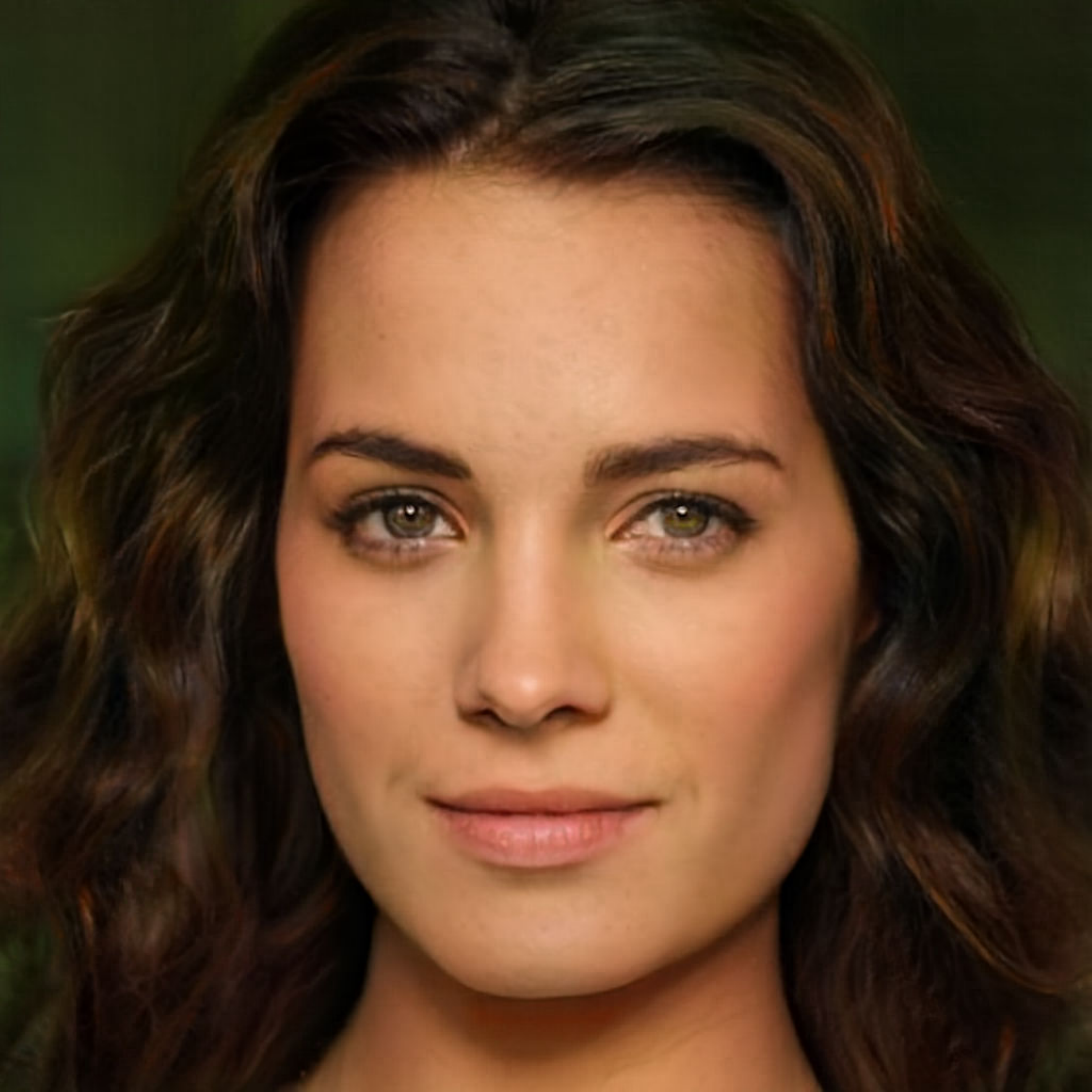} & \includegraphics[width=0.24\columnwidth]{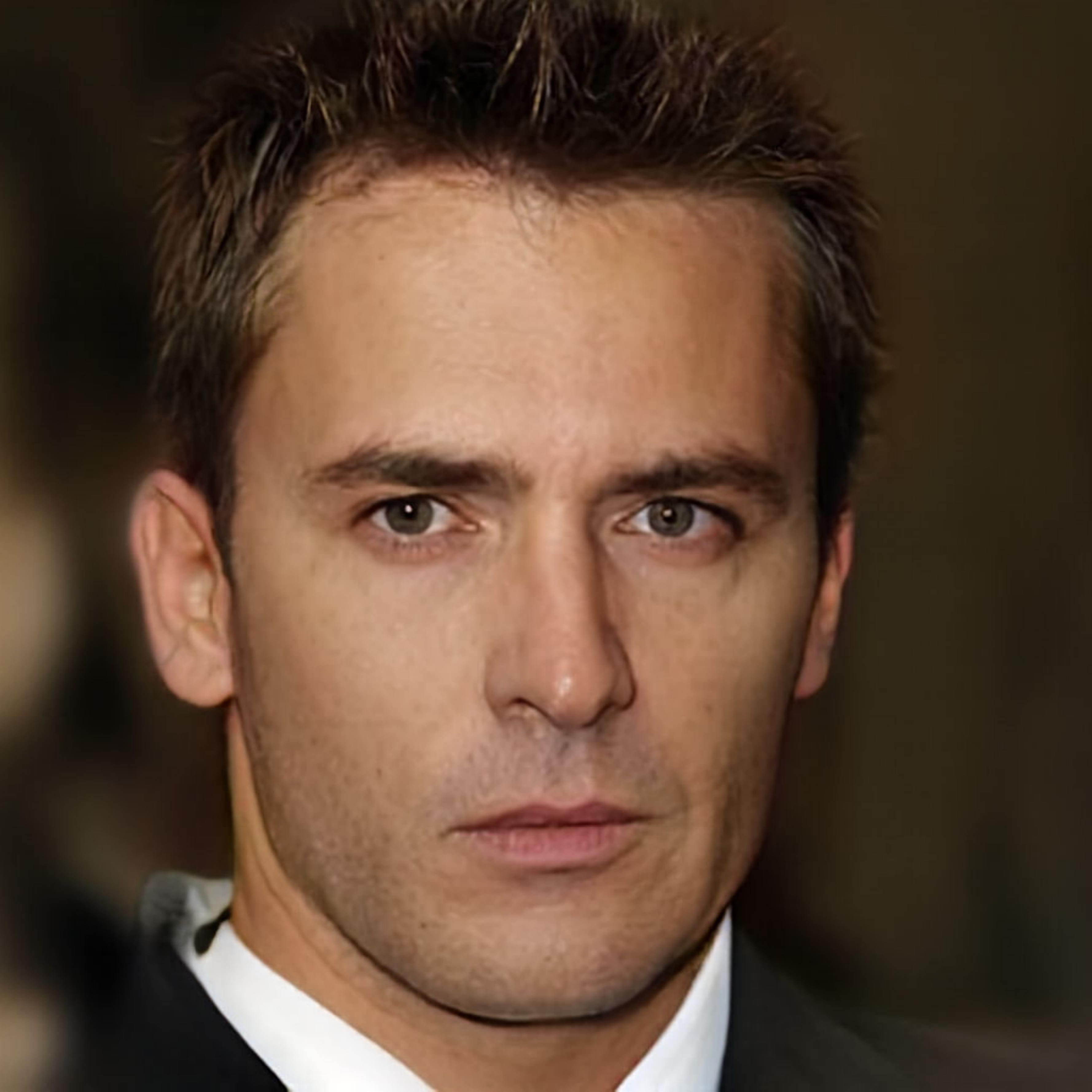} &
         \includegraphics[width=0.24\columnwidth]{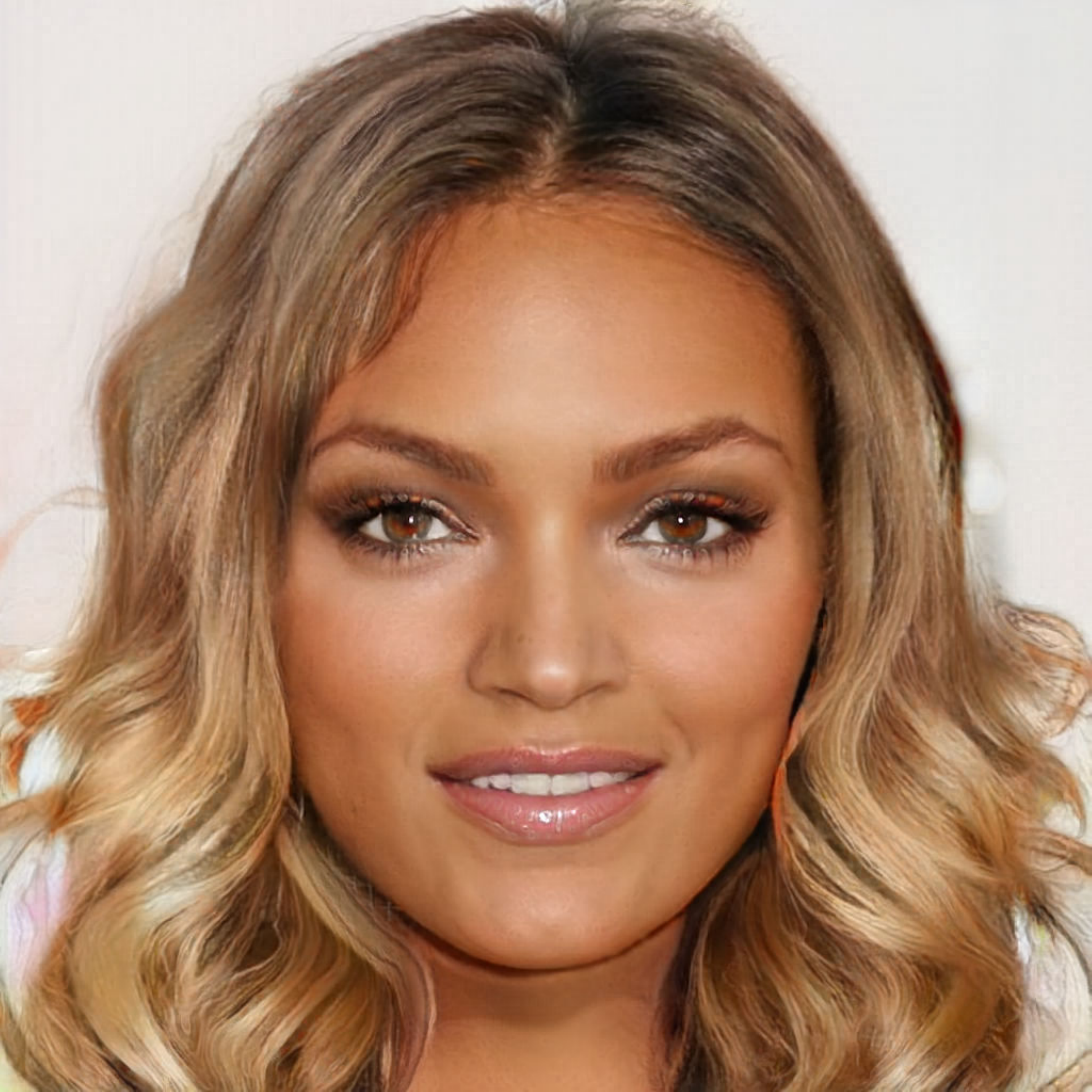} & \includegraphics[width=0.24\columnwidth]{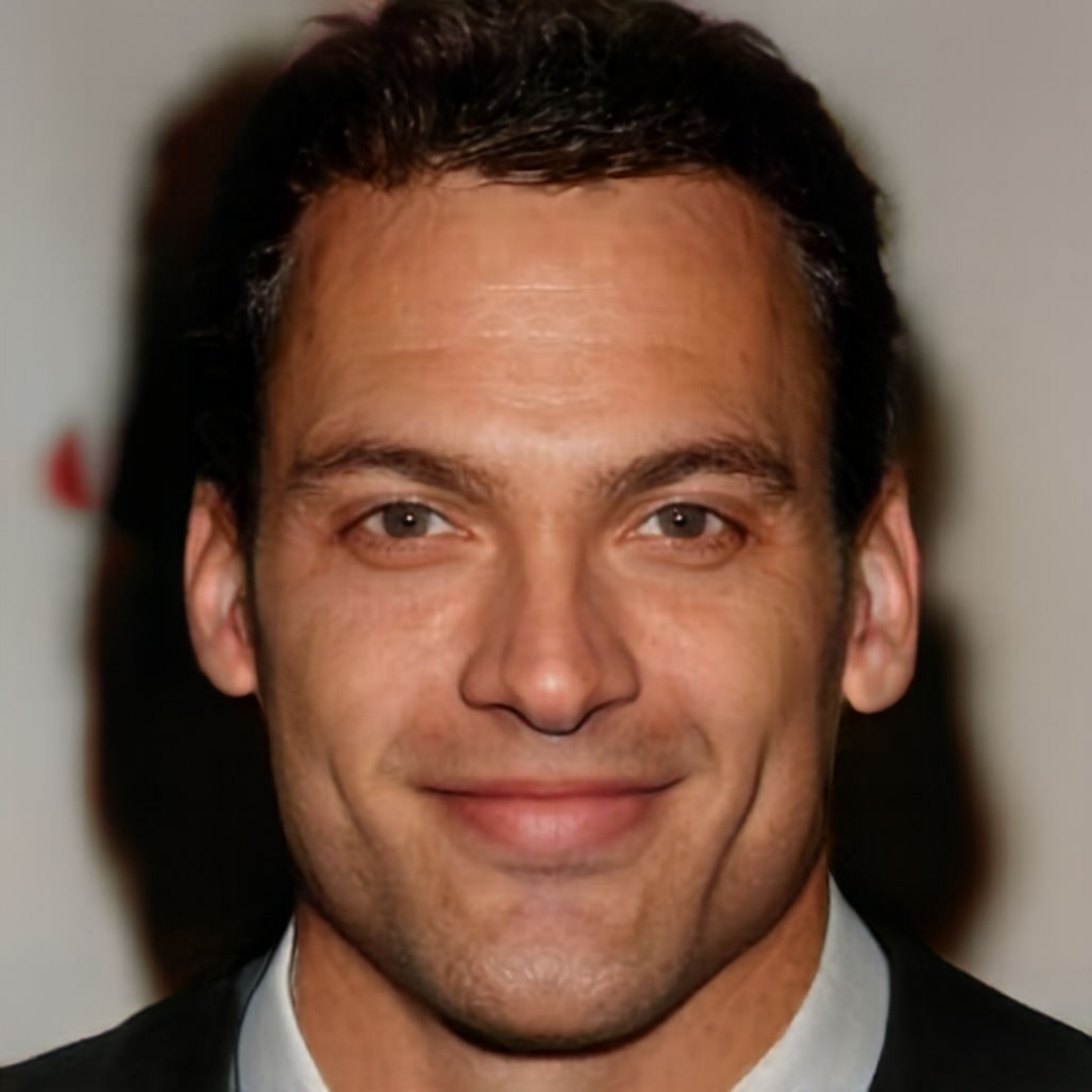}\\
    \end{tabular}
    }
    \caption{Image samples of CelebA-HQ $1024\times 1024$. }
    \label{fig:CelebAHQ1024_supp}
\end{figure*}

\newpage
\begin{figure*}[h]
    \center
    \small
    \setlength\tabcolsep{0pt}
    \renewcommand{\arraystretch}{0}
    {
    \begin{tabular}{@{}cccc@{}}
         \includegraphics[width=0.24\columnwidth]{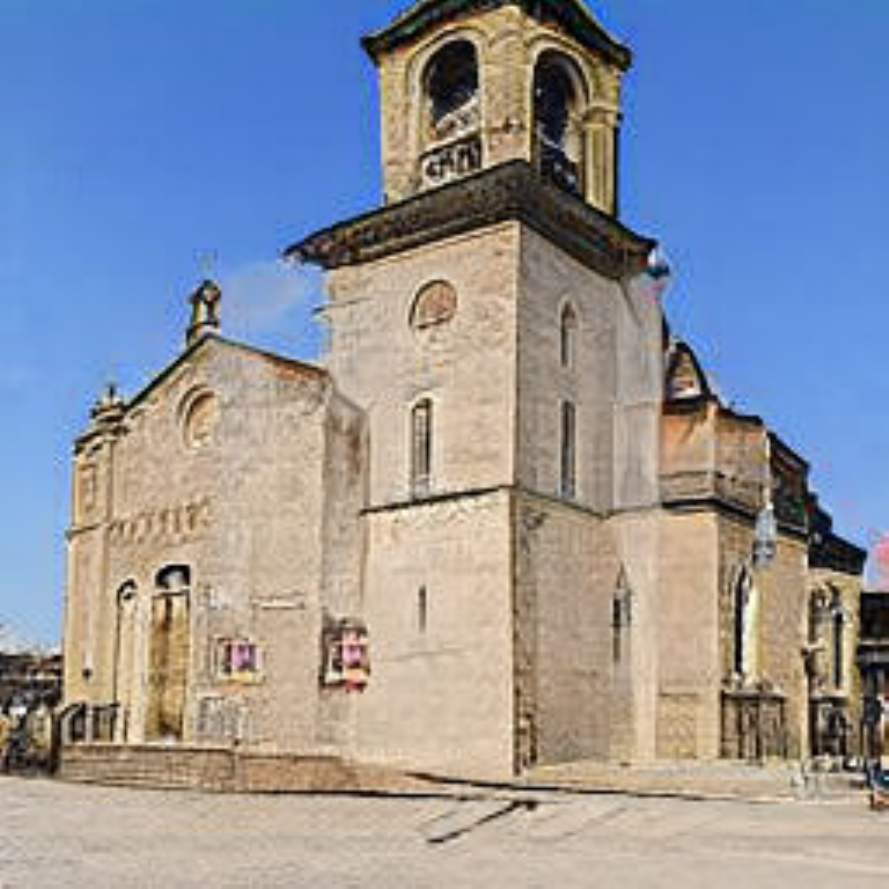} & \includegraphics[width=0.24\columnwidth]{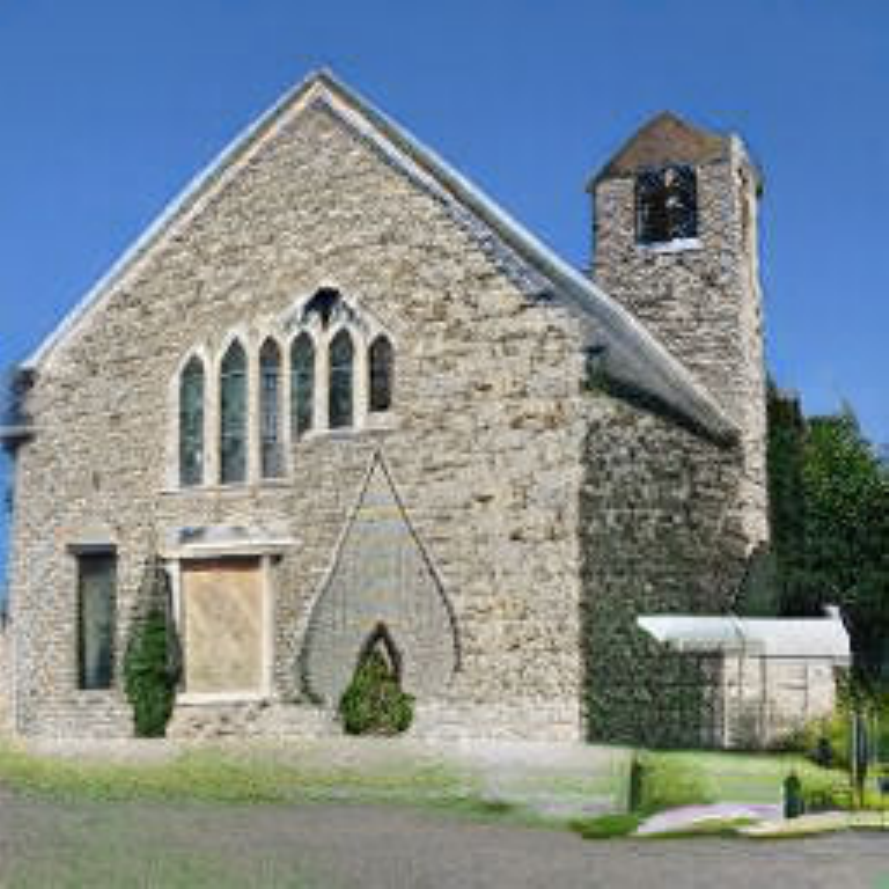} &
         \includegraphics[width=0.24\columnwidth]{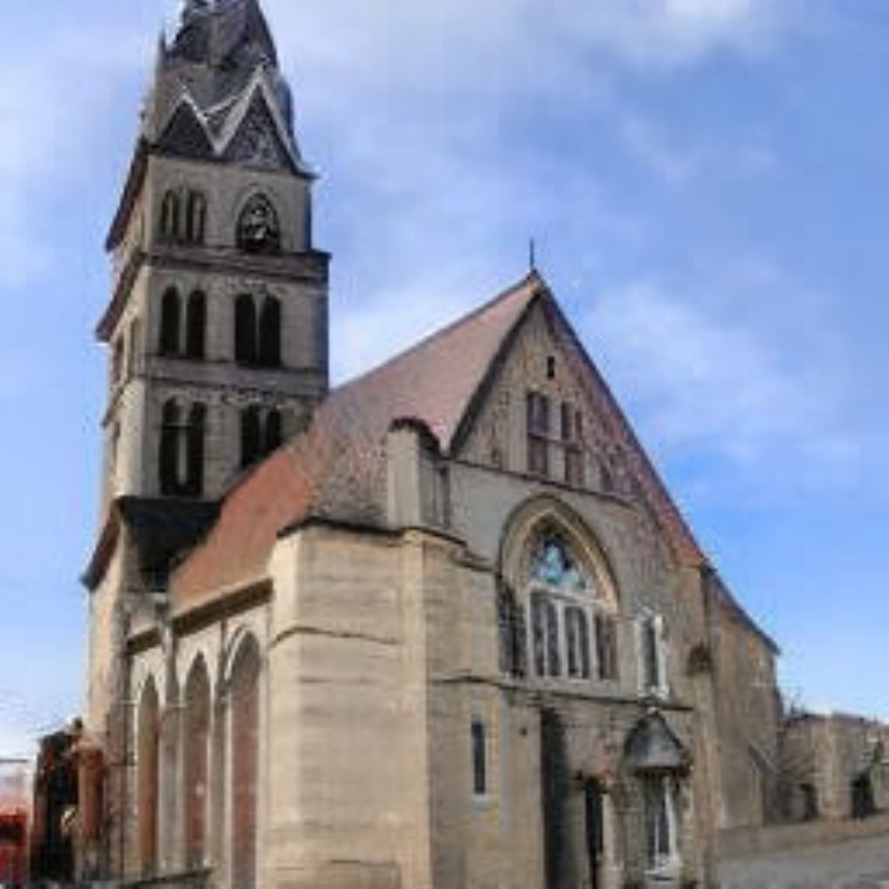} & \includegraphics[width=0.24\columnwidth]{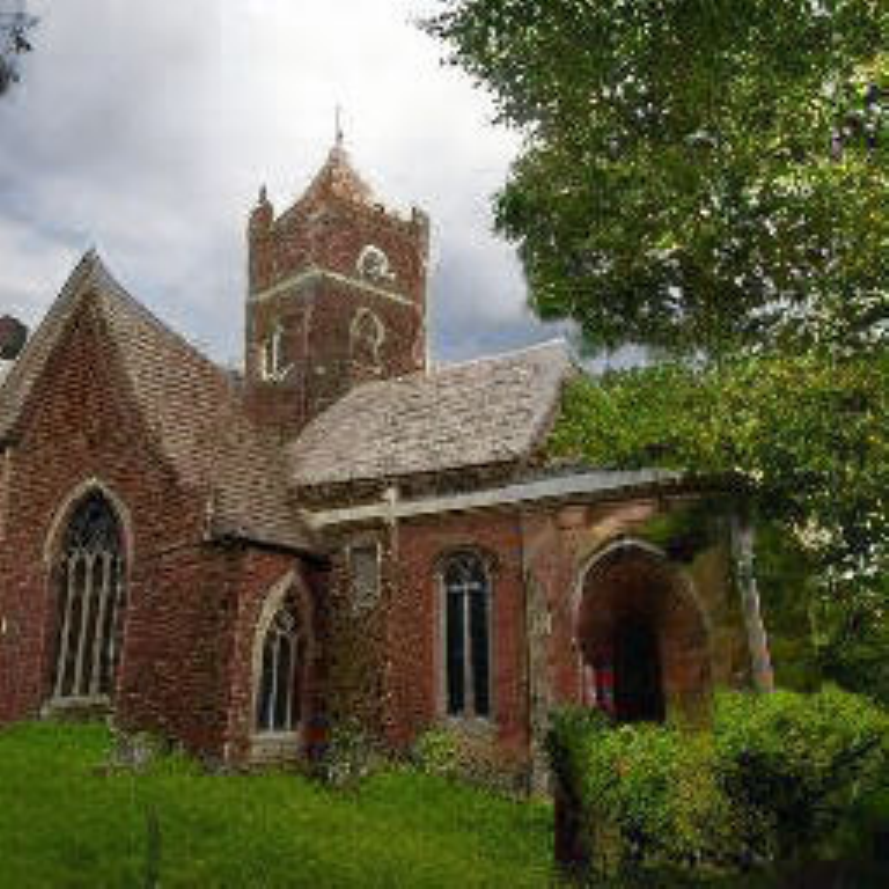}\\
         \includegraphics[width=0.24\columnwidth]{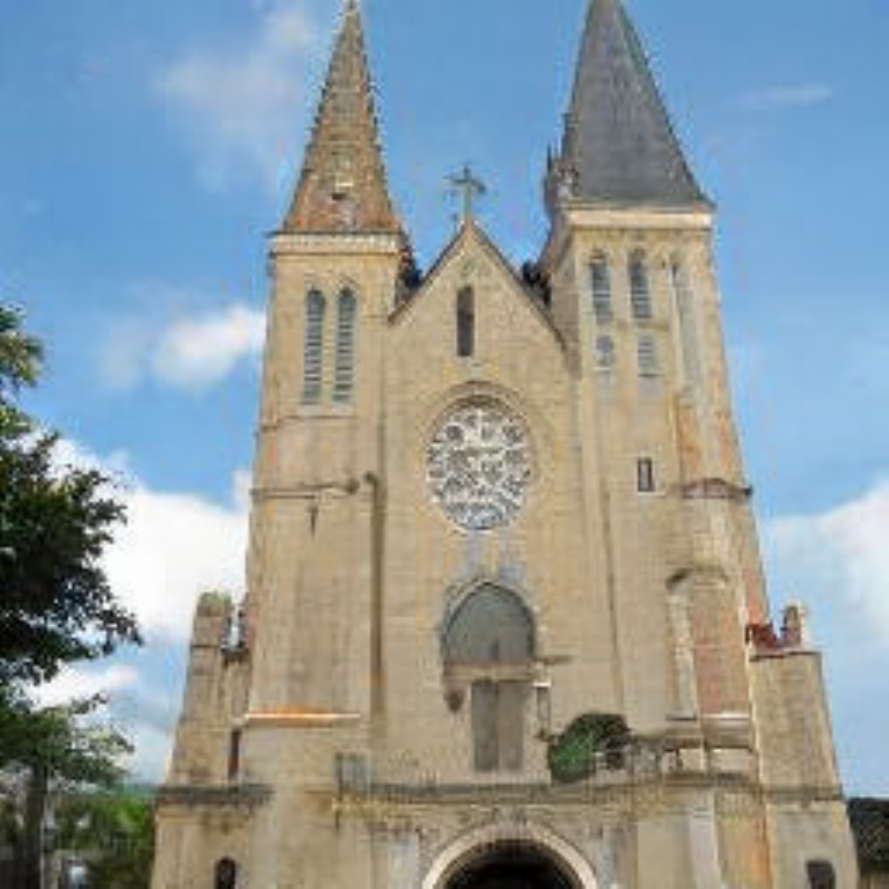} & \includegraphics[width=0.24\columnwidth]{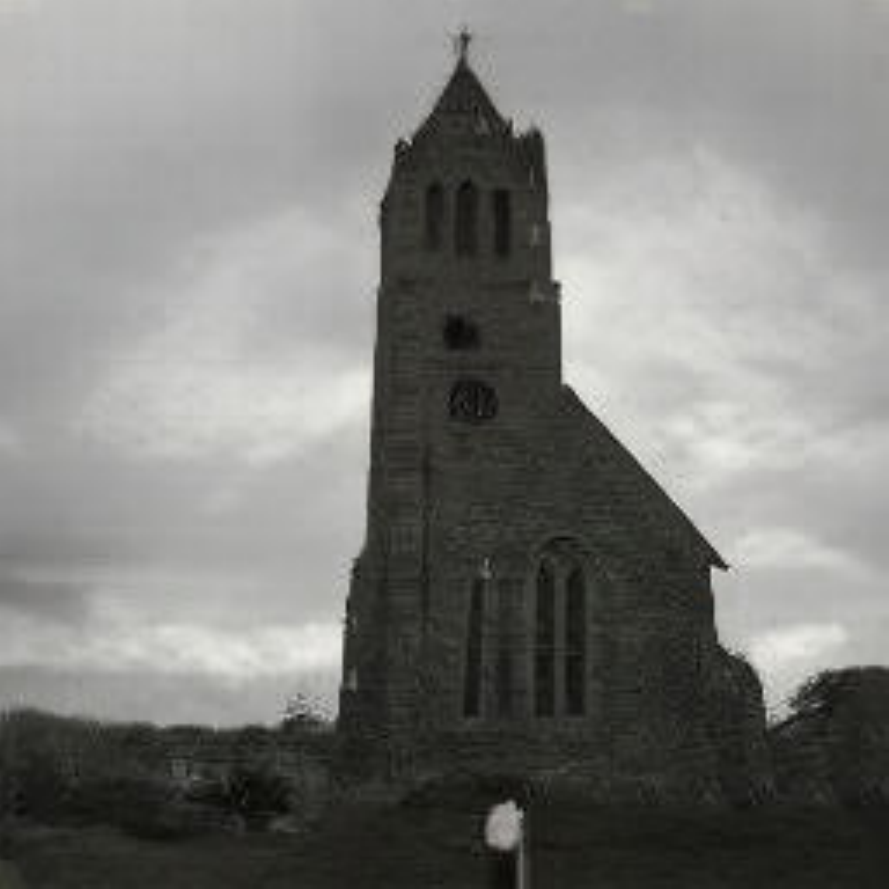} &
         \includegraphics[width=0.24\columnwidth]{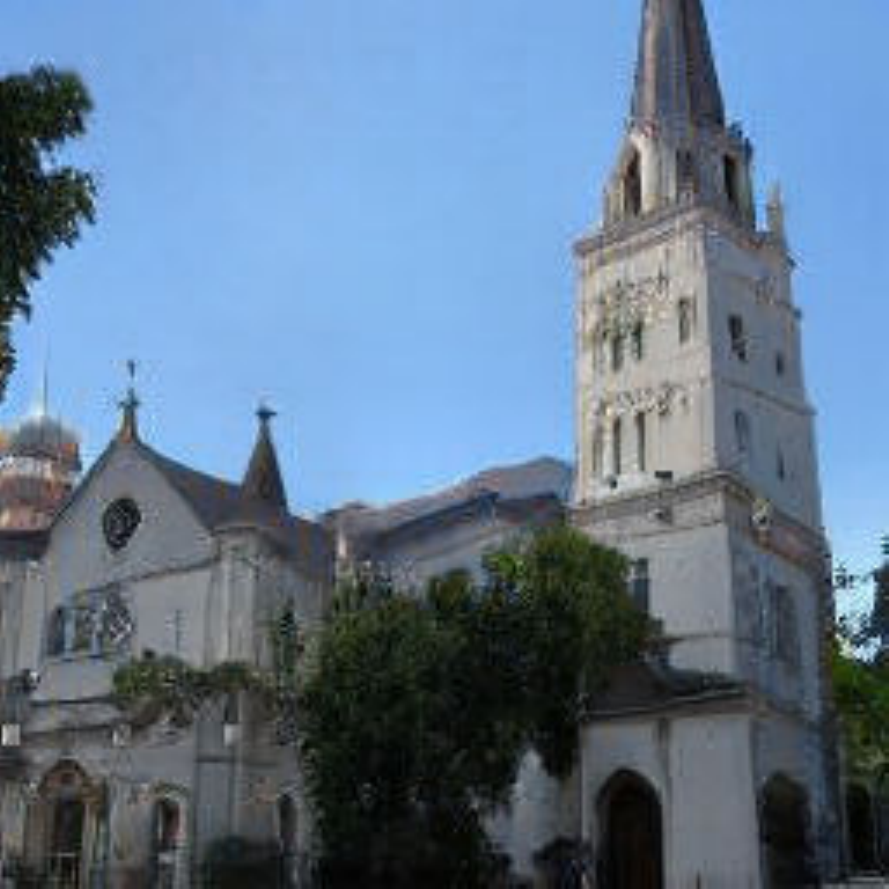} & \includegraphics[width=0.24\columnwidth]{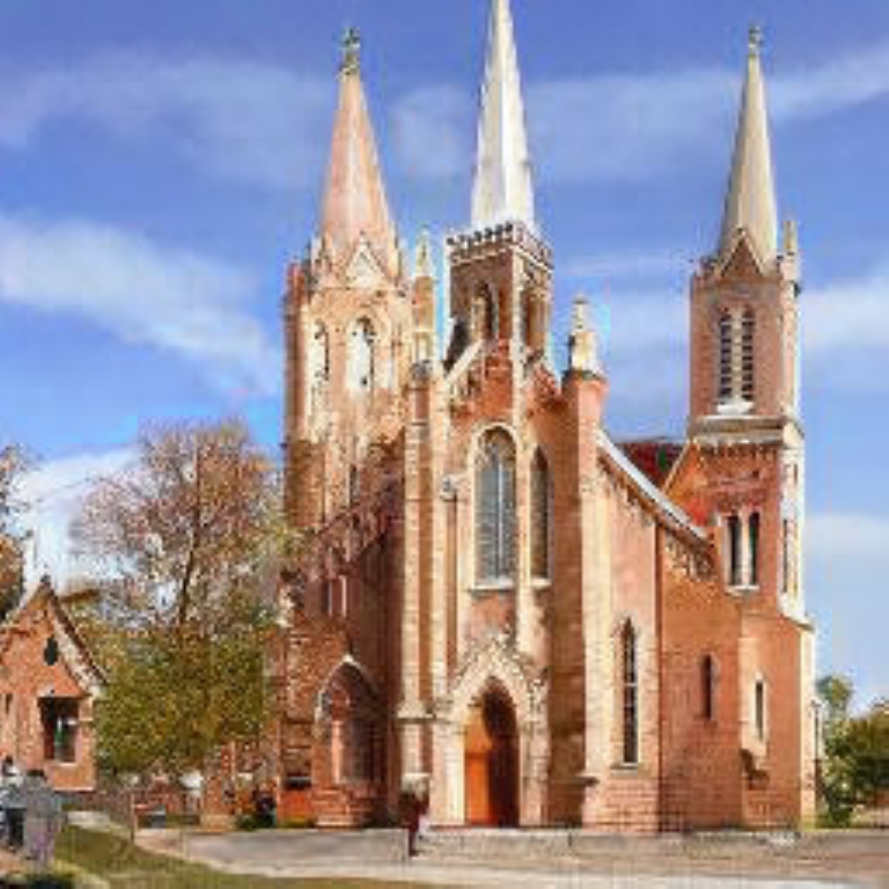}\\
         \includegraphics[width=0.24\columnwidth]{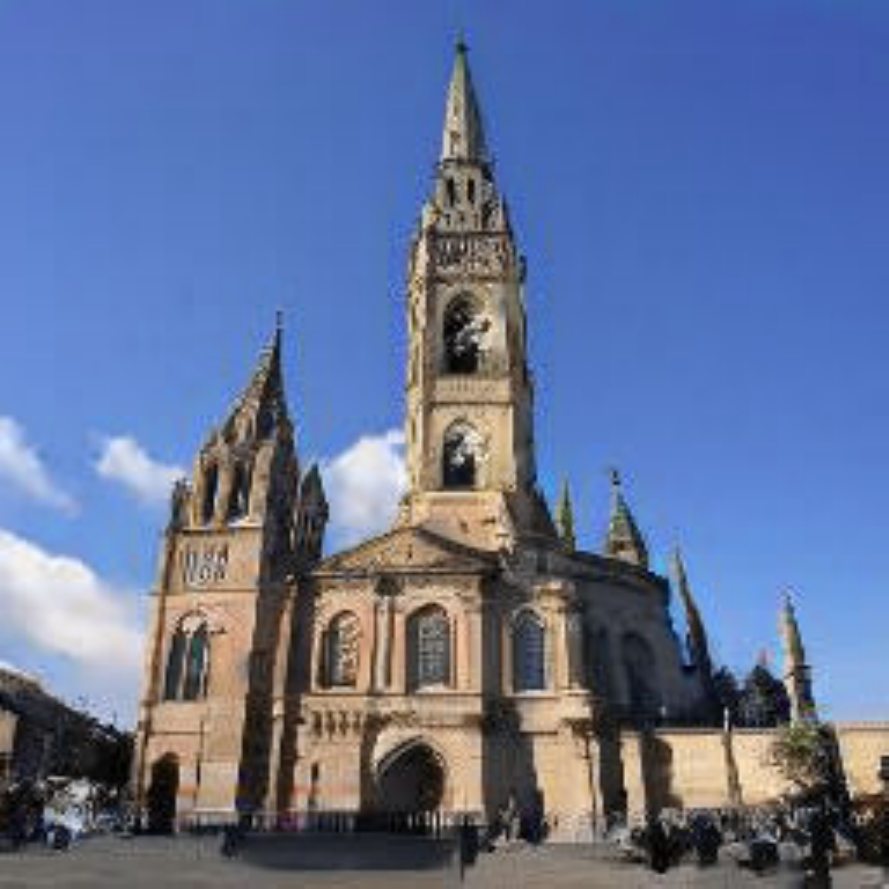} & \includegraphics[width=0.24\columnwidth]{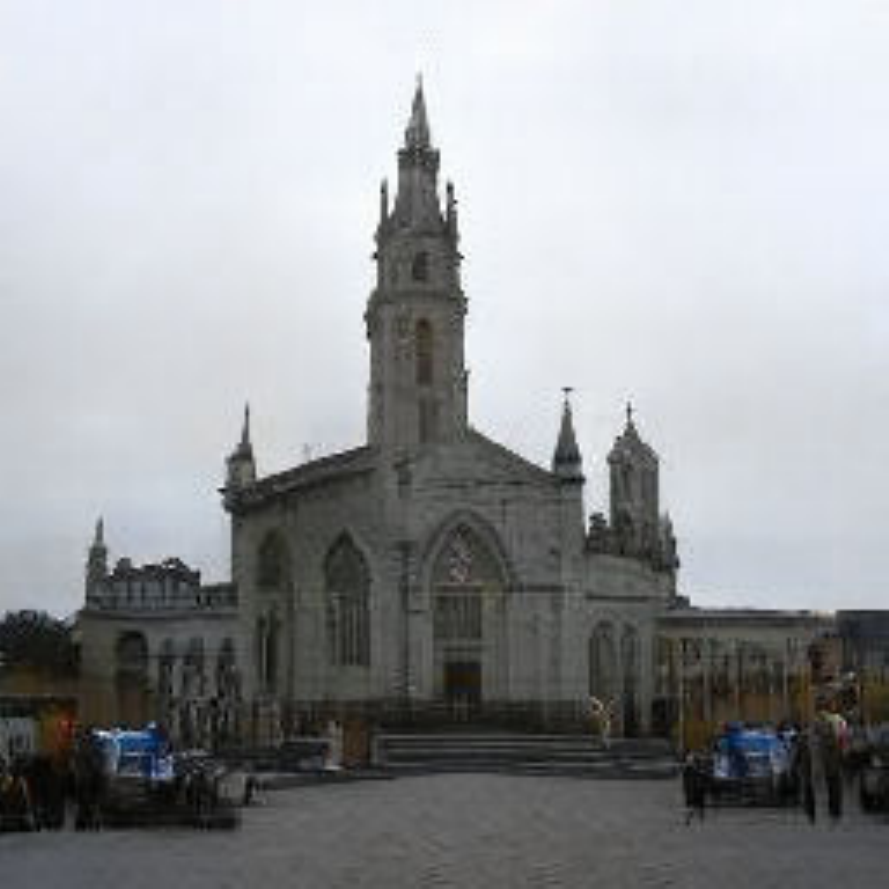} &
         \includegraphics[width=0.24\columnwidth]{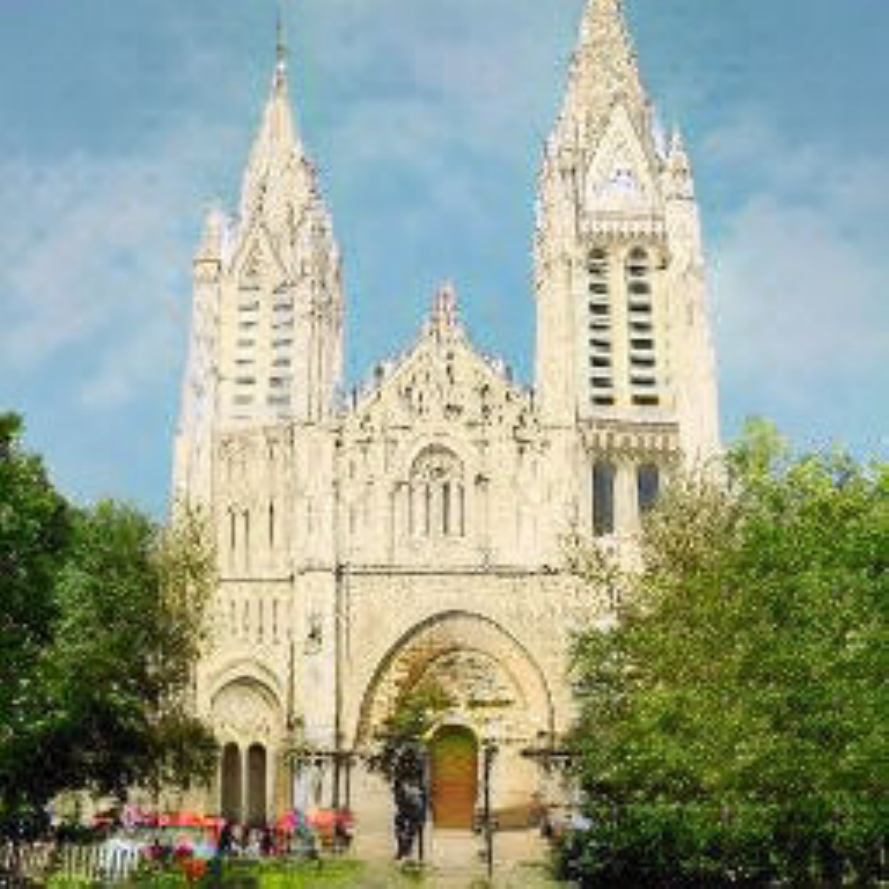} & \includegraphics[width=0.24\columnwidth]{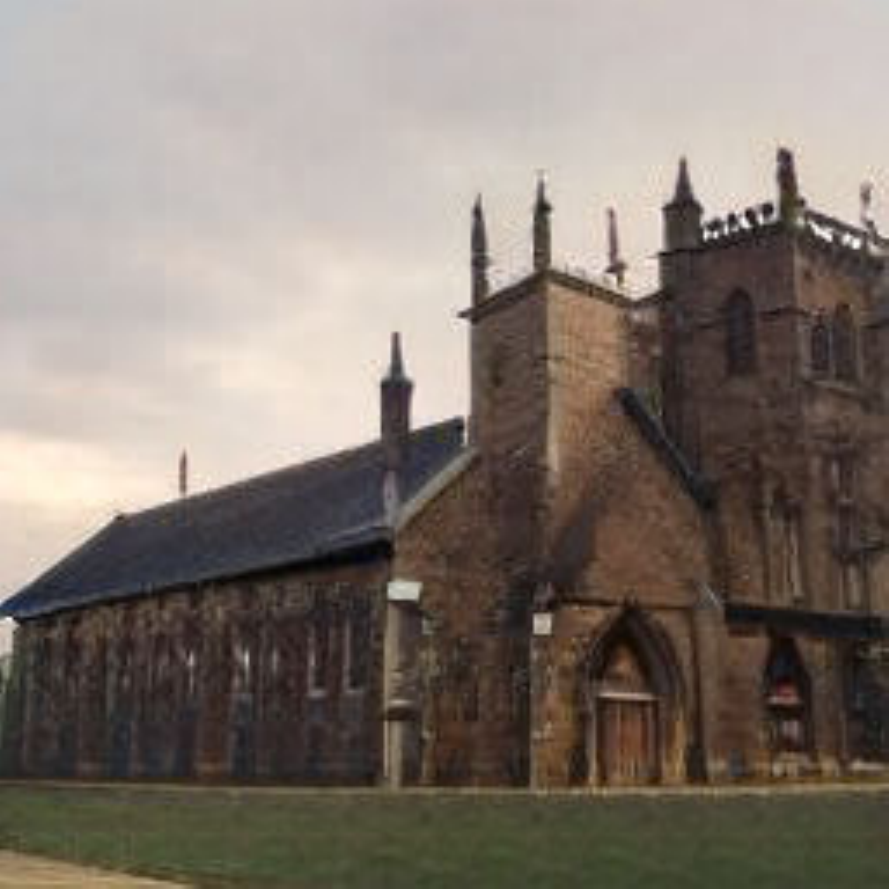}\\
         \includegraphics[width=0.24\columnwidth]{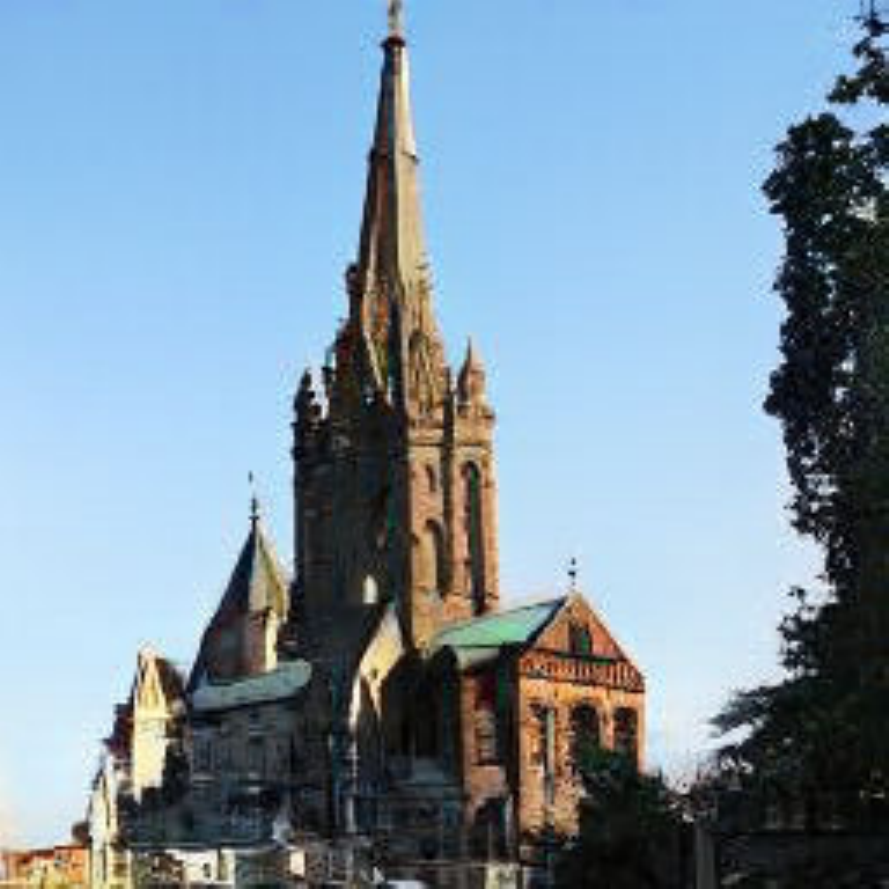} & \includegraphics[width=0.24\columnwidth]{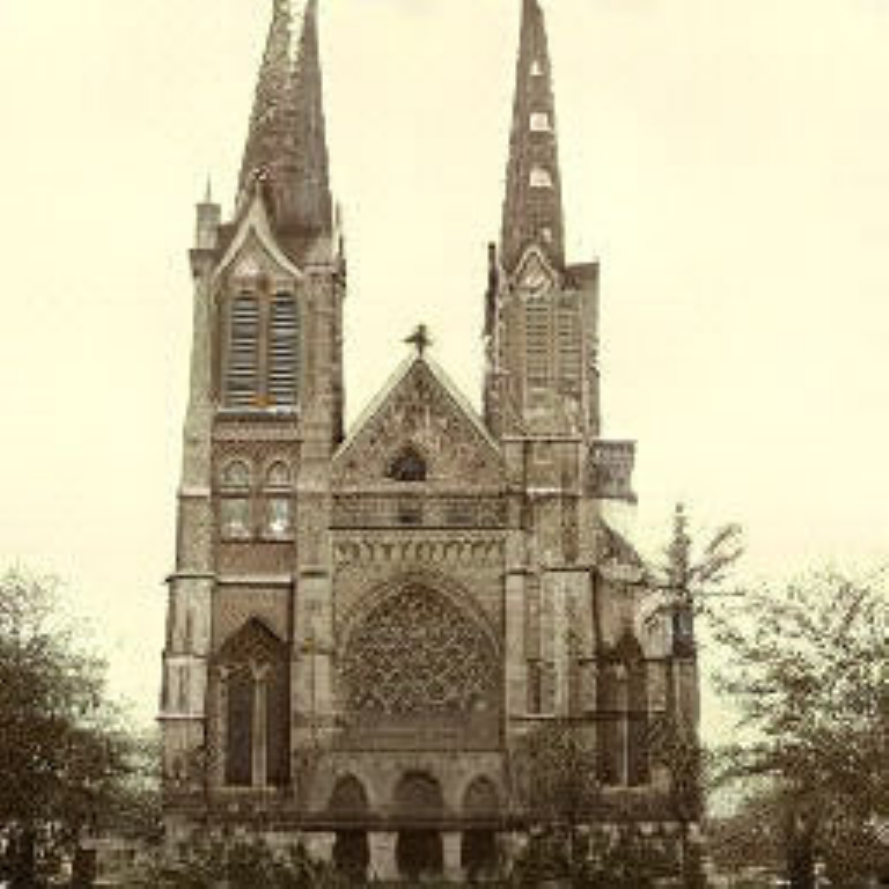} &
         \includegraphics[width=0.24\columnwidth]{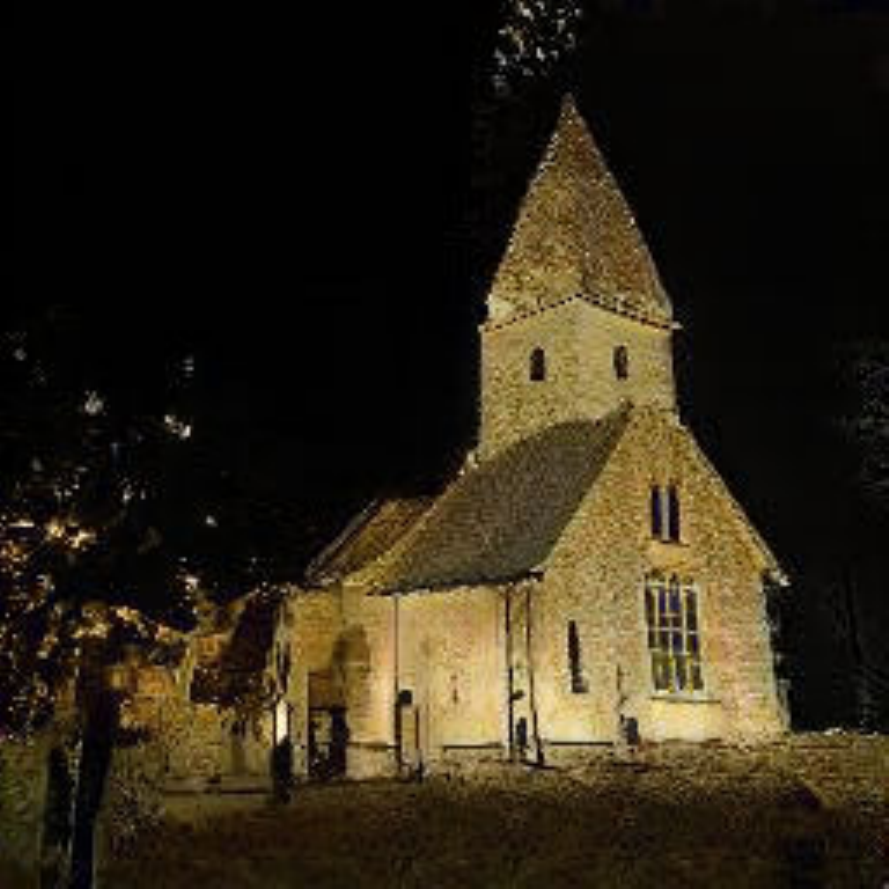} & \includegraphics[width=0.24\columnwidth]{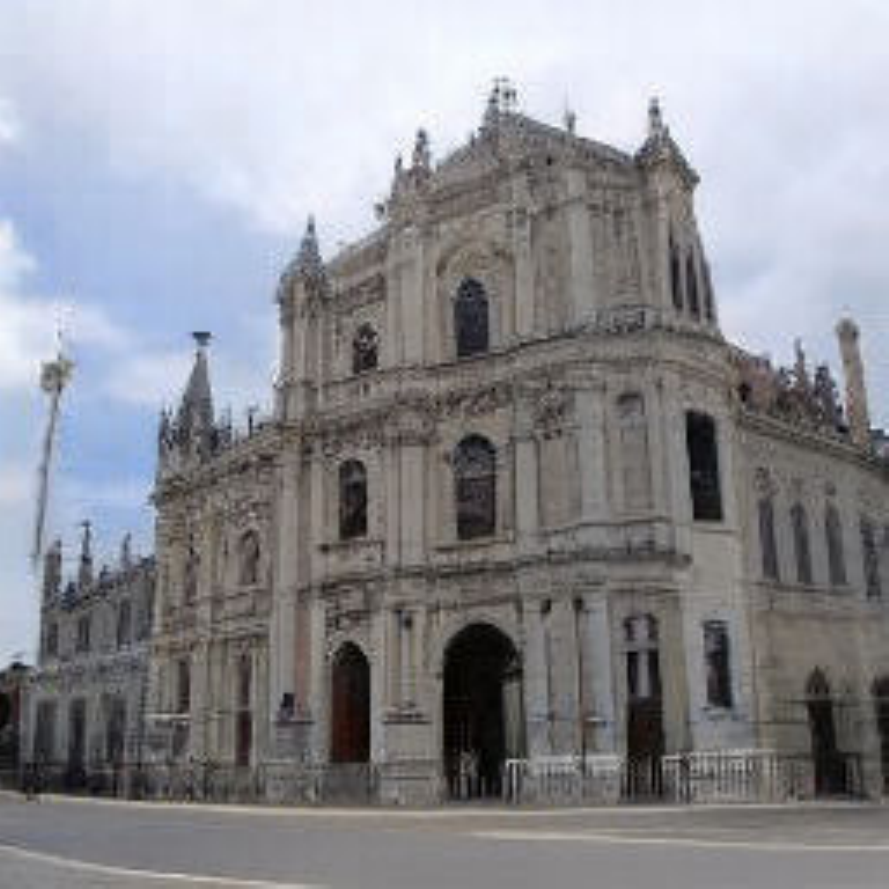}\\
         \includegraphics[width=0.24\columnwidth]{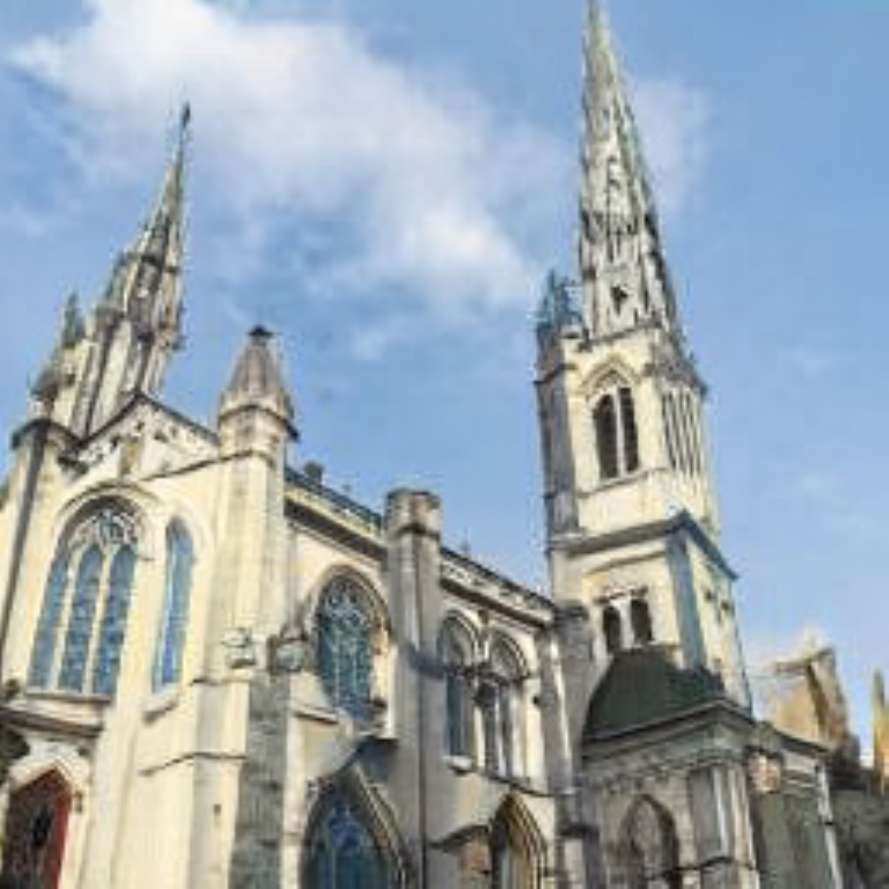} & \includegraphics[width=0.24\columnwidth]{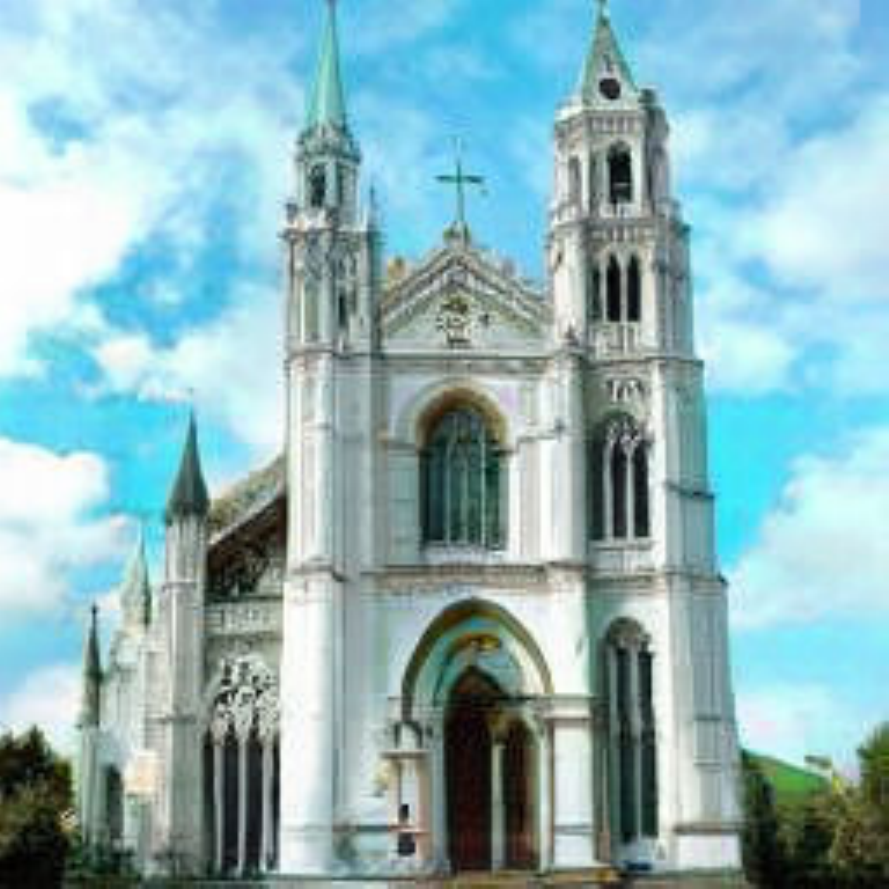} &
         \includegraphics[width=0.24\columnwidth]{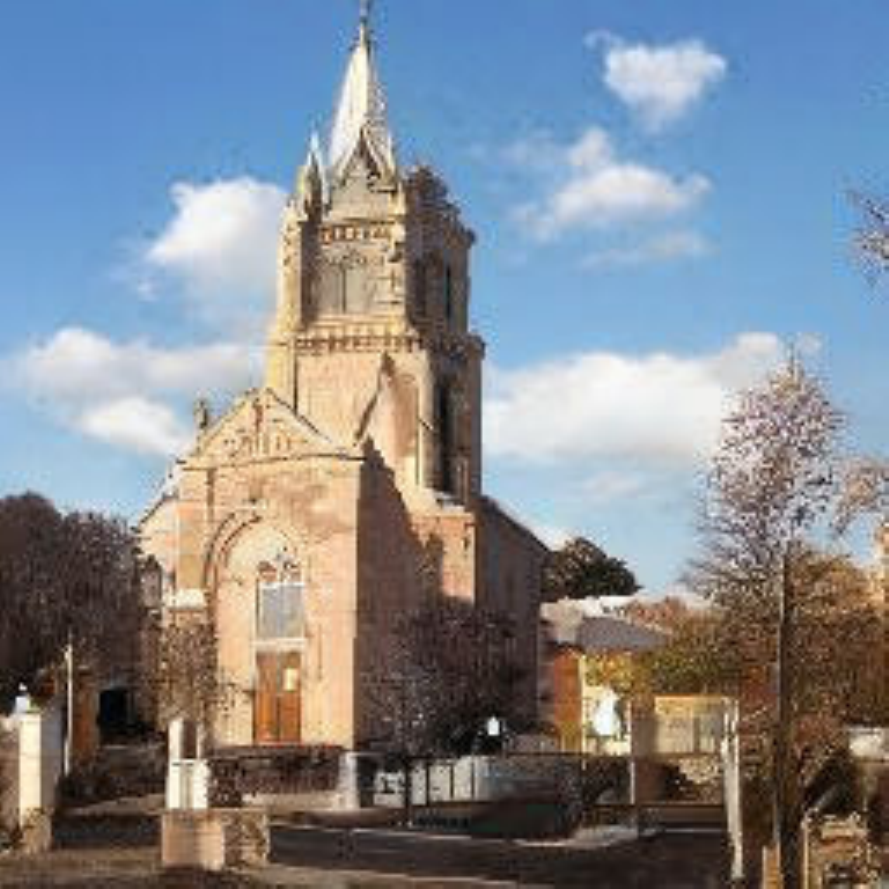} & \includegraphics[width=0.24\columnwidth]{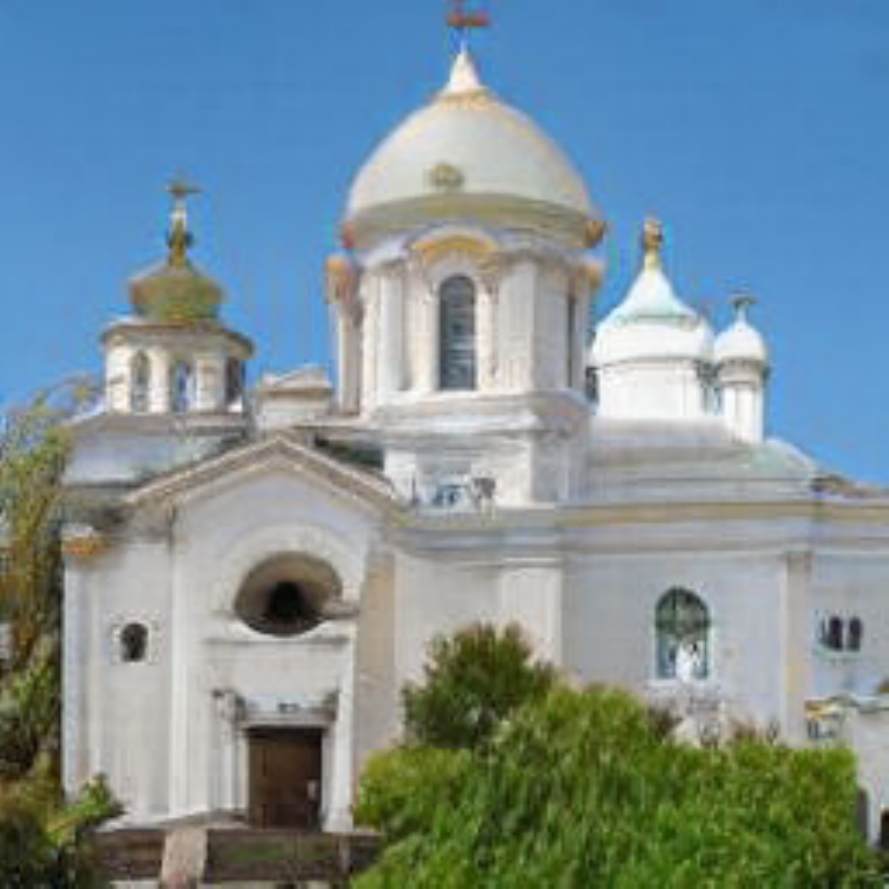}\\
    \end{tabular}
    }
    \caption{Image samples of LSUN Church $256\times 256$.}
    \label{fig:LSUNChurch256_supp}
\end{figure*}

\newpage
\begin{figure*}[h]
    \center
    \small
    \setlength\tabcolsep{0pt}
    \renewcommand{\arraystretch}{0}
    {
    \begin{tabular}{@{}cccc@{}}
         \includegraphics[width=0.24\columnwidth]{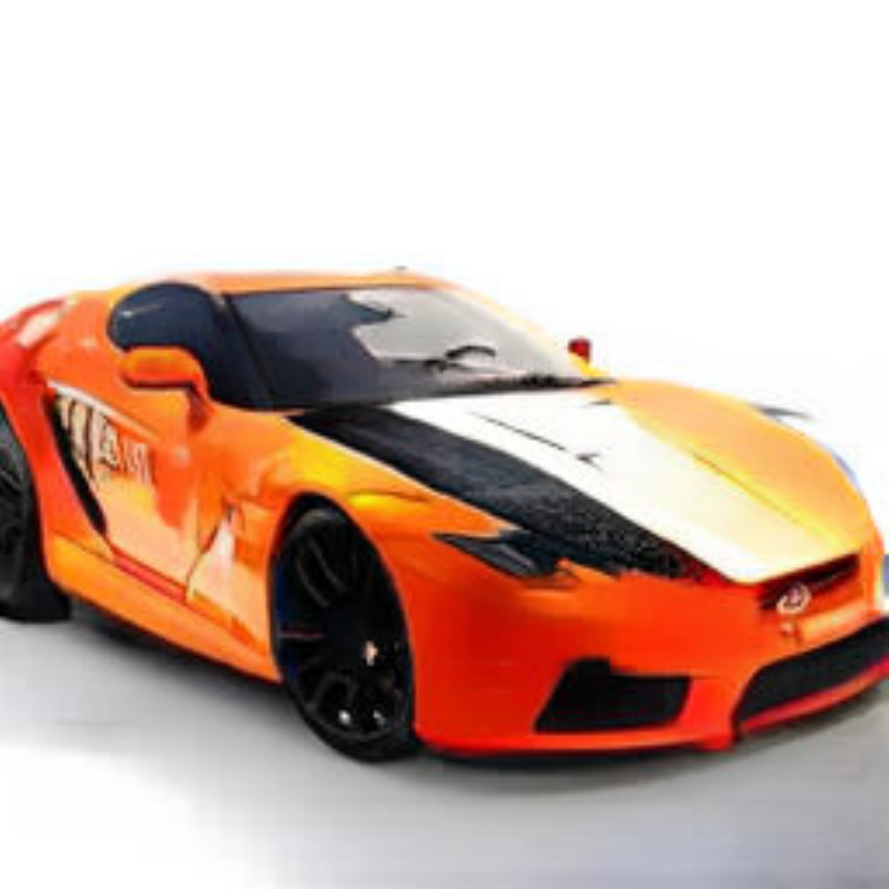} & \includegraphics[width=0.24\columnwidth]{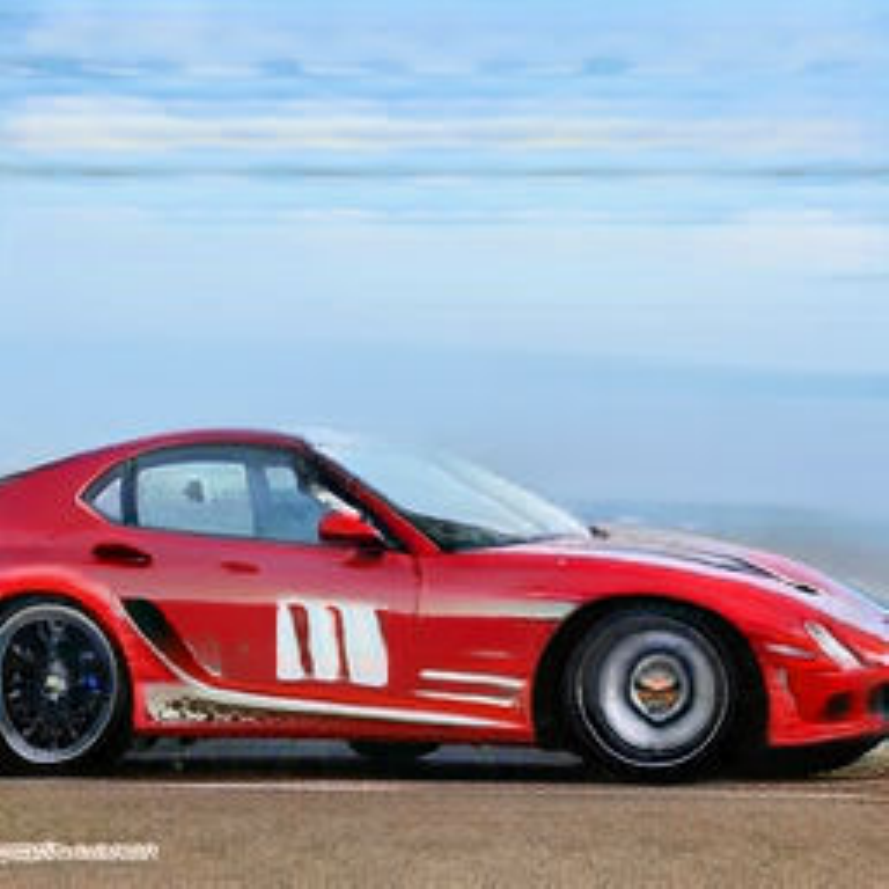} &
         \includegraphics[width=0.24\columnwidth]{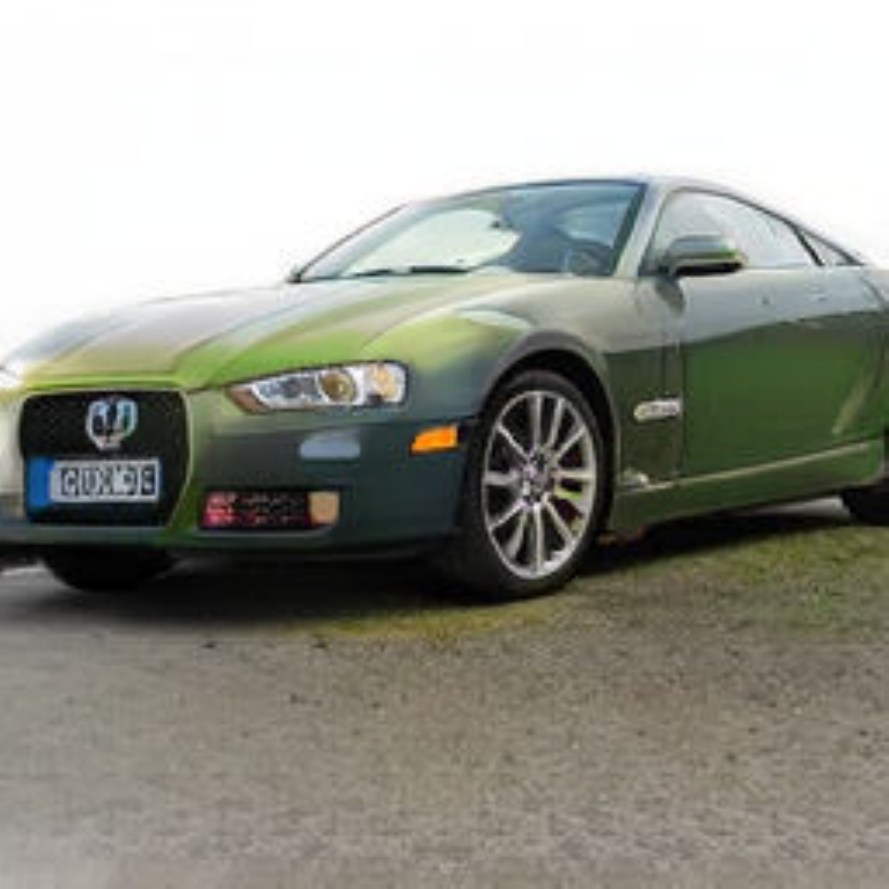} & \includegraphics[width=0.24\columnwidth]{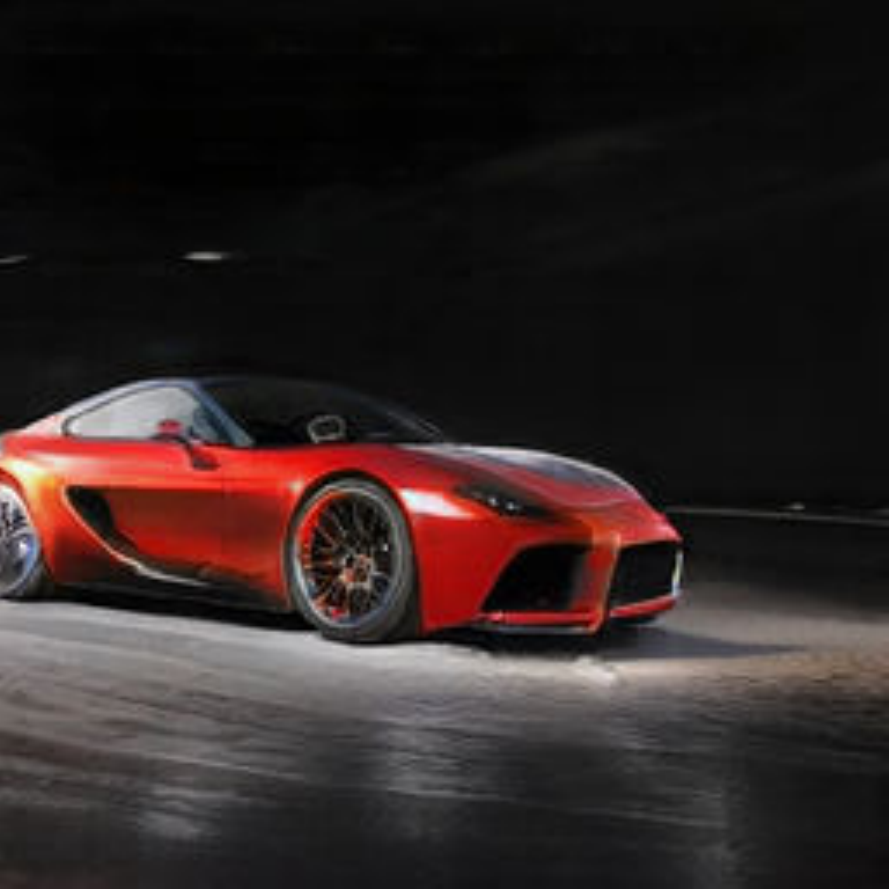}\\
         \includegraphics[width=0.24\columnwidth]{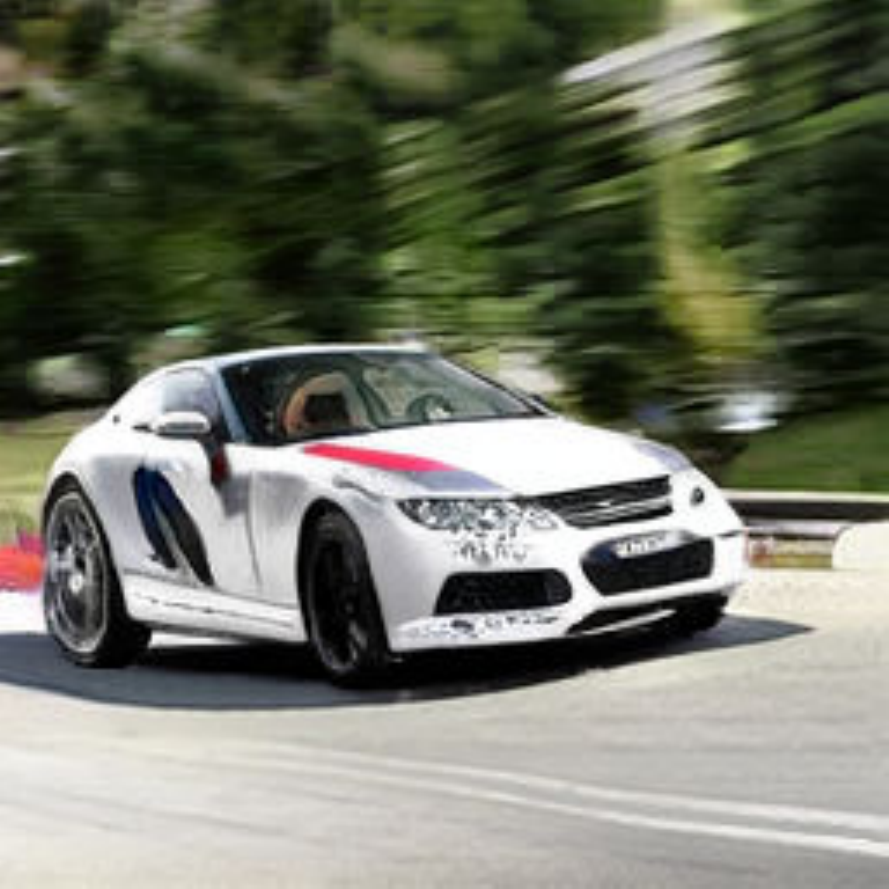} & \includegraphics[width=0.24\columnwidth]{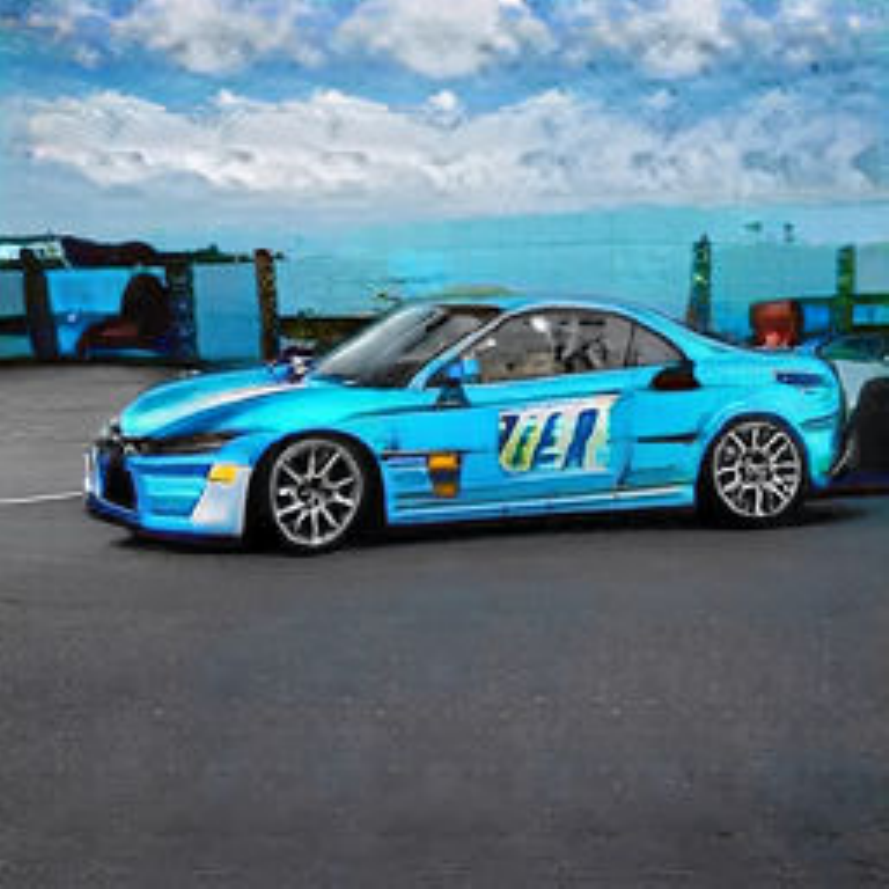} &
         \includegraphics[width=0.24\columnwidth]{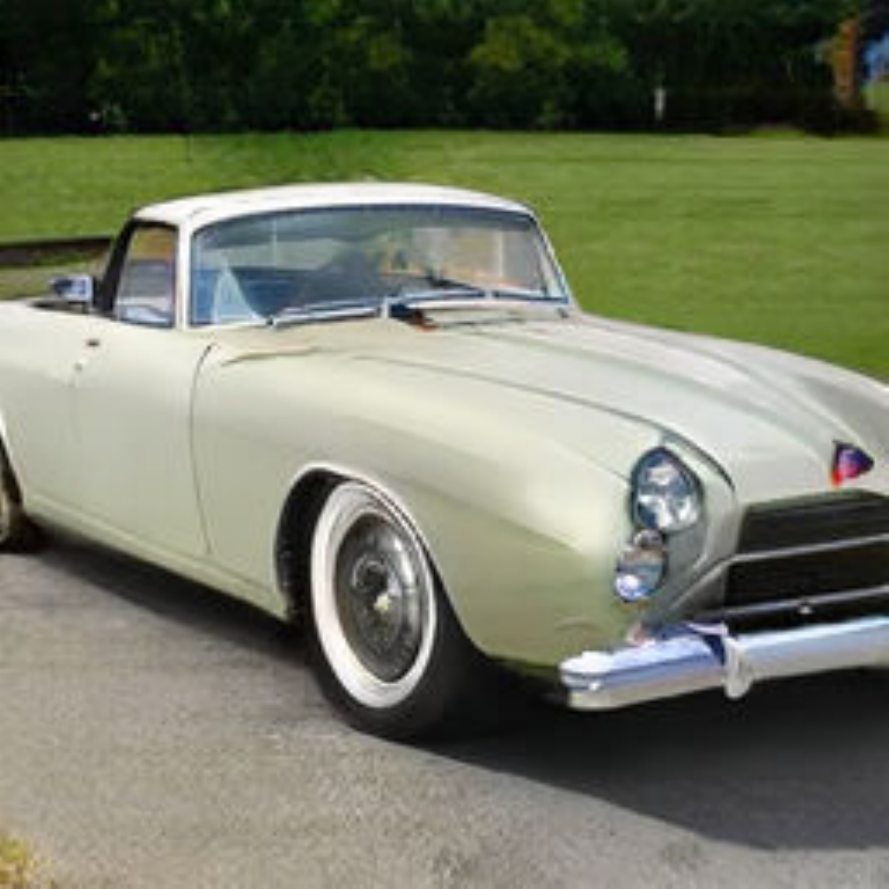} & \includegraphics[width=0.24\columnwidth]{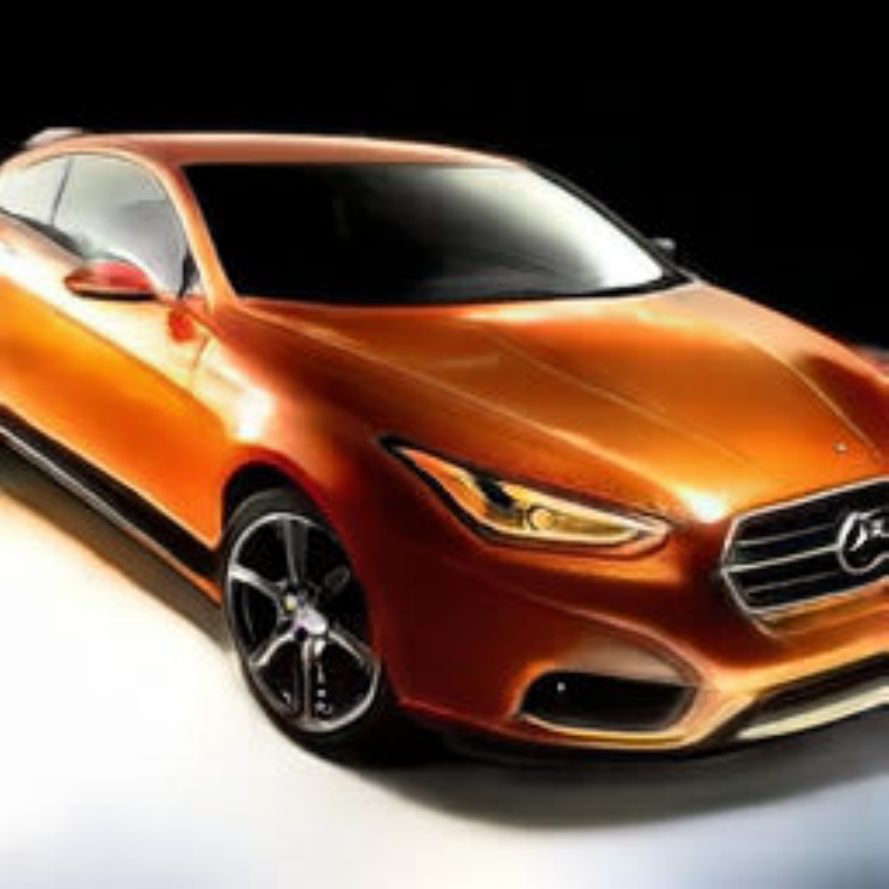}\\
         \includegraphics[width=0.24\columnwidth]{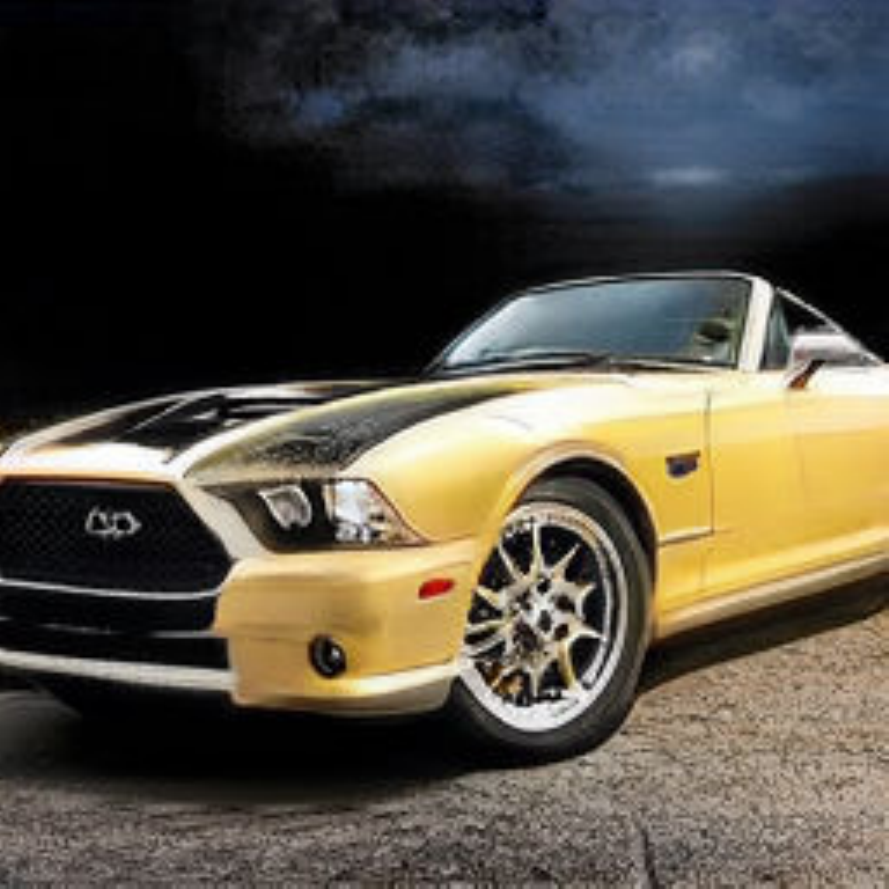} & \includegraphics[width=0.24\columnwidth]{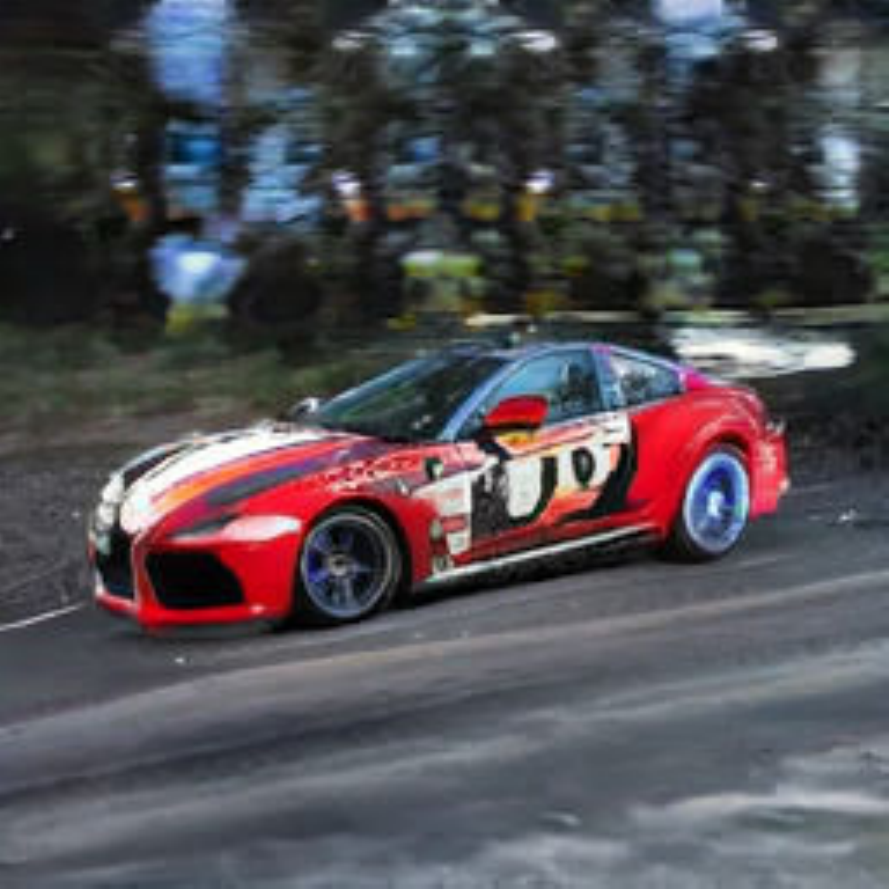} &
         \includegraphics[width=0.24\columnwidth]{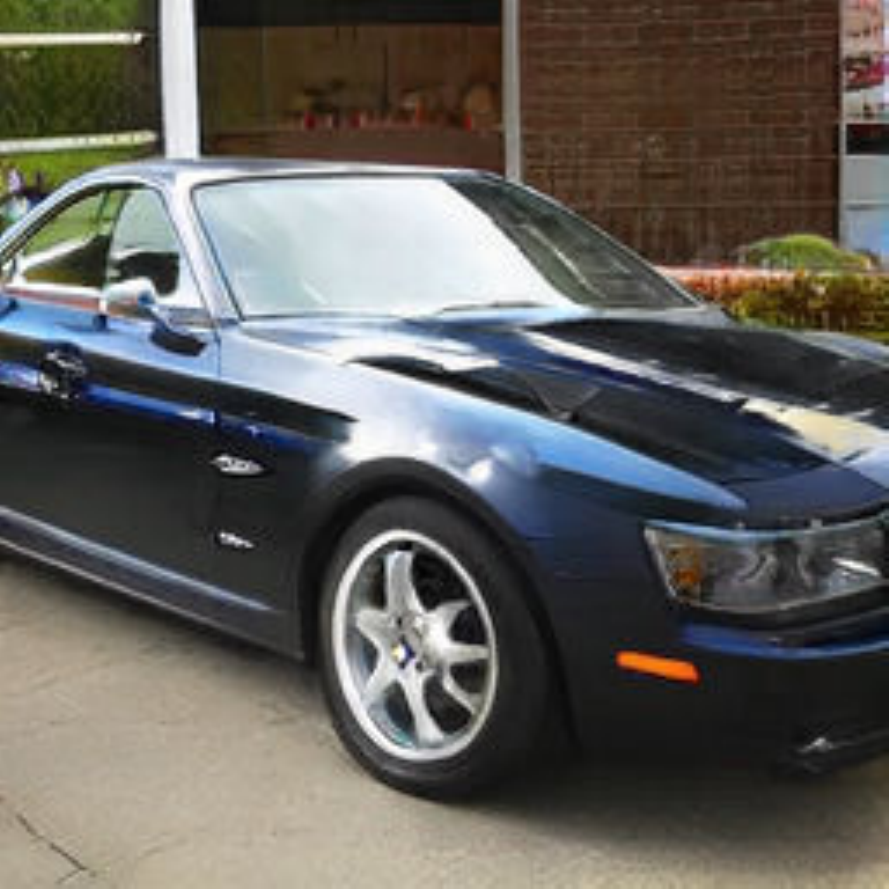} & \includegraphics[width=0.24\columnwidth]{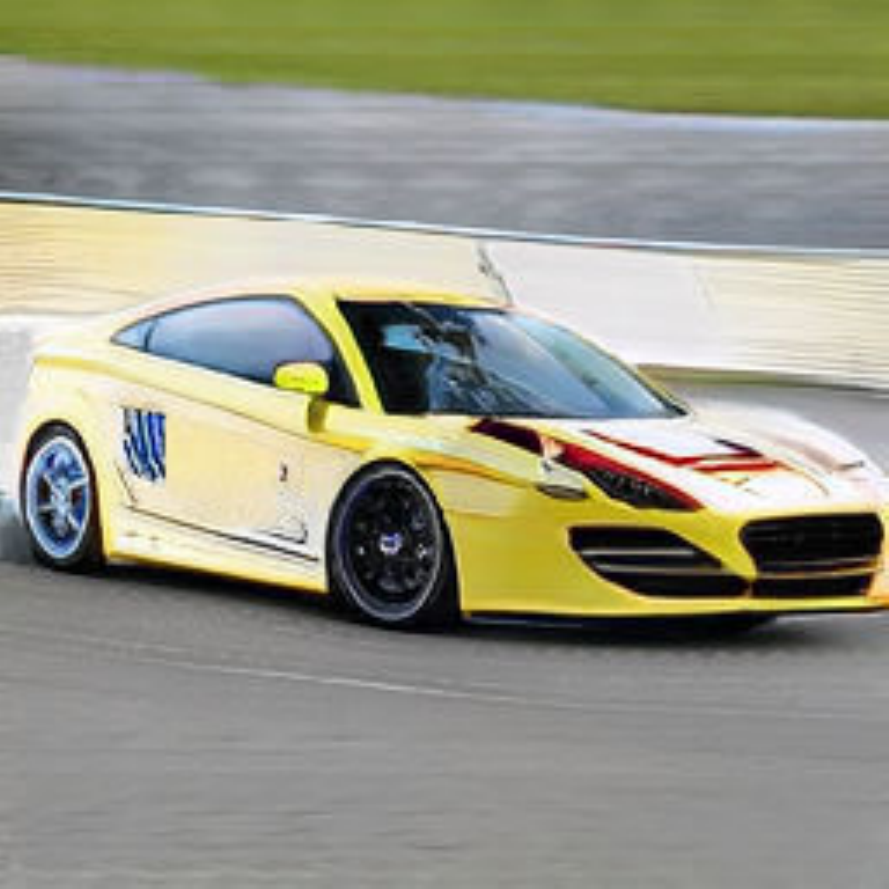}\\
         \includegraphics[width=0.24\columnwidth]{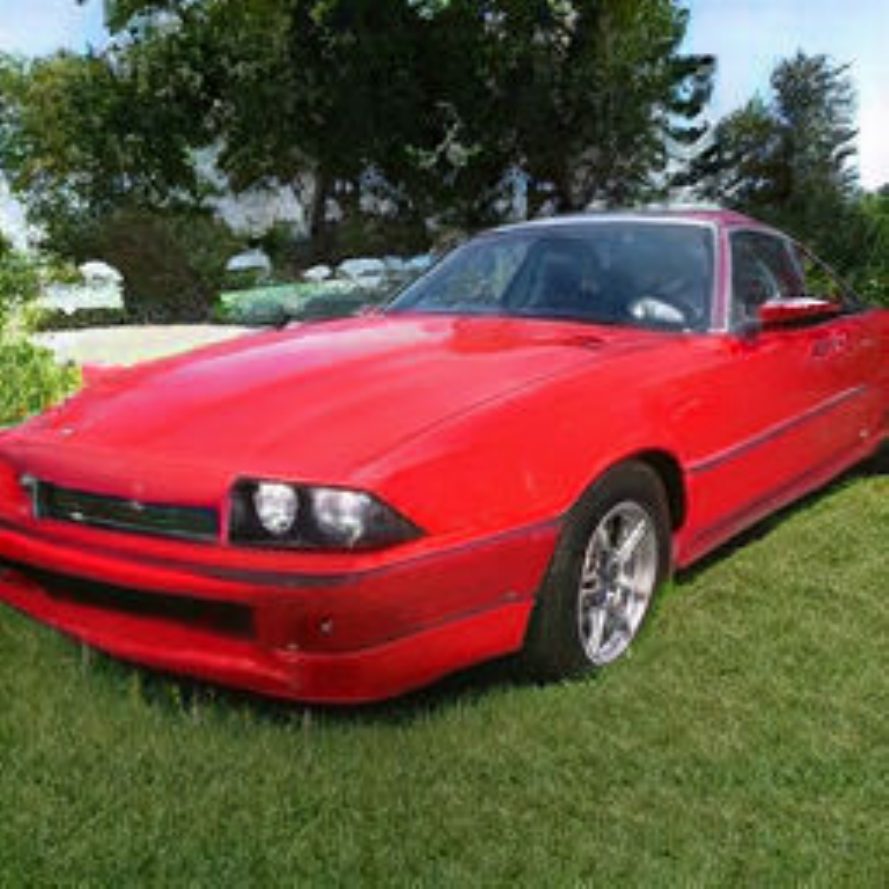} & \includegraphics[width=0.24\columnwidth]{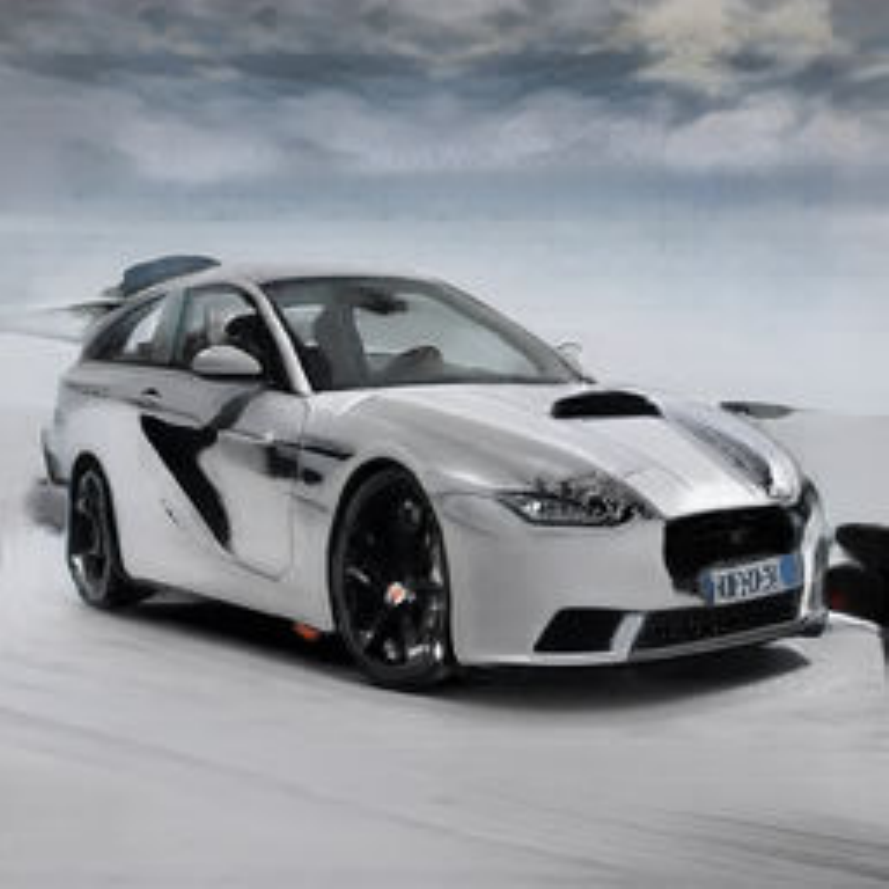} &
         \includegraphics[width=0.24\columnwidth]{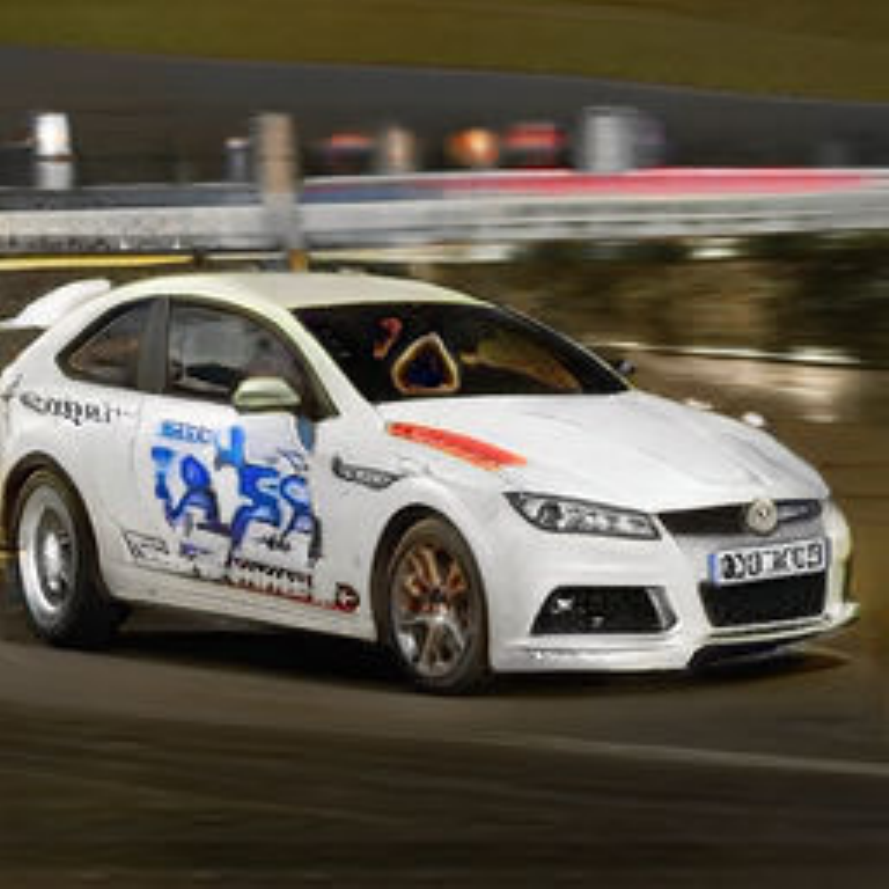} & \includegraphics[width=0.24\columnwidth]{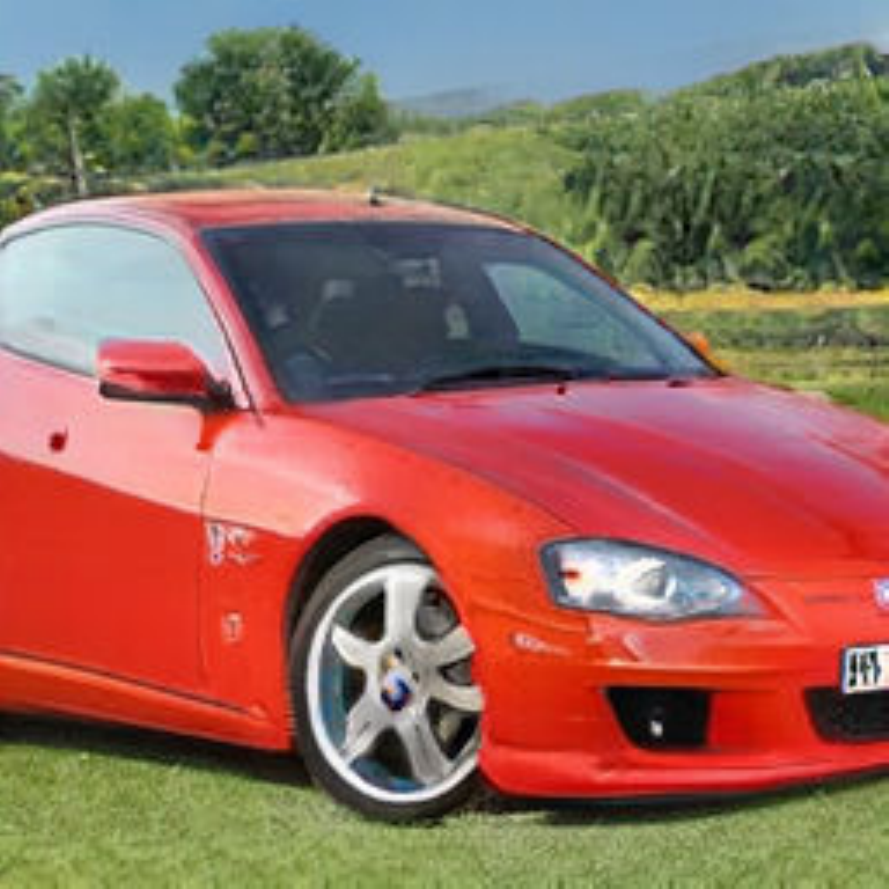}\\
         \includegraphics[width=0.24\columnwidth]{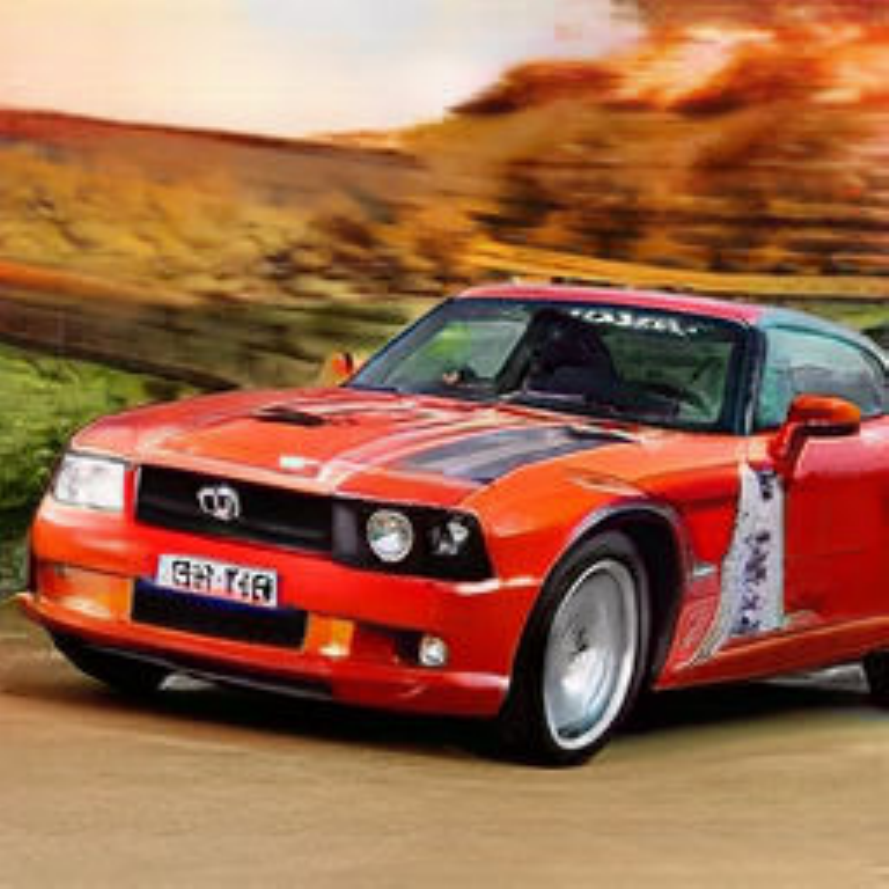} & \includegraphics[width=0.24\columnwidth]{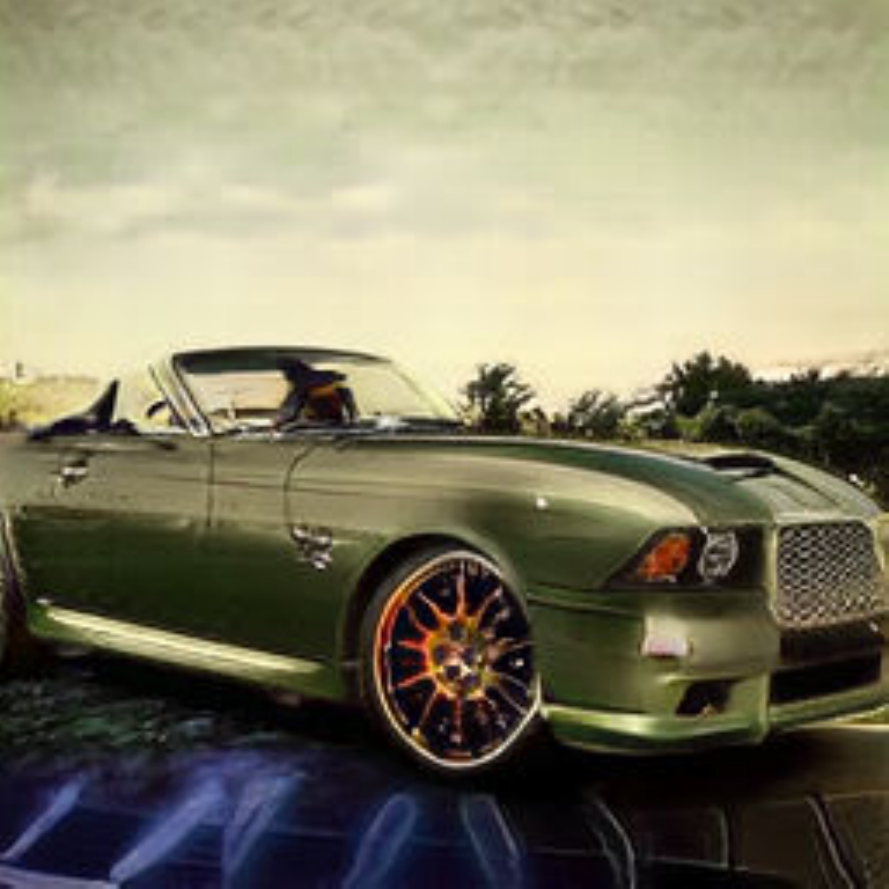} &
         \includegraphics[width=0.24\columnwidth]{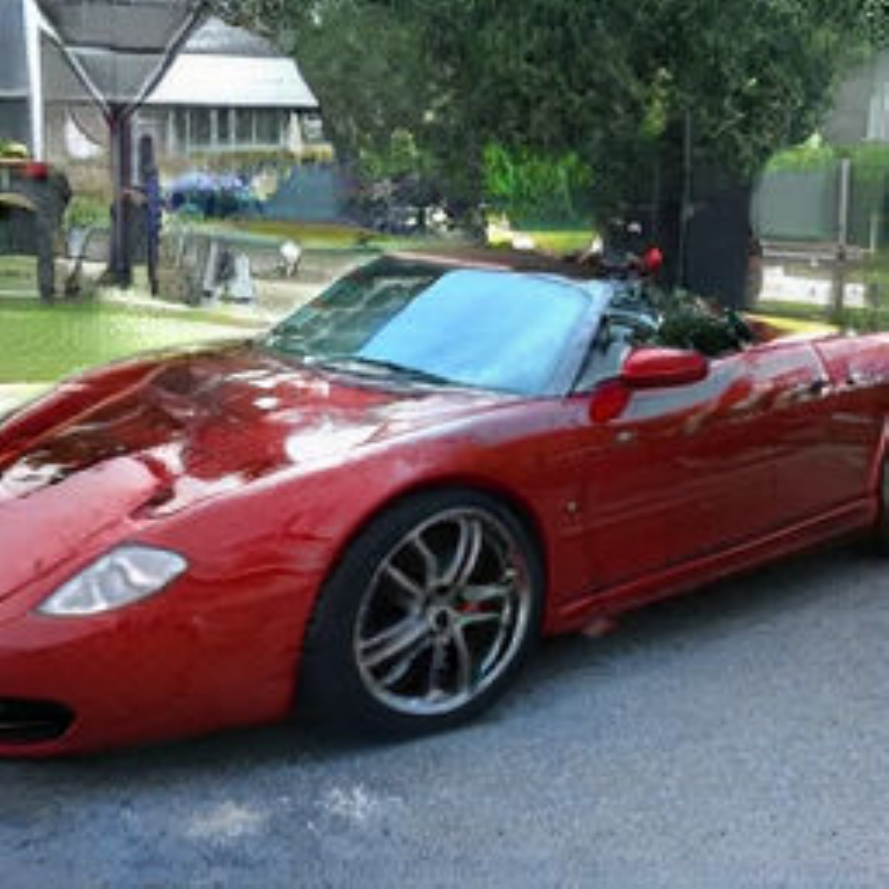} & \includegraphics[width=0.24\columnwidth]{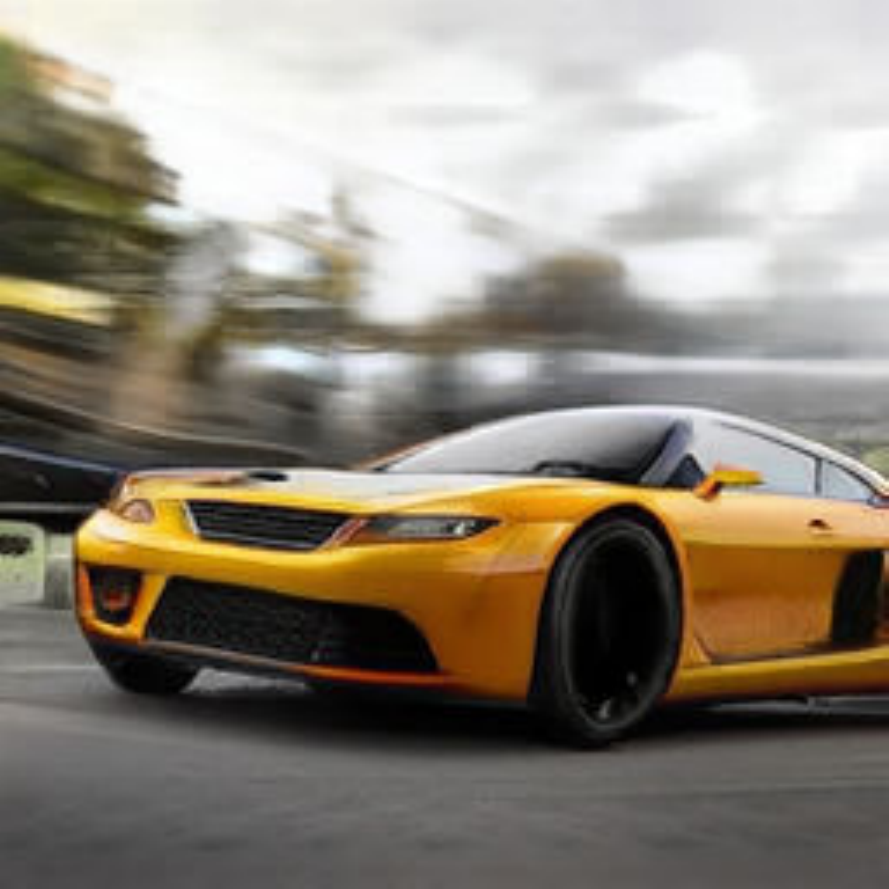}\\
    \end{tabular}
    }
    \caption{Image samples of LSUN Car $256\times 256$.}
    \label{fig:LSUNCar256_supp}
\end{figure*}

\end{appendices}
}

\end{document}